# Minimalist Data Wrangling with Python

**Marek Gagolewski**

**Marek Gagolewski**
Deakin University, Australia
Warsaw University of Technology, Poland
https://www.gagolewski.com

Product and company names mentioned herein may be the trademarks of their respective owners. Rather than use a trademark symbol with every occurrence of a trademarked name, the names are used in an editorial fashion to the benefit of the trademark owner, with no intention of infringement of the trademark.

Weird is the world we live in, but the following had to be written.

Every effort has been made in the preparation of this book to ensure the accuracy of the information presented. However, the information contained in this book is provided without warranty, either express or implied. The author will of course not be held liable for any damages caused or alleged to be caused directly or indirectly by this book.

Anyway, any bug reports/corrections/feature requests are welcome. To make this textbook even better, please file them at https://github.com/gagolews/datawranglingpy.

Typeset with XeLaTeX. Please be understanding: it was an algorithmic process, hence the results are $\in$ [good enough, perfect).

# *Contents*

## III  Multidimensional Data

## 7  Multidimensional Numeric Data at a Glance

## 8  Processing Multidimensional Data

## V   Other Data Types                                             343

## 14   Text Data                                                   345

*Minimalist Data Wrangling with Python* is envisaged as a student's first **introduction to data science**, providing a high-level overview as well as discussing key concepts in detail. We explore methods for cleaning data gathered from different sources, transforming, selecting, and extracting features, performing exploratory data analysis and dimensionality reduction, identifying naturally occurring data clusters, modelling patterns in data, comparing data between groups, and reporting the results.

For many students around the world, educational resources are hardly affordable. Therefore, I have decided that this book should **remain an independent, non-profit, open-access project** (available both in PDF[1] and HTML[2] forms). Whilst, for some people, the presence of a "designer tag" from a major publisher might still be a proxy for quality, it is my hope that this publication will prove useful to those who seek knowledge for knowledge's sake.

**Please spread the news** about it by sharing the above URL with your mates/peers/students. Thanks.

Also, consider citing this book as: Gagolewski M. (2022), *Minimalist Data Wrangling with Python*, Zenodo, Melbourne, DOI: 10.5281/zenodo.6451068[3], ISBN: 978-0-6455719-1-2, URL: https://datawranglingpy.gagolewski.com/.

Any bug/typo reports/fixes[4] are appreciated. Although available online, this is a whole course, and should be read from the beginning to the end. In particular, refer to the *Preface* for general introductory remarks.

---

*Preface*

## 0.1 The Art of Data Wrangling

*Data science*[5] aims at making sense of and generating predictions from data that have[6] been collected in copious quantities from various sources, such as physical sensors, surveys, online forms, access logs, and (pseudo)random number generators, to name a few. They can take diverse forms, e.g., vectors, matrices, and other tensors, graphs/networks, audio/video streams, or text.

Researchers in psychology, economics, sociology, agriculture, engineering, cybersecurity, biotechnology, pharmacy, sports science, medicine, and genetics, amongst many others, need statistical methods to make discoveries as well as confirm or falsify existing theories. What is more, with the increased availability of open data, everyone can do remarkable work for the common good, e.g., by volunteering for non-profit NGOs or debunking false news and overzealous acts of wishful thinking on any side of the political spectrum.

Furthermore, data scientists, machine learning engineers, statisticians, and business analysts are the most well-paid specialists[7]. This is because data-driven decision-making, modelling, and prediction proved themselves especially effective in many domains, including healthcare, food production, pharmaceuticals, transportation, financial services (banking, insurance, investment funds), real estate, and retail.

Overall, data science (and its different flavours, including operational research, machine learning, business and artificial intelligence) can be applied wherever we have some relevant data at hand and there is a need to improve or understand the underlying processes.

**Exercise 0.1** *Miniaturisation, increased computing power, cheaper storage, and the popularity of various internet services all caused data to become ubiquitous. Think about how much information people consume and generate when they interact with different news feeds or social media on their phones.*

Data usually do not come in a *tidy* and *tamed* form. *Data wrangling* is the very broad process of appropriately curating raw information chunks and then exploring the underlying data structure so that they become *analysable*.

---

[5] Traditionally known as *statistics*.
[6] Yes, *data* are plural (*datum* is singular).
[7] https://insights.stackoverflow.com/survey/2021#other-frameworks-and-libraries

## 0.2   Aims, Scope, and Design Philosophy

This course is envisaged as a student's first exposure to data science[8], providing a high-level overview as well as discussing key concepts at a healthy level of detail.

By no means do we have the ambition to be comprehensive with regard to any topic we cover. Time for that will come later in separate lectures on calculus, matrix algebra, probability, mathematical statistics, continuous and combinatorial optimisation, information theory, stochastic processes, statistical/machine learning, algorithms and data structures, take a deep breath, databases and big data analytics, operational research, graphs and networks, differential equations and dynamical systems, time series analysis, signal processing, etc.

Instead, we lay very solid groundwork for the above by introducing all the objects at an appropriate level of generality and building the most crucial connections between them. We provide the necessary intuitions behind the more advanced methods and concepts. This way, further courses do not need to waste our time introducing the most elementary definitions and answering the metaphysical questions like "but why do we need that (e.g., matrix multiplication) at all".

For those reasons, in this book, we explore methods for:

- performing exploratory data analysis (e.g., aggregation and visualisation),
- working with different types of data (e.g., numerical, categorical, text, time series),
- cleaning data gathered from structured and unstructured sources, e.g., by identifying outliers, normalising strings, extracting numbers from text, imputing missing data,
- transforming, selecting, and extracting features, dimensionality reduction,
- identifying naturally occurring data clusters,
- discovering patterns in data via approximation/modelling approaches using the most popular probability distributions and the easiest to understand statistical/machine learning algorithms,
- testing whether two data distributions differ significantly from each other,
- reporting the results of data analysis.

We primarily focus on methods and algorithms that have stood the test of time and that continue to inspire researchers and practitioners. They all meet a reality check that is comprised of the three following properties, which we believe are essential in practice:

---

[8] We might have entitled it *Introduction to Data Science (with Python)*.

- simplicity (and thus interpretability, being equipped with no or only a few underlying tunable parameters; being based on some sensible intuitions that can be explained in our own words),

- mathematical analysability (at least to some extent; so that we can understand their strengths and limitations),

- implementability (not too abstract on the one hand, but also not requiring any advanced computer-y hocus-pocus on the other).

---

**Note**  Many *more complex* algorithms are merely variations on or clever combinations of the more basic ones. This is why we need to study the fundamentals in great detail. We might not see it now, but this will become evident as we progress.

---

### 0.2.1  We Need Maths

The maths we introduce is the most elementary possible, in a good sense. Namely, we do not go beyond:

- simple analytic functions (affine maps, logarithms, cosines),

- the natural linear ordering of points on the real line (and the lack thereof in the case of multidimensional data),

- the sum of squared differences between things (including the Euclidean distance between points),

- linear vector/matrix algebra, e.g., to represent the most useful geometric transforms (rotation, scaling, translation),

- the frequentist interpretation (as in: *in samples of large sizes, we expect that...*) of some common objects from probability theory and statistics.

This is the kind of toolkit that we believe is a *sine qua non* requirement for every prospective data scientist. We cannot escape falling in love with it.

### 0.2.2  We Need Some Computing Environment

We no longer practice data analysis solely using a piece of paper and a pencil[9]. Over the years, dedicated computer programs that solve the *most common* problems arising in the most straightforward scenarios were developed (e.g., spreadsheet-like click-here-click-there stand-alone statistical packages). Still, *we* need a tool that will enable us to respond to *any* challenge in a manner that is scientifically rigorous (and hence well organised and reproducible).

---

[9] We acknowledge that some more theoretically inclined readers might ask the question: *but why do we need programming at all?* Unfortunately, some mathematicians forgot that probability and statistics are deeply rooted in the so-called real world. We should remember that theory beautifully supplements practice and provides us with very deep insights, but we still need to get our hands dirty from time to time.

In this course, we will be writing code in Python, which we shall introduce from scratch. Consequently, we do not require any prior programming experience.

The 2021 StackOverflow Developer Survey[10] lists it as the 2nd most popular programming language nowadays (slightly behind JavaScript, whose primary use is in Web development). Over the last few years, Python has proven to be a very robust choice for learning and applying data wrangling techniques. This is possible thanks to the famous[11] high-quality packages written by the devoted community of open-source programmers, including but not limited to `numpy`, `scipy`, `pandas`, `matplotlib`, `seaborn`, and `scikit-learn`.

Nevertheless, Python and its third-party packages are amongst *many* software tools which can help gain new knowledge from data. Other[12] open-source choices include, e.g., R[13] and Julia[14]. And many new ones will emerge in the future.

---

**Important**   We will therefore emphasise developing *transferable skills*: most of what we learn here can be applied (using a different syntax but the same kind of reasoning) in other environments. In other words, this is a course on data wrangling (*with* Python), not a course *on* Python (with examples in data wrangling).

---

We want the reader to become an *independent* user of this computing environment. Somebody who is not overwhelmed when they are faced with any intermediate-level data analysis problem. A user whose habitual response to a new challenge is not to look everything up on the internet even in the simplest possible scenarios. In other words, we value creative thinking.

We believe we have found a good trade-off between the minimal set of tools that need to be mastered and the less frequently used ones that can later be found in the documentation or on the internet. In other words, the reader will discover the joy of programming and using their logical thinking to tinker with things.

### 0.2.3   We Need Data and Domain Knowledge

There is no data science or machine learning without *data*, and data's purpose is to represent a given problem domain. Mathematics allows us to study different processes at a healthy level of abstractness/specificity. Still, we should always be familiar with the reality behind the numbers we have at hand, for example, by working closely with various experts in the field of our interest or pursuing our own study/research therein.

Courses such as this one, out of necessity, must use some generic datasets that are quite familiar to most readers (e.g., data on life expectancy and GDP in different countries, time to finish a marathon, yearly household incomes).

---

[10] https://insights.stackoverflow.com/survey/2021#technology-most-popular-technologies
[11] https://insights.stackoverflow.com/survey/2021#other-frameworks-and-libraries
[12] There are also some commercial solutions available on the market, but we believe that ultimately all software should be free. Consequently, we are not going to talk about them here at all.
[13] https://www.r-project.org/
[14] https://julialang.org/

Yet, many textbooks introduce statistical concepts using carefully crafted datasets where everything runs smoothly, and all models work out of the box. This gives a false sense of security. In practice, however, most datasets are not only unpolished but also (even after some careful treatment) uninteresting. Such is life. We will not be avoiding the *more difficult* problems during our journey.

## 0.3  Structure

This book is a whole course and should be read from the beginning to the end.

The material has been divided into five parts.

1. Introducing Python:

   - Chapter 1 discusses how to execute the first code chunks in Jupyter Notebooks, which are a flexible tool for the reproducible generation of reports from data analyses.

   - Chapter 2 introduces the basic scalar types in base Python, ways to call existing and to write our own functions, and control a code chunk's execution flow.

   - Chapter 3 mentions sequential and other iterable types in base Python; more advanced data structures (vectors, matrices, data frames) that we introduce below will build upon these concepts.

2. Unidimensional Data:

   - Chapter 4 introduces vectors from `numpy`, which we use for storing data on the real line (think: individual columns in a tabular dataset). Then, we look at the most common types of empirical distributions of data (e.g., bell-shaped, right-skewed, heavy-tailed ones).

   - In Chapter 5, we list the most basic ways for processing sequences of numbers, including methods for data aggregation, transformation (e.g., standardisation), and filtering. We also mention that a computer's floating-point arithmetic is imprecise and what we can do about it.

   - Chapter 6 reviews the most common probability distributions (normal, lognormal, Pareto, uniform, and mixtures thereof), methods for assessing how well they fit empirical data, and pseudorandom number generation that is crucial for experiments based on simulations.

3. Multidimensional Data:

   - Chapter 7 introduces matrices from `numpy`. They are a convenient means of storing multidimensional quantitative data (many points described by possibly many numerical features). We also present some methods for their visualisation (and the problems arising from our being three-dimensional creatures).

- Chapter 8 is devoted to basic operations on matrices. We will see that some of them simply extend on what we learned in Chapter 5, but there is more: for instance, we discuss how to determine the set of each point's nearest neighbours.

- Chapter 9 discusses ways to explore the most basic relationships between the variables in a dataset: the Pearson and Spearman correlation coefficients (and what it means that correlation is not causation), $k$-nearest neighbour and linear regression (including the sad cases where a model matrix is ill-conditioned), and finding interesting combinations of variables that can help reduce the dimensionality of a problem (via the so-called principal component analysis).

4. Heterogeneous Data:

- Chapter 10 introduces `Series` and `DataFrame` objects from **pandas**, which we can think of as vectors and matrices on steroids. For instance, they allow rows and columns to be labelled and columns to be of different types. We emphasise that most of what we learned in the previous chapters still applies, but there is even more: for example, methods for joining (merging) many datasets, converting between long and wide formats, etc.

- In Chapter 11, we introduce the ways to represent and handle categorical data as well as how (not) to lie with statistics.

- Chapter 12 covers the case of aggregating, transforming, and visualising data in groups defined by one or more qualitative variables, including classification with $k$-nearest neighbours (when we are asked to fill the gaps in a categorical variable). We will also try to discover the naturally occurring partitions using the $k$-means method, which is an example of a computationally hard optimisation problem that needs to be tackled with some imperfect heuristics.

- Chapter 13 is an interlude where we solve some pleasant exercises on data frames and learn the basics of SQL. This will be handy when we are faced with datasets that do not fit into a computer's memory.

5. Other Data Types:

- Chapter 14 discusses ways to handle text data and extract information from them, e.g., through regular expressions. We also briefly mention the challenges related to the processing of non-English text, including phrases like *pozdro dla ziomali z Bródna*, *Viele Grüße und viel Spaß*, and *χαίρετε*.

- Chapter 15 emphasises that some data may be missing or be questionable (e.g., censored, incorrect, rare) and what we can do about them.

- In Chapter 16, we cover the most basic methods for the processing of time

series, because, ultimately, everything changes, and we should be able to track the evolution of things.

---

**Note**  (*) The parts marked with a single or double asterisk can be skipped the first time we read this book. They are of increased difficulty and are less essential for beginner students.

---

## 0.4   The Rules

Our goal here, in the long run, is for you, dear reader, to become a skilled expert who is independent, ethical, and capable of critical thinking; one who hopefully will make a small contribution towards making this world a slightly better place.

To guide you through it, we have a few tips for you.

1. *Follow the rules.*

2. *Technical textbooks are not belletristic – purely for shallow amusement.* Sometimes a single page will be very meaning-intense. Do not try to consume too much at the same time. Go for a walk, reflect on what you learned, and build connections between different concepts. In case of any doubt, go back to one of the previous sections. Learning is an iterative process, not a linear one.

3. *Solve all the suggested exercises.* We might be introducing new concepts or developing crucial intuitions there as well. Also, try implementing most of the methods you learn about instead of looking for copy-paste solutions on the internet. How else will you be able to master the material and develop the necessary programming skills?

4. *Code is an integral part of the text.* Each piece of good code is worth 1234 words (on average). Do not skip it. On the contrary, you should play and experiment with it. Run every major line of code, inspect the results generated, and read more about the functions you use in the official documentation. What is the type (class) of the object returned? If it is an array or a data frame, what is its shape? What would happen if we replaced X with Y? Do not fret; your computer will not blow up.

5. *Harden up*[15]. Your journey towards expertise will take years, there are no shortcuts, but it will be quite enjoyable every now and then, so don't give up. Still, sitting all day in front of your computer is unhealthy – exercise and socialise between 28 and 31 times per month, because you're not, nor will ever be, a robot.

6. *Learn maths.* Our field has a very long history and stands on the shoulders of many giants; many methods we use these days are merely minor variations on the classical, fundamental results that date back to Newton, Leibniz, Gauss, and Laplace.

---

[15] Cyclists know.

Eventually, you will need some working knowledge of mathematics to understand them (linear algebra, calculus, probability and statistics). Remember that software products/APIs seem to change frequently, but they are just a facade, a flashy wrapping around the methods we were using for quite a while.

7. *Use only methods that you can explain.* You should refrain from working with algorithms/methods/models whose definitions (pseudocode, mathematical formulae, objective functions they are trying to optimise) and properties you do not know, understand, or cannot rephrase in your own words. That they might be accessible or easy to use should not make any difference to you. Also, prefer simple models over black boxes.

8. *Compromises are inevitable*[16]. There will never be a single best metric, algorithm, or way to solve all the problems. Even though some solutions might be better than others with regard to specific criteria, this will only be true under certain assumptions (*if* they fit a theoretical model). Beware that focusing too much on one aspect leads to undesirable consequences with respect to other factors, especially those that cannot be measured easily. Refraining from improving things might sometimes be better than pushing too hard. Always apply common sense.

9. *Be scientific and ethical.* Make your reports reproducible, your toolkit well-organised, and all the assumptions you make explicit. Develop a dose of scepticism and impartiality towards everything, from marketing slogans, through your ideological biases, to all hotly debated topics. Most data analysis exercises end up with conclusions like: "it's too early to tell", "data don't show it's either way", "there is a difference, but it is hardly significant", "yeah, but our sample is not representative for the entire population" – and there is nothing wrong with this. Remember that it is highly unethical to use statistics to tell lies; this includes presenting only one side of the overly complex reality and totally ignoring all the other ones (compare Rule#8). Using statistics for doing dreadful things (tracking users to find their vulnerabilities, developing products and services which are addictive) is a huge no-no!

10. *The best things in life are free.* These include the open-source software and open-access textbooks (such as this one) we use in our journey. Spread the good word about them and – if you can – don't only be a taker: contribute something valuable yourself (even as small as reporting typos in their documentation or helping others in different forums when they are stuck). After all, it is our shared responsibility.

---

[16] Some people would refer to this rule as *There is no free lunch*, but in our – overall friendly – world, many things are actually free (see Rule #9). Therefore, this name is misleading.

## 0.5   About the Author

I, Marek Gagolewski[17] (pronounced like Ma'rek Gong-olive-ski), am currently a Senior Lecturer[18] in Applied AI at Deakin University in Melbourne, VIC, Australia and an Associate Professor in Data Science (on long-term leave) at the Faculty of Mathematics and Information Science, Warsaw University of Technology, Poland and Systems Research Institute of the Polish Academy of Sciences.

My research interests are related to data science, in particular: modelling complex phenomena, developing usable, general purpose algorithms, studying their analytical properties, and finding out how people use, misuse, understand, and misunderstand methods of data analysis in research, commercial, and decision making settings. I'm an author of 85+ publications, including journal papers in outlets such as *Proceedings of the National Academy of Sciences (PNAS)*, *Information Fusion*, *International Journal of Forecasting*, *Statistical Modelling*, *Journal of Statistical Software*, *Information Sciences*, *Knowledge-Based Systems*, *IEEE Transactions on Fuzzy Systems*, and *Journal of Informetrics*.

In my "spare" time, I write books for my students and develop open-source (libre) data analysis software, such as **stringi**[19] – one of the most often downloaded R packages, **genieclust**[20] – a fast and robust clustering algorithm in both Python and R, and many others[21].

## 0.6   Acknowledgements

*Minimalist Data Wrangling with Python* is based on my experience as an author of a quite successful textbook *Przetwarzanie i analiza danych w języku Python* (Data Processing and Analysis in Python), [30] that I wrote (in Polish, 2016, published by PWN) with my former (successful) PhD students, Maciej Bartoszuk and Anna Cena – thanks! The current book is an entirely different work; however, its predecessor served as an excellent testbed for many ideas conveyed here.

The teaching style exercised in this book has proven successful in many similar courses that yours truly has been responsible for, including at Warsaw University of Technology, Data Science Retreat (Berlin), and Deakin University (Geelong/Melbourne). I thank all my students for the feedback given over the last 10 or so years.

A thank-you to all the authors and contributors of the Python packages that we use throughout this course: `numpy` [40], `scipy` [82], `matplotlib` [45], `seaborn` [83], and `pandas` [56], amongst others (as well as the many C/C++/Fortran libraries they provide wrappers for). Their version numbers are given in Section 1.4.

---

[17] https://www.gagolewski.com
[18] https://en.wikipedia.org/wiki/Senior_lecturer
[19] https://stringi.gagolewski.com
[20] https://genieclust.gagolewski.com
[21] https://github.com/gagolews

This book has been prepared in a Markdown superset called MyST[22], Sphinx[23], and TeX (XeLaTeX). Python code chunks were processed with the R (sic!) package `knitr` [89]. A little help from Makefiles, custom shell scripts, and `Sphinx` plugins (`sphinxcontrib-bibtex`[24], `sphinxcontrib-proof`[25]) dotted the $j$'s and crossed the $f$'s. The `Ubuntu Mono`[26] font is used for the display of `code`. Typesetting of the main text relies upon the *Alegreya*[27] and *Lato*[28] typefaces.

To my friends: Ania, Basia, Grzesiek, Fizz Grady, and Tessa – thanks for being so patient and for your comments about different things!

---

[22] https://myst-parser.readthedocs.io/en/latest/index.html
[23] https://www.sphinx-doc.org/
[24] https://pypi.org/project/sphinxcontrib-bibtex/
[25] https://pypi.org/project/sphinxcontrib-proof/
[26] https://design.ubuntu.com/font/
[27] https://www.huertatipografica.com/en
[28] https://www.latofonts.com/

# Part I

# Introducing Python

# 1

## *Getting Started with Python*

## 1.1 Installing Python

Python[1] was designed and implemented by the Dutch programmer Guido van Rossum in the late 1980s. It is an immensely popular[2] object-oriented programming language, particularly suitable for rapid prototyping. Its name is a tribute to the funniest British comedy troupe ever. We will surely be having a jolly good laugh[3] along our journey.

We will be using the reference implementation of the Python language (called `CPython`[4]), version 3.10 (or any later one).

Users of Unix-like operating systems (GNU/Linux[5], FreeBSD, etc.) may download Python via their native package manager (e.g., `sudo apt install python3` in Debian and Ubuntu). Then, additional Python packages (see Section 1.4) can be installed[6] by the said manager or directly from the Python Package Index (PyPI[7]) via the `pip` tool.

Users of other operating systems can download Python from the project's website or some other distribution available on the market, e.g., Anaconda or Miniconda.

**Exercise 1.1** *Install Python on your computer.*

## 1.2 Working with Jupyter Notebooks

Jupyter[8] brings a web-based development environment supporting numerous[9] programming languages. Even though, in the long run, it is not the most convenient space for exercising data science in Python (writing standalone scripts in some more advanced editors is the preferred option), we chose it here because of its educative advantages (interactive, easy to start with, etc.).

---

[1] https://www.python.org/

[2] https://insights.stackoverflow.com/survey/2021#technology-most-popular-technologies

[3] When we are all in tears because of mathematics and programming, those that we shed are often tears of joy.

[4] (*) `CPython` was written in the C programming language. Many Python packages are just convenient wrappers around code programmed in C, C++, or Fortran.

[5] GNU/Linux is the operating system of choice for machine learning engineers and data scientists both on the desktop and in the cloud. Switching to a free system at some point cannot be recommended highly enough.

[6] https://packaging.python.org/en/latest/tutorials/installing-packages/

[7] https://pypi.org/

[8] https://jupyterlab.readthedocs.io/en/stable/

[9] https://github.com/jupyter/jupyter/wiki/Jupyter-kernels

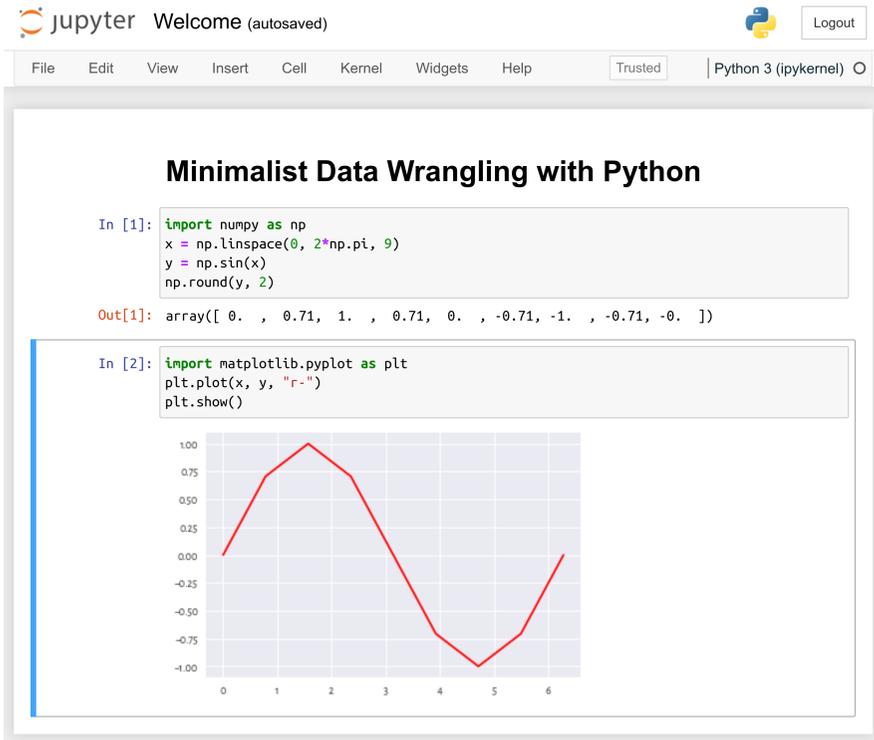

Figure 1.1: Jupyter Notebook at a glance

In Jupyter, we can work with:

- Jupyter notebooks[10] — `.ipynb` documents combining code, text, plots, tables, and other rich outputs; importantly, code chunks can be created, modified, and run interactively, which makes it a good reporting tool for our basic data science needs; see Figure 1.1;

- code consoles — terminals for running code chunks interactively (read-eval-print loop);

- source files in many different languages — with syntax highlighting and the ability to send code to the associated consoles;

and many more.

**Exercise 1.2** *Head to the official documentation[11] of the Jupyter project. Watch the introductory video linked in the* Overview *section.*

---

[10] https://jupyterlab.readthedocs.io/en/stable/user/notebook.html
[11] https://jupyterlab.readthedocs.io/en/stable/index.html

**Note** (*) More advanced students might consider, for example, **jupytext**[12] as a means to create `.ipynb` files directly from Markdown documents.

## 1.2.1 Launching JupyterLab

How we launch JupyterLab (or its lightweight version, Jupyter Notebook) will vary from system to system. Everyone needs to determine the best way to do it by themself.

Some users will be able to start JupyterLab via their start menu/application launcher. Alternatively, we can open the system terminal (**bash**, **zsh**, etc.) and type:

```
cd our/favourite/directory   # change directory
jupyter lab   # or jupyter-lab, depending on the system
```

This should launch the JupyterLab server and open the corresponding web app in the default web browser.

**Note** Some commercial cloud-hosted instances or forks of the open-source Jupyter-Lab project are available on the market, but we endorse none of them (even though they might be provided gratis, there are always strings attached). It is best to run our applications locally, where we are free[13] to be in control over the software environment.

## 1.2.2 First Notebook

Here is how we can create our first notebook.

1. From JupyterLab, create a new notebook running a Python 3 kernel (for example, by selecting File → New → Notebook from the menu).

2. Select File → Rename Notebook and change the filename to `HelloWorld.ipynb`.

   **Important** The file is stored relative to the current working directory of the running JupyterLab server instance. Make sure you can locate `HelloWorld.ipynb` on your disk using your favourite file explorer (by the way, `.ipynb` is just a JSON file that can also be edited using an ordinary text editor).

3. Input the following in the code cell:

   ```
   print("G'day!")
   ```

---

[12] https://jupytext.readthedocs.io/en/latest/
[13] https://www.youtube.com/watch?v=Ag1AKIl_2GM

4. Press `Ctrl+Enter` (or `Cmd+Return` on macOS) to execute the code cell and display the result; see Figure 1.2.

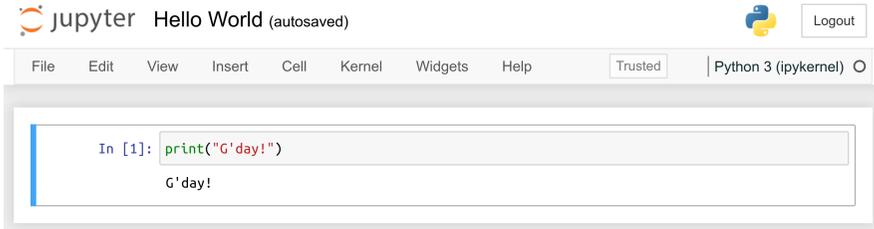

Figure 1.2: "Hello World" in a Jupyter notebook

### 1.2.3   More Cells

Time for some more cells.

1. By pressing `Enter`, we can enter the *Edit mode*. Modify the cell's contents so that it now reads:

```python
# My first code cell (this is a comment)
print("G'day!")  # prints a message (this is a comment too)
print(2+5)  # prints a number
```

2. Press `Ctrl+Enter` to execute the code and replace the previous outputs with the new ones.

3. Enter a command to print some other message that is to your liking. Note that character strings in Python must be enclosed in either double quotes or apostrophes.

4. Press `Shift+Enter` to execute the code cell, create a new one below, and then enter the edit mode.

5. In the new cell, enter and then execute the following:

```python
import matplotlib.pyplot as plt  # basic plotting library
plt.bar(
    ["Python", "JavaScript", "HTML", "CSS"],  # a list of strings
    [80, 30, 10, 15]  # a list of integers (the corresponding bar heights)
)
plt.title("What makes you happy?")
plt.show()
```

6. Add three more code cells, displaying some text or creating other bar plots.

**Exercise 1.3**  *Change **print**(2+5) to **PRINT**(2+5). Execute the code chunk and see what happens.*

**Note** In the *Edit* mode, JupyterLab behaves like an ordinary text editor. Most keyboard shortcuts known from elsewhere are available, for example:

- `Shift+LeftArrow, DownArrow, UpArrow,` or `RightArrow` – select text,
- `Ctrl+c` – copy,
- `Ctrl+x` – cut,
- `Ctrl+v` – paste,
- `Ctrl+z` – undo,
- `Ctrl+]` – indent,
- `Ctrl+[` – dedent,
- `Ctrl+/` – toggle comment.

### 1.2.4 Edit vs Command Mode

Moreover:

1. By pressing `ESC`, we can enter the *Command mode*.

2. In the *Command mode*, we can use the arrow `DownArrow` and `UpArrow` keys to move between the code cells.

3. In the *Command mode*, pressing `d,d` (`d` followed by another `d`) deletes the currently selected cell.

4. Press `z` to undo the last operation.

5. Press `a` and `b` to insert a new blank cell, respectively, above and below the current one.

6. Note a simple drag and drop can relocate cells.

**Important** `ESC` and `Enter` switch between the *Command* and *Edit* modes, respectively.

### 1.2.5 Markdown Cells

So far we have only been playing with *code* cells. We can convert the current cell to a Markdown block by pressing `m` in the *Command mode* (note that by pressing `y` we can turn it back to a *code* cell).

Markdown is a lightweight, human-readable markup language widely used for formatting text documents.

1. Enter the following into a new Markdown cell:

```
# Section

## Subsection

This ~~was~~ *is* **really** nice.

* one
* two
    1. aaa
    2. bbbb
* three

```python
# some code to display (but not execute)
2+2
```

```

2. Press `Ctrl+Enter` to display the formatted text.

3. Notice that Markdown cells can be modified by entering the *Edit mode* as usual (`Enter` key).

**Exercise 1.4**  *Read the official introduction[14] to the Markdown syntax.*

**Exercise 1.5**  *Follow this[15] interactive Markdown tutorial.*

**Exercise 1.6**  *Apply what you learned by making the current Jupyter notebook more readable. Add a header at the beginning of the report featuring your name and email address. Before and after each code cell, explain (in your own words) its purpose and how we can interpret the obtained results.*

## 1.3  The Best Note-Taking App

Learning, and this is what we are here for, will not be effective without making notes of the concepts that we come across during this course – many of them will be new to us. We will need to write down some definitions and noteworthy properties of the methods we discuss, draw simple diagrams and mind maps to build connections between different topics, check intermediate results, or derive simple mathematical formulae ourselves.

---

[14] https://daringfireball.net/projects/markdown/syntax
[15] https://commonmark.org/help/tutorial/

Let us not waste our time finding the best app for our computers, phones, or tablets. The best and most versatile note-taking solution is an ordinary piece of A4 paper and a pen or a pencil. Loose sheets of paper, 5 mm grid-ruled for graphs and diagrams, work nicely. They can be held together using a cheap landscape clip folder (the one with a clip on the long side). An advantage of this solution is that it can be browsed through like an ordinary notebook. Also, new pages can be added anywhere, and their ordering altered arbitrarily.

---

## 1.4    Initialising Each Session and Getting Example Data (!)

From now on, we assume that the following commands are issued at the beginning of each session:

```python
# import key packages – required:
import numpy as np
import scipy.stats
import pandas as pd
import matplotlib.pyplot as plt
import seaborn as sns

# further settings – optional:
pd.set_option("display.notebook_repr_html", False)  # disable "rich" output

import os
os.environ["COLUMNS"] = "74"   # output width, in characters
np.set_printoptions(linewidth=74)
pd.set_option("display.width", 74)

plt.style.use("seaborn")  # overall plot style

_colours = [  # the "R4" palette
    "#000000", "#DF536B", "#61D04F", "#2297E6",
    "#28E2E5", "#CD0BBC", "#F5C710", "#999999"
]

_linestyles = [
    "solid", "dashed", "dashdot", "dotted"
]

plt.rcParams["axes.prop_cycle"] = plt.cycler(
    # each plotted line will have a different plotting style
```

```
    color=_colours, linestyle=_linestyles*2
)
plt.rcParams["patch.facecolor"] = _colours[0]

np.random.seed(123)  # initialise the pseudorandom number generator
```

The above imports the most frequently used packages (together with their usual aliases, we will get to that later). Then, it sets up some further options that yours truly is particularly fond of. On a side note, for the discussion on the reproducible pseudorandom number generation, please see Section 6.4.2.

The software we use regularly receives feature upgrades, API changes, and bug fixes. It is good to know which version of the Python environment was used to evaluate all the code included in this book:

```
import sys
print(sys.version)
## 3.10.4 (main, Jun 29 2022, 12:14:53) [GCC 11.2.0]
```

The versions of the packages that we use in this course are given below. They can usually be fetched by calling, for example, **print**(np.__version__), etc.

| Package | Version |
|---|---|
| **numpy** | 1.23.1 |
| **scipy** | 1.8.1 |
| **pandas** | 1.4.3 |
| **matplotlib** | 3.5.2 |
| **seaborn** | 0.11.2 |
| **sklearn** (**scikit-learn**) (*) | 1.1.1 |
| **icu** (**PyICU**) (*) | 2.9 |
| **IPython** (*) | 8.4.0 |
| **mplfinance** (*) | 0.12.9b1 |

We expect 99% of the code listed in this book to work in future versions of our environment. If the kind reader discovers that this is not the case, filing a bug report at https://github.com/gagolews/datawranglingpy/issues will be much appreciated (for the benefit of other students).

---

**Important**  All example datasets that we use throughout this course are available for download at https://github.com/gagolews/teaching-data.

---

**Exercise 1.7**  *Ensure you are comfortable accessing raw data files from the above repository. Chose any file, e.g., nhanes_adult_female_height_2020.txt in the marek folder, and then* *click* Raw. *It is the URL that you were redirected to, not the previous one, that includes the link to be referred to from within your Python session.*

*Note that each dataset starts with several comment lines explaining its structure, the meaning of the variables, etc.*

## 1.5   Exercises

**Exercise 1.8**  *What is the difference between the* Edit *and the* Command *mode in Jupyter?*

**Exercise 1.9**  *How can we format a table in Markdown? How can we insert an image?*

# 2

---

## *Scalar Types and Control Structures in Python*

In this part, we introduce the basics of the Python language itself. As it is a general-purpose tool, various packages supporting data wrangling operations will provided as third-party extensions. In further chapters, based on the concepts discussed here, we will be able to use `numpy`, `scipy`, `pandas`, `matplotlib`, `seaborn`, and other packages with some healthy degree of confidence.

---

## 2.1 Scalar Types

The five ubiquitous scalar types (i.e., *single* or *atomic* values) are:

- `bool` – logical,

- `int`, `float`, `complex` – numeric,

- `str` – character.

### 2.1.1 Logical Values

There are only two possible logical (Boolean) values: `True` and `False`. We can type:

```
True
## True
```

to instantiate one of them. This might seem boring; unless, when trying to play with the above code, we fell into the following pitfall.

---

**Important**   Python is case-sensitive. Writing "`TRUE`" or "`true`" instead of "`True`" is an error.

---

### 2.1.2 Numeric Values

The three numeric scalar types are:

- `int` – integers, e.g., 1, -42, 1_000_000;

- `float` – floating-point (real) numbers, e.g., `-1.0`, `3.14159`, `1.23e-4`;

- `complex` (*) – complex numbers, e.g., `1+2j` (these are infrequently used in our applications; however, see Section 4.1.4).

In practice, numbers of type `int` and `float` often interoperate seamlessly. We usually do not have to think about them as being of distinctive types.

**Exercise 2.1** *`1.23e-4` and `9.8e5` are examples of numbers entered using the so-called scientific notation, where "e" stands for "times 10 to the power of". Additionally, `1_000_000` is a decorated (more human-readable) version of `1000000`. Use the `print` function to check their values.*

**Arithmetic Operators**

Here is the list of available arithmetic operators:

```
1 + 2     # addition
## 3
1 - 7     # subtraction
## -6
4 * 0.5   # multiplication
## 2.0
7 / 3     # float division (the result is always of type float)
## 2.3333333333333335
7 // 3    # integer division
## 2
7 % 3     # division remainder
## 1
2 ** 4    # exponentiation
## 16
```

The precedence of these operators is quite predictable[1], e.g., exponentiation has higher priority than multiplication and division, which in turn bind more strongly than addition and subtraction. Consequently:

```
1 + 2 * 3 ** 4   # the same as 1+(2*(3**4))
## 163
```

is different from, e.g., `((1+2)*3)**4`.

---

**Note** Keep in mind that computers' floating-point arithmetic is precise only up to a few significant digits. As a consequence, the result of `7/3` is only approximate (`2.3333333333333335`). We will get back to this topic in Section 5.5.6.

---

[1] https://docs.python.org/3/reference/expressions.html#operator-precedence

**Creating Named Variables**

Named variables can be introduced using the *assignment operator*, `=`. They can store arbitrary Python objects and be referred to at any time. Names of variables can include any lower- and uppercase letters, underscores, and digits (but not at the beginning). It is best to make them self-explanatory, like:

```
x = 7  # read: let `x` from now on be equal to 7 (or: `x` becomes 7)
```

We can check that x (great name, by the way: it means *something of general interest* in mathematics) is now available for further reference by printing out the value that is bound therewith:

```
print(x)  # or just `x`
## 7
```

New variable can easily be created based on existing ones:

```
my_2nd_variable = x/3 - 2  # creates `my_2nd_variable`
print(my_2nd_variable)
## 0.3333333333333335
```

Also, existing variables can be re-bound to any other value whenever we please:

```
x = x/3  # let the new `x` be equal to the old `x` (7) divided by 3
print(x)
## 2.3333333333333335
```

**Exercise 2.2** *Create two named variables* `height` *(in centimetres) and* `weight` *(in kilograms). Based on them, determine your* BMI[2].

---

**Note** (*) Augmented assignments are also available. For example:

```
x *= 3
print(x)
## 7.0
```

In this context, the above is equivalent to x = x*3, i.e., a new variable has been created. Nevertheless, in other scenarios, augmented assignments modify the objects they act upon in place; compare Section 3.5.

---

### 2.1.3 Character Strings

Character strings (objects of type `str`), which can consist of arbitrary text, are created using either double quotes or apostrophes:

---
[2] https://en.wikipedia.org/wiki/Body_mass_index

```python
print("spam, spam, #, bacon, and spam")
## spam, spam, #, bacon, and spam
print("Cześć! ¿Qué tal?")
## Cześć! ¿Qué tal?
print('"G\'day, howya goin\'," he asked.\n"Fine, thanks," she responded.\\')
## "G'day, howya goin'," he asked.
## "Fine, thanks," she responded.\
```

Above, "\'" (a way to include an apostrophe in an apostrophe-delimited string), "\\" (a backslash), and "\n" (a newline character) are examples of *escape sequences*[3].

Multiline strings are also possible:

```python
"""
spam\\spam
tasty\t"spam"
lovely\t'spam'
"""
## '\nspam\\spam\ntasty\t"spam"\nlovely\t\'spam\'\n'
```

**Exercise 2.3** *Call the* `print` *function on the above object to reveal the special meaning of the included escape sequences.*

---

**Important**   Many string operations are available. They are related, for example to formatting, pattern searching, or extracting matching chunks. They are especially important in the art of data wrangling as oftentimes information comes to us in textual form. We shall be covering this topic in detail in Chapter 14.

---

**F-Strings (Formatted String Literals)**

Also, the so-called *f-strings* (formatted string literals) can be used to prepare nice output messages:

```python
x = 2
f"x is {x}"
## 'x is 2'
```

Notice the "f" prefix. The "{x}" part was replaced with the value stored in the x variable.

There are many options available. As usual, it is best to study the documentation[4] in search of interesting features. Here, let us just mention that we will frequently be referring to placeholders like "{variable:width}" and "{variable:width.precision}", which specify the field width and the number of fractional digits of a number. This can result in a series of values nicely aligned one below another.

---

[3] https://docs.python.org/3/reference/lexical_analysis.html#string-and-bytes-literals
[4] https://docs.python.org/3/reference/lexical_analysis.html#f-strings

```python
π = 3.14159265358979323846
e = 2.71828182845904523536
print(f"""
π = {π:10.8f}
e = {e:10.8f}
""")
##
## π =  3.14159265
## e =  2.71828183
```

`10.8f` means that a value should be formatted as a `float`, be of at least width 10, and use eight fractional digits.

## 2.2 Calling Built-in Functions

There are quite a few built-in functions ready for use. For instance:

```python
e = 2.718281828459045
round(e, 2)
## 2.72
```

Rounds e to 2 decimal digits.

**Exercise 2.4** *Call* `help("round")` *to access the function's manual. Note that the second argument, called* `ndigits`, *which we set to 2, has a default value of* `None`. *Check what happens when we omit it during the call.*

### 2.2.1 Positional and Keyword Arguments

As **round** has two parameters, `number` and `ndigits`, the following (and no other) calls are equivalent:

```python
print(
    round(e, 2),   # two arguments matched positionally
    round(e, ndigits=2),   # positional and keyword argument
    round(number=e, ndigits=2),   # two keyword arguments
    round(ndigits=2, number=e)   # the order does not matter for keyword args
)
## 2.72 2.72 2.72 2.72
```

That no other form is allowed is left as an exercise, i.e., positionally matched arguments must be listed before the keyword ones.

### 2.2.2  Modules and Packages

Other functions are available in numerous Python modules and packages (which are collections of modules).

For example, `math` features many mathematical functions:

```python
import math   # the math module must be imported prior its first use
print(math.log(2.718281828459045))  # the natural logarithm (base e)
## 1.0
print(math.floor(-7.33))  # the floor function
## -8
print(math.sin(math.pi))  # sin(pi) equals 0 (with some numeric error)
## 1.2246467991473532e-16
```

See the official documentation[5] for the comprehensive list of objects defined therein. On a side note, all floating-point computations in any programming language are subject to round-off errors and other inaccuracies. This is why the result of $\sin \pi$ is not exactly 0, but some value very close thereto. We will elaborate on this topic in Section 5.5.6.

Packages can be given aliases, for the sake of code readability or due to our being lazy. For instance, we are used to importing the `numpy` package under the `np` alias:

```python
import numpy as np
```

And now, instead of writing, for example, `numpy.random.rand()`, we can call instead:

```python
np.random.rand()  # a pseudorandom value in [0.0, 1.0)
## 0.6964691855978616
```

### 2.2.3  Slots and Methods

Python is an object-oriented programming language. Each object is an instance of some *class* whose name we can reveal by calling the **type** function:

```python
x = 1+2j
type(x)
## <class 'complex'>
```

---

**Important**  Classes define the following kinds of *attributes*:

- *slots* – associated data,
- *methods* – associated functions.

---

**Exercise 2.5** *Call* **help(***"complex"***)** *to reveal that the* complex *class features, amongst others, the* **conjugate** *method and the* real *and* imag *slots.*

Here is how we can read the two slots:

```python
print(x.real)  # access slot `real` of object `x` of class `complex`
## 1.0
print(x.imag)
## 2.0
```

And here is an example of a method call:

```python
x.conjugate()  # equivalently: complex.conjugate(x)
## (1-2j)
```

Notably, the documentation of this function can be accessed by typing **help(**"complex.conjugate")* *(class name – dot – method name)*.

## 2.3 Controlling Program Flow

### 2.3.1 Relational and Logical Operators

We have several operators which return a single logical value:

```python
1 == 1.0  # is equal to?
## True
2 != 3  # is not equal to?
## True
"spam" < "egg" # is less than? (with respect to the lexicographic order)
## False
```

Some more examples:

```python
math.sin(math.pi) == 0.0  # well, numeric error...
## False
abs(math.sin(math.pi)) <= 1e-9  # is close to 0?
## True
```

Logical results might be combined using **and** (conjunction; for testing if both operands are true) and **or** (alternative; for determining whether at least one operand is true). Likewise, **not** (negation) is available too.

```python
3 <= math.pi and math.pi <= 4
## True
```

```
not (1 > 2 and 2 < 3) and not 100 <= 3
## True
```

Notice that **not** 100 <= 3 is equivalent to 100 > 3. Also, based on the de Morgan's laws, **not** (1 > 2 **and** 2 < 3) is true if and only if 1 <= 2 **or** 2 >= 3 holds.

**Exercise 2.6** *Assuming that p, q, r are logical and a, b, c, d are float-type variables, simplify the following expressions:*

- *not not p,*
- *not p and not q,*
- *not (not p or not q or not r),*
- *not a == b,*
- *not (b > a and b < c),*
- *not (a>=b and b>=c and a>=c),*
- *(a>b and a<c) or (a<c and a>d).*

### 2.3.2    The `if` Statement

The `if` statement allows us to execute a chunk of code conditionally, based on whether the provided expression is true or not.

For instance, given some variable:

```
x = np.random.rand()   # a pseudorandom value in [0.0, 1.0)
```

we can react enthusiastically to its being less than 0.5 (note the colon after the tested condition):

```
if x < 0.5: print("spam!")
```

which did not happen, because it is equal to:

```
print(x)
## 0.6964691855978616
```

Multiple `elif` (*else-if*) parts can also be added, followed by an optional `else` part, which is executed if all the conditions tested are not true.

```
if x < 0.25:   print("spam!")
elif x < 0.5:  print("ham!")    # i.e., x in [0.25, 0.5)
elif x < 0.75: print("bacon!")  # i.e., x in [0.5, 0.75)
```

*(continued from previous page)*

```python
else:           print("eggs!")   # i.e., x >= 0.75
## bacon!
```

If more than one statement is to be executed conditionally, an indented code block can be introduced.

```python
if x >= 0.25 and x <= 0.75:
    print("spam!")
    print("I love it!")
else:
    print("I'd rather eat spam!")
print("more spam!")  # executed regardless of the condition's state
## spam!
## I love it!
## more spam!
```

---

**Important**  The indentation must be neat and consistent. We recommend using four spaces. The reader is encouraged to try to execute the following code chunk and note what kind of error is generated:

```python
if x < 0.5:
    print("spam!")
   print("ham!")     # :(
```

---

**Exercise 2.7**  *For a given BMI, print out the corresponding category as defined by the WHO (underweight if below 18.5, normal range up to 25.0, etc.). Let us bear in mind that the BMI is a simplistic measure. Both the medical and statistical communities point out its inherent limitations. Read the Wikipedia article thereon for more details (and appreciate the amount of data wrangling required for its preparation – tables, charts, calculations; something that we will be able to do quite soon, given good reference data, of course).*

**Exercise 2.8**  *(\*) Check if it is easy to find on the internet (in reliable sources) some raw data sets related to the body mass studies, e.g., measuring subjects' height, weight, body fat and muscle percentage, etc.*

### 2.3.3  The `while` Loop

The `while` loop executes a given statement or a series of statements as long as a given condition is true.

For example, here is a simple simulator determining how long we have to wait until drawing the first number not greater than 0.01 whilst generating numbers in the unit interval:

```
count = 0
while np.random.rand() > 0.01:
    count = count + 1
print(count)
## 117
```

**Exercise 2.9** *Using the* `while` *loop, determine the arithmetic mean of 10 randomly generated numbers (i.e., the sum of the numbers divided by 10).*

## 2.4   Defining Own Functions

We can also define our own functions as a means for code reuse. For instance, below is one that computes the minimum (with respect to the `<` relation) of three given objects:

```
def min3(a, b, c):
    """
    A function to determine the minimum of three given inputs.

    By the way, this is a docstring (documentation string);
    call help("min3") later.
    """
    if a < b:
        if a < c:
            return a
        else:
            return c
    else:
        if b < c:
            return b
        else:
            return c
```

Example calls:

```
print(min3(10, 20, 30),
      min3(10, 30, 20),
      min3(20, 10, 30),
      min3(20, 30, 10),
      min3(30, 10, 20),
      min3(30, 20, 10))
## 10 10 10 10 10 10
```

Note that the function *returns* a value. The result can be fetched and used in further computations:

```python
x = min3(np.random.rand(), 0.5, np.random.rand())  # minimum of 3 numbers
x = round(x, 3)  # do something with the result
print(x)
## 0.5
```

**Exercise 2.10** *Write a function named **bmi** which computes and returns a person's BMI, given their weight (in kilograms) and height (in centimetres). As documenting functions constitutes a good development practice, do not forget about including a docstring.*

We can also introduce new variables inside a function's body. This can help the function perform what it has been designed to do.

```python
def min3(a, b, c):
    """
    A function to determine the minimum of three given inputs
    (alternative version).
    """
    m = a  # a local (temporary/auxiliary) variable
    if b < m:
        m = b
    if c < m:   # be careful! no `else` or `elif` here – it's a separate `if`
        m = c
    return m
```

Example call:

```python
m = 7
n = 10
o = 3
min3(m, n, o)
## 3
```

All *local variables* cease to exist after the function is called. Notice that `m` inside the function is a variable independent of `m` in the global (calling) scope.

```python
print(m)  # this is still the global `m` from before the call
## 7
```

**Exercise 2.11** *Write a function **max3** which determines the maximum of three given values.*

**Exercise 2.12** *Write a function **med3** which defines the median of three given values (the one value that is in-between the other ones).*

**Exercise 2.13** (*) *Write a function **min4** to compute the minimum of four values.*

---

**Note** *Lambda expressions* give us an uncomplicated way to define functions using a single line of code. Their syntax is: `lambda argument_name: return_value`.

```python
square = lambda x: x**2  # i.e., def square(x): return x**2
square(4)
## 16
```

---

Objects generated through lambda expressions do not have to be assigned a name – they can be anonymous. This is useful when calling methods that take other functions as their arguments. With lambdas, the latter can be generated on the fly.

```python
def print_x_and_fx(x, f):
    """
    Arguments: x - some object; f - a function to be called on x
    """
    print(f"x = {x} and f(x) = {f(x)}")

print_x_and_fx(4, lambda x: x**2)
## x = 4 and f(x) = 16
print_x_and_fx(math.pi/4, lambda x: round(math.cos(x), 5))
## x = 0.7853981633974483 and f(x) = 0.70711
```

## 2.5 Exercises

**Exercise 2.14** *What does* `import xxxxxx as x` *mean?*

**Exercise 2.15** *What is the difference between* `if` *and* `while`?

**Exercise 2.16** *Name the scalar types we introduced in this chapter.*

**Exercise 2.17** *What is a docstring and how can we create and access it?*

**Exercise 2.18** *What are keyword arguments?*

# 3

## Sequential and Other Types in Python

## 3.1 Sequential Types

*Sequential* objects store data items that can be accessed by index (position). The three main types of sequential objects are: lists, tuples, and ranges.

As a matter of fact, strings (which we often treat as scalars) can also be classified as such. Therefore, amongst sequential objects are such diverse classes as:

- lists,
- tuples,
- ranges, and
- strings.

### 3.1.1 Lists

*Lists* consist of arbitrary Python objects. They are created using square brackets:

```python
x = [True, "two", 3, [4j, 5, "six"], None]
print(x)
## [True, 'two', 3, [4j, 5, 'six'], None]
```

Above is an example list featuring objects of type: `bool`, `str`, `int`, `list` (yes, it is possible to have a list inside another list), and `None` (the `None` object is the only of this kind, it represents a placeholder for nothingness), in this order.

---

**Note** We will often be using lists when creating vectors in **numpy** or data frame columns in **pandas**. Further, lists of lists of equal lengths can be used to create matrices.

---

Each list is *mutable*. Consequently, its state may be changed arbitrarily. For instance, we can append a new object at its end:

```python
x.append("spam")
print(x)
## [True, 'two', 3, [4j, 5, 'six'], None, 'spam']
```

The **`list.append`** method modified x in place.

### 3.1.2 Tuples

Next, *tuples* are like lists, but they are *immutable* (read-only) – once created, they cannot be altered.

```
("one", [], (3j, 4))
## ('one', [], (3j, 4))
```

This gave us a triple (a 3-tuple) featuring a string, an empty list, and a pair (a 2-tuple). Let us stress that we can drop the round brackets and still get a tuple:

```
1, 2, 3  # the same as `(1, 2, 3)`
## (1, 2, 3)
```

Also:

```
42,  # equivalently: `(42, )`
## (42,)
```

Note the trailing comma; the above notation defines a singleton (a 1-tuple). It is not the same as the simple 42 or (42), which is an object of type `int`.

---

**Note**  Having a separate data type representing an immutable sequence makes sense in certain contexts. For example, a data frame's *shape* is its inherent property that should not be tinkered with. If a tabular dataset has 10 rows and 5 columns, we should not allow the user to set the former to 15 (without making further assumptions, providing extra data, etc.).

When creating collections of items, we usually prefer lists, as they are more flexible a data type. Yet, in Section 3.4.2, we will mention that many functions return tuples. We should be able to handle them with confidence.

---

### 3.1.3 Ranges

Objects defined by calling **`range`**(from, to) or **`range`**(from, to, by) represent arithmetic progressions of integers. For the sake of illustration, let us convert a few of them to ordinary lists:

```
list(range(0, 5))  # i.e., range(0, 5, 1) – from 0 to 5 (exclusive) by 1
## [0, 1, 2, 3, 4]
list(range(10, 0, -1))  # from 10 to 0 (exclusive) by -1
## [10, 9, 8, 7, 6, 5, 4, 3, 2, 1]
```

Let us point out that the rightmost boundary (to) is exclusive and that by defaults to 1.

### 3.1.4 Strings (Again)

Recall that we discussed character strings in Section 2.1.3.

```
print("lovely\nspam")
## lovely
## spam
```

Strings are often treated as scalars (atomic entities, as in: a string as a whole). However, as we will soon find out, their individual characters can also be accessed by index.

Furthermore, in Chapter 14, we will discuss a plethora of operations on text.

## 3.2 Working with Sequences

### 3.2.1 Extracting Elements

The index operator, `` `[...]` ``, can be applied on any sequential object to extract an element at a position specified by a single integer.

```
x = ["one", "two", "three", "four", "five"]
x[0]  # the first element
## 'one'
x[1]  # the second element
## 'two'
x[len(x)-1]  # the last element
## 'five'
```

The valid indexes are 0, 1, ..., n-2, n-1, where n is the length (size) of the sequence, which can be fetched by calling **len**.

---

**Important** Think of an index as the distance from the start of a sequence. For example, x[3] means "3 items away from the beginning", i.e., the 4th element.

---

Negative indexes count from the end:

```
x[-1]  # the last element (ultimate)
## 'five'
x[-2]  # the next to last (the last but one, penultimate)
## 'four'
x[-len(x)]  # the first element
## 'one'
```

The index operator can be applied on any sequential object:

```
"string"[3]
## 'i'
```

Indexing a string returns a string – that is why we classified strings as scalars too.

More examples:

```
range(0, 10)[-1]  # the last item in an arithmetic progression
## 9
(1, )[0]  # extract from a 1-tuple
## 1
```

### 3.2.2 Slicing

We can also use slices of the form from:to or from:to:by to select a subsequence of a given sequence. Slices are similar to ranges, but `:` can only be used within square brackets.

```
x = ["one", "two", "three", "four", "five"]
x[1:4]  # from 2nd to 5th (exclusive)
## ['two', 'three', 'four']
x[-1:0:-2]  # from last to first (exclusive) by every 2nd backwards
## ['five', 'three']
```

In fact, from and to are optional – when omitted, they default to one of the sequence boundaries.

```
x[3:]  # from 3rd to end
## ['four', 'five']
x[:2]  # first two
## ['one', 'two']
x[:0]  # none (first zero)
## []
x[::2]  # every 2nd from the start
## ['one', 'three', 'five']
x[::-1]  # elements in reverse order
## ['five', 'four', 'three', 'two', 'one']
```

And, of course, they can be applied on other sequential objects as well:

```
"spam, bacon, spam, and eggs"[13:17]  # fetch a substring
## 'spam'
```

---

**Important**  Knowing the difference between element extraction and subsetting a sequence (creating a subsequence) is crucial.

---

For example:

```python
x[0]  # extraction (indexing with a single integer)
## 'one'
```

gives the object *at* that index.

```python
x[0:1]  # subsetting (indexing with a slice)
## ['one']
```

gives the object of the same type as x (here, a list) featuring the items at that indexes (in this case, only the first object, but a slice can potentially select any number of elements, including none).

**pandas** data frames and **numpy** arrays will behave similarly, but there will be many more indexing options (as discussed in Section 5.4, Section 8.2, and Section 10.5).

### 3.2.3  Modifying Elements

Lists are *mutable* – their state may be changed. The index operator can be used to replace the elements at given indexes.

```python
x = ["one", "two", "three", "four", "five"]
x[0] = "spam"  # replace the first element
x[-3:] = ["bacon", "eggs"]  # replace last three with given two
print(x)
## ['spam', 'two', 'bacon', 'eggs']
```

**Exercise 3.1** *There are quite a few methods that we can use to modify list elements: not only the aforementioned* **append***, but also* **insert***,* **remove***,* **pop***, etc. Invoke* **help("list")** *to access their descriptions and call them on a few example lists.*

**Exercise 3.2** *Verify that we cannot perform anything similar to the above on tuples, ranges, and strings.*

### 3.2.4  Searching for Elements

The **in** operator and its negation, **not in**, determine whether an element exists in a given sequence:

```python
7 in range(0, 10)
## True
```

```
[2, 3] in [ 1, [2, 3], [4, 5, 6] ]
## True
```

For strings, `in` tests whether a string features a specific *substring*, so we do not have to restrict ourselves to single characters:

```
"spam" in "lovely spams"
## True
```

**Exercise 3.3** *Check out the **count** and **index** methods for the `list` and other classes.*

### 3.2.5   Arithmetic Operators

Some arithmetic operators were *overloaded* for certain sequential types, but they carry different meanings than those for integers and floats.

In particular, `+` can be used to join (concatenate) strings, lists, and tuples:

```
"spam" + " " + "bacon"
## 'spam bacon'
[1, 2, 3] + [4]
## [1, 2, 3, 4]
```

and `*` duplicates (recycles) a given sequence:

```
"spam" * 3
## 'spamspamspam'
(1, 2) * 4
## (1, 2, 1, 2, 1, 2, 1, 2)
```

In each case, a new object has been returned.

## 3.3   Dictionaries

Dictionaries (objects of type `dict`) are sets of `key:value` pairs, where the `values` (any Python object) can be accessed by `key` (usually a string).

```
x = {
    "a": [1, 2, 3],
    "b": 7,
    "z": "spam!"
}
```

```
print(x)
## {'a': [1, 2, 3], 'b': 7, 'z': 'spam!'}
```

We can also create a dictionary with string keys using the `dict` function which accepts any keyword arguments:

```
dict(a=[1, 2, 3], b=7, z="spam!")
## {'a': [1, 2, 3], 'b': 7, 'z': 'spam!'}
```

The index operator can be used to extract specific elements:

```
x["a"]
## [1, 2, 3]
```

In this context, `x[0]` is not valid – it is not an object of sequential type; a key of `0` does not exist in a given dictionary.

The `in` operator checks whether a given key exists:

```
"a" in x, 0 not in x, "z" in x, "w" in x  # a tuple of 4 tests' results
## (True, True, True, False)
```

We can also add new elements to a dictionary:

```
x["f"] = "more spam!"
print(x)
## {'a': [1, 2, 3], 'b': 7, 'z': 'spam!', 'f': 'more spam!'}
```

**Example 3.4** *(\*) In practice, we often import JSON files (which is a popular data exchange format on the internet) exactly in the form of Python dictionaries. Let us demo it quickly:*

```
import requests
x = requests.get("https://api.github.com/users/gagolews/starred").json()
```

*Now `x` is a sequence of dictionaries giving the information on the repositories starred by yours truly on GitHub. As an exercise, the reader is encouraged to inspect its structure.*

## 3.4    Iterable Types

All the objects we discussed here are *iterable*. In other words, we can iterate through each element contained therein.

In particular, the **list** and **tuple** *functions* take any iterable object and convert it to a sequence of the corresponding type, for instance:

```python
list("spam")
## ['s', 'p', 'a', 'm']
tuple(range(0, 10, 2))
## (0, 2, 4, 6, 8)
list({ "a": 1, "b": ["spam", "bacon", "spam"] })
## ['a', 'b']
```

**Exercise 3.5** *Take a look at the documentation of the **extend** method for the `list` class. The manual page suggests that this operation takes any iterable object. Feed it with a list, tuple, range, and a string and see what happens.*

The notion of iterable objects is essential, as they appear in many contexts. There are quite a few other iterable types that are, for example, non-sequential (we cannot access their elements at random using the index operator).

**Exercise 3.6** *(\*) Check out the **enumerate**, **zip**, and **reversed** functions and what kind of iterable objects they return.*

### 3.4.1   The for Loop

The **for** loop iterates over every element in an iterable object, allowing us to perform a specific action. For example:

```python
x = [1, "two", ["three", 3j, 3], False]  # some iterable object
for el in x:   # for every element in `x`, let's call it `el`
    print(el)  # do something on `el`
## 1
## two
## ['three', 3j, 3]
## False
```

Another example:

```python
for i in range(len(x)):
    print(i, x[i], sep=": ")  # sep=" " is the default (element separator)
## 0: 1
## 1: two
## 2: ['three', 3j, 3]
## 3: False
```

One more example – computing the elementwise multiply of two vectors of equal lengths:

```python
x = [1,  2,   3,    4,     5] # for testing
```

```python
y = [1, 10, 100, 1000, 10000]  # just a test
z = []  # result list – start with an empty one
for i in range(len(x)):
    z.append(x[i] * y[i])
print(z)
## [1, 20, 300, 4000, 50000]
```

Yet another example: here is a function that determines the minimum of a given iterable object (compare the built-in **min** function, see **help**("min")).

```python
import math
def mymin(x):
    """
    The smallest element in an iterable object x.
    We assume that x consists of numbers only.
    """
    curmin = math.inf  # infinity is greater than any other number
    for e in x:
        if e < curmin:
            curmin = e  # a better candidate for the minimum
    return curmin
```

**Exercise 3.7** *Write your own basic versions (using the **for** loop) of the built-in **max**, **sum**, **any**, and **all** functions.*

**Exercise 3.8** *(\*) The **glob** function in the **glob** module can be used to list all files in a given directory whose names match a specific wildcard, e.g., **glob.glob**("~/Music/\*.mp3") ("~" points to the current user's home directory, see Section 13.6.1). Moreover, **getsize** from the **os.path** module returns the size of a given file, in bytes. Write a function that determines the total size of all the files in a given directory.*

### 3.4.2   Tuple Assignment

We can create many variables in one line of code by using the syntax tuple_of_ids = iterable_object, which unpacks the iterable:

```python
a, b, c = [1, "two", [3, 3j, "three"]]
print(a)
## 1
print(b)
## two
print(c)
## [3, 3j, 'three']
```

This is useful, for example, when the swapping of two elements is needed:

```python
a, b = 1, 2   # the same as (a, b) = (1, 2)
a, b = b, a   # swap a and b
print(a)
## 2
print(b)
## 1
```

Another use case is where we fetch outputs of functions that return many objects at once. For instance, later we will learn about `numpy.unique` which (depending on arguments passed) may return a tuple of arrays:

```python
import numpy as np
result = np.unique([1, 2, 1, 2, 1, 1, 3, 2, 1], return_counts=True)
print(result)
## (array([1, 2, 3]), array([5, 3, 1]))
```

That this is indeed a tuple of length two (which we should be able to tell already by merely looking at the result: note the round brackets and two objects separated by a comma) can be verified as follows:

```python
type(result), len(result)
## (<class 'tuple'>, 2)
```

Now, instead of:

```python
values = result[0]
counts = result[1]
```

we can write:

```python
values, counts = np.unique([1, 2, 1, 2, 1, 1, 3, 2, 1], return_counts=True)
```

This gives two separate variables, each storing a different array:

```python
print(values)
## [1 2 3]
print(counts)
## [5 3 1]
```

If only the second item was of our interest, we could have written:

```python
counts = np.unique([1, 2, 1, 2, 1, 1, 3, 2, 1], return_counts=True)[1]
print(counts)
## [5 3 1]
```

because a tuple is a sequential object.

**Example 3.9**  (*) *Knowing that the* `dict.items` *method generates an iterable object that can be used to traverse through all the* (key, value) *pairs:*

```python
x = { "a": 1, "b": ["spam", "bacon", "spam"] }
print(list(x.items()))  # just a demo
## [('a', 1), ('b', ['spam', 'bacon', 'spam'])]
```

*we can utilise tuple assignments in contexts such as:*

```python
for k, v in x.items():   # or: for (k, v) in x.items()...
    print(k, v, sep=": ")
## a: 1
## b: ['spam', 'bacon', 'spam']
```

---

**Note**  (**) If there are too many values to unpack, we can use the notation like `*name` inside the `tuple_of_identifiers`. This will serve as a placeholder that gathers all the remaining values and wraps them up in a list:

```python
a, b, *c, d = range(10)
print(a, b, c, d, sep="\n")
## 0
## 1
## [2, 3, 4, 5, 6, 7, 8]
## 9
```

This placeholder may appear only once on the lefthand side of the assignment operator.

---

### 3.4.3  Argument Unpacking (*)

Sometimes we will need to call a function with many parameters or call a series of functions with similar arguments (e.g., when plotting many objects using the same plotting style like colour, shape, font). In such scenarios, it may be convenient to pre-prepare the data to be passed as their inputs beforehand.

Consider the following function that takes four arguments and prints them out:

```python
def test(a, b, c, d):
    "It is just a test – simply prins the arguments passed"
    print("a = ", a, ", b = ", b, ", c = ", c, ", d = ", d, sep="")
```

Arguments to be matched positionally can be wrapped inside any iterable object and then unpacked using the asterisk operator:

```python
args = [1, 2, 3, 4]  # merely an example
```

```
test(*args)  # just like test(1, 2, 3, 4)
## a = 1, b = 2, c = 3, d = 4
```

Keyword arguments can be wrapped inside a dictionary and unpacked with a double asterisk:

```
kwargs = dict(a=1, c=3, d=4, b=2)
test(**kwargs)
## a = 1, b = 2, c = 3, d = 4
```

The unpackings can be intertwined. For this reason, the following calls are equivalent:

```
test(1, *range(2, 4), 4)
## a = 1, b = 2, c = 3, d = 4
test(1, **dict(d=4, c=3, b=2))
## a = 1, b = 2, c = 3, d = 4
test(*range(1, 3), **dict(d=4, c=3))
## a = 1, b = 2, c = 3, d = 4
```

### 3.4.4  Variadic Arguments: *args and **kwargs (*)

We can also construct a function that takes any number of positional or keyword arguments by including *args or **kwargs (those are customary names) in their parameter list:

```
def test(a, b, *args, **kwargs):
    "simply prints the arguments passed"
    print(
        "a = ", a, ", b = ", b,
        ", args = ", args, ", kwargs = ", kwargs, sep=""
    )
```

For example:

```
test(1, 2, 3, 4, 5, spam=6, eggs=7)
## a = 1, b = 2, args = (3, 4, 5), kwargs = {'spam': 6, 'eggs': 7}
```

We see that *args gathers all the positionally matched arguments (except a and b, which were set explicitly) into a tuple. On the other hand, **kwargs is a dictionary that stores all keyword arguments not featured in the function's parameter list.

**Exercise 3.10** *From time to time, we will be coming across *args and **kwargs in various contexts. Study what* **matplotlib.pyplot.plot** *uses them for (by calling* **help(plt.plot)**).

## 3.5 Object References and Copying (*)

### 3.5.1 Copying References

It is important to always keep in mind that when writing:

```python
x = [1, 2, 3]
y = x
```

the assignment operator does not create a copy of x; both x and y refer to the same object in the computer's memory.

---

**Important** If x is mutable, any change made to it will affect y (as, again, they are two different means to access the same object). This will also be true for **numpy** arrays and **pandas** data frames.

---

For example:

```python
x.append(4)
print(y)
## [1, 2, 3, 4]
```

That now a call to **print**(x) gives the same result as above is left as an exercise.

### 3.5.2 Pass by Assignment

Arguments are passed to functions by assignment too. In other words, they behave as if `=` was used – what we get is another reference to the existing object.

```python
def myadd(z, i):
    z.append(i)
```

And now:

```python
myadd(x, 5)
myadd(y, 6)
print(x)
## [1, 2, 3, 4, 5, 6]
```

### 3.5.3 Object Copies

If we find the above behaviour undesirable, we can always make a copy of an object. It is customary for the mutable objects to be equipped with a relevant method:

```
x = [1, 2, 3]
y = x.copy()
x.append(4)
print(y)
## [1, 2, 3]
```

This did not change the object referred to as y, because it is now a different entity.

### 3.5.4   Modify in Place or Return a Modified Copy?

We now know that we *can* have functions or methods that change the state of a given object. Consequently, for all the functions we apply, it is important to read their documentation to determine if they modify their inputs in place or if they return an entirely new object.

Consider the following examples. The `sorted` function returns a sorted version of the input iterable:

```
x = [5, 3, 2, 4, 1]
print(sorted(x)) # returns a sorted copy of x (does not change x)
## [1, 2, 3, 4, 5]
print(x) # unchanged
## [5, 3, 2, 4, 1]
```

The **list.sorted** method modifies the list it is applied on in place:

```
x = [5, 3, 2, 4, 1]
x.sort() # modifies x in place and returns nothing
print(x)
## [1, 2, 3, 4, 5]
```

Additionally, **random.shuffle** is a *function* (not: a method) that changes the state of the argument:

```
x = [5, 3, 2, 4, 1]
import random
random.shuffle(x) # modifies x in place, returns nothing
print(x)
## [5, 4, 2, 1, 3]
```

Later we will learn about the `Series` class in **pandas**, which represents data frame columns. It has the **sort_values** method which by default returns a sorted copy of the object it acts upon:

```
import pandas as pd
x = pd.Series([5, 3, 2, 4, 1])
```

```python
print(list(x.sort_values()))  # inplace=False
## [1, 2, 3, 4, 5]
print(list(x))  # unchanged
## [5, 3, 2, 4, 1]
```

This behaviour might be changed:

```python
x = pd.Series([5, 3, 2, 4, 1])
x.sort_values(inplace=True)  # note the argument now
print(list(x))  # changed
## [1, 2, 3, 4, 5]
```

**Important** We should always study the official[1] documentation of every function we call. Although surely some patterns arise (such as: a method is likely to modify an object in place whereas a similar standalone function will be returning a copy), ultimately, the functions' developers are free to come up with some exceptions to them if they deem it more sensible or convenient.

## 3.6 Further Reading

Our overview of the Python language is by no means exhaustive. Still, it touches upon the most important topics from the perspective of data wrangling.

We will mention a few additional language elements in this course (list comprehensions, file handling, string formatting, regular expressions, etc.). Yet, we have deliberately decided *not* to introduce some language constructs which we can easily do without (e.g., `else` clauses on `for` and `while` loops, the `match` statement) or are perhaps too technical for an introductory course (`yield`, `iter` and `next`, sets, name binding scopes, deep copying of objects, defining own classes, overloading operators, function factories and closures).

Also, we skipped the constructs that do not work well with the third-party packages we will soon be using (e.g., a notation like x < y < z is not valid if the three involved variables are `numpy` vectors of lengths greater than 1).

The said simplifications were brought in so the reader is not overwhelmed. We strongly advocate for minimalism in software development. Python is the basis for one of many possible programming environments for exercising data science. In the long run, it is best to focus on developing the most *transferable* skills, as other software solutions might not enjoy all the Python's syntactic sugar, and vice versa.

---

[1] And not some random tutorial on the internet displaying numerous ads.

The reader is encouraged to skim through at least the following chapters in the official Python 3 tutorial[2]:

- 3. An Informal Introduction to Python[3],
- 4. More Control Flow Tools[4],
- 5. Data Structures[5].

## 3.7 Exercises

**Exercise 3.11** *Name the sequential objects we introduced.*

**Exercise 3.12** *Is every iterable object sequential?*

**Exercise 3.13** *Is `dict` an instance of a sequential type?*

**Exercise 3.14** *What is the meaning of `+` and `*` operations on strings and lists?*

**Exercise 3.15** *Given a list x featuring numeric scalars, how can we create a new list of the same length giving the squares of all the elements in the former?*

**Exercise 3.16** *(\*) How can we make an object copy and when should we do so?*

**Exercise 3.17** *What is the difference between `x[0]`, `x[1]`, `x[:0]`, and `x[:1]`, where x is a sequential object?*

---

# Part II

# Unidimensional Data

# 4

## *Unidimensional Numeric Data and Their Empirical Distribution*

Our data wrangling adventure starts the moment we get access to, or decide to collect, dozens of data points representing some measurements, such as sensor readings for some industrial processes, body measures for patients in a clinic, salaries of employees, sizes of cities, etc.

For instance, consider the heights of adult females (>= 18 years old, in cm) in the longitudinal study called National Health and Nutrition Examination Survey (NHANES[1]) conducted by the US Centres for Disease Control and Prevention.

```
heights = np.loadtxt("https://raw.githubusercontent.com/gagolews/" +
    "teaching-data/master/marek/nhanes_adult_female_height_2020.txt")
```

Let us preview a few observations:

```
heights[:6]  # first six
## array([160.2, 152.7, 161.2, 157.4, 154.6, 144.7])
```

This is an example of *quantitative* (numeric) data. They are in the form of a series of numbers. It makes sense to apply various mathematical operations on them, including subtraction, division, taking logarithms, comparing which one is greater than the other, and so forth.

Most importantly, here, all the observations are *independent* of each other. Each value represents a different person. Our data sample consists of 4,221 points on the real line (a *bag* of points whose actual ordering does not matter). In Figure 4.1, we see that merely looking at the numbers themselves tells us nothing. There are too many of them.

This is why we are interested in studying a multitude of methods that can bring some insight into the reality behind the numbers. For example, inspecting their distribution.

---

[1] https://wwwn.cdc.gov/nchs/nhanes/search/datapage.aspx

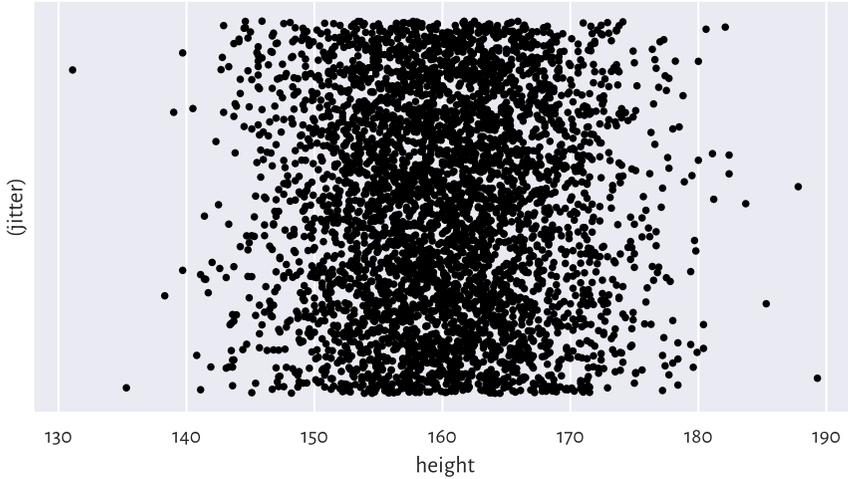

Figure 4.1: The `heights` dataset is comprised of independent points on the real line; we added some jitter on the y-axis for dramatic effects only: the points are too plentiful

## 4.1   Creating Vectors in `numpy`

In this chapter, we introduce basic ways to create `numpy` vectors, which are an efficient data structure for storing and operating on numeric data just like the ones above.

`numpy`[2] [40] is an open-source add-on for numerical computing written by Travis Oliphant and other developers in 2005 (although the project has a much longer history[3] and stands on the shoulders of many giants (e.g., concepts from the APL and Fortran languages). It adds support for multi-dimensional arrays and numerous operations on them, similar to those available in R, S, GNU Octave, Scilab, Julia, Perl (via **Perl Data Language**), and some numerical analysis libraries such as **LAPACK**, **GNU GSL**, etc.

Many other packages are built on top of **numpy**, including: **scipy** [82], **pandas** [56], and **sklearn** [64]. This is why we should study it in great detail. Whatever we learn about vectors will be beautifully transferable to the case of the processing of data frame columns.

It is customary to import the **numpy** package under the `np` alias:

```python
import numpy as np
```

Our code can now refer to the objects defined therein as `np.spam`, `np.bacon`, or `np.spam`.

---

[2] https://numpy.org/doc/stable/reference/index.html
[3] https://scipy.github.io/old-wiki/pages/History_of_SciPy

### 4.1.1   Enumerating Elements

One way to create a vector is by calling the **numpy.array** function:

```
x = np.array([10, 20, 30, 40, 50, 60])
x
## array([10, 20, 30, 40, 50, 60])
```

Here, the vector elements were specified by means of an ordinary list. Ranges and tuples can also be used as content providers, which the kind reader is encouraged to check themself.

A vector of length (size) *n* is often used to represent a point in an *n*-dimensional space (for example, GPS coordinates of a place on Earth assume *n*=2) or *n* readings of some one-dimensional quantity (e.g., recorded heights of *n* people).

The said length can either be read using the previously mentioned **len** function:

```
len(x)
## 6
```

or by reading the array's `shape` slot:

```
x.shape
## (6,)
```

A vector is a *one-dimensional array*. Accordingly, its shape is stored as a tuple of length 1 (the number of dimensions is given by querying `x.ndim`). We can therefore fetch its length by accessing `x.shape[0]`.

On a side note, matrices (two-dimensional arrays), which we shall study in Chapter 7, will be of shape like (`number_of_rows, number_of_columns`).

Recall that Python lists, e.g., [1, 2, 3], represent simple sequences of objects of any kind. Their use cases are very broad, which is both an advantage and something quite the opposite. *Vectors* in **numpy** are like lists, but on steroids. They are powerful in scientific computing because of the underlying assumption that each object they store is of the same type[4]. Although it is possible to save references to arbitrary objects therein, in most scenarios we will be dealing with vectors of logical values, integers, and floating-point numbers. Thanks to this, a wide range of methods could have been defined to enable the performing of the most popular mathematical operations.

And so, above we created a sequence of integers:

---

[4] (*) Vectors are directly representable as simple arrays in the C programming language, in which **numpy** procedures are written. Operations on vectors will be very fast provided that we are using the functions that process them *as a whole*. The readers with some background in other lower-level languages will need to get out of the habit of acting on *individual* elements using a **for**-like loop.

```
x.dtype   # data type
## dtype('int64')
```

But other element types are possible too. For instance, we can convert the above to a float vector:

```
x.astype(float)   # or np.array(x, dtype=float)
## array([10., 20., 30., 40., 50., 60.])
```

Let us emphasise that the above is now printed differently (compare the output of **print**(x) above).

Furthermore:

```
np.array([True, False, False, True])
## array([ True, False, False,  True])
```

gave a logical vector. The constructor detected that the common type of all the elements is bool. Also:

```
np.array(["spam", "spam", "bacon", "spam"])
## array(['spam', 'spam', 'bacon', 'spam'], dtype='<U5')
```

This yielded an array of strings in Unicode (i.e., capable of storing any character in any alphabet, emojis, mathematical symbols, etc.), each of no more than five code points in length. We will point out in Chapter 14 that replacing any element with new content will result in the too-long strings' truncation. We will see that this can be remedied by calling x.astype("<U10").

### 4.1.2  Arithmetic Progressions

**numpy**'s **arange** is similar to the built-in **range**, but outputs a vector:

```
np.arange(0, 10, 2)   # from 0 to 10 (exclusive) by 2
## array([0, 2, 4, 6, 8])
```

**numpy.linspace** (*linear space*) creates a sequence of equidistant points in a given interval:

```
np.linspace(0, 1, 5)   # from 0 to 1 (inclusive), 5 equispaced values
## array([0.  , 0.25, 0.5 , 0.75, 1.  ])
```

**Exercise 4.1** *Call **help**(np.linspace) or **help**("numpy.linspace") to study the meaning of the* endpoint *argument. Find the same documentation page on the **numpy** project's web-site[5]. Another way is to use your favourite search engine such as DuckDuckGo and query "lin-*

---

[5] https://numpy.org/doc/stable/reference/index.html

*space site:numpy.org*"[6]. *Always remember to gather information from first-hand sources. You should become a frequent visitor of this page (and similar ones). In particular, every so often it is a good idea to check out for significant updates at* https://numpy.org/news/.

### 4.1.3 Repeating Values

**numpy.repeat** repeats *each* value a given number of times:

```
np.repeat(5, 6)
## array([5, 5, 5, 5, 5, 5])
np.repeat([1, 2], 3)
## array([1, 1, 1, 2, 2, 2])
np.repeat([1, 2], [3, 5])
## array([1, 1, 1, 2, 2, 2, 2, 2])
```

In each case, every element from the list passed as the 1st argument was repeated the *corresponding* number of times, as defined by the 2nd argument. The kind reader should not expect us to elaborate upon the obtained results any further, because everything is evident: they need to look at the example calls, carefully study all the displayed outputs, and make the conclusions by themself. If something is unclear, they should consult the official documentation and apply Rule #4.

Moving on. **numpy.tile**, on the other hand, repeats a whole sequence with *recycling*:

```
np.tile([1, 2], 3)
## array([1, 2, 1, 2, 1, 2])
```

Notice the difference between the above and the result of **numpy.repeat**([1, 2], 3).

See also[7] **numpy.zeros** and **numpy.ones** for some specialised versions of the above.

### 4.1.4 **numpy.r_** (*)

**numpy.r_** gives perhaps the most flexible means for creating vectors involving quite a few of the aforementioned scenarios. Yet, it has a quirky syntax.

For example:

```
np.r_[1, 2, 3, np.nan, 5, np.inf]
## array([ 1.,  2.,  3., nan,  5., inf])
```

Here, nan stands for a *not-a-number* and is used as a placeholder for missing values (discussed in Section 15.1) or *wrong* results, such as the square root of -1 in the domain of reals. The `inf` object, on the other hand, means *infinity*, $\infty$. We can think of it as a value that is too large to be represented in the set of floating-point numbers.

---

[6] DuckDuckGo also supports *search bangs* like "!numpy linspace" which redirect to the official documentation automatically.

[7] When we write *See also*, it means that this is an exercise for the reader (Rule #3), in this case: to look something up in the official documentation.

We see that `numpy.r_` uses square brackets instead of the round ones. This is smart, because we mentioned in Section 3.2.2 that slices (`` `:` ``) cannot be used outside them. And so:

```
np.r_[0:10:2]  # like np.arange(0, 10, 2)
## array([0, 2, 4, 6, 8])
```

What is more, it accepts the following syntactic sugar:

```
np.r_[0:1:5j]  # like np.linspace(0, 1, 5)
## array([0.  , 0.25, 0.5 , 0.75, 1.  ])
```

Here, `5j` does not have a literal meaning (a complex number). By an arbitrary convention, and only in this context, it denotes the output length of the sequence to be generated. Could the **numpy** authors do that? Well, they could, and they did. End of story.

Finally, we can combine many chunks into one:

```
np.r_[1, 2, [3]*2, 0:3, 0:3:3j]
## array([1. , 2. , 3. , 3. , 0. , 1. , 2. , 0. , 1.5, 3. ])
```

### 4.1.5   Generating Pseudorandom Variates

The automatically attached **numpy.random** module defines many functions to generate pseudorandom numbers. We will be discussing the reason for our using the *pseudo* prefix in Section 6.4, so now let us only quickly take note of a way to sample from the uniform distribution on the unit interval:

```
np.random.rand(5)  # 5 pseudorandom observations in [0, 1]
## array([0.49340194, 0.41614605, 0.69780667, 0.45278338, 0.84061215])
```

and to pick a few values from a given set with replacement (so that any number can be generated multiple times):

```
np.random.choice(np.arange(1, 10), 20)  # replace=True
## array([7, 7, 4, 6, 6, 2, 1, 7, 2, 1, 8, 9, 5, 5, 9, 8, 1, 2, 6, 6])
```

### 4.1.6   Loading Data from Files

We will usually be reading whole heterogeneous tabular data sets using **pandas. read_csv**, being the topic we shall cover in Chapter 10.

It is worth knowing, though, that arrays with elements of the same type can be read efficiently from text files (e.g., CSV) using `numpy.loadtxt`. See the code chunk at the beginning of this chapter for an example.

**Exercise 4.2** *Use* `numpy.loadtxt` *to read the* `population_largest_cities_unnamed`[8] *dataset from GitHub (click* Raw *to get access to its contents and use the URL you were redirected to, not the original one).*

## 4.2    Mathematical Notation

Mathematically, we will be denoting number sequences with:

$$\boldsymbol{x} = (x_1, x_2, \dots, x_n),$$

where $x_i$ is the $i$-th element therein and $n$ is the length (size) of the tuple. Using the programming syntax, $n$ corresponds to `len(x)` or, equivalently, `x.shape[0]`. Furthermore, $x_i$ is `x[i-1]` (because the first element is at index 0).

The bold font (hopefully visible) is to emphasise that $\boldsymbol{x}$ is not an atomic entity ($x$), but rather a collection thereof. For brevity, instead of saying "let $\boldsymbol{x}$ be a real-valued sequence[9] of length $n$", we shall write "let $\boldsymbol{x} \in \mathbb{R}^n$". Here:

- the "$\in$" symbol stands for "*is in*" or "*is a member of*",

- $\mathbb{R}$ denotes the set of real numbers (the very one that features, 0, $-358745.2394$, 42 and $\pi$, amongst uncountably many others), and

- $\mathbb{R}^n$ is the set of real-valued sequences of length $n$ (i.e., $n$ such numbers considered at a time); e.g., $\mathbb{R}^2$ includes pairs such as $(1, 2)$, $(\pi/3, \sqrt{2}/2)$, and $(1/3, 10^3)$.

**Note**  Mathematical notation is pleasantly *abstract* (general) in the sense that $\boldsymbol{x}$ can be anything, e.g., data on the incomes of households, sizes of the largest cities in some country, or heights of participants in some longitudinal study. At first glance, such a representation of objects from the so-called *real world* might seem overly simplistic, especially if we wish to store information on very complex entities. Nonetheless, in most cases, expressing them as *vectors* (i.e., establishing a set of numeric attributes that best describe them in a task at hand) is not only natural but also perfectly sufficient for achieving whatever we aim at.

**Exercise 4.3** *Consider the following problems:*

- *How would you represent a patient in a clinic (for the purpose of conducting research in cardiology)?*

---

[8] https://github.com/gagolews/teaching-data/blob/-/marek/population_largest_cities_unnamed.txt
[9] If $\boldsymbol{x} \in \mathbb{R}^n$, then we often say that $\boldsymbol{x}$ is *a sequence of n numbers, a (numeric) n-tuple, a n-dimensional real vector, a point in a n-dimensional real space,* or *an element of a real n-space,* etc. In many contexts, they are synonymic.

- *How would you represent a car in an insurance company's database (to determine how much a driver should pay annually for the mandatory policy)?*

- *How would you represent a student in a university (to grant them scholarships)?*

*In each case, list a few numeric features that best describe the reality of concern. On a side note, descriptive (categorical) labels can always be encoded as numbers, e.g., female=1, male=2, but this will be the topic of Chapter 11.*

By $x_{(i)}$ (notice the bracket[10]) we will denote the $i$-th smallest value in $x$ (also called the $i$-th *order statistic*). In particular, $x_{(1)}$ is the sample minimum and $x_{(n)}$ is the maximum. The same in Python:

```python
x = np.array([5, 4, 2, 1, 3])  # just an example
x_sorted = np.sort(x)
x_sorted[0], x_sorted[-1]  # the minimum and the maximum
## (1, 5)
```

To avoid the clutter of notation, in certain formulae (e.g., in the definition of the type-7 quantiles in Section 5.1.1), we will be assuming that $x_{(0)}$ is the same as $x_{(1)}$ and $x_{(n+1)}$ is equivalent to $x_{(n)}$.

## 4.3    Inspecting the Data Distribution with Histograms

*Histograms* are one of the most intuitive tools for depicting the empirical distribution of a data sample. We will be drawing them using the statistical data visualisation package called **seaborn**[11] [83] (written by Michael Waskom), which was built on top of the classic plotting library **matplotlib**[12] [45] (originally developed by John D. Hunter). Let us import both packages and set their traditional aliases:

```python
import matplotlib.pyplot as plt
import seaborn as sns
```

**Note** It is customary to call a single function from **seaborn** and then perform a series of additional calls to **matplotlib** to tweak the display details. It is important to remember that the former uses the latter to achieve its goals, not the other way around. In many exercises, **seaborn** might not even have the required functionality at all, and we will be using **matplotlib** only, and nothing else.

---

[10] Some textbooks denote the $i$-th order statistic with $x_{i:n}$, but we will not.
[11] https://seaborn.pydata.org
[12] https://matplotlib.org/

### 4.3.1 `heights`: A Bell-Shaped Distribution

Let us draw a histogram of the `heights` dataset, see Figure 4.2.

```
sns.histplot(heights, bins=11, color="lightgray")
plt.show()
```

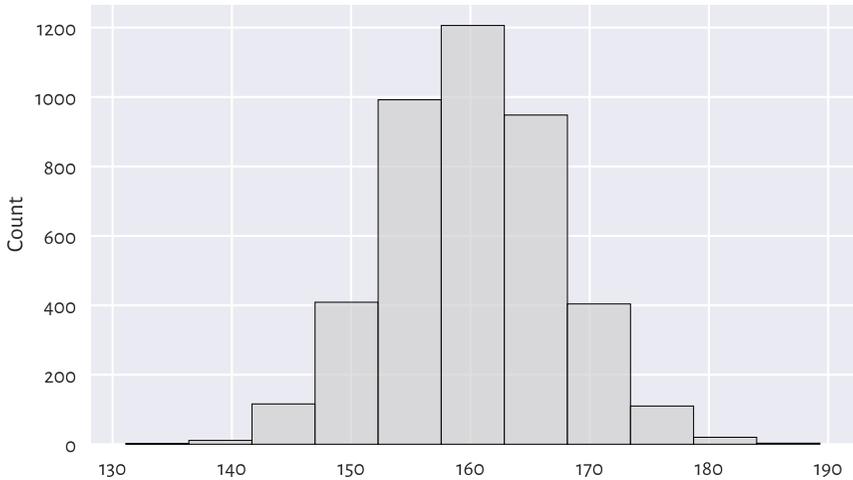

Figure 4.2: Histogram of the `heights` dataset; the empirical distribution is nicely bell-shaped

The data were split into 11 bins and plotted in such a way that the bar heights are proportional to the number of observations falling into each interval. The bins are non-overlapping, adjacent to each other, and of equal lengths. We can read their coordinates by looking at the bottom side of each rectangular bar. For example, ca. 1200 observations fall into the interval [158, 163] (more or less) and ca. 400 into [168, 173] (approximately).

This distribution is bell-shaped[13] – nicely symmetrical around about 160 cm. The most typical (*normal*) observations are somewhere in the middle, and the probability mass decreases quickly on both sides. As a matter of fact, in Chapter 6, we will model this dataset using a *normal* distribution and obtain an excellent fit. In particular, we will mention that observations outside the interval [139, 181] are very rare (probability less than 1%; via the *3σ rule*, i.e., expected value ± 3 standard deviations).

### 4.3.2 `income`: A Right-Skewed Distribution

For some of us, a normal distribution is a prototypical one – we might expect (wishfully think) that many phenomena yield similar regularities. And that is indeed the case[14], e.g., in psychology (IQ or personality tests), physiology (the above heights), or when measuring stuff with not-so-precise devices (distribution of errors). We might be tempted to think now that *everything* is normally distributed, but this is very much untrue.

---

[13] A rather traditional name, but we will get used to it.

Let us consider another dataset. In Figure 4.3, we depict the distribution of a simulated[15] sample of disposable income of 1,000 randomly chosen UK households, in British Pounds, for the financial year ending 2020.

```
income = np.loadtxt("https://raw.githubusercontent.com/gagolews/" +
    "teaching-data/master/marek/uk_income_simulated_2020.txt")
sns.histplot(income, stat="percent", bins=20, color="lightgray")
plt.show()
```

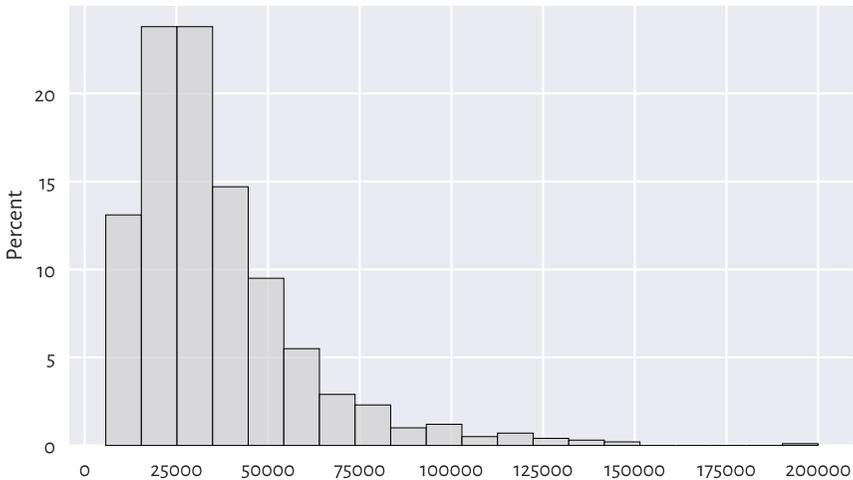

Figure 4.3: Histogram of the `income` dataset; the distribution is right-skewed

We normalised (`stat="percent"`) the bar heights so that they all sum to 1 (or, equivalently, 100%), which resulted in a *probability histogram*.

We notice that the probability density quickly increases, reaches its peak at around £15,500–£35,000, and then slowly goes down. We say that it has a *long tail* on the right or that it is *right- or positive-skewed*. Accordingly, there are several people earning good money. It is quite a non-normal distribution. Most people are rather unwealthy: their income is way below the per-capita revenue (being the average income for the whole population).

---

[14] In fact, we have a proposition stating that the sum or average of many observations or otherwise simpler components of some more complex entity, assuming that they are independent and follow the same (any!) distribution with finite variance, is approximately normally distributed. This is called the Central Limit Theorem and it is a very strong mathematical result.

[15] For privacy and other reasons, the UK Office for National Statistics does not publish details on individual taxpayers. This is why we needed to guesstimate them based on data from a report published at https://www.ons.gov.uk/peoplepopulationandcommunity.

---

**Note** Looking at Figure 4.3, we might have taken note of the relatively higher bars, as compared to their neighbours, at ca. £100,000 and £120,000. We might be tempted to try to invent a *story* about why there can be some difference in the relative probability mass, but we should refrain from it. As our data sample is quite small, they might merely be due to some natural variability (Section 6.4.4). Of course, there might be some reasons behind it (theoretically), but we cannot read this only by looking at a single histogram. In other words, it is a tool that we use to identify some rather general features of the data distribution (like the overall shape), not the specifics.

---

**Exercise 4.4** *There is also the `nhanes_adult_female_weight_2020`[16] dataset in our data repository, giving corresponding weights (in kilograms) of the NHANES study participants. Draw a histogram. Does its shape resemble the `income` or `heights` distribution more?*

### 4.3.3 How Many Bins?

Unless some stronger assumptions about the data distribution are made, choosing the right number of bins is more art than science:

- too many will result in a rugged histogram,

- too few might result in our missing important details.

Figure 4.4 illustrates this.

```
plt.subplot(1, 2, 1)    # 1 row, 2 columns, 1st plot
sns.histplot(income, bins=5, color="lightgray")
plt.subplot(1, 2, 2)    # 1 row, 2 columns, 2nd plot
sns.histplot(income, bins=200, color="lightgray")
plt.ylabel(None)
plt.show()
```

For example, in the histogram with five bins, we miss the information that the ca. £20,000 income is more popular than the ca. £10,000 one. (as given by the first two bars in Figure 4.3).

On the other hand, the histogram with 200 bins already seems too fine-grained.

---

**Important** Usually, the "truth" is probably somewhere in-between. When preparing histograms for publication (e.g., in a report or on a webpage), we might be tempted to think "one must choose one and only one bin count". In fact, we do not have to. Even though some people will insist on it, remember that it is we who are responsible for the data being presented in the most unambiguous fashion possible. Providing two or three histograms can sometimes be a much better idea.

---

[16] https://github.com/gagolews/teaching-data/raw/master/marek/nhanes_adult_female_weight_2020.txt

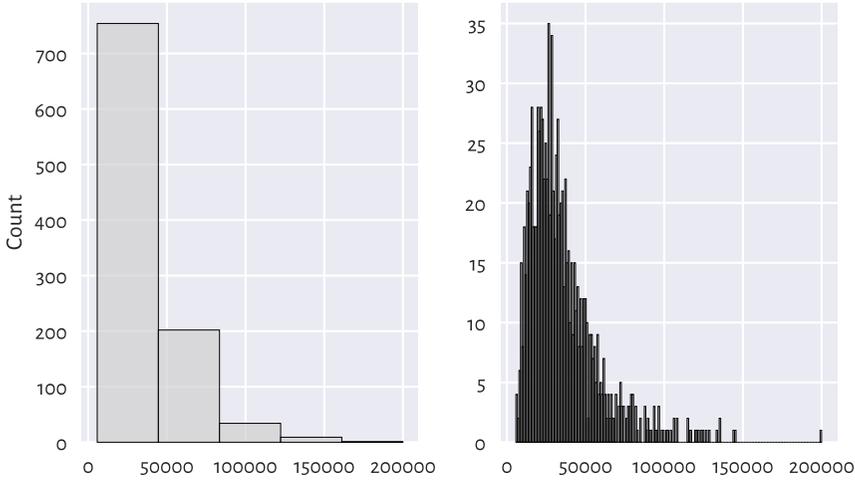

Figure 4.4: Too few and too many histogram bins (the `income` dataset)

Further, let us be aware that someone might want to trick us by choosing the number of bins that depict the reality in good light, when the truth is quite the opposite. For instance, the histogram on the left above hides the poorest households inside the first bar – the first income bracket is very wide. If we cannot request access to the original data, the best thing we can do is to simply ignore such a data visualisation instance and warn others not to trust it. A true data scientist must be sceptical.

The documentation of **seaborn.histplot** (and the underlying **matplotlib.pyplot. hist**), states that the `bins` argument is passed to **numpy.histogram_bin_edges** to determine the intervals into which our data are to be split. **numpy.histogram** uses the same function and additionally returns the corresponding counts (how many observations fall into each bin) instead of plotting them.

```
counts, bins = np.histogram(income, 20)
counts
## array([131, 238, 238, 147,  95,  55,  29,  23,  10,  12,   5,   7,   4,
##          3,   2,   0,   0,   0,   0,   1])
bins
## array([  5750.  ,  15460.95,  25171.9 ,  34882.85,  44593.8 ,  54304.75,
##          64015.7 ,  73726.65,  83437.6 ,  93148.55, 102859.5 , 112570.45,
##         122281.4 , 131992.35, 141703.3 , 151414.25, 161125.2 , 170836.15,
##         180547.1 , 190258.05, 199969.  ])
```

Thus, there are 238 observations both in the [15,461; 25,172) and [25,172; 34,883) intervals.

---

**Note** A table of ranges and the corresponding counts can be effective for data reporting. It is more informative and takes less space than a series of raw numbers, especially if we present them like in the table below.

Table 4.1: Incomes of selected British households; the bin edges are pleasantly round numbers

| income bracket [£1000s] | count |
|---|---|
| 0−200 | 236 |
| 200−400 | 459 |
| 400−600 | 191 |
| 600−800 | 64 |
| 800−1000 | 26 |
| 1000−1200 | 11 |
| 1200−1400 | 10 |
| 1400−1600 | 2 |
| 1600−1800 | 0 |
| 1800−2000 | 1 |

Reporting data in tabular form can also increase the privacy of the subjects (making subjects less identifiable, which is good) or hide some uncomfortable facts (which is not so good; "there are ten people in our company earning *more* than £200,000 p.a." – this can be as much as £10,000,000, but shush).

---

**Exercise 4.5** *Find how we can provide the* `seaborn.histplot` *and* `numpy.histogram` *functions with custom bin breaks. Plot a histogram where the bin edges are 0, 20,000, 40,000, etc. (just like in the above table). Also let us highlight the fact that bins do not have to be of equal sizes: set the last bin to [140,000; 200,000].*

**Example 4.6** *Let us also inspect the bin edges and counts that we see in* Figure 4.2:

```
counts, bins = np.histogram(heights, 11)
counts
## array([  2,  11, 116, 409, 992, 1206, 948, 404, 110,  20,   3])
bins
## array([131.1       , 136.39090909, 141.68181818, 146.97272727,
##        152.26363636, 157.55454545, 162.84545455, 168.13636364,
##        173.42727273, 178.71818182, 184.00909091, 189.3       ])
```

**Exercise 4.7** (*) *There are quite a few heuristics to determine the number of bins automagically, see* `numpy.histogram_bin_edges` *for a few formulae. Check out how different values of the* `bins` *argument (e.g.,* `"sturges"`*,* `"fd"`*) affect the histogram shapes on both* `income` *and* `heights` *datasets. Each has its limitations, none is perfect, but some might be a good starting point for further fine-tuning.*

We will get back to the topic of manual data binning in Section 11.1.4.

### 4.3.4  `peds`: A Bimodal Distribution (Already Binned)

Here are the December 2021 hourly average pedestrian counts[17] near the Southern Cross Station in Melbourne:

```
peds = np.loadtxt("https://raw.githubusercontent.com/gagolews/" +
    "teaching-data/master/marek/southern_cross_station_peds_2019_dec.txt")
peds
## array([  31.22580645,    18.38709677,    11.77419355,     8.48387097,
##            8.58064516,    58.70967742,   332.93548387,  1121.96774194,
##         2061.87096774,  1253.41935484,   531.64516129,   502.35483871,
##          899.06451613,   775.        ,   614.87096774,   825.06451613,
##         1542.74193548,  1870.48387097,   884.38709677,   345.83870968,
##          203.48387097,   150.4516129 ,   135.67741935,    94.03225806])
```

This time, data have already been binned by somebody else. Consequently, we cannot use **seaborn.histplot** to depict them. Instead, we can rely on a more low-level function, **matplotlib.pyplot.bar**; see Figure 4.5.

```
plt.bar(np.arange(0, 24), width=1, height=peds,
    color="lightgray", edgecolor="black", alpha=0.8)
plt.show()
```

This is an example of a bimodal (or even trimodal) distribution: there is a morning peak and an evening peak (and some analysts probably would distinguish a lunchtime one too).

### 4.3.5  `matura`: A Bell-Shaped Distribution (Almost)

Figure 4.6 depicts a histogram of another interesting dataset which comes in an already pre-summarised form.

```
matura = np.loadtxt("https://raw.githubusercontent.com/gagolews/" +
    "teaching-data/master/marek/matura_2019_polish.txt")
plt.bar(np.arange(0, 71), width=1, height=matura,
    color="lightgray", edgecolor="black", alpha=0.8)
plt.show()
```

---

[17] http://www.pedestrian.melbourne.vic.gov.au/

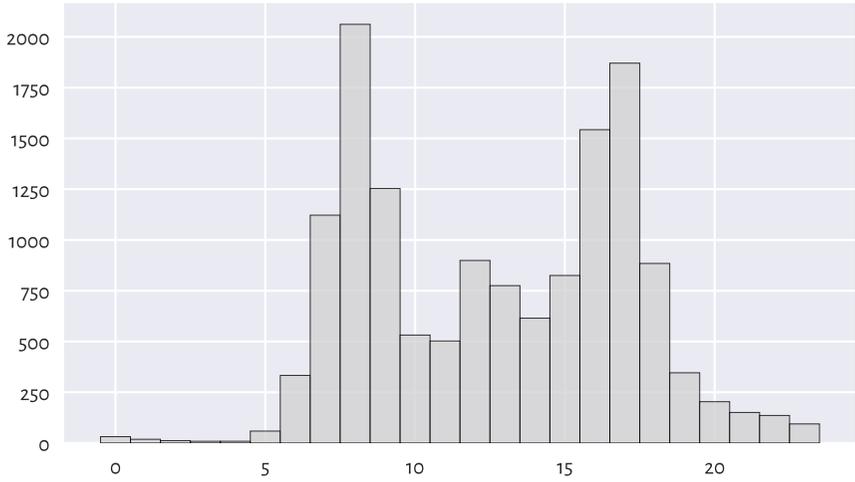

Figure 4.5: Histogram of the `peds` dataset; a bimodal (trimodal?) distribution

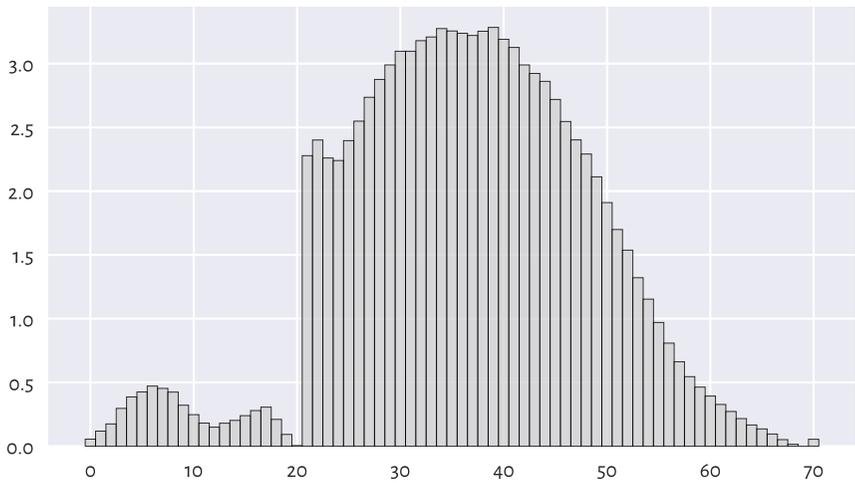

Figure 4.6: Histogram of the `matura` dataset; a bell-shaped distribution... almost

This gives the distribution[18] of the 2019 Matura (end of high school) exam scores in Poland (in %) – Polish literature[19] at the basic level.

It seems that the distribution should be bell-shaped, but someone tinkered with it. Still, knowing that:

- the examiners are good people – we teachers love our students,

- 20 points were required to pass,

- 50 points were given for an essay – and beauty is in the eye of the beholder,

it all starts to make sense. Without graphically depicting this dataset, we would not know that such (albeit lucky for some students) *anomalies* occurred.

### 4.3.6   `marathon` (Truncated – Fastest Runners): A Left-Skewed Distribution

Next, let us consider the 37th PZU Warsaw Marathon (2015) results.

```
marathon = np.loadtxt("https://raw.githubusercontent.com/gagolews/" +
    "teaching-data/master/marek/37_pzu_warsaw_marathon_mins.txt")
```

Here are the top five gun times (in minutes):

```
marathon[:5]  # preview first 5 (data are already sorted increasingly)
## array([129.32, 130.75, 130.97, 134.17, 134.68])
```

Plotting the histogram of the data on the participants who finished the 42.2 km run in less than three hours, i.e., a *truncated* version of this dataset, reveals that the data are highly *left*-skewed, see Figure 4.7.

```
sns.histplot(marathon[marathon < 180], color="lightgray")
plt.show()
```

This was of course expected – there are only a few elite runners in the game. Yours truly wishes his personal best will be less than 180 minutes someday. We shall see. Running is fun, and so is walking; why not take a break for an hour and go outside?

**Exercise 4.8**  *Plot the histogram of the untruncated (complete) version of this dataset.*

### 4.3.7   Log-Scale and Heavy-Tailed Distributions

Consider the dataset on the populations of cities in the 2000 US Census:

```
cities = np.loadtxt("https://raw.githubusercontent.com/gagolews/" +
    "teaching-data/master/other/us_cities_2000.txt")
```

---

[18] https://cke.gov.pl/images/_EGZAMIN_MATURALNY_OD_2015/Informacje_o_wynikach/2019/sprawozdanie/Sprawozdanie%202019%20-%20J%C4%99zyk%20polski.pdf

[19] Gombrowicz, Nałkowska, Miłosz, Tuwim, etc.; I recommend.

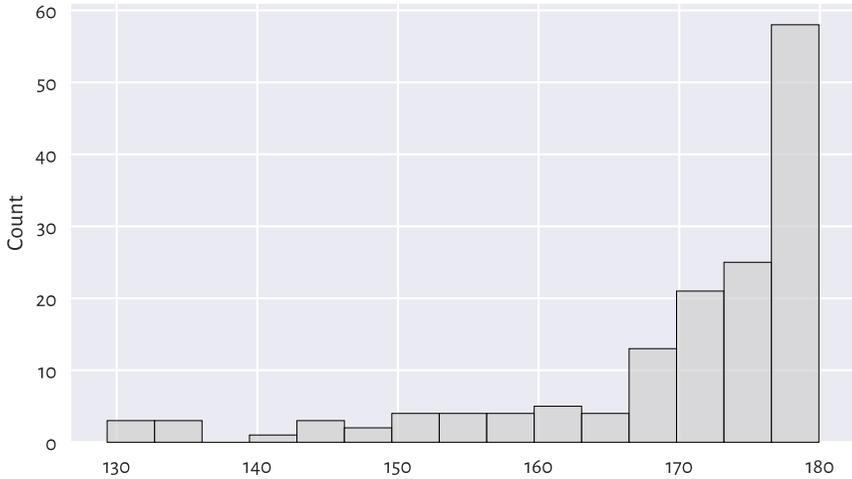

Figure 4.7: Histogram of a truncated version of the `marathon` dataset; the distribution is left-skewed

Let us restrict ourselves only to the cities whose population is not less than 10,000 (another instance of truncating, this time on the other side of the distribution). It turns out that, even though they constitute ca. 14% of all the US settlements, as much as about 84% of all the citizens live there.

```
large_cities = cities[cities >= 10000]
```

Here are the populations of the five largest cities (can we guess which ones are they?):

```
large_cities[-5:]  # preview last 5 – data are sorted increasingly
## array([1517550., 1953633., 2896047., 3694742., 8008654.])
```

The histogram is depicted in Figure 4.8. It is virtually unreadable because the distribution is not just right-skewed; it is extremely *heavy-tailed*: most cities are small, and those that are large – such as New York – are *really* unique. Had we plotted the whole dataset (`cities` instead of `large_cities`), the results' intelligibility would be even worse.

```
sns.histplot(large_cities, bins=20, color="lightgray")
plt.show()
```

This is why we should rather draw such a distribution on the *logarithmic* scale, see Figure 4.9.

```
sns.histplot(large_cities, bins=20, log_scale=True, color="lightgray")
plt.show()
```

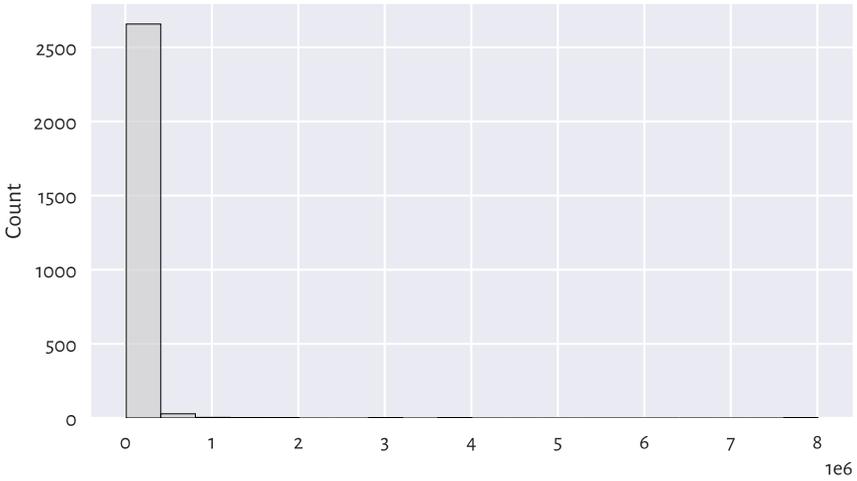

Figure 4.8: Histogram of the `large_cities` dataset; it is extremely heavy-tailed

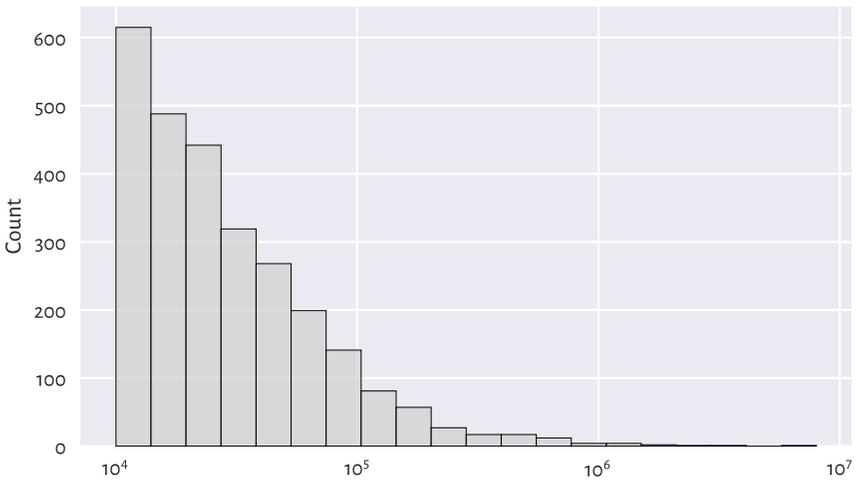

Figure 4.9: Another histogram of the `large_cities` dataset; it is right-skewed even on a logarithmic scale

The log-scale on the x-axis does not increase linearly: it is not based on steps of equal sizes like 0, 1,000,000, 2,000,000, ..., and so forth. Instead, now the increases are geometrical: 10,000, 100,000, 1,000,000, etc.

This is a right-skewed distribution even on the logarithmic scale. Many real-world datasets have similar behaviour, for instance, the frequencies of occurrences of words in books. On a side note, in Chapter 6, we will discuss the Pareto distribution family which yields similar histograms.

**Exercise 4.9**  *Draw the histogram of* `income` *on the logarithmic scale. Does it resemble a bell-shaped distribution?*

**Exercise 4.10**  *(\*) Use* `numpy.geomspace` *and* `numpy.histogram` *to apply logarithmic binning of the* `large_cities` *dataset manually, i.e., to create bins of equal lengths on the log-scale.*

### 4.3.8    Cumulative Counts and the Empirical Cumulative Distribution Function

Let us go back to the `heights` dataset. The histogram in Figure 4.2 told us that, amongst others, 28.6% (1,206 of 4,221) of women are approximately 160.2±2.65 cm tall.

Still, sometimes we might be more interested in *cumulative* counts; see Figure 4.10. They have a different interpretation: we can read that, e.g., 80% of all women are *no more than* ca. 166 cm tall (or that only 20% are taller than this height).

```
sns.histplot(heights, stat="percent", cumulative=True, color="lightgray")
plt.show()
```

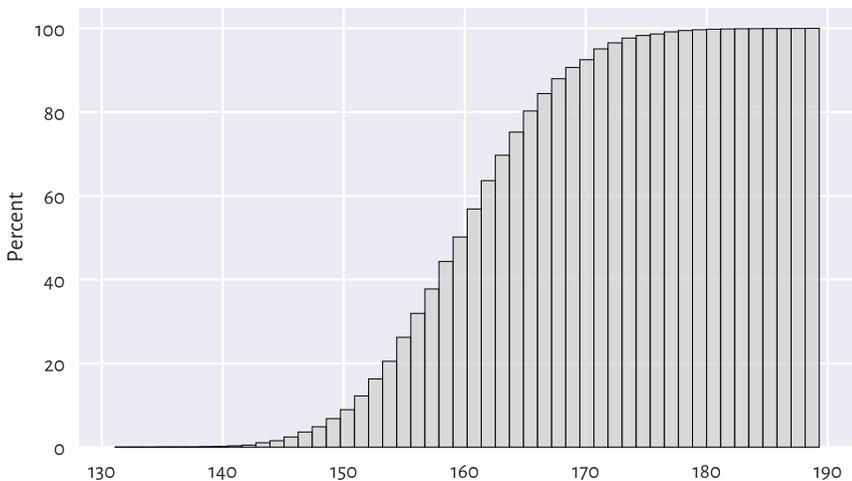

Figure 4.10: Cumulative histogram of the `heights` dataset

Very similar is the plot of the *empirical cumulative distribution function* (ECDF), which for a sample $\boldsymbol{x} = (x_1, \ldots, x_n)$ we denote as $\hat{F}_n$. And so, at any given point $t \in \mathbb{R}$, $\hat{F}_n(t)$ is a step function[20] that gives the proportion of observations in our sample that are not greater than $t$:

$$\hat{F}_n(t) = \frac{|i : x_i \leq t|}{n}.$$

We read $|i : x_i \leq t|$ as the number of indexes like $i$ such that the corresponding $x_i$ is less than or equal to $t$. It can be shown that, given the ordered inputs $x_{(1)} \leq x_{(2)} \leq \ldots \leq x_{(n)}$, it holds:

$$\hat{F}_n(t) = \begin{cases} 0 & \text{for } t < x_{(1)}, \\ k/n & \text{for } x_{(k)} \leq t < x_{(k+1)}, \\ 1 & \text{for } t \geq x_{(n)}. \end{cases}$$

Let us underline the fact that drawing the ECDF does not involve binning – we only need to arrange the observations in an ascending order. Then, assuming that all observations are unique (there are no ties), the arithmetic progression $1/n, 2/n, \ldots, n/n$ is plotted against them; see Figure 4.11[21].

```python
n = len(heights)
heights_sorted = np.sort(heights)
plt.plot(heights_sorted, np.arange(1, n+1)/n, drawstyle="steps-post")
plt.xlabel("$t$")  # LaTeX maths
plt.ylabel("$\\hat{F}_n(t)$, i.e., Prob(height $\\leq$ t)")
plt.show()
```

**Exercise 4.11** *Check out* ***seaborn.ecdfplot*** *for a built-in method implementing the drawing of an ECDF.*

---

**Note**  (*) Quantiles (which we introduce in Section 5.1.1) can be considered a generalised inverse of the ECDF.

---

## 4.4    Exercises

**Exercise 4.12** *What is the difference between* ***numpy.arange*** *and* ***numpy.linspace***?

**Exercise 4.13** *(*) What happens when we convert a logical vector to a numeric one? And what about when we convert a numeric vector to a logical one? We will discuss that later, but you might want to check it yourself now.*

---

[20] We cannot see the steps in Figure 4.11, because the points are too plentiful.
[21] (*) We are using (La)TeX maths typesetting within "$...$" to obtain nice plot labels, see [61] for a good introduction.

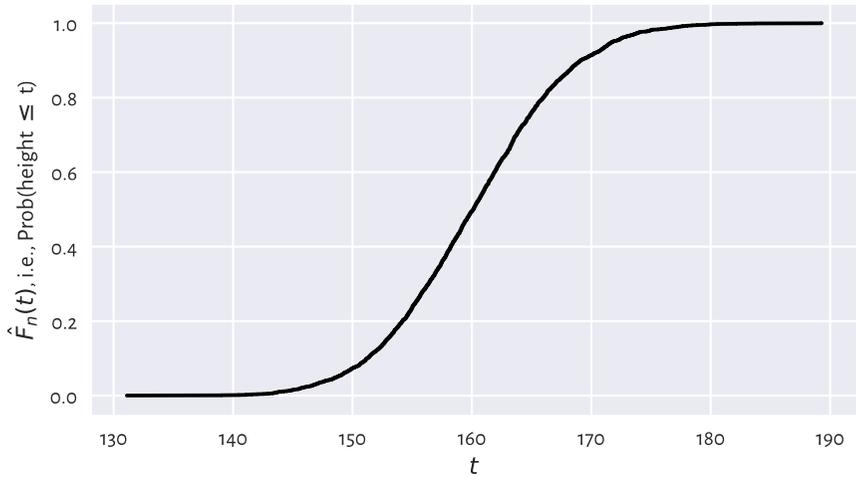

Figure 4.11: Empirical cumulative distribution function for the `heights` dataset

**Exercise 4.14** *Check what happens when we try to create a vector featuring a mix of logical, integer, and floating-point values.*

**Exercise 4.15** *What is a bell-shaped distribution?*

**Exercise 4.16** *What is a right-skewed distribution?*

**Exercise 4.17** *What is a heavy-tailed distribution?*

**Exercise 4.18** *What is a multi-modal distribution?*

**Exercise 4.19** *(*) When does logarithmic binning make sense?*

# 5

## *Processing Unidimensional Data*

It is extremely rare for our datasets to bring interesting and valid insights out of the box. The ones we are using for illustrational purposes in the first part of our book have already been curated. After all, this is an introductory course, and we need to build up the necessary skills and not overwhelm the kind reader with too much information at the same time. We learn simple things first, learn them well, and then we move to more complex matters with a healthy level of confidence.

In real life, various *data cleansing* and *feature engineering* techniques will need to be performed on data. Most of them are based on the simple operations on vectors that we cover in this chapter:

- summarising data (for example, computing the median or sum),

- transforming values (applying mathematical operations on each element, such as subtracting a scalar or taking the natural logarithm),

- filtering (selecting or removing observations that meet specific criteria, e.g., those that are larger than the arithmetic mean ± 5 standard deviations).

---

**Important** The same operations we are going to be applying on individual data frame columns in Chapter 10.

---

## 5.1 Aggregating Numeric Data

Recall that in the previous chapter we discussed the `heights` and `income` datasets whose histograms we gave in Figure 4.2 and Figure 4.3, respectively. Such graphical summaries are based on binned data. They provide us with snapshots of how much probability mass is allocated in different parts of the data domain.

Instead of dealing with large datasets, we obtained a few counts. The process of binning and its textual or visual depictions is valuable in determining whether the distribution is unimodal or multimodal, skewed or symmetric around some point, what range of values contains most of the observations, and how small or large extreme values are.

Too much information may sometimes be overwhelming. Besides, revealing it might not be a clever idea for privacy or confidentiality reasons[1]. Consequently, we might be interested in even more synthetic descriptions – data aggregates which reduce the whole dataset into a *single* number reflecting one of its many characteristics. They can provide us with a kind of bird's eye view on some aspect of it. We refer to such processes as data *aggregation*; see [26, 38].

In this part, we discuss a few noteworthy *measures* of:

- *location*; e.g., central tendency measures such as mean and median;

- *dispersion*; e.g., standard deviation and interquartile range;

- distribution *shape*; e.g., skewness.

We also introduce *box and whisker plots*.

### 5.1.1 Measures of Location

**Arithmetic Mean and Median**

Two main measures of *central tendency* are:

- *the arithmetic mean* (sometimes for simplicity called the mean or average), defined as the sum of all observations divided by the sample size:

$$\bar{x} = \frac{(x_1 + x_2 + \cdots + x_n)}{n} = \frac{1}{n} \sum_{i=1}^{n} x_i,$$

- *the median*, being the middle value in a sorted version of the sample if its length is odd or the arithmetic mean of the two middle values otherwise:

$$m = \begin{cases} x_{(n+1)/2} & \text{if } n \text{ is odd,} \\ \frac{x_{(n/2)} + x_{(n/2+1)}}{2} & \text{if } n \text{ is even.} \end{cases}$$

They can be computed using the `numpy.mean` and `numpy.median` functions.

```
heights = np.loadtxt("https://raw.githubusercontent.com/gagolews/" +
    "teaching-data/master/marek/nhanes_adult_female_height_2020.txt")
np.mean(heights), np.median(heights)
## (160.1367922932953, 160.1)
income = np.loadtxt("https://raw.githubusercontent.com/gagolews/" +
    "teaching-data/master/marek/uk_income_simulated_2020.txt")
np.mean(income), np.median(income)
## (35779.994, 30042.0)
```

---

[1] Nevertheless, we strongly advocate for all information of concern to the public to be openly available, so that *experienced* statisticians can put them to good use.

Let us underline what follows:

- for symmetric distributions, the arithmetic mean and the median are expected to be more or less equal,

- for skewed distributions, the arithmetic mean will be biased towards the heavier tail.

**Exercise 5.1** *Get the arithmetic mean and median for the* `37_pzu_warsaw_marathon_mins` *dataset mentioned in Chapter 4.*

**Exercise 5.2** *(\*) Write a function that computes the median without the use of* `numpy.median` *(based on its mathematical definition and* `numpy.sort`*).*

---

**Note** (\*) Technically, the arithmetic mean can also be computed using the **mean** *method* for the `numpy.ndarray` class – it will sometimes be the case that we have many ways to perform the same operation. We can even "implement" it manually using the **sum** function. Thus, all the following expressions are equivalent:

```
print(
    np.mean(income),
    income.mean(),
    np.sum(income)/len(income),
    income.sum()/income.shape[0]
)
## 35779.994 35779.994 35779.994 35779.994
```

On the other hand, there exists the **numpy.median** function but, unfortunately, the **median** method for vectors is not available. This is why we prefer sticking with functions.

---

### Sensitive to Outliers vs Robust

The arithmetic mean is strongly influenced by very large or very small observations (which in some contexts we refer to as *outliers*). For instance, assume that we are inviting a billionaire to the `income` dataset:

```
income2 = np.append(income, [1_000_000_000])
print(np.mean(income), np.mean(income2))
## 35779.994 1034745.2487512487
```

Comparing this new result to the previous one, oh we all feel much richer now, right? In fact, the arithmetic mean reflects the income each of us would get if all the wealth were gathered inside a single Santa Claus's (or Robin Hood's) sack and then distributed equally amongst all of us. A noble idea provided that everyone contributes equally to the society, which unfortunately is not the case.

On the other hand, the median is the value such that 50% of the observations are less than or equal to it and 50% of the remaining ones are not less than it. Hence, it is completely insensitive to most of the data points – on both the left and the right side of the distribution:

```
print(np.median(income), np.median(income2))
## 30042.0 30076.0
```

Because of this, we cannot say that one measure is better than the other. Certainly, for symmetrical distributions with no outliers (e.g., `heights`), the mean will be better as it uses *all* data (and its efficiency can be proven for certain statistical models). For skewed distributions (e.g., `income`), the median has a nice interpretation, as it gives the value in the middle of the ordered sample. Let us still remember that these data summaries allow us to look at a single data aspect only, and there can be many different, valid perspectives. The reality is complex.

**Sample Quantiles**

Quantiles generalise the notions of the sample median and of the inverse of the empirical cumulative distribution function (Section 4.3.8). They provide us with the value that is not exceeded by the elements in a given sample with a predefined probability.

Before proceeding with a formal definition, which is quite technical, let us point out that for larger sample sizes, we have the following rule of thumb.

---

**Important**    For any $p$ between 0 and 1, the $p$-quantile, denoted $q_p$, is a value dividing the sample in such a way that approximately $100p\%$ of observations are not greater than $q_p$, and the remaining ca. $100(1 - p)\%$ are not less than $q_p$.

---

Quantiles appear under many different names, but they all refer to the same concept. In particular, we can speak about the $100p$-th *percentiles*, e.g., the 0.5-quantile is the same as the 50th percentile.

Furthermore:

- 0-quantile ($q_0$) – the minimum (also: `numpy.min`),
- 0.25-quantile ($q_{0.25}$) – the 1st quartile (denoted $Q_1$),
- 0.5-quantile ($q_{0.5}$) – the 2nd quartile a.k.a. median (denoted $m$ or $Q_2$),
- 0.75-quantile ($q_{0.75}$) – the 3rd quartile (denoted $Q_3$),
- 1.0-quantile ($q_1$) – the maximum (also: `numpy.max`).

Here are the above five aggregates for our two datasets:

```
np.quantile(heights, [0, 0.25, 0.5, 0.75, 1])
## array([131.1, 155.3, 160.1, 164.8, 189.3])
```

```
np.quantile(income, [0, 0.25, 0.5, 0.75, 1])
## array([  5750.  ,  20669.75,  30042.  ,  44123.75, 199969.  ])
```

**Exercise 5.3** *What is the income bracket for 95% of the most typical UK taxpayers? In other words, determine the 2.5- and 97.5-percentiles.*

**Exercise 5.4** *Compute the midrange of `income` and `heights`, being the arithmetic mean of the minimum and the maximum (this measure is extremely sensitive to outliers).*

---

**Note** (*) As we do not like the *approximately* part in the "asymptotic definition" given above, in this course we shall assume that for any $p \in [0, 1]$, the $p$-quantile is given by

$$q_p = x_{(\lfloor k \rfloor)} + (k - \lfloor k \rfloor)(x_{(\lfloor k \rfloor + 1)} - x_{(\lfloor k \rfloor)}),$$

where $k = (n-1)p + 1$ and $\lfloor k \rfloor$ is the floor function, i.e., the greatest integer less than or equal to $k$ (e.g., $\lfloor 2.0 \rfloor = \lfloor 2.001 \rfloor = \lfloor 2.999 \rfloor = 2$, $\lfloor 3.0 \rfloor = \lfloor 3.999 \rfloor = 3$, $\lfloor -3.0 \rfloor = \lfloor -2.999 \rfloor = \lfloor -2.001 \rfloor = -3$, and $\lfloor -2.0 \rfloor = \lfloor -1.001 \rfloor = -2$).

$q_p$ is a function that linearly interpolates between the points featuring the consecutive order statistics, $((k-1)/(n-1), x_{(k)})$ for $k = 1, \dots, n$. For instance, for $n = 5$, we connect the points $(0, x_{(1)})$, $(0.25, x_{(2)})$, $(0.5, x_{(3)})$, $(0.75, x_{(4)})$, $(1, x_{(5)})$. For $n = 6$, we do the same for $(0, x_{(1)})$, $(0.2, x_{(2)})$, $(0.4, x_{(3)})$, $(0.6, x_{(4)})$, $(0.8, x_{(5)})$, $(1, x_{(6)})$. See Figure 5.1 for an illustration.

Notice that for $p = 0.5$ we get the median regardless of whether $n$ is even or not.

---

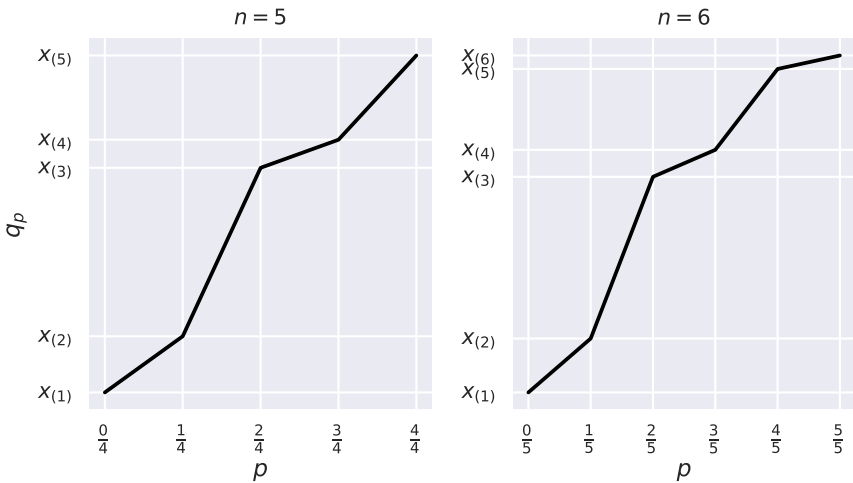

Figure 5.1: $q_p$ as a function of $p$ for example vectors of length 5 (left subfigure) and 6 (right)

---

**Note** (\*\*) There are many definitions of quantiles across statistical software; see the `method` argument to `numpy.quantile`. They were nicely summarised in [47] as well as in the corresponding Wikipedia[2] article. They are all approximately equivalent for large sample sizes (i.e., asymptotically), but the best practice is to be explicit about which variant we are using in the computations when reporting data analysis results. Accordingly, in our case, we say that we are relying on the type-7 quantiles as described in [47], see also [39].

In fact, simply mentioning that our computations are done with `numpy` version 1.xx (as specified in Section 1.4) implicitly implies that the default method parameters are used everywhere, unless otherwise stated. In many contexts, that is good enough.

---

### 5.1.2 Measures of Dispersion

Measures of central tendency quantify the location of the most *typical* value (whatever that means, and we already know it is complicated). That of dispersion (spread), on the other hand, will tell us how much *variability* is in our data. After all, when we say that the height of a group of people is 160 cm (on average) ± 14 cm (here, 2 standard deviations), the latter piece of information is a valuable addition (and is very different from the imaginary ± 4 cm case).

Some degree of variability might be good in certain contexts and bad in other ones. A bolt factory should keep the variance of the fasteners' diameters as low as possible: this is how we define quality products (assuming that they all meet, on average, the required specification). On the other hand, too much diversity in human behaviour, where everyone feels that they are special, is not really sustainable (but lack thereof would be extremely boring), and so forth.

Let us explore the different ways in which we can quantify this data aspect.

**Standard Deviation (and Variance)**

The standard deviation[3], is the average distance to the arithmetic mean:

$$s = \sqrt{\frac{(x_1 - \bar{x})^2 + (x_2 - \bar{x})^2 + \cdots + (x_n - \bar{x})^2}{n}} = \sqrt{\frac{1}{n} \sum_{i=1}^{n} (x_i - \bar{x})^2}.$$

Computing the above with `numpy`:

---

[2] https://en.wikipedia.org/wiki/Quantile
[3] (\*\*) Based on the so-called *uncorrected for bias* version of the sample variance. We prefer it here for didactical reasons (simplicity, better interpretability). Plus, it is the default one in `numpy`. Passing `ddof=1` (*delta degrees of freedom*) to `numpy.std` will apply division by $n - 1$ instead of by $n$. When used as an estimator of the distribution's standard deviation, the latter has slightly better statistical properties (which we normally explore in a course on mathematical statistics, which this one is not). On the other hand, we will see later that the `std` methods in the `pandas` package have `ddof=1` by default. Therefore, we might be interested in setting `ddof=0` therein.

```
np.std(heights), np.std(income)
## (7.062021850008261, 22888.77122437908)
```

The standard deviation quantifies the typical amount of spread around the arithmetic mean. It is overall adequate for making comparisons across different samples measuring similar things (e.g., heights of males vs of females, incomes in the UK vs in South Africa). However, without further assumptions, it is quite difficult to express the meaning of a particular value of $s$ (e.g., the statement that the standard deviation of income is £22,900 is hard to interpret). This measure makes therefore most sense for data distributions that are symmetric around the mean.

---

**Note** (*) For bell-shaped data (more precisely: for normally-distributed samples; see the next chapter) such as `heights`, we sometimes report $\bar{x} \pm 2s$, because the theoretical expectancy is that ca. 95% of data points fall into the $[\bar{x} - 2s, \bar{x} + 2s]$ interval (the so-called $2\sigma$ rule).

---

Further, the *variance* is the square of the standard deviation, $s^2$. Mind that if data are expressed in centimetres, then the variance is in centimetres *squared*, which is not very intuitive. The standard deviation does not have this drawback. Mathematicians find the square root annoying though (for many reasons); that is why we will come across the $s^2$ every now and then too.

**Interquartile Range**

The interquartile range (IQR) is another popular measure of dispersion. It is defined as the difference between the 3rd and the 1st quartile:

$$\mathrm{IQR} = q_{0.75} - q_{0.25} = Q_3 - Q_1.$$

Computing the above is almost effortless:

```
np.quantile(heights, 0.75) - np.quantile(heights, 0.25)
## 9.5
np.quantile(income, 0.75) - np.quantile(income, 0.25)
## 23454.0
```

The IQR has an appealing interpretation: it is the range comprised of the 50% *most typical* values. It is a quite robust measure, as it ignores the 25% smallest and 25% largest observations. Standard deviation, on the other hand, is extremely sensitive to outliers.

Furthermore, by *range* (or support) we will mean a measure based on extremal quantiles: it is the difference between the maximal and minimal observation.

### 5.1.3   Measures of Shape

From a histogram, we can easily read whether a dataset is symmetric or skewed. It turns out that we can easily quantify such a characteristic. Namely, the *skewness* is given by:

$$g = \frac{\frac{1}{n} \sum_{i=1}^{n} (x_i - \bar{x})^3}{\left( \sqrt{\frac{1}{n} \sum_{i=1}^{n} (x_i - \bar{x})^2} \right)^3}.$$

For symmetric distributions, skewness is approximately zero. Positive and negative skewness indicates a heavier right and left tail, respectively.

For example, heights are an instance of an almost-symmetric distribution:

```
scipy.stats.skew(heights)
## 0.0811184528074054
```

Income, on the other hand, is right-skewed:

```
scipy.stats.skew(income)
## 1.9768735693998942
```

Now we have them expressed as a single number.

---

**Note** (*) It is worth stressing that $g > 0$ does not imply that the sample mean is greater than the median. As an alternative measure of skewness, sometimes the practitioners use:

$$g' = \frac{\bar{x} - m}{s}.$$

*Yule's coefficient* is an example of a robust skewness measure:

$$g'' = \frac{Q_3 + Q_1 - 2m}{Q_3 - Q_1}.$$

The computation thereof on our example datasets is left as an exercise.

Furthermore, *kurtosis* (or Fisher's excess kurtosis, compare `scipy.stats.kurtosis`) can be used to describe whether an empirical distribution is heavy- or thin-tailed.

---

### 5.1.4   Box (and Whisker) Plots

A *box and whisker* plot (*box plot* for short) depicts some noteworthy features of a data sample.

```python
plt.subplot(2, 1, 1)  # 2 rows, 1 column, 1st subplot
sns.boxplot(data=heights, orient="h", color="lightgray")
plt.yticks([0], ["heights"])
plt.subplot(2, 1, 2)  # 2 rows, 1 column, 2nd subplot
sns.boxplot(data=income, orient="h", color="lightgray")
plt.yticks([0], ["income"])
plt.show()
```

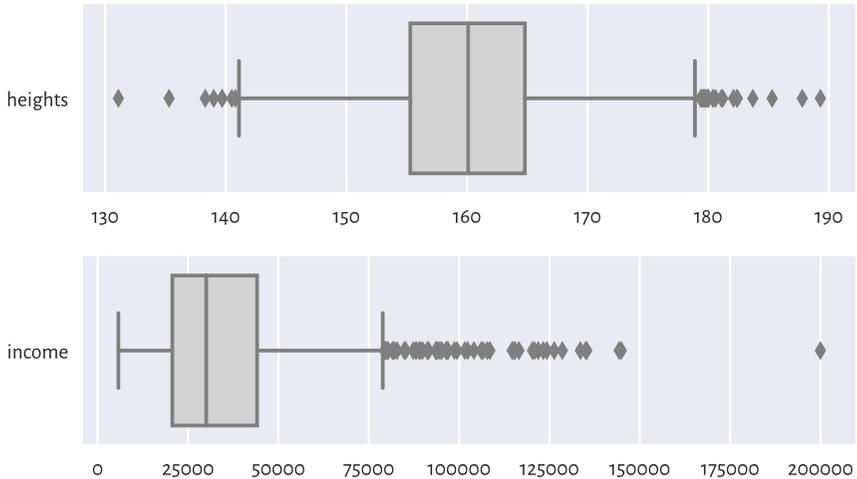

Figure 5.2: Example box plots

Each box plot (compare Figure 5.2) consists of:

- the box, which spans between the 1st and the 3rd quartile:

  - the median is clearly marked by a vertical bar inside the box;

  - the width of the box corresponds to the IQR;

- the whiskers, which span[4] between:

  - the smallest observation (the minimum) or $Q_1 - 1.5\text{IQR}$ (the left side of the box minus 3/2 of its width), whichever is larger, and

  - the largest observation (the maximum) or $Q_3 + 1.5\text{IQR}$ (the right side of the box plus 3/2 of its width), whichever is smaller.

Additionally, all observations that are less than $Q_1 - 1.5\text{IQR}$ (if any) or greater than $Q_3 + 1.5\text{IQR}$ (if any) are separately marked.

---

[4] The 1.5IQR rule is the most popular in the statistical literature, but some plotting software may use different whisker ranges. See Section 15.4.1 for further discussion.

**Note**  We are used to referring to the individually marked points as *outliers*. Still, it does not automatically mean there is anything *anomalous* about them. They are *atypical* in the sense that they are considerably farther away from the box. But this might as well indicate some problems in data quality (e.g., when someone made a typo entering the data). Actually, box plots are calibrated (via the nicely round magic constant 1.5) in such a way that we expect there to be no or only few outliers if the data are normally distributed. For skewed distributions, there will naturally be many outliers on either side; see Section 15.4 for more details.

Box plots are based solely on sample quantiles. Most of the statistical packages *do not* draw the arithmetic mean. If they do, it is marked with a distinctive symbol.

**Exercise 5.5**  *Call* `matplotlib.pyplot.plot(numpy.mean(..data..), 0, "wX")` *after* `seaborn.boxplot` *to mark the arithmetic mean with a white cross.*

Box plots are particularly beneficial for comparing data samples with each other (e.g., body measures of men and women separately), both in terms of the relative shift (location) as well as spread and skewness; see, e.g., Figure 12.1.

**Example 5.6**  *(\*) We may also sometimes be interested in a* violin plot *(Figure 5.3), which combines a box plot (although with no outliers marked) and the so-called kernel density estimator (which is a smoothened version of a histogram; see Section 15.4.2).*

```
plt.subplot(2, 1, 1)  # 2 rows, 1 column, 1st subplot
sns.violinplot(data=heights, orient="h", color="lightgray")
plt.yticks([0], ["heights"])
plt.subplot(2, 1, 2)  # 2 rows, 1 column, 2nd subplot
sns.violinplot(data=income, orient="h", color="lightgray")
plt.yticks([0], ["income"])
plt.show()
```

### 5.1.5  Further Methods (\*)

We said that the arithmetic mean is overly sensitive to extreme observations. The sample median is an example of a *robust* aggregate — it ignores all but 1–2 middle observations (we would say it has a high *breakdown point*). Some measures of central tendency that are in-between the mean-median extreme include:

- *trimmed means* – the arithmetic mean of all the observations except several, say $p$, the smallest and the greatest ones,

- *winsorised means* – the arithmetic mean with $p$ smallest and $p$ greatest observations replaced with the $(p + 1)$-th smallest the $(p + 1)$-th largest one.

The arithmetic mean is not the only mean of interest. The two other famous means are the *geometric* and *harmonic* ones. The former is more meaningful for averaging growth rates and speedups whilst the latter can be used for computing the average speed from speed measurements at sections of identical lengths; see also the notion of the F-measure in Section 12.3.2. Also, the quadratic mean is featured in the definition of the standard deviation (it is the quadratic mean of the distances to the mean).

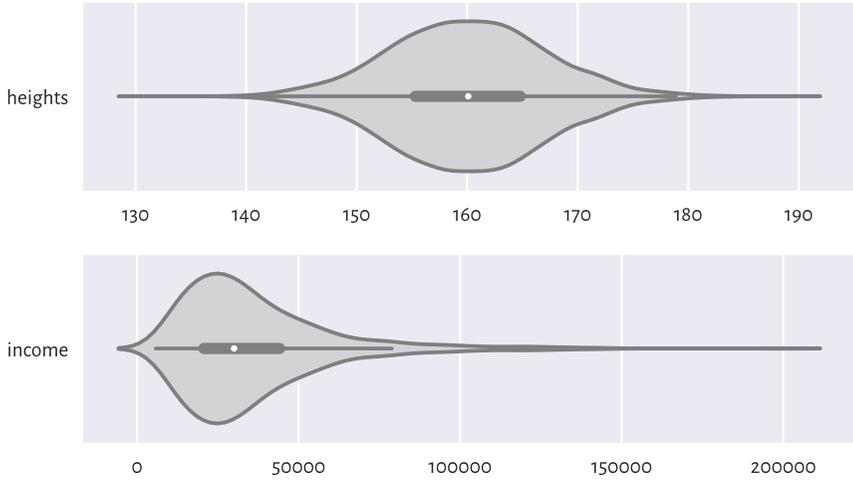

Figure 5.3: Example violin plots

As far as spread measures are concerned, the interquartile range (IQR) is a robust statistic. If necessary, the standard deviation might be replaced with:

- mean absolute deviation from the mean: $\frac{1}{n} \sum_{i=1}^{n} |x_i - \bar{x}|$,

- mean absolute deviation from the median: $\frac{1}{n} \sum_{i=1}^{n} |x_i - m|$,

- median absolute deviation from the median: the median of $(|x_1 - m|, |x_2 - m|, \dots, |x_n - m|)$.

The *coefficient of variation*, being the standard deviation dived by the arithmetic mean, is an example of a *relative* (or normalised) spread measure. It can be appropriate for comparing data on different scales, as it is unit-less (think how standard deviation changes when you convert between metres and centimetres).

The *Gini index*, widely used in economics, can also serve as a measure of relative dispersion, but assumes that all data points are nonnegative:

$$G = \frac{\sum_{i=1}^{n} \sum_{j=1}^{n} |x_i - x_j|}{2(n-1)n\,\bar{x}} = \frac{\sum_{i=1}^{n} (n - 2i + 1)x_{(n-i+1)}}{(n-1) \sum_{i=1}^{n} x_i}.$$

It is normalised so that it takes values in the unit interval. An index of 0 reflects the situation where all values in a sample are the same (0 variance; perfect equality). If there is a single entity in possession of all the "wealth", and the remaining ones are 0, then the index is equal to 1.

For a more generic (algebraic) treatment of aggregation functions for unidimensional data; see, e.g., [9, 26, 27, 38]. Some measures might be better than others under certain (often strict) assumptions usually explored in a course on mathematical statistics, e.g., [35].

Overall, numerical aggregates should be used in cases where data are unimodal. For multimodal mixtures or data in groups, they should rather be applied to summarise each cluster/class separately; compare Chapter 12. Also, in Chapter 8, we will extend consider a few summaries for multidimensional data.

## 5.2  Vectorised Mathematical Functions

**numpy**, just like any other comprehensive numerical computing package, library, or environment (e.g., R, GNU Octave, Scilab, and Julia), defines many basic mathematical functions:

- absolute value: **numpy.abs**,

- square and square root: **numpy.square** and **numpy.sqrt**, respectively,

- (natural) exponential function: **numpy.exp**,

- logarithms: **numpy.log** (the natural logarithm, i.e., base $e$), **numpy.log10** (base 10), etc.,

- trigonometric functions: **numpy.cos**, **numpy.sin**, **numpy.tan**, etc., and their inverses: **numpy.arccos**, etc.

- rounding and truncating: **numpy.round**, **numpy.floor**, **numpy.ceil**, **numpy.trunc**.

Each of these functions is *vectorised*: if we apply, say, $f$, on a vector like $\boldsymbol{x} = (x_1, \dots, x_n)$, we obtain a sequence of the same size with all elements appropriately transformed:

$$f(\boldsymbol{x}) = (f(x_1), f(x_2), \dots, f(x_n)).$$

We will frequently be using such operations for adjusting the data, e.g., as in Figure 6.7, where we discover that the *logarithm* of the UK incomes has a bell-shaped distribution.

An example call to the vectorised version of the rounding function:

```
np.round([-3.249, -3.151, 2.49, 2.51, 3.49, 3.51], 1)
## array([-3.2, -3.2,  2.5,  2.5,  3.5,  3.5])
```

The input list has been automatically converted to a **numpy** vector.

**Important**  Thanks to the vectorised functions, our code is not only more readable,

but also runs faster: we do not have to employ the generally slow Python-level **while** or **for** loops to traverse through each element in a given sequence.

---

Here are some significant properties of the natural logarithm and its inverse, the exponential function. By convention, Euler's number $e \simeq 2.718$, $\log x = \log_e x$, and $\exp(x) = e^x$.

- $\log 1 = 0$, $\log e = 1$; note that logarithms are only defined for $x > 0$: in the limit as $x \to 0$, we have that $\log x \to -\infty$,

- $\log x^y = y \log x$ and hence $\log e^x = x$,

- $\log(xy) = \log x + \log y$ and thus $\log(x/y) = \log x - \log y$,

- $e^0 = 1$, $e^1 = e$, and $e^x \to 0$ as $x \to -\infty$,

- $e^{\log x} = x$,

- $e^{x+y} = e^x e^y$ and so $e^{x-y} = e^x/e^y$.

Both functions are strictly increasing. For $x \geq 1$, the logarithm grows very slowly whereas the exponential function increases very rapidly; see Figure 5.4.

```python
plt.subplot(1, 2, 1)
x = np.linspace(np.exp(-2), np.exp(3), 1001)
plt.plot(x, np.log(x), label="$y=\\log x$")
plt.legend()
plt.subplot(1, 2, 2)
x = np.linspace(-2, 3, 1001)
plt.plot(x, np.exp(x), label="$y=\\exp(x)$")
plt.legend()
plt.show()
```

Logarithms of different bases and non-natural exponential functions are also available. In particular, when drawing plots, we used the base-10 logarithmic scales on the axes. It holds $\log_{10} x = \frac{\log x}{\log 10}$ and its inverse is $10^x$. For example:

```python
10.0**np.array([-1, 0, 1, 2])  # see below
## array([  0.1,    1. ,   10. ,  100. ])
np.log10([-1, 0.01, 0.1, 1, 2, 5, 10, 100, 1000, 10000])
## array([     nan, -2.    , -1.    ,  0.    ,  0.30103,  0.69897,
##         1.    ,  2.    ,  3.    ,  4.    ])
##
## <string>:1: RuntimeWarning: invalid value encountered in log10
```

Take note of the warning and the not-a-number (NaN) generated.

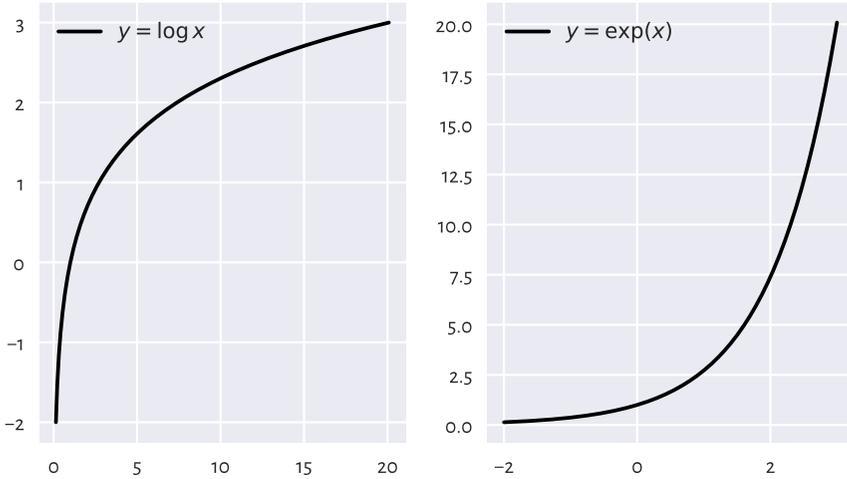

Figure 5.4: The natural logarithm (left) and the exponential function (right)

**Exercise 5.7** *Check that when using the log-scale on the x-axis (`plt.xscale("log")`), the plot of the logarithm (of any base) is a straight line. Similarly, the log-scale on the y-axis (`plt.yscale("log")`) makes the exponential function linear.*

Moving on, the trigonometric functions in **numpy** accept angles in radians. If $x$ is the degree in angles, then to compute its cosine, we should instead write $\cos(x\pi/180)$; see Figure 5.5.

```
x = np.linspace(-2*np.pi, 4*np.pi, 1001)
plt.plot(x, np.cos(x))
plt.xticks(
    [-2*np.pi, -np.pi, 0, np.pi/2, np.pi, 3*np.pi/2, 2*np.pi, 4*np.pi],
    ["$-2\\pi$", "$-\\pi$", "$0$", "$\\pi/2$", "$\\pi$",
     "$3\\pi/2$", "$2\\pi$", "$4\\pi$"]
)
plt.show()
```

Some identities worth memorising:

- $\sin x = \cos(\pi/2 - x)$,
- $\cos(-x) = \cos x$,
- $\cos^2 x + \sin^2 x = 1$, where $\cos^2 x = (\cos x)^2$,
- $\cos(x + y) = \cos x \cos y - \sin x \sin y$,
- $\cos(x - y) = \cos x \cos y + \sin x \sin y$.

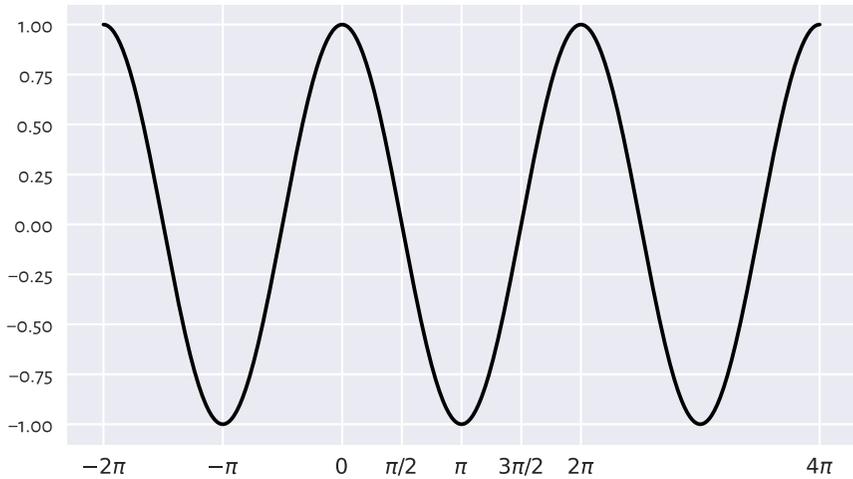

Figure 5.5: The cosine

We will refer to them later.

---

**Important**  The classical handbook of mathematical functions and the (in)equalities related to them is [1], see [62] for its updated version.

---

## 5.3  Arithmetic Operators

We can apply the standard binary (two-argument) arithmetic operators `+`, `-`, `*`, `/`, `**`, `%`, and `//` on vectors too. Below we discuss the possible cases of the operands' lengths.

### 5.3.1  Vector-Scalar Case

Often, we will be referring to the binary operators in contexts where one operand is a vector and the other one is a single value (scalar), for example:

```
np.array([-2, -1, 0, 1, 2, 3])**2
## array([4, 1, 0, 1, 4, 9])
(np.array([-2, -1, 0, 1, 2, 3])+2)/5
## array([0. , 0.2, 0.4, 0.6, 0.8, 1. ])
```

In such a case, each element in the vector is being operated upon (e.g., squared, di-vided by 5) and we get a vector of the same length in return. Hence, in this case, the operators behave just like the vectorised mathematical functions discussed above.

Mathematically, it is common to assume that the scalar multiplication and, less commonly, the addition are performed in this way.

$$c\boldsymbol{x} + t = (cx_1 + t, cx_2 + t, \dots, cx_n + t).$$

We will also become used to writing, $(\boldsymbol{x}-t)/c$ which is of course equivalent to $(1/c)\boldsymbol{x} + (-t/c)$.

### 5.3.2   Application: Feature Scaling

Vector-scalar operations and aggregation functions are the basis for the most commonly applied *feature scalers*:

- standardisation,
- normalisation,
- min-max scaling and clipping.

They can increase the interpretability of data points by bringing them onto a common, unitless scale. They might also be essential when computing pairwise distances between multidimensional points; see Section 8.4.

All the above are linear transforms of the form $y = cx + t$, which geometrically we can interpret as scaling (stretching or shrinking) and shifting (translating), see Figure 5.6 for an illustration.

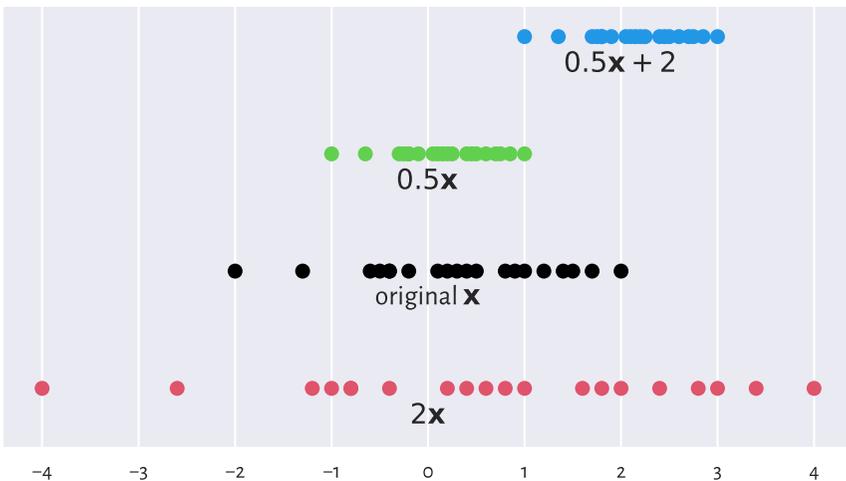

Figure 5.6: Scaled and shifted versions of an example vector

**Note**  Let $y = cx + t$ and let $\bar{x}, \bar{y}, s_x, s_y$ denote the vectors' arithmetic means and standard deviations. The following properties hold.

- The arithmetic mean and all the quantiles (including, of course, the median), are equivariant with respect to translation and scaling; it holds, for instance, $\bar{y} = c\bar{x} + t$.

- The standard deviation, the interquartile range, and the range are invariant to translations and equivariant to scaling; e.g., $s_y = cs_x$.

As a by-product, for the variance, we get... $s_y^2 = c^2 s_x^2$.

**Standardisation and Z-Scores**

A *standardised* version of a vector $x = (x_1, \dots, x_n)$ consists in subtracting, from each element, the sample arithmetic mean (which we call *centring*) and then dividing it by the standard deviation, i.e., $z = (x - \bar{x})/s$.

Thus, we transform each $x_i$ to obtain:

$$z_i = \frac{x_i - \bar{x}}{s}.$$

Consider again the female heights dataset:

```
heights[-5:] # preview
## array([157. , 167.4, 159.6, 168.5, 147.8])
```

whose mean $\bar{x}$ and standard deviation $s$ are equal to:

```
np.mean(heights), np.std(heights)
## (160.13679222932953, 7.062021850008261)
```

With **numpy**, standardisation is as simple as applying two aggregation functions and two arithmetic operations:

```
heights_std = (heights-np.mean(heights))/np.std(heights)
heights_std[-5:]  # preview
## array([-0.44417764,  1.02848843, -0.07601113,  1.18425119, -1.74692071])
```

What we obtained is sometimes referred to as the *z-scores*. They are nicely interpretable:

- z-score of 0 corresponds to an observation equal to the sample mean (perfectly average);

- z-score of 1 is obtained for a datum 1 standard deviation above the mean;

- z-score of -2 means that it is a value 2 standard deviations below the mean;

and so forth.

Because of the way they emerge, the mean of the z-scores is always 0 and standard deviation is 1 (up to a tiny numerical error, as usual; see Section 5.5.6):

```
np.mean(heights_std), np.std(heights_std)
## (1.8920872660373198e-15, 1.0)
```

Even though the original `heights` were measured in centimetres, the z-scores are *unitless* (centimetres divided by centimetres).

---

**Important**    Standardisation enables the comparison of measurements on different scales (think: height in centimetres vs weight in kilograms or apples vs oranges). It makes the most sense for bell-shaped distributions, in particular normally-distributed ones. In the next chapter we will note that, due to the 2σ rule for the normal family, we *expect* that 95% of observations should have z-scores between -2 and 2. Further, z-scores less than -3 and greater than 3 are highly unlikely.

---

**Exercise 5.8**  *We have a patient whose height z-score is 1 and weight z-score is -1. How can we interpret this information?*

**Exercise 5.9**  *How about a patient whose weight z-score is 0 but BMI z-score is 2?*

On a side note, sometimes we might be interested in performing some form of *robust* standardisation (e.g., for skewed data or those that feature some outliers). In such a case, we can replace the mean with the median and the standard deviation with the IQR.

**Min-Max Scaling and Clipping**

A less frequently but still noteworthy transformation is called *min-max scaling* and involves subtracting the minimum and then dividing by the range, $(x - x_{(1)})/(x_{(n)} - x_{(1)})$.

```
x = np.array([-1.5, 0.5, 3.5, -1.33, 0.25, 0.8])
(x - np.min(x))/(np.max(x)-np.min(x))
## array([0.   , 0.4  , 1.   , 0.034, 0.35 , 0.46 ])
```

Here, the smallest value is mapped to 0 and the largest becomes equal to 1. Let us stress that, in this context, 0.5 does not represent the value which is equal to the mean (unless we are incredibly lucky).

Also, *clipping* can be used to replace all values less than 0 with 0 and those greater than 1 with 1.

```
np.clip(x, 0, 1)
## array([0.   , 0.5  , 1.   , 0.   , 0.25, 0.8 ])
```

The function is of course flexible; another popular choice is clipping to $[-1, 1]$. This can also be implemented manually by means of the vectorised pairwise minimum and maximum functions.

```
np.minimum(1, np.maximum(0, x))
## array([0.  , 0.5 , 1.  , 0.  , 0.25, 0.8 ])
```

**Normalisation ($l_2$; Dividing by Magnitude)**

*Normalisation* is the scaling of a given vector so that it is of *unit length*. Usually, by *length* we mean the square root of the sum of squares, i.e., the Euclidean ($l_2$) norm also known as the *magnitude*:

$$\|(x_1, \dots, x_n)\| = \sqrt{x_1^2 + x_2^2 + \dots + x_n^2} = \sqrt{\sum_{i=1}^{n} x_i^2}.$$

Its special case for $n = 2$ we know well from high school: the length of a vector $(a, b)$ is $\sqrt{a^2 + b^2}$, e.g., $\|(1, 2)\| = \sqrt{5} \simeq 2.236$. Also, it is good to program our brains so that when next time we see $\|x\|^2$, we immediately think of the *sum of squares*.

Consequently, a normalised vector:

$$\frac{x}{\|x\|} = \left( \frac{x_1}{\|x\|}, \frac{x_2}{\|x\|}, \dots, \frac{x_n}{\|x\|} \right),$$

can be determined via:

```
x = np.array([1, 5, -4, 2, 2.5])  # example vector
x/np.sqrt(np.sum(x**2))  # x divided by the Euclidean norm of x
## array([ 0.13834289,  0.69171446, -0.55337157,  0.27668579,  0.34585723])
```

**Exercise 5.10** *Normalisation is similar to standardisation if data are already centred (when the mean was subtracted). Show that we can obtain one from the other via the scaling by $\sqrt{n}$.*

---

**Important** A common confusion is that normalisation is supposed to make data *more normally* distributed. This is not the case[5], as we only scale (stretch or shrink) the observations here.

---

**Normalisation ($l_1$; Dividing by Sum)**

At other times, we might be interested in considering the Manhattan ($l_1$) norm:

$$\|(x_1, \dots, x_n)\|_1 = |x_1| + |x_2| + \dots + |x_n| = \sum_{i=1}^{n} |x_i|,$$

being the sum of absolute values.

---

[5] (*) A Box–Cox transformation can help achieve this in some datasets; see [8].

```
x / np.sum(np.abs(x))
## array([ 0.06896552,  0.34482759, -0.27586207,  0.13793103,  0.17241379])
```

$l_1$ normalisation is frequently applied on vectors of nonnegative values, whose normalised versions can be interpreted as *probabilities* or *proportions*: values between 0 and 1 which sum to 1 (or, equivalently, 100%).

**Example 5.11**  *Given some binned data:*

```
c, b = np.histogram(heights, [-np.inf, 150, 160, 170, np.inf])
print(c)  # counts
## [ 306 1776 1773  366]
```

*We can convert the counts to empirical probabilities:*

```
p = c/np.sum(c)    # np.abs is not needed here
print(p)
## [0.07249467 0.42075338 0.42004264 0.08670931]
```

*We did not apply* **numpy.abs**, *because the values were already nonnegative.*

### 5.3.3   Vector-Vector Case

So far we have been applying `*`, `+`, etc., on vectors and scalars only. All arithmetic operators can also be applied on two vectors of equal lengths. In such a case, they will act *elementwisely*: taking each element from the first operand and combining it with the *corresponding* element from the second argument:

```
np.array([2, 3, 4, 5]) * np.array([10, 100, 1000, 10000])
## array([   20,   300,  4000, 50000])
```

We see that the first element in the left operand (2) was multiplied by the first element in the right operand (10). Then, we multiplied 3 by 100 (the second corresponding elements), and so forth.

Such a behaviour of the binary operators is inspired by the usual convention in vector algebra where applying $+$ (or $-$) on $\boldsymbol{x} = (x_1, \dots, x_n)$ and $\boldsymbol{y} = (y_1, \dots, y_n)$ means exactly:

$$\boldsymbol{x} + \boldsymbol{y} = (x_1 + y_1, x_2 + y_2, \dots, x_n + y_n).$$

Using other operators this way (elementwisely) is less standard in mathematics (for instance multiplication might denote the dot product), but in **numpy** it is really convenient.

**Example 5.12**  *Let us compute the value of the expression* $h = -(p_1 \log p_1 + \dots + p_n \log p_n)$, *i.e.,* $h = -\sum_{i=1}^{n} p_i \log p_i$ *(the so-called entropy):*

```
p = np.array([0.1, 0.3, 0.25, 0.15, 0.12, 0.08])  # example vector
-np.sum(p*np.log(p))
## 1.6790818544987114
```

The above involves the use of a unary vectorised minus (change sign), an aggregation function (sum), a vectorised mathematical function (log), and an elementwise multiplication of two vectors of the same lengths.

**Example 5.13** Let us assume that – for whatever reason – we would like to plot two mathematical functions, the sine, $f(x) = \sin x$, and a polynomial of degree 7, $g(x) = x - x^3/6 + x^5/120 - x^7/5040$ for $x$ in the interval $[-\pi, 3\pi/2]$.

To do this, we can probe the values of $f$ and $g$ at sufficiently many points using the vectorised operations discussed so far and then use the **matplotlib.pyplot.plot** function to draw what we see in Figure 5.7.

```
x = np.linspace(-np.pi, 1.5*np.pi, 1001)  # many points in the said interval
yf = np.sin(x)
yg = x - x**3/6 + x**5/120 - x**7/5040
plt.plot(x, yf, 'k-', label="f(x)")  # black solid line
plt.plot(x, yg, 'r:', label="g(x)")  # red dotted line
plt.legend()
plt.show()
```

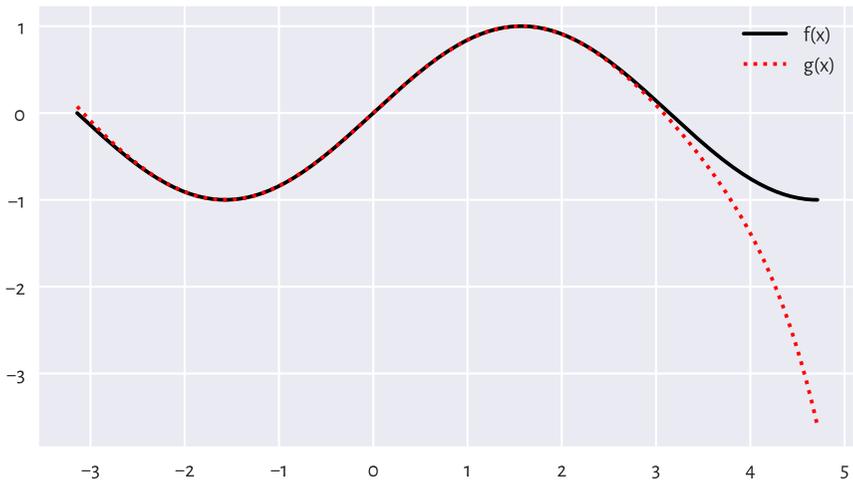

Figure 5.7: With vectorised functions, it is easy to generate plots like this one; we used different line styles so that the plot is readable also when printed in black and white

*Decreasing the number of points in x will reveal that the plotting function merely draws a series of straight-line segments. Computer graphics is essentially discrete.*

**Exercise 5.14** *Using a single line of code, compute the vector of BMIs of all persons based on the* `nhanes_adult_female_height_2020`[6] *and* `nhanes_adult_female_weight_2020`[7] *datasets. It is assumed that the* i*-th elements therein both refer to the same person.*

---

## 5.4    Indexing Vectors

Recall from Section 3.2.1 and Section 3.2.2 that sequential objects in Python (lists, tuples, strings, ranges) support indexing using scalars and slices:

```
x = [10, 20, 30, 40, 50]
x[1]  # scalar index – extract
## 20
x[1:2]  # slice index – subset
## [20]
```

**numpy** vectors support two *additional* indexing schemes: using integer and boolean sequences.

### 5.4.1    Integer Indexing

Indexing with a single integer *extracts* a particular element:

```
x = np.array([10, 20, 30, 40, 50])
x[0]  # first
## 10
x[1]  # second
## 20
x[-1]  # last
## 50
```

We can also use lists of vectors of integer indexes, which return a subvector with elements at the specified indexes:

```
x[ [0] ]
## array([10])
x[ [0, 1, -1, 0, 1, 0, 0] ]
## array([10, 20, 50, 10, 20, 10, 10])
x[ [] ]
## array([], dtype=int64)
```

---

[6] https://github.com/gagolews/teaching-data/raw/master/marek/nhanes_adult_female_height_2020.txt

[7] https://github.com/gagolews/teaching-data/raw/master/marek/nhanes_adult_female_weight_2020.txt

We added some spaces between the square brackets, but only because, for example, x[[0]] might seem slightly more enigmatic. (What are these double square brackets? Nah, it is a list inside the index operator.)

## 5.4.2  Logical Indexing

Subsetting using a logical vector of the same length as the indexed vector is possible too:

```
x[ [True, False, True, True, False] ]
## array([10, 30, 40])
```

This returned the 1st, 3rd, and 4th element (select 1st, omit 2nd, select 3rd, select 4th, omit 5th).

This is particularly useful as a *data filtering* technique. Knowing that the relational operators `<`, `<=`, `==`, `!=`, `>=`, and `>` on vectors are performed elementwisely (just like `+`, `*`, etc.), for instance:

```
x >= 30
## array([False, False,  True,  True,  True])
```

we can write:

```
x[ x >= 30 ]
## array([30, 40, 50])
```

to mean "select the elements in x which are not less than 30".

Of course, the indexed vector and the vector specifying the *filter* do not[8] have to be the same:

```
y = (x/10) % 2  # whatever
y  # equal to 0 if a number is a multiply of 10 times an even number
## array([1., 0., 1., 0., 1.])
x[ y == 0 ]
## array([20, 40])
```

---

**Important**  If we would like to combine many logical vectors, sadly we cannot use the **and**, **or**, and **not** operators, because they are not vectorised (this is a limitation of our language per se).

In **numpy**, we use the `&`, `|`, and `~` operators instead. Unfortunately, they have a

---

[8] (*) This is because the indexer is computed first, and its *value* is passed as an argument to the index operator. Python neither is a symbolic programming language, nor does it feature any nonstandard evaluation techniques. In other words, [...] does not care how the indexer was obtained.

lower order of precedence than `<`, `<=`, `==`, etc. Therefore, the bracketing of the comparisons is obligatory.

---

For example:

```
x[ (20 <= x) & (x <= 40) ]  # check what happens if we skip the brackets
## array([20, 30, 40])
```

means "elements in x between 20 and 40" (greater than or equal to 20 and less than or equal to 40).

**Exercise 5.15** *Compute the BMIs only of the women whose height is between 150 and 170 cm.*

### 5.4.3 Slicing

Just as with ordinary lists, slicing with ":" can be used to fetch the elements at indexes in a given range like `from:to` or `from:to:by`.

```
x[::-1]
## array([50, 40, 30, 20, 10])
x[3:]
## array([40, 50])
x[1:4]
## array([20, 30, 40])
```

---

**Important** For efficiency reasons, slicing returns a *view* on existing data. It does not have to make an independent copy of the subsetted elements, because sliced ranges are *regular* by definition.

---

In other words, both x and its sliced version share the same memory. This is important when we apply operations which modify a given vector in place, such as the `sort` method.

```
y = np.array([6, 4, 8, 5, 1, 3, 2, 9, 7])
y[::2] *= 10  # modifies parts of y in place
y  # has changed
## array([60,  4, 80,  5, 10,  3, 20,  9, 70])
```

This multiplied every second element in y by 10 (i.e., [6, 8, 1, 2, 7]). On the other hand, indexing with an integer or logical vector always returns a copy.

```
y[ [1, 3, 5, 7] ] *= 10  # modifies a new object and then forgets about it
y  # has not changed since the last modification
## array([60, 40, 80, 50, 10, 30, 20, 90, 70])
```

This *did not* modify the original vector, because we applied `*=` on a different object, which has not even been memorised after that operation took place.

## 5.5 Other Operations

### 5.5.1 Cumulative Sums and Iterated Differences

Recall that the `+` operator acts on two vectors elementwisely and that the `numpy.sum` function aggregates all values into a single one. We have a similar function, but vectorised in a slightly different fashion. Namely, `numpy.cumsum` returns the vector of *cumulative sums*:

```
np.cumsum([5, 3, -4, 1, 1, 3])
## array([5, 8, 4, 5, 6, 9])
```

This gave, in this order: the first element, the sum of first two elements, the sum of first three elements, …, the sum of all elements.

*Iterative differences* are a somewhat inverse operation:

```
np.diff([5, 8, 4, 5, 6, 9])
## array([ 3, -4,  1,  1,  3])
```

returned the difference between the 2nd and 1st element, then the difference between the 3rd and the 2nd, and so forth. The resulting vector is one element shorter than the input one.

We often make use of cumulative sums and iterated differences when processing time series, e.g., stock exchange data (e.g., by how much the price changed since the previous day?; Section 16.3.1) or determining cumulative distribution functions (Section 4.3.8).

### 5.5.2 Sorting

The `numpy.sort` function returns a sorted copy of a given vector, i.e., determines the order statistics.

```
x = np.array([40, 10, 20, 40, 40, 30, 20, 40, 50, 10, 10, 70, 30, 40, 30])
np.sort(x)
## array([10, 10, 10, 20, 20, 30, 30, 30, 40, 40, 40, 40, 40, 50, 70])
```

The `sort` method (as in: `x.sort()`), on the other hand, sorts the vector in place (and returns nothing).

**Exercise 5.16** *Readers interested more in chaos than in bringing order should give* `numpy.random.permutation` *a try. This function shuffles the elements in a given vector.*

### 5.5.3  Dealing with Tied Observations

Some statistical methods, especially for continuous data[9], assume that all observations in a vector are unique, i.e., there are no *ties*. In real life, however, some values might be recorded multiple times. For instance, two marathoners might finish their runs in exactly the same time, data can be rounded up to a certain number of fractional digits, or it just happens that observations are inherently integer. Therefore, we should be able to detect duplicated entries.

`numpy.unique` is a workhorse for dealing with tied observations.

```
x = np.array([40, 10, 20, 40, 40, 30, 20, 40, 50, 10, 10, 70, 30, 40, 30])
np.unique(x)
## array([10, 20, 30, 40, 50, 70])
```

Returns a *sorted*[10] version of a given vector with duplicates removed.

We can also get the corresponding counts:

```
np.unique(x, return_counts=True)  # returns a tuple of length 2
## (array([10, 20, 30, 40, 50, 70]), array([3, 2, 3, 5, 1, 1]))
```

This can be used to determine if all the values in a vector are unique:

```
np.all(np.unique(x, return_counts=True)[1] == 1)
## False
```

**Exercise 5.17** *Play with the* `return_index` *argument to* `numpy.unique` *that allows pinpointing the indexes of the first occurrences of each unique value.*

### 5.5.4  Determining the Ordering Permutation and Ranking

`numpy.argsort` returns a sequence of indexes that lead to an ordered version of a given vector (i.e., an ordering permutation).

```
x = np.array([40, 10, 20, 40, 40, 30, 20, 40, 50, 10, 10, 70, 30, 40, 30])
np.argsort(x)
## array([ 1,  9, 10,  2,  6,  5, 12, 14,  0,  3,  4,  7, 13,  8, 11])
```

Which means that the smallest element is at index 1, then the 2nd smallest is at index 9, 3rd smallest at index 10, etc. Therefore:

---

[9] Where, theoretically, the probability of obtaining a tie is equal to 0.

[10] Later we will mention `pandas.unique` which lists the values in the order of appearance.

```
x[np.argsort(x)]
## array([10, 10, 10, 20, 20, 30, 30, 30, 40, 40, 40, 40, 40, 50, 70])
```

is equivalent to **numpy.sort**(x).

---

**Note** (**) If there are tied observations in a vector x, **numpy.argsort**(x, kind="stable") will use a *stable* sorting algorithm (timsort[11], a variant of mergesort), which guarantees that the ordering permutation is unique: tied elements are placed in the order of appearance.

---

Next, **scipy.stats.rankdata** returns a vector of *ranks*.

```
x = np.array([10, 40, 50, 20, 30])
scipy.stats.rankdata(x)
## array([1., 4., 5., 2., 3.])
```

Element 10 is the smallest ("the winner", say, the quickest racer). Hence, it ranks first. Element 40 is the 4th on the podium. Thus, its rank is 4. And so on.

On a side note, there are many methods in nonparametric statistics (those that do not make any too particular assumptions about the underlying data distribution) that are based on ranks. In particular, in Section 9.1.4, we cover the Spearman correlation coefficient.

**Exercise 5.18** *Consult the manual page of **scipy.stats.rankdata** and test various methods for dealing with ties.*

---

**Note** (**) Readers with some background in discrete mathematics will be interested in the fact that calling **numpy.argsort** on a vector representing a permutation of elements in fact generates its inverse. In particular, **np.argsort(np.argsort(**x, kind="stable"))+1 is equivalent to **scipy.stats.rankdata**(x, method="ordinal").

---

### 5.5.5 Searching for Certain Indexes (Argmin, Argmax)

**numpy.argmin** and **numpy.argmax** return the index at which we can find the smallest and the largest observation in a given vector.

```
x = np.array([10, 30, 50, 40, 20, 50])
np.argmin(x), np.argmax(x)
## (0, 2)
```

If there are tied observations, the smallest index is returned.

---

[11] https://github.com/python/cpython/blob/3.7/Objects/listsort.txt

Using mathematical notation, we can denote the former with:

$$i = \arg\min_j x_j,$$

and read it as: let $i$ be the index of the smallest element in the sequence. Alternatively, it is the *argument of the minimum*, whenever:

$$x_i = \min_j x_j,$$

i.e., the $i$-th element is the smallest.

We can use `numpy.flatnonzero` to fetch the indexes where a logical vector has elements equal to `True` (in Section 11.1.2, we mention that a value equal to zero is treated as the logical `False`, and as `True` in all other cases). For example:

```
np.flatnonzero(x == np.max(x))
## array([2, 5])
```

It is a version of `numpy.argmax` that lets us decide what we would like to do with the tied maxima (there are two).

**Exercise 5.19** *Let x be a vector with possible ties. Write an expression that returns a randomly chosen index pinpointing one of the sample maxima.*

### 5.5.6 Dealing with Round-off and Measurement Errors

Mathematics tells us (the easy proof is left as an exercise to the reader) that a centred version of a given vector $x$, $y = x - \bar{x}$, has the arithmetic mean of 0, i.e., $\bar{y} = 0$.

Of course, it is also true on a computer. Or is it?

```
heights_centred = (heights - np.mean(heights))
np.mean(heights_centred) == 0
## False
```

The average is actually equal to:

```
np.mean(heights_centred)
## 1.3359078775153175e-14
```

which is *almost* zero (0.000000000000134), but not *exactly* zero (it is zero for an engineer, not a mathematician). We saw a similar result when performing standardisation (which involves centring) in Section 5.3.2.

---

**Important**  All floating-point operations on a computer[12] (not only in Python) are performed with *finite* precision of 15–17 decimal digits.

---

We know it from school – for example, some fractions cannot be represented as decimals. When asked to add or multiply them, we will always have to apply some rounding that ultimately leads to precision loss. We know that $1/3+1/3+1/3 = 1$, but using a decimal representation with one fractional digit, we get $0.3 + 0.3 + 0.3 = 0.9$. With two digits, we obtain $0.33 + 0.33 + 0.33 = 0.99$. And so on. This sum will never be equal exactly to 1 when using a finite precision.

Moreover, errors induced in one operation will propagate onto further ones. Most often they cancel out, but in extreme cases, they can lead to undesirable consequences (like for some model matrices in linear regression; see Section 9.2.9).

There is no reason to panic, though. The rule to remember is:

---

**Important**  As the floating-point values are precise up to a few decimal digits, we should refrain from comparing them using the `==` operator, because it tests exact equality.

---

When a comparison is needed, we need to take some error margin into account. Ideally, instead of testing x == y, we should either inspect the *absolute error*:

$$|x - y| \le \varepsilon,$$

or, assuming $y \neq 0$, the *relative error*:

$$\frac{|x - y|}{|y|} \le \varepsilon,$$

where $\varepsilon$ is some small error margin.

For instance, **numpy.allclose**(x, y) checks (by default) if for all corresponding elements in both vectors it holds **numpy.abs**(x-y) <= 1e-8 + 1e-5***numpy.abs**(y), which is a combination of both tests.

```
np.allclose(np.mean(heights_centred), 0)
## True
```

To avoid sorrow surprises, even the testing of inequalities like x >= 0 should rather be performed as, say, x >= 1e-8.

---

**Note**  Our data are often imprecise by nature. When asked about people's heights,

---

[12] Double precision float64 format as defined by the IEEE Standard for Floating-Point Arithmetic (IEEE 754).

rarely will they provide a non-integer answer (assuming they know how tall they are and are not lying about it, but it is a different story). We will most likely get data rounded to 0 decimal digits. In our dataset the precision is a bit higher:

```
heights[:6]   # preview
## array([160.2, 152.7, 161.2, 157.4, 154.6, 144.7])
```

But still, there is an inherent *observational error*. Even if, for example, the mean thereof was computed exactly, the fact that the inputs themselves are not necessarily ideal makes the estimate *approximate* as well. We can only hope that these errors will more or less cancel out in the computations.

---

**Exercise 5.20** *Compute the BMIs of all females in the NHANES study. Determine their arithmetic mean. Compare it to the arithmetic mean computed for BMIs rounded to 1, 2, 3, 4, etc., decimal digits.*

---

**Note** (*)Another problem is related to the fact that floats on a computer use the binary base, not the decimal one. Therefore, some fractional numbers that we *believe* to be representable exactly, require an infinite number of bits. As a consequence, they are subject to rounding.

```
0.1 + 0.1 + 0.1 == 0.3   # obviously
## False
```

This is because `0.1`, `0.1+0.1+0.1`, and `0.3` is literally represented as, respectively:

```
print(f"{0.1:.19f}, {0.1+0.1+0.1:.19f}, and {0.3:.19f}.")
## 0.1000000000000000056, 0.3000000000000000444, and 0.2999999999999999889.
```

---

A good introductory reference to the topic of numerical inaccuracies is [36]; see also [43, 50] for a more comprehensive treatment of numerical analysis.

### 5.5.7 Vectorising Scalar Operations with List Comprehensions

*List comprehensions* of the form [ `expression` **for** `name` **in** `iterable` ] are part of base Python. They allow us to create lists based on transformed versions of individual elements in a given iterable object. Hence, they might work in cases where a task at hand cannot be solved by means of vectorised **numpy** functions.

For example, here is a way to generate the squares of a few positive natural numbers:

```
[ i**2 for i in range(1, 11) ]
## [1, 4, 9, 16, 25, 36, 49, 64, 81, 100]
```

The result can be passed to **numpy.array** to convert it to a vector.

Further, given an example vector:

```
x = np.round(np.random.rand(10)*2-1, 2)
x
## array([ 0.86, -0.37, -0.63, -0.59,  0.14,  0.19,  0.93,  0.31,  0.5 ,
##          0.31])
```

If we wish to filter our all elements that are not greater than 0, we can write:

```
[ e for e in x if e > 0 ]
## [0.86, 0.14, 0.19, 0.93, 0.31, 0.5, 0.31]
```

We can also use the ternary operator of the form x_true **if** cond **else** x_false to return either x_true or x_false depending on the truth value of cond.

```
e = -2
e**0.5 if e >= 0 else (-e)**0.5
## 1.4142135623730951
```

Combined with a list comprehension, we can write, for instance:

```
[ round(e**0.5 if e >= 0 else (-e)**0.5, 2) for e in x ]
## [0.93, 0.61, 0.79, 0.77, 0.37, 0.44, 0.96, 0.56, 0.71, 0.56]
```

This gave the square root of absolute values.

There is also a tool which vectorises a scalar function so that it can be used on **numpy** vectors:

```
def clip01(x):
    """clip to the unit inverval"""
    if x < 0:    return 0
    elif x > 1:  return 1
    else:        return x

clip01s = np.vectorize(clip01)  # returns a function object
clip01s([0.3, -1.2, 0.7, 4, 9])
## array([0.3, 0. , 0.7, 1. , 1. ])
```

In the above cases, it is much better (faster, more readable code) to rely on vectorised **numpy** functions. Still, if the corresponding operations are unavailable (e.g., string processing, reading many files), list comprehensions provide a reasonable replacement therefor.

**Exercise 5.21** *Write equivalent versions of the above expressions using vectorised* **numpy** *functions.*

**Exercise 5.22**  *Write equivalent versions of the above expressions using base Python lists, the* **for** *loop and the* `list.append` *method (start from an empty list that will store the result).*

## 5.6  Exercises

**Exercise 5.23**  *What are some benefits of using a* **numpy** *vector over an ordinary Python list? What are the drawbacks?*

**Exercise 5.24**  *How can we interpret the possibly different values of the arithmetic mean, median, standard deviation, interquartile range, and skewness, when comparing between heights of men and women?*

**Exercise 5.25**  *There is something scientific and magical about* numbers *that make us approach them with some kind of respect. However, taking into account that there are many possible data aggregates, there is a risk that a party may be cherry-picking – reporting the one that portrays the analysed entity in a good or bad light. For instance, reporting the mean instead of the median or vice versa. Is there anything that can be done about it?*

**Exercise 5.26**  *Even though, mathematically speaking, all measures can be computed on all data, it does not mean that it always makes sense to do so. For instance, some distributions will have skewness of 0, but we should not automatically assume that they are delightfully symmetric and bell-shaped (e.g., this can be a bimodal distribution). This is why we always need to visualise our data. Give some examples of datasets and measures where we should be critical of the obtained results.*

**Exercise 5.27**  *Give some examples where simple data preprocessing can drastically change the values of chosen sample aggregates.*

**Exercise 5.28**  *Give the mathematical definitions, use cases, and interpretations of standardisation, normalisation, and min-max scaling.*

**Exercise 5.29**  *How are* `numpy.log` *and* `numpy.exp` *related to each other? How about* `numpy.log` *vs* `numpy.log10`, `numpy.cumsum` *vs* `numpy.diff`, `numpy.min` *vs* `numpy.argmin`, `numpy.sort` *vs* `numpy.argsort`, *and* `scipy.stats.rankdata` *vs* `numpy.argsort`?

**Exercise 5.30**  *What is the difference between* `numpy.trunc`, `numpy.floor`, `numpy.ceil`, *and* `numpy.round`?

**Exercise 5.31**  *What happens when we apply* `` `+` `` *on two vectors of different lengths?*

**Exercise 5.32**  *List the four ways to index a vector.*

**Exercise 5.33**  *What is wrong with the expression* `x[ x >= 0 and x <= 1 ]`, *where x is a numeric vector? How about* `x[ x >= 0 & x <= 1 ]`?

**Exercise 5.34**  *What does it mean that slicing returns a view on existing data?*

**Exercise 5.35**   (\*\*) *Reflect on the famous*[13] *saying:* not everything that can be counted counts, and not everything that counts can be counted.

**Exercise 5.36**   (\*\*) *Being a data scientist can be a frustrating job, especially when you care for some causes. Reflect on:* some things that count can be counted, but we will not count them, because there's no budget for them.

**Exercise 5.37**   (\*\*) *Being a data scientist can be a frustrating job, especially when you care for the truth. Reflect on:* some things that count can be counted, but we will not count them, because some people might be offended or find it unpleasant.

**Exercise 5.38**   (\*\*) *Assume you were to establish your own nation on some island and become the benevolent dictator thereof. How would you* measure *if your people are happy or not? Let us say that you need to come up with 3 quantitative measures (key performance indicators). What would happen if your policy-making was solely focused on optimising those KPIs? How about the same problem but with regard to your company and employees? Think about what can go wrong in other areas of life.*

---

[13] https://quoteinvestigator.com/2010/05/26/everything-counts-einstein/

# 6

## *Continuous Probability Distributions*

Each successful data analyst will deal with hundreds or thousands of datasets in their lifetime. In the long run, at some level, most of them will be deemed *boring*. This is because only a few common patterns will be occurring over and over again.

In particular, the previously mentioned bell-shapedness and right-skewness are quite prevalent in the so-called real world. Surprisingly, however, this is exactly when things become scientific and interesting – allowing us to study various phenomena at an appropriate level of generality.

Mathematically, such idealised patterns in the histogram shapes can be formalised using the notion of a *probability density function* (PDF) of a *continuous, real-valued random variable*.

Intuitively[1], a PDF is a smooth curve that would arise if we drew a histogram for the entire *population* (e.g., all women living currently on Earth and beyond or otherwise an extremely large data sample obtained by independently querying the same underlying data generating process) in such a way that the total area of all the bars is equal to 1 and the bin sizes are very small.

As stated at the beginning, we do not intend this to be a course in probability theory and mathematical statistics. Rather, it precedes and motivates them (e.g., [19, 33, 35]). Therefore, our definitions are out of necessity simplified so that they are digestible. For the purpose of our illustrations, we will consider the following characterisation.

---

**Important**  (*) We call an integrable function $f : \mathbb{R} \to \mathbb{R}$ a *probability density function* if $f(x) \geq 0$ for all $x$ and $\int_{-\infty}^{\infty} f(x)\,dx = 1$, i.e., it is nonnegative and normalised in such a way that the total area under the whole curve is 1.

For any $a < b$, we treat the area under the fragment of the $f(x)$ curve for $x$ between $a$ and $b$, i.e., $\int_{a}^{b} f(x)\,dx$, as the probability of the underlying real-valued random variable's (theoretical data generating process') falling into the $[a, b]$ interval.

---

Some distributions appear more frequently than others and appear to fit empirical data or parts thereof particularly well; compare [23]. In this chapter, we review a few noteworthy probability distributions: the normal, log-normal, Pareto, and uniform

---
[1] (*) This intuition is of course theoretically grounded and is based on the asymptotic behaviour of the histograms as the estimators of the underlying probability density function, see, e.g., [24] and the many references therein.

families (we will also mention the chi-squared, Kolmogorov, and exponential ones in this course).

## 6.1   Normal Distribution

A *normal (Gaussian) distribution* has a prototypical, nicely symmetric, bell-shaped density. It is described by two parameters: $\mu \in \mathbb{R}$ (the expected value, at which the PDF is centred) and $\sigma > 0$ (the standard deviation, saying how much the distribution is dispersed around $\mu$); compare Figure 6.1.

The probability density function of $N(\mu, \sigma)$ is given by:

$$f(x) = \frac{1}{\sqrt{2\pi\sigma^2}} \exp\left(-\frac{(x - \mu)^2}{2\sigma^2}\right).$$

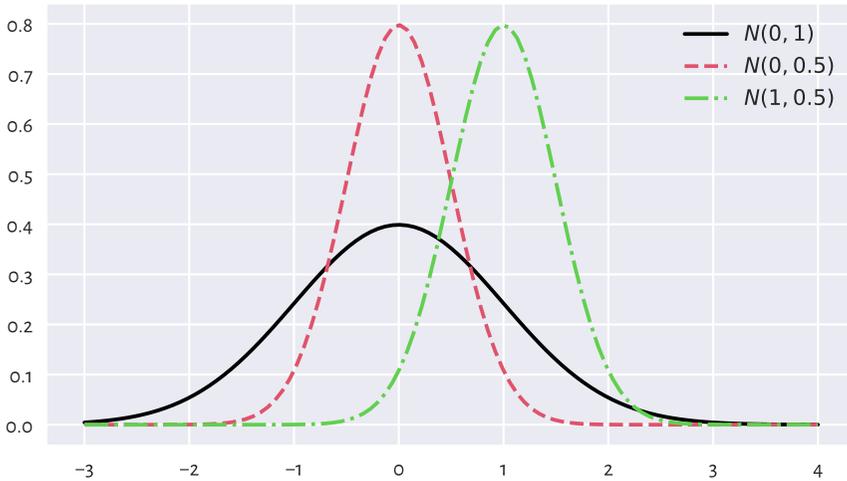

Figure 6.1: The probability density functions of some normal distributions $N(\mu, \sigma)$; note that $\mu$ is responsible for shifting and $\sigma$ affects scaling/stretching of the probability mass

### 6.1.1   Estimating Parameters

A course in statistics (which, again, this one is not, we are merely making an illustration here), may tell us that the sample arithmetic mean $\bar{x}$ and standard deviation $s$ are natural, statistically well-behaving *estimators* of the said parameters: if all obser-vations would really be drawn independently from N($\mu$, $\sigma$) each, then we *expect* $\bar{x}$ and $s$ be equal to, more or less, $\mu$ and $\sigma$ (the larger the sample size, the smaller the error).

Recall the `heights` (females from the NHANES study) dataset and its bell-shaped histogram in Figure 4.2.

```python
heights = np.loadtxt("https://raw.githubusercontent.com/gagolews/" +
    "teaching-data/master/marek/nhanes_adult_female_height_2020.txt")
n = len(heights)
n
## 4221
```

Let us estimate the said parameters for this sample:

```python
mu = np.mean(heights)
sigma = np.std(heights, ddof=1)
mu, sigma
## (160.13679222932953, 7.062858532891359)
```

Mathematically, we will denote these two with $\hat{\mu}$ and $\hat{\sigma}$ (mu and sigma with a hat) to emphasise that they are merely guesstimates[2] of the unknown respective parameters $\mu$ and $\sigma$. On a side note, we use `ddof=1`, because in this context this estimator has slightly better statistical properties.

Let us draw the fitted density function (i.e., the PDF of N(160.1, 7.06) which we can compute using **scipy.stats.norm.pdf**), on top of the histogram; see Figure 6.2. We pass `stat="density"` to **seaborn.histplot** so that the histogram bars are normalised (i.e., the total area of these rectangles sums to 1).

```python
sns.histplot(heights, stat="density", color="lightgray")
x = np.linspace(np.min(heights), np.max(heights), 1000)
plt.plot(x, scipy.stats.norm.pdf(x, mu, sigma), "r--",
    label=f"PDF of N({mu:.1f}, {sigma:.2f})")
plt.legend()
plt.show()
```

At first glance, this is a genuinely nice match. Before proceeding with an overview of the ways to assess the goodness-of-fit more rigorously, we should praise the potential benefits of having an idealised *model* of our dataset at our disposal.

---

[2] (*) It might be the case that we will have to obtain the estimates of the probability distribution's parameters by numerical optimisation, for example, by minimising $\mathcal{L}(\mu, \sigma) = \sum_{i=1}^{n} \left( \frac{(x_i - \mu)^2}{\sigma^2} + \log \sigma^2 \right)$ with respect to $\mu$ and $\sigma$ (corresponding to the objective function in the maximum likelihood estimation problem for the normal distribution family). In our case, however, we are lucky; there exist open-form formulae expressing the solution to the above, exactly in the form of the sample mean and standard deviation. For other distributions, things can get a little trickier, though. Furthermore, sometimes we will have many options for point estimators to choose from, which might be more suitable if data are not of top quality (e.g., contain outliers). For instance, in the normal model, it can be shown that we can also estimate $\mu$ and $\sigma$ via the sample median and IQR/1.349.

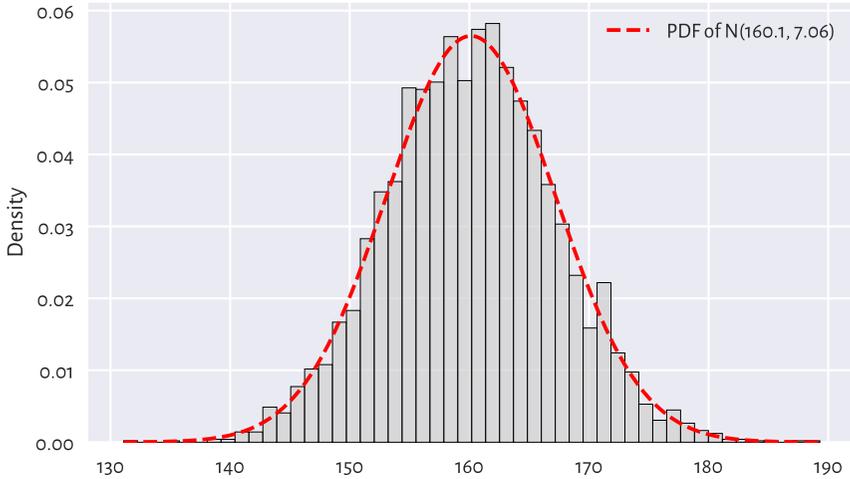

Figure 6.2: A histogram and the probability density function of the fitted normal distribution for the `heights` dataset

### 6.1.2 Data Models Are Useful

*If* (provided that, assuming that, on condition that) our sample is a realisation of independent random variables following a given distribution, or a data analyst judges that such an approximation might be justified or beneficial, then we have a set of many numbers *reduced* to merely a few parameters.

In the above case, we might want to risk the statement that data follow the normal distribution (assumption 1) with parameters $\mu = 160.1$ and $\sigma = 7.06$ (assumption 2). Still, the choice of the distribution family is one thing, and the way we estimate the underlying parameters (in our case, we use $\hat{\mu}$ and $\hat{\sigma}$) is another.

This not only saves storage space and computational time, but also – based on what we can learn from a course in probability and statistics (by appropriately integrating the PDF) – we can imply facts such as for normally distributed data:

- ca. 68% of (i.e., a *majority*) women are $\mu \pm \sigma$ tall (the 1σ rule),

- ca. 95% of (i.e., *most typical*) women are $\mu \pm 2\sigma$ tall (the 2σ rule),

- ca. 99.7% of (i.e., *almost all*) women are $\mu \pm 3\sigma$ tall (the 3σ rule).

Also, if we knew that the distribution of heights of men is also normal with some other parameters, we could be able to make some comparisons between the two samples. For example, we could compute the probability that a woman randomly selected from the crowd is taller than a male passer-by.

Furthermore, there is a range of *parametric* (assuming some distribution family) stat-istical methods that could *additionally* be used if we assumed the data normality, e.g., the *t*-test to compare the expected values.

**Exercise 6.1** *How different manufacturing industries (e.g., clothing) can make use of such models? Are simplifications necessary when dealing with complexity? What are the alternatives?*

---

**Important** We should always verify the assumptions of a model that we wish to apply in practice. In particular, we will soon note that incomes are not normally distributed. Therefore, we must not refer to the above $2\sigma$ or $3\sigma$ rule in their case. A cow neither barks nor can it serve as a screwdriver. Period.

---

## 6.2 Assessing Goodness-of-Fit

### 6.2.1 Comparing Cumulative Distribution Functions

In the previous subsection, we were comparing densities and histograms. It turns out that there is a better way of assessing the extent to which a sample deviates from a hypothesised distribution. Namely, we can measure the discrepancy between some theoretical *cumulative distribution function* (CDF) and the empirical one ($\hat{F}_n$; see Section 4.3.8).

---

**Important** If $f$ is a PDF, then the corresponding theoretical CDF is defined as $F(x) = \int_{-\infty}^{x} f(t)\, dt$, i.e., the probability of the underlying random variable's taking a value less than or equal to $x$.

By definition[3], each CDF takes values in the unit interval ($[0, 1]$) and is nondecreasing.

---

For the normal distribution family, the values of the theoretical CDF can be computed by calling **scipy.stats.norm.cdf**; see Figure 6.3.

```
x = np.linspace(np.min(heights), np.max(heights), 1001)
probs = scipy.stats.norm.cdf(x, mu, sigma)  # sample the CDF at many points
plt.plot(x, probs, "r--", label=f"CDF of N({mu:.1f}, {sigma:.2f})")
heights_sorted = np.sort(heights)
plt.plot(heights_sorted, np.arange(1, n+1)/n,
    drawstyle="steps-post", label="Empirical CDF")
plt.xlabel("$x$")
```

*(continues on next page)*

---

[3] The probability distribution of any real-valued random variable $X$ can be uniquely defined by means of a nondecreasing, right (upward) continuous function $F : \mathbb{R} \rightarrow [0, 1]$ such that $\lim_{x \to -\infty} F(x) = 0$ and $\lim_{x \to \infty} F(x) = 1$, in which case $\Pr(X \leq x) = F(x)$. The probability density function only exists for continuous random variables and is defined as the derivative of $F$.

```
plt.ylabel("Prob(height $\\leq$ x)")
plt.legend()
plt.show()
```

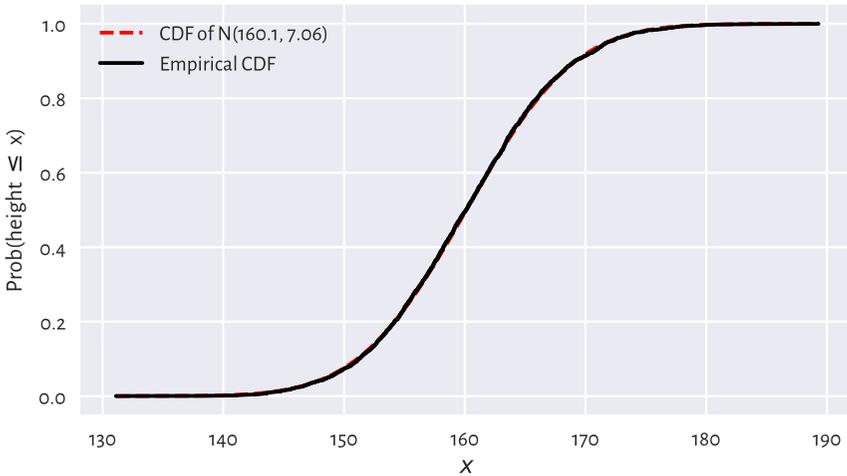

Figure 6.3: The empirical CDF and the fitted normal CDF for the `heights` dataset; the fit is superb

This *looks* like a superb match.

**Example 6.2** $F(b) - F(a) = \int_a^b f(t)\,dt$ is the probability of generating a value in the interval $[a, b]$.

*Let us empirically verify the $3\sigma$ rule:*

```
F = lambda x: scipy.stats.norm.cdf(x, mu, sigma)
F(mu+3*sigma) - F(mu-3*sigma)
## 0.9973002039367398
```

*Indeed, almost all observations are within $[\mu - 3\sigma, \mu + 3\sigma]$, if data are normally distributed.*

---

**Note**  A common way to summarise the discrepancy between the empirical and a given theoretical CDF is by computing the greatest absolute deviation:

$$\hat{D}_n = \sup_{t \in \mathbb{R}} |\hat{F}_n(t) - F(t)|,$$

where the supremum is a continuous version of the maximum.

---

It holds:

$$\hat{D}_n = \max\left\{\max_{k=1,\dots,n}\left\{\left|\frac{k-1}{n} - F(x_{(k)})\right|\right\}, \max_{k=1,\dots,n}\left\{\left|\frac{k}{n} - F(x_{(k)})\right|\right\}\right\},$$

i.e., $F$ needs to be probed only at the $n$ points from the sorted input sample.

```python
def compute_Dn(x, F):  # equivalent to scipy.stats.kstest(x, F)[0]
    Fx = F(np.sort(x))
    n = len(x)
    k = np.arange(1, n+1)  # 1, 2, ..., n
    Dn1 = np.max(np.abs((k-1)/n - Fx))
    Dn2 = np.max(np.abs(k/n - Fx))
    return max(Dn1, Dn2)

Dn = compute_Dn(heights, F)
Dn
## 0.010470976524201148
```

If the difference is *sufficiently[4] small*, then we can assume that a normal model describes data quite well. This is indeed the case here: we may estimate the probability of someone being as tall as any given height with an error less than about 1.05%.

## 6.2.2  Comparing Quantiles

A *Q-Q plot* (quantile-quantile or probability plot) is another graphical method for comparing two distributions. This time, instead of working with a cumulative distribution function $F$, we will be dealing with its (generalised) inverse, i.e., the quantile function $Q$.

Given a CDF $F$, the corresponding *quantile function* is defined for any $p \in (0, 1)$ as:

$$Q(p) = \inf\{x : F(x) \ge p\},$$

i.e., the smallest $x$ such that the probability of drawing a value not greater than $x$ is at least $p$.

---

**Important**  If a CDF $F$ is continuous, and this is the assumption in the current chapter, then $Q$ is exactly its inverse, i.e., it holds $Q(p) = F^{-1}(p)$ for all $p \in (0, 1)$; compare Figure 6.4.

---

The theoretical quantiles can be generated by the **scipy.stats.norm.ppf** function. Here, *ppf* stands for the percent point function which is another (yet quite esoteric) name for the above $Q$.

---

[4] The larger the sample size, the less tolerant we should be regarding the size of this disparity; see Section 6.2.3.

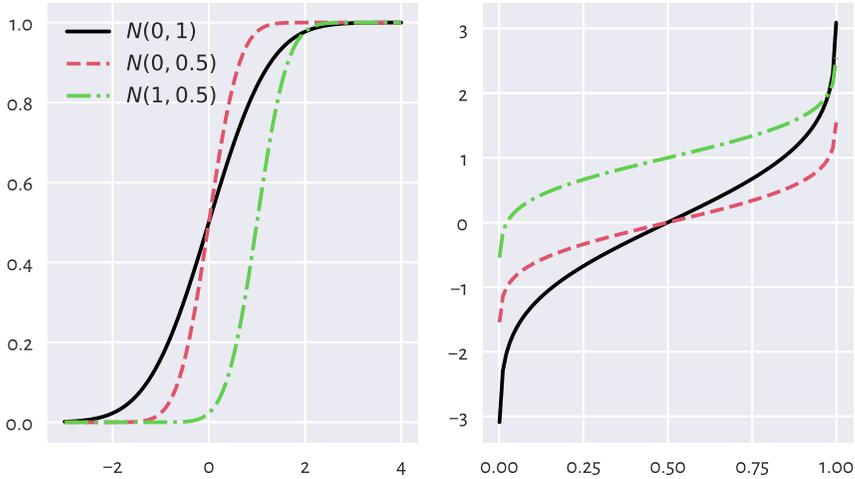

Figure 6.4: The cumulative distribution functions (left) and the quantile functions (being the inverse of the CDF; right) of some normal distributions

For instance, in our N(160.1, 7.06)-distributed `heights` dataset, $Q(0.9)$ is the height not exceeded by 90% of the female population. In other words, only 10% of American women are taller than:

```
scipy.stats.norm.ppf(0.9, mu, sigma)
## 169.18820963937648
```

A Q-Q plot draws a version of sample quantiles as a function of the corresponding theoretical quantiles. The sample quantiles that we introduced in Section 5.1.1 are natural estimators of the theoretical quantile function. However, we also mentioned that there are quite a few possible definitions thereof in the literature; compare [47].

For simplicity, instead of using `numpy.quantile`, we will assume that the $\frac{i}{n+1}$-quantile[5] is equal to $x_{(i)}$, i.e., the $i$-th smallest value in a given sample $(x_1, x_2, \dots, x_n)$ and consider only $i = 1, 2, \dots, n$.

Our simplified setting avoids the problem which arises when the 0- or 1-quantiles of the theoretical distribution, i.e., $Q(0)$ or $Q(1)$, are infinite (and this is the case for the normal distribution family).

```
def qq_plot(x, Q):
    """
```

*(continues on next page)*

---

[5] (*) `scipy.stats.probplot` uses a slightly different definition (there are many other ones in common use).

```
    Draws a Q-Q plot, given:
    * x - a data sample (vector)
    * Q - a theoretical quantile function
    """
    n = len(x)
    q = np.arange(1, n+1)/(n+1)      # 1/(n+1), 2/(n+2), ..., n/(n+1)
    x_sorted = np.sort(x)            # sample quantiles
    quantiles = Q(q)                 # theoretical quantiles
    plt.plot(quantiles, x_sorted, "o")
    plt.axline((x_sorted[n//2], x_sorted[n//2]), slope=1,
        linestyle=":", color="gray")  # identity line
```

Figure 6.5 depicts the Q-Q plot for our example dataset.

```
qq_plot(heights, lambda q: scipy.stats.norm.ppf(q, mu, sigma))
plt.xlabel(f"Quantiles of N({mu:.1f}, {sigma:.2f})")
plt.ylabel("Sample quantiles")
plt.show()
```

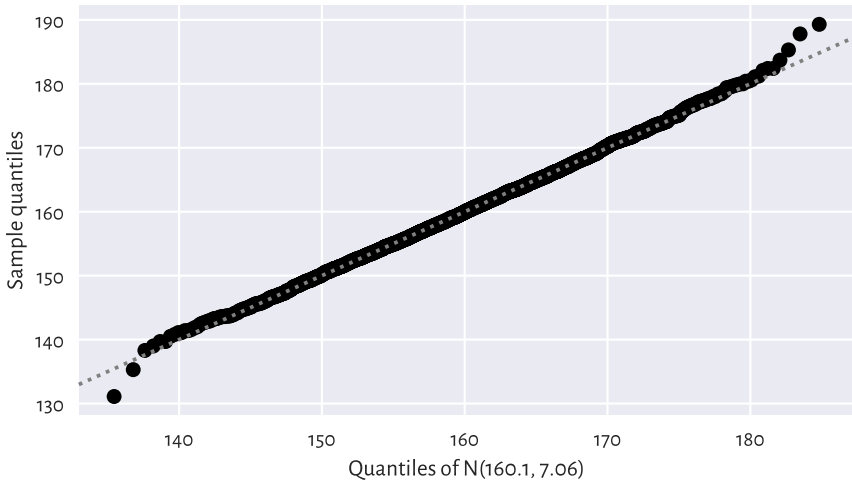

Figure 6.5: Q-Q plot for the `heights` dataset; it's a nice fit

Ideally, the points are expected to be arranged on the $y = x$ line (which was added for readability). This would happen if the sample quantiles matched the theoretical ones perfectly. In our case, there are small discrepancies[6] in the tails (e.g., the smallest observation was slightly smaller than expected, and the largest one was larger than expected), although it is quite a *normal* behaviour for small samples and certain distribution families. Still, we can say that we observe a very good fit.

---

[6] (*) We can quantify (informally) the goodness of fit by using the Pearson linear correlation coefficient; see Section 9.1.1.

### 6.2.3  Kolmogorov–Smirnov Test (*)

To be scientific, we should yearn for some more formal method that will enable us to test the null hypothesis stating that a given empirical distribution $\hat{F}_n$ does not differ *significantly* from the theoretical continuous CDF $F$:

$$\begin{cases} H_0: & \hat{F}_n = F \quad \text{(null hypothesis)} \\ H_1: & \hat{F}_n \neq F \quad \text{(two-sided alternative)} \end{cases}$$

The popular goodness-of-fit test by Kolmogorov and Smirnov can give us a conservative interval of the acceptable values of $\hat{D}_n$ (again: the largest deviation between the empirical and theoretical CDF) as a function of $n$ (within the framework of frequentist hypothesis testing).

Namely, if the *test statistic* $\hat{D}_n$ is smaller than some *critical value* $K_n$, then we shall deem the difference insignificant. This is to take into account the fact that reality might deviate from the ideal. In Section 6.4.4, we mention that even for samples that truly come from a hypothesised distribution, there is some inherent variability. We need to be somewhat tolerant.

A good textbook in statistics will tell us (and prove) that, under the assumption that $\hat{F}_n$ is the ECDF of a sample of $n$ independent variables *really* generated from a continuous CDF $F$, the random variable $\hat{D}_n = \sup_{t \in \mathbb{R}} |\hat{F}_n(t) - F(t)|$ follows the Kolmogorov distribution with parameter $n$ (available via `scipy.stats.kstwo`).

In other words, if we generate many samples of length $n$ from $F$, and compute $\hat{D}_n$s for each of them, we expect it to be distributed like in Figure 6.6.

The choice $K_n$ involves a trade-off between our desire to:

- accept the null hypothesis when it is true (data *really* come from $F$), and

- reject it when it is false (data follow some other distribution, i.e., the difference is significant enough).

These two needs are, unfortunately, mutually exclusive.

In practice, we assume some fixed upper bound (*significance level*) for making the former kind of mistake, the so-called *type-I error*. A nicely conservative (in a good way[7]) value that we suggest employing is $\alpha = 0.001 = 0.1\%$, i.e., only 1 out of 1,000 samples that really come from $F$ will be rejected as not coming from $F$.

Such a $K_n$ may be determined by considering the inverse of the CDF of the Kolmogorov distribution, $\Xi_n$. Namely, $K_n = \Xi_n^{-1}(1 - \alpha)$:

---

[7] See Section 12.2.6 for more details.

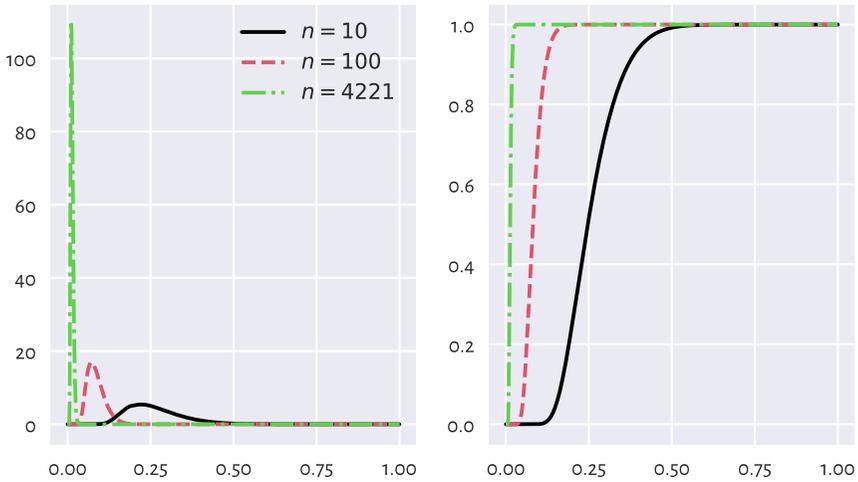

Figure 6.6: Densities (left) and cumulative distribution functions (right) of some Kolmogorov distributions; the greater the sample size, the smaller the acceptable deviations between the theoretical and empirical CDFs

```
alpha = 0.001  # significance level
scipy.stats.kstwo.ppf(1-alpha, n)
## 0.029964456376393188
```

In our case $\hat{D}_n < K_n$, because $0.01047 < 0.02996$. We conclude that our empirical (`heights`) distribution does not differ significantly (at significance level 0.1%) from the assumed one, i.e., $N(160.1, 7.06)$. In other words, we do not have enough evidence against the statement that data are normally distributed. It is the presumption of innocence: they are normal enough.

We will go back to this discussion in Section 6.4.4 and Section 12.2.6.

## 6.3 Other Noteworthy Distributions

### 6.3.1 Log-Normal Distribution

We say that a sample is *log-normally distributed*, if its logarithm is normally distributed.

In particular, it is sometimes observed that the income of most[8] individuals is distrib-uted, at least approximately, log-normally. Let us investigate whether this is the case for UK taxpayers.

---

[8] Except for the few richest, who are interesting on their own; see Section 6.3.2 where we discuss the Pareto distribution.

```
income = np.loadtxt("https://raw.githubusercontent.com/gagolews/" +
    "teaching-data/master/marek/uk_income_simulated_2020.txt")
```

The plotting of the histogram of the logarithm of income is left as an exercise (we can pass `log_scale=True` to **seaborn.histplot**; we will plot it soon anyway in a different way). We proceed directly with the fitting of a log-normal model, LN($\mu$, $\sigma$). The fitting process is similar to the normal case, but this time we determine the mean and standard deviation based on the logarithms of data:

```
lmu = np.mean(np.log(income))
lsigma = np.std(np.log(income), ddof=1)
lmu, lsigma
## (10.314409794364623, 0.581658519780816)
```

Figure 6.7 depicts the fitted probability density function together with the histograms on the log- and original scale. When creating this plot, there are two pitfalls, though. Firstly, **scipy.stats.lognorm** encodes the distribution via the parameter $s$ equal to $\sigma$ and scale equal to $e^\mu$. Computing the PDF at different points is done as follows:

```
x  = np.linspace(np.min(income), np.max(income), 101)
fx = scipy.stats.lognorm.pdf(x, s=lsigma, scale=np.exp(lmu))
```

Second, passing both `log_scale=True` and `stat="density"` to **seaborn.histplot** does not normalise the values on the y-axis correctly. To make the histogram on the log-scale comparable with the true density, we need to turn on the log-scale and pass the manually generated bins that are equidistant on the logarithmic scale (via **numpy.geomspace**).

```
b = np.geomspace(np.min(income), np.max(income), 30)
```

And now:

```
plt.subplot(1, 2, 1)
sns.histplot(income, stat="density", bins=b, color="lightgray")  # own bins!
plt.xscale("log")  # log-scale on the x-axis
plt.plot(x, fx, "r--")
plt.subplot(1, 2, 2)
sns.histplot(income, stat="density", color="lightgray")
plt.plot(x, fx, "r--", label=f"PDF of LN({lmu:.1f}, {lsigma:.2f})")
plt.legend()
plt.show()
```

Overall, this fit is not too bad. Nonetheless, we are only dealing with a sample of 1,000

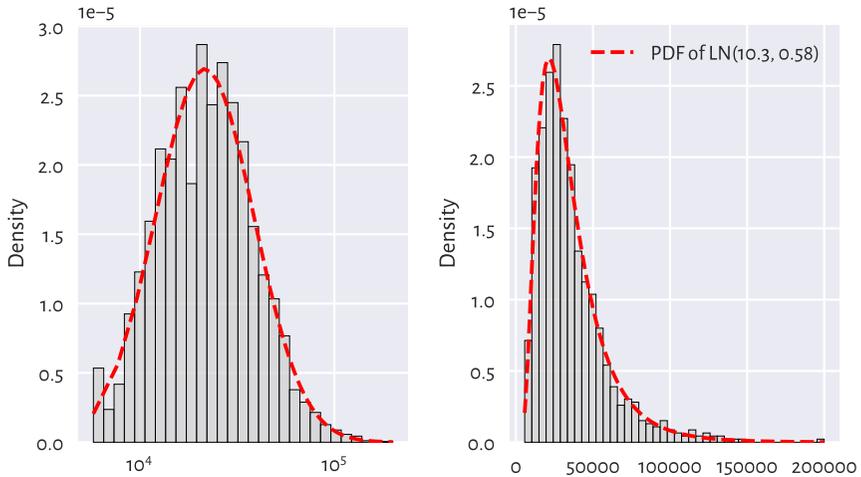

Figure 6.7: A histogram and the probability density function of the fitted log-normal distribution for the `income` dataset, on log- (left) and original (right) scale

households; the original UK Office of National Statistics data[9] could tell us more about the quality of this model in general, but it is beyond the scope of our simple exercise.

Furthermore, Figure 6.8 gives the quantile-quantile plot on a double logarithmic scale for the above log-normal model. Additionally, we (empirically) verify the hypothesis of normality (using a "normal" normal distribution, not its "log" version).

```python
plt.subplot(1, 2, 1)
qq_plot(  # see above for the definition
    income,
    lambda q: scipy.stats.lognorm.ppf(q, s=lsigma, scale=np.exp(lmu))
)
plt.xlabel(f"Quantiles of LN({lmu:.1f}, {lsigma:.2f})")
plt.ylabel("Sample quantiles")
plt.xscale("log")
plt.yscale("log")

plt.subplot(1, 2, 2)
mu = np.mean(income)
sigma = np.std(income, ddof=1)
qq_plot(income, lambda q: scipy.stats.norm.ppf(q, mu, sigma))
plt.xlabel(f"Quantiles of N({mu:.1f}, {sigma:.2f})")
```

*(continues on next page)*

---

[9] https://www.ons.gov.uk/peoplepopulationandcommunity/personalandhouseholdfinances/incomeandwealth/bulletins/householddisposableincomeandinequality/financialyear2020

```
plt.show()
```

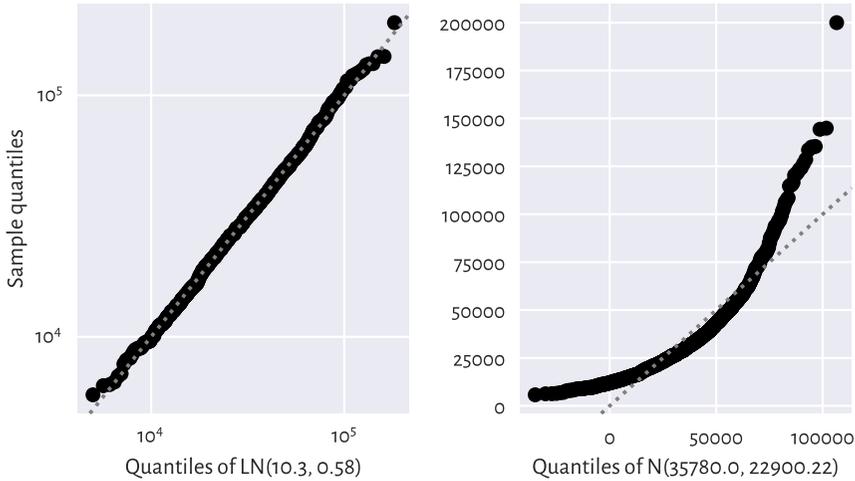

Figure 6.8: Q-Q plots for the `income` dataset vs a fitted log-normal (good fit; left) and normal (bad fit; right) distribution

**Exercise 6.3** *Graphically compare the empirical CDF for `income` and the theoretical CDF of LN(10.3, 0.58).*

**Exercise 6.4** (*) *Perform the Kolmogorov–Smirnov goodness-of-fit test as in Section 6.2.3, to verify that the hypothesis of log-normality is not rejected at the $\alpha = 0.001$ significance level. At the same time, the income distribution significantly differs from a normal one.*

The hypothesis that our data follow a normal distribution is most likely false. On the other hand, the log-normal model, might be quite adequate. It again reduced the whole dataset to merely two numbers, μ and σ, based on which (and probability theory), we may deduce that:

- the expected average (mean) income is $e^{\mu+\sigma^2/2}$,

- median is $e^{\mu}$,

- most probable one (mode) in $e^{\mu-\sigma^2}$,

etc.

**Note** Recall again that for skewed distributions such as this one, reporting the mean might be misleading. This is why *most* people get angry when they read the news about the prospering economy ("yeah, we'd like to see that kind of money in our pockets"). Hence, it is not only μ that matters, it is also σ that quantifies the discrepancy between the rich and the poor (too much inequality is bad, but also too much uniformity is to be avoided).

For a normal distribution, the situation is vastly different, because the mean, the median, and the most probable outcomes tend to be the same – the distribution is symmetric around μ.

**Exercise 6.5**  *What is the fraction of people with earnings below the mean in our LN(10.3, 0.58) model? Hint: use* `scipy.stats.lognorm.cdf` *to get the answer.*

### 6.3.2 Pareto Distribution

Consider again the dataset on the populations of the US cities in the 2000 US Census:

```
cities = np.loadtxt("https://raw.githubusercontent.com/gagolews/" +
    "teaching-data/master/other/us_cities_2000.txt")
len(cities), sum(cities)  # number of cities, total population
## (19447, 175062893.0)
```

Figure 6.9 gives the histogram of the city sizes with the populations on the log-scale. It kind of looks like a log-normal distribution again, which the reader can inspect themself when they are feeling playful.

```
sns.histplot(cities, bins=20, log_scale=True, color="lightgray")
plt.show()
```

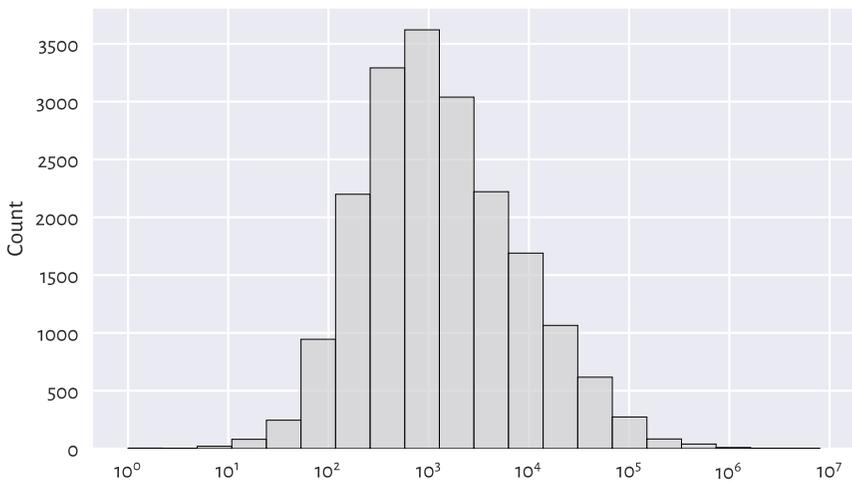

Figure 6.9: Histogram of the unabridged `cities` dataset; note the log-scale on the x-axis

This time, however, will be interested in not what is *typical*, but what is in some sense *anomalous* or *extreme*. Let us look again at the *truncated* version of the city size distribution by considering the cities with 10,000 or more inhabitants (i.e., we will only study the right tail of the original data, just like in Section 4.3.7).

```
s = 10_000
large_cities = cities[cities >= s]
len(large_cities), sum(large_cities)  # number of cities, total population
## (2696, 146199374.0)
```

Plotting the above on a double logarithmic scale can be performed by passing `log_scale=(True, True)` to **seaborn.histplot**, which is left as an exercise. Anyway, doing so will lead to a similar picture as in Figure 6.10. This reveals something remarkable: the bar tops on the double log-scale are arranged more or less in a straight line. There are many datasets that exhibit this behaviour; we say that they follow a *power law* (power in the arithmetic sense, not social one); see [12, 60] for discussion.

Let us introduce the *Pareto distribution* family which has a prototypical power law-like density. It is identified by two parameters:

- the (what **scipy** calls it) scale parameter $s > 0$ is equal to the shift from 0,

- the shape parameter, $\alpha > 0$, controls the slope of the said line on the double log-scale.

The probability density function of P($\alpha$, s) is given for $x \geq s$ by:

$$f(x) = \frac{\alpha s^{\alpha}}{x^{\alpha+1}},$$

and $f(x) = 0$ otherwise.

$s$ is usually taken as the sample minimum (i.e., 10,000 in our case). $\alpha$ can be estimated through the reciprocal of the mean of the scaled logarithms of our observations:

```
alpha = 1/np.mean(np.log(large_cities/s))
alpha
## 0.9496171695997675
```

We noticed that comparing the theoretical densities and an empirical histogram on a log-scale is quite problematic for **seaborn**. Therefore, to generate what we see in Figure 6.10, we must apply the logarithmic binning manually again.

```
b = np.geomspace(s, np.max(large_cities), 21)  # bin boundaries
sns.histplot(large_cities, bins=b, stat="density", color="lightgray")
plt.plot(b, scipy.stats.pareto.pdf(b, alpha, scale=s), "r--",
    label=f"PDF of P({alpha:.3f}, {s})")
plt.xscale("log")
plt.yscale("log")
```

*(continues on next page)*

```
plt.legend()
plt.show()
```

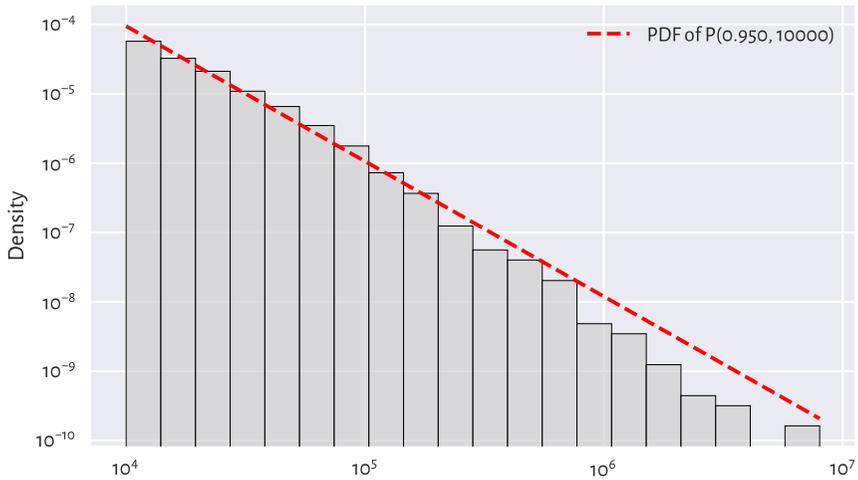

Figure 6.10: Histogram of the `large_cities` dataset and the fitted density on a double log-scale

Figure 6.11 gives the corresponding Q-Q plot on a double logarithmic scale.

```
qq_plot(  # defined above
    large_cities,
    lambda q: scipy.stats.pareto.ppf(q, alpha, scale=s)
)
plt.xlabel(f"Quantiles of P({alpha:.3f}, {s})")
plt.ylabel("Sample quantiles")
plt.xscale("log")
plt.yscale("log")
plt.show()
```

We see that the populations of the largest cities are overestimated. The model could be better, but the cities are still growing, right?

**Example 6.6**  *(\*) It might also be interesting to see how well we can predict the probability of a randomly selected city being at least a given size. Let us denote with $S(x) = 1 − F(x)$ the complementary cumulative distribution function (CCDF; sometimes referred to as the survival function), and with $\hat{S}_n(x) = 1 − \hat{F}_n(x)$ its empirical version. Figure 6.12 compares the empirical and the fitted CCDFs with probabilities on the linear- and log-scale.*

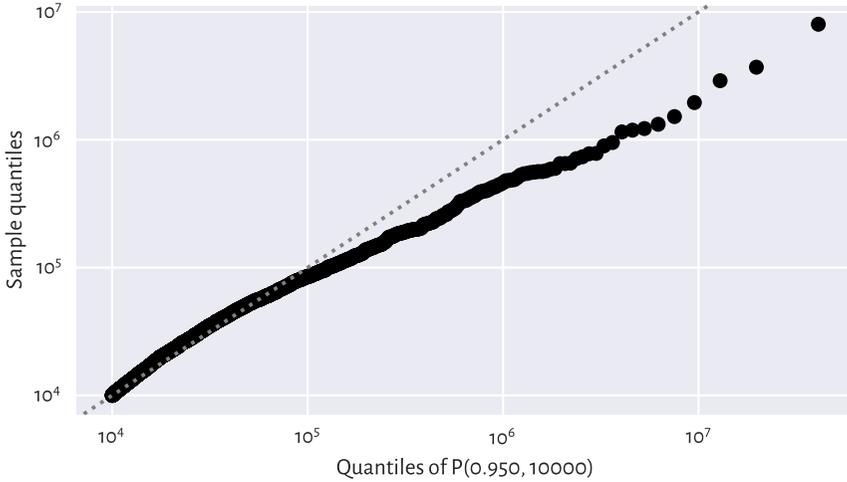

Figure 6.11: Q-Q plot for the `large_cites` dataset vs the fitted Paretian model

```
x = np.geomspace(np.min(large_cities), np.max(large_cities), 1001)
probs = scipy.stats.pareto.cdf(x, alpha, scale=s)
n = len(large_cities)
for i in [1, 2]:
    plt.subplot(1, 2, i)
    plt.plot(x, 1-probs, "r--", label=f"CCDF of P({alpha:.3f}, {s})")
    plt.plot(np.sort(large_cities), 1-np.arange(1, n+1)/n,
        drawstyle="steps-post", label="Empirical CCDF")
    plt.xlabel("$x$")
    plt.xscale("log")
    plt.yscale(["linear", "log"][i-1])
    if i == 1:
        plt.ylabel("Prob(city size > x)")
        plt.legend()
plt.show()
```

In terms of the maximal absolute distance between the two functions, $\hat{D}_n$, from the left plot we see that the fit looks fairly good (let us stress that the log-scale overemphasises the relatively minor differences in the right tail and should not be used for judging the value of $\hat{D}_n$).

That the Kolmogorov–Smirnov goodness-of-fit test rejects the hypothesis of Paretianity (at a significance level 0.1%) is left as an exercise to the reader.

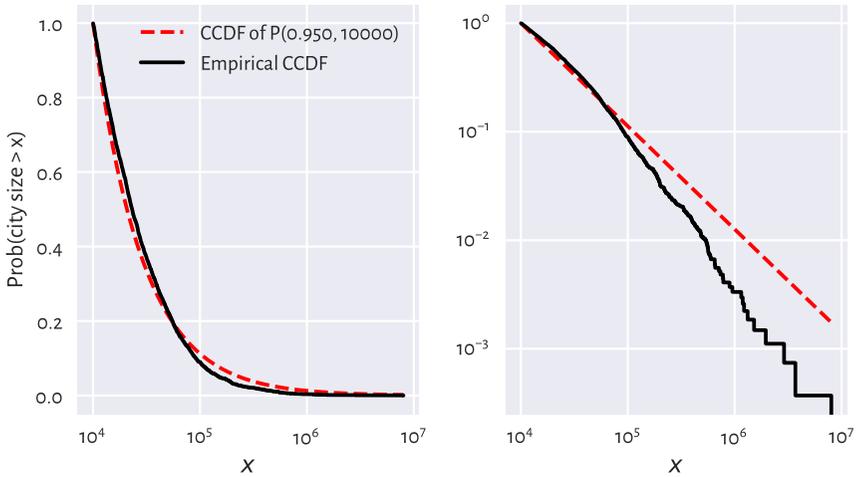

Figure 6.12: Empirical and theoretical complementary cumulative distribution functions for the `large_cities` dataset with probabilities on the linear- (left) and log-scale (right) and city sizes on the log-scale

### 6.3.3 Uniform Distribution

Consider the Polish *Lotto* lottery, where six numbered balls $\{1, 2, \dots, 49\}$ are drawn without replacement from an urn. We have a dataset that summarises the number of times each ball has been drawn in all the games in the period 1957–2016.

```
lotto = np.loadtxt("https://raw.githubusercontent.com/gagolews/" +
    "teaching-data/master/marek/lotto_table.txt")
lotto
## array([720., 720., 714., 752., 719., 753., 701., 692., 716., 694., 716.,
##        668., 749., 713., 723., 693., 777., 747., 728., 734., 762., 729.,
##        695., 761., 735., 719., 754., 741., 750., 701., 744., 729., 716.,
##        768., 715., 735., 725., 741., 697., 713., 711., 744., 652., 683.,
##        744., 714., 674., 654., 681.])
```

Each event seems to occur more or less with the same probability. Of course, the numbers on the balls are integer, but in our idealised scenario, we may try modelling this dataset using a continuous *uniform distribution*, which yields arbitrary real numbers on a given interval $(a, b)$, i.e., between some $a$ and $b$. We denote such a distribution with $\mathrm{U}(a, b)$. It has the probability density function given for $x \in (a, b)$ by:

$$f(x) = \frac{1}{b - a},$$

and $f(x) = 0$ otherwise.

Notice that **scipy.stats.uniform** uses parameters `a` and `scale` equal to $b - a$ instead.

In our case, it makes sense to set $a = 1$ and $b = 50$ and interpret an outcome like 49.1253 as representing the 49th ball (compare the notion of the floor function, $\lfloor x \rfloor$).

```
x = np.linspace(1, 50, 1001)
plt.bar(np.arange(1, 50), width=1, height=lotto/np.sum(lotto),
    color="lightgray", edgecolor="black", alpha=0.8, align="edge")
plt.plot(x, scipy.stats.uniform.pdf(x, 1, scale=49), "r--",
    label="PDF of U(1, 50)")
plt.ylim(0, 0.025)
plt.legend()
plt.show()
```

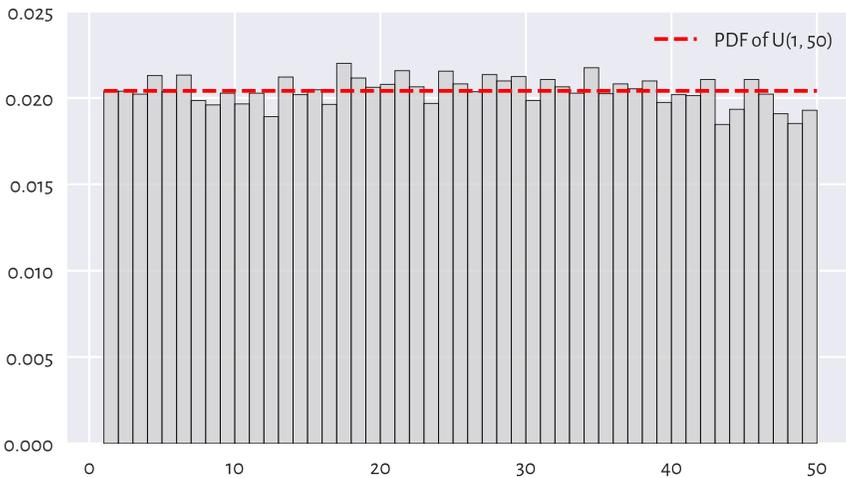

Figure 6.13: Histogram of the `lotto` dataset

Visually, see Figure 6.13, this model makes much sense, but again, some more rigorous statistical testing would be required to determine if someone has not been tampering with the lottery results, i.e., if data does not deviate from the uniform distribution significantly.

Unfortunately, we cannot use the Kolmogorov–Smirnov test in the version defined above, because data are not continuous. See, however, Section 11.4.3 for the Pearson chi-squared test that is applicable here.

**Exercise 6.7** *Does playing lotteries and engaging in gambling make* rational *sense at all, from the perspective of an individual player? Well, we see that 16 is the most frequently occurring outcome in* Lotto, *maybe there's some magic in it? Also, some people sometimes became millionaires, right?*

---

**Note** In data modelling (e.g., Bayesian statistics), sometimes a uniform distribution is chosen as a placeholder for "we know nothing about a phenomenon, so let us just assume that every event is equally likely". Nonetheless, it is quite fascinating that the real world tends to be structured after all. Emerging patterns are plentiful, most often they are far from being uniformly distributed. Even more strikingly, they are subject to quantitative analysis.

### 6.3.4   Distribution Mixtures (*)

Some datasets may fail to fit through simple models such as the ones described above. It may sometimes be due to their non-random behaviour: statistics gives just one means to create data idealisations, we also have partial differential equations, approximation theory, graphs and complex networks, agent-based modelling, and so forth, which might be worth giving a study (and then try).

Another reason may be that what we observe is in fact a *mixture* (creative combination) of simpler processes.

The dataset representing the December 2021 hourly averages pedestrian counts near the Southern Cross Station in Melbourne might be a good instance of such a scenario; compare Figure 4.5.

```
peds = np.loadtxt("https://raw.githubusercontent.com/gagolews/" +
    "teaching-data/master/marek/southern_cross_station_peds_2019_dec.txt")
```

It might not be a bad idea to try to fit a probabilistic (convex) combination of three normal distributions $f_1, f_2, f_3$, corresponding to the morning, lunch-time, and evening pedestrian count peaks. This yields the PDF:

$$f(x) = w_1 f_1(x) + w_2 f_2(x) + w_3 f_3(x),$$

for some coefficients $w_1, w_2, w_3 \geq 0$ such that $w_1 + w_2 + w_3 = 1$.

Figure 6.14 depicts a mixture of N(8, 1), N(12, 1), and N(17, 2) with the corresponding weights of 0.35, 0.1, and 0.55. This dataset is quite coarse-grained (we only have 24 bar heights at our disposal). Consequently, the estimated coefficients should be taken with a pinch of chilli pepper.

```
plt.bar(np.arange(24), width=1, height=peds/np.sum(peds),
    color="lightgray", edgecolor="black", alpha=0.8)
x = np.arange(0, 25, 0.1)
p1 = scipy.stats.norm.pdf(x, 8, 1)
p2 = scipy.stats.norm.pdf(x, 12, 1)
p3 = scipy.stats.norm.pdf(x, 17, 2)
p = 0.35*p1 + 0.1*p2 + 0.55*p3  # weighted combination of 3 densities
plt.plot(x, p, "r--", label="PDF of a normal mixture")
plt.legend()
plt.show()
```

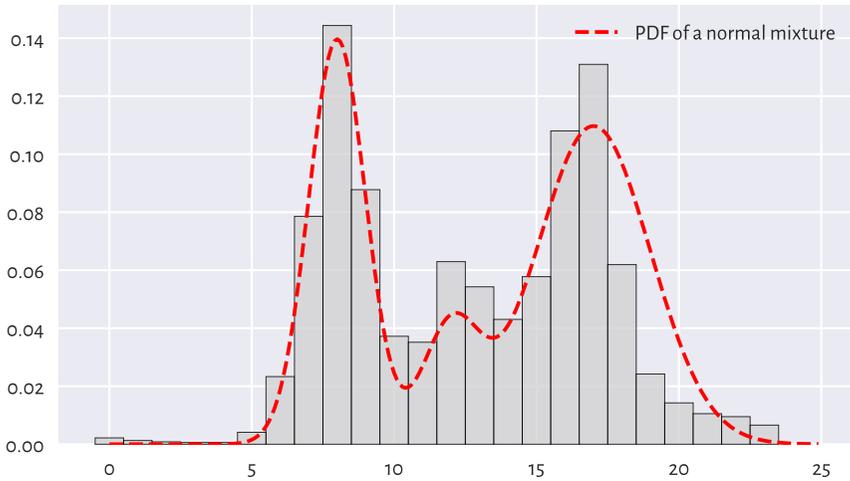

Figure 6.14: Histogram of the `peds` dataset and a guesstimated mixture of three normal distributions

---

**Important** It will frequently be the case in data wrangling that more complex entities (models, methods) will be arising as combinations of simpler (primitive) components. This is why we should spend a great deal of time studying the *fundamentals*.

---

**Note** Some data clustering techniques (in particular, the *k*-means algorithm that we briefly discuss later in this course) could be used to split a data sample into disjoint chunks corresponding to different mixture components.

Also, it might be the case that the mixture components can in fact be explained by another categorical variable that divides the dataset into natural groups; compare Chapter 12.

---

## 6.4 Generating Pseudorandom Numbers

A probability distribution is useful not only for describing a dataset. It also enables us to perform many experiments on data that we do not currently have, but we might obtain in the future, to test various scenarios and hypotheses.

To do this, we can generate a random sample of independent (not related to each other) observations.

### 6.4.1 Uniform Distribution

When most people say *random*, they implicitly mean *uniformly distributed*. For example:

```
np.random.rand(5)
## array([0.69646919, 0.28613933, 0.22685145, 0.55131477, 0.71946897])
```

gives five observations sampled independently from the uniform distribution on the unit interval, i.e., U(0, 1).

The same with **scipy**, but this time the support will be *(-10, 15)*.

```
scipy.stats.uniform.rvs(-10, scale=25, size=5)  # from -10 to -10+25
## array([ 0.5776615 , 14.51910496,  7.12074346,  2.02329754, -0.19706205])
```

Alternatively, we could do that ourselves by shifting and scaling the output of the random number generator on the unit interval using the formula `numpy.random.rand(5)*25-10`.

### 6.4.2 Not Exactly Random

We generate numbers using a computer, which is purely deterministic. Hence, we shall refer to them as *pseudorandom* or random-like ones (albeit they are non-distinguishable from truly random, when subject to rigorous tests for randomness).

To prove it, we can set the initial state of the generator (the *seed*) via some number and see what values are produced:

```
np.random.seed(123)  # set seed
np.random.rand(5)
## array([0.69646919, 0.28613933, 0.22685145, 0.55131477, 0.71946897])
```

Then, we set the seed once again via the same number and see how "random" the next values are:

```
np.random.seed(123)  # set seed
np.random.rand(5)
## array([0.69646919, 0.28613933, 0.22685145, 0.55131477, 0.71946897])
```

This enables us to perform completely *reproducible* numerical experiments, and this is a very good feature: truly scientific inquiries should lead to identical results under the same conditions.

---

**Note**  If we do not set the seed manually, it will be initialised based on the current wall time, which is different every... time. As a result, the numbers will *seem* random to us.

---

Many Python packages that we will be using in the future, including **pandas** and **sk-**

**learn**, rely on **numpy**'s random number generator. We will become used to calling **numpy.random.seed** to make them predictable.

Additionally, some of them (e.g., **sklearn.model_selection.train_test_split** or **pandas.DataFrame.sample**) are equipped with the `random_state` argument, which behaves as if we *temporarily* changed the seed (for just one call to that function). For instance:

```
scipy.stats.uniform.rvs(size=5, random_state=123)
## array([0.69646919, 0.28613933, 0.22685145, 0.55131477, 0.71946897])
```

This gives the same sequence as above.

### 6.4.3 Sampling from Other Distributions

Generating data from other distributions is possible too; there are many **rvs** methods implemented in **scipy.stats**. For example, here is a sample from N(100, 16):

```
scipy.stats.norm.rvs(100, 16, size=3, random_state=50489)
## array([113.41134015,  46.99328545, 157.1304154 ])
```

Pseudorandom deviates from the *standard* normal distribution, i.e., N(0, 1), can also be generated using **numpy.random.randn**. As N(100, 16) is a scaled and shifted version thereof, the above is equivalent to:

```
np.random.seed(50489)
np.random.randn(3)*16 + 100
## array([113.41134015,  46.99328545, 157.1304154 ])
```

---

**Important** Conclusions based on simulated data are trustworthy, because they cannot be manipulated. Or can they?

The pseudorandom number generator's seed used above, 50489, is quite suspicious. It might suggest that someone wanted to *prove* some point (in this case, the violation of the $3\sigma$ rule).

This is why we recommend sticking to only one seed most of the time, e.g., 123, or – when performing simulations – setting consecutive seeds for each iteration: 1, 2, ....

---

**Exercise 6.8** *Generate 1,000 pseudorandom numbers from the log-normal distribution and draw a histogram thereof.*

---

**Note** (*) Having a good pseudorandom number generator from the uniform distribution on the unit interval is crucial, because sampling from other distributions usually involves transforming independent U(0, 1) variates.

For instance, realisations of random variables following any continuous cumulative distribution function $F$ can be constructed through the *inverse transform sampling* (see [32, 70]):

1. Generate a sample $x_1, \dots, x_n$ independently from U(0, 1).

2. Transform each $x_i$ by applying the quantile function, $y_i = F^{-1}(x_i)$.

Now $y_1, \dots, y_n$ follows the CDF $F$.

---

**Exercise 6.9**   (*) *Generate 1,000 pseudorandom numbers from the log-normal distribution using inverse transform sampling.*

**Exercise 6.10**   (**) *Generate 1,000 pseudorandom numbers from the distribution mixture discussed in Section 6.3.4.*

### 6.4.4   Natural Variability

Even a sample truly generated from a specific distribution will deviate from it, sometimes considerably. Such effects will be especially visible for small sample sizes, but they usually disappear[10] when the availability of data increases.

For example, Figure 6.15 depicts the histograms of nine different samples of size 100, all drawn independently from the standard normal distribution.

```
plt.figure(figsize=(plt.rcParams["figure.figsize"][0], )*2)  # width=height
for i in range(9):
    plt.subplot(3, 3, i+1)
    sample = scipy.stats.norm.rvs(size=100, random_state=i+1)
    sns.histplot(sample, stat="density", bins=10, color="lightgray")
    plt.ylabel(None)
    plt.xlim(-4, 4)
    plt.ylim(0, 0.6)
plt.legend()
plt.show()
```

There is some ruggedness in the bar sizes that a naïve observer might try to interpret as something meaningful. A competent data scientist must train their eye to ignore such impurities (but should always be ready to detect those which are worth attention). In this case, they are only due to random effects.

**Exercise 6.11**   *Repeat the above experiment for samples of sizes 10, 1,000, and 10,000.*

**Example 6.12**   (*) *Using a simple Monte Carlo simulation, we can verify (approximately) that the Kolmogorov–Smirnov goodness-of-fit test introduced in Section 6.2.3 has been calibrated properly, i.e., that for samples that really follow the assumed distribution, the null hypothesis is rejected only in ca. 0.1% of the cases.*

*Let us say we are interested in the null hypothesis referencing the standard normal distribution,*

---

[10] Compare the Fundamental Theorem of Statistics (the Glivenko–Cantelli theorem).

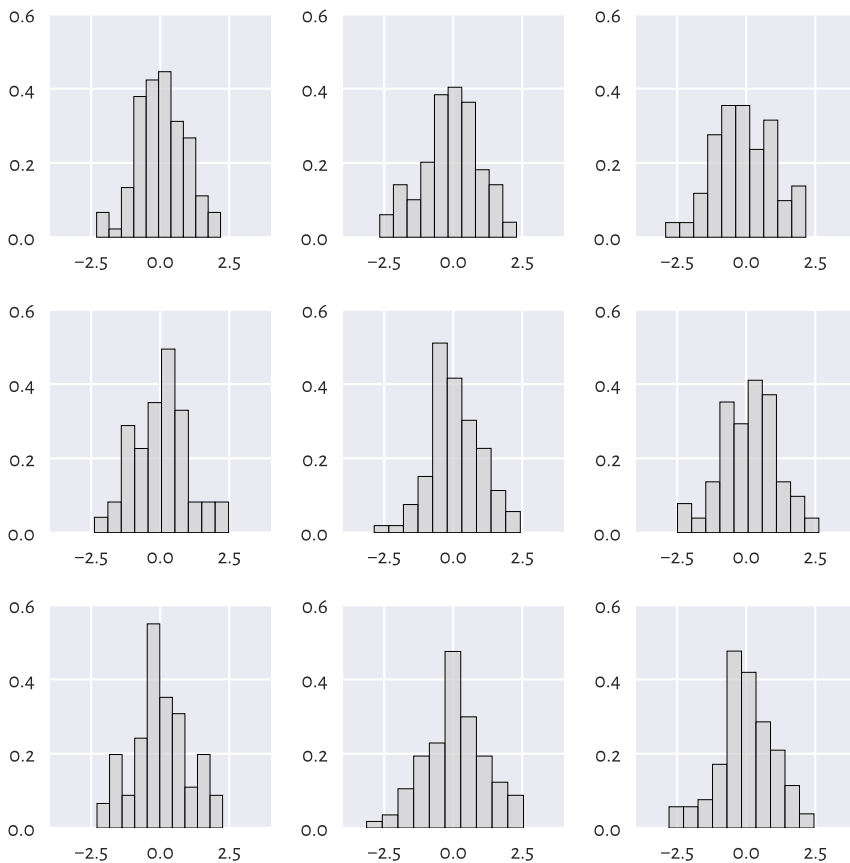

Figure 6.15: All nine samples are normally distributed

*N(0, 1), and sample size n = 100. We need to generate many (we assume 10,000 below) such samples for each of which we compute and store the maximal absolute deviation from the theoretical CDF, i.e., $\hat{D}_n$.*

```
n = 100
distrib = scipy.stats.norm(0, 1)  # assumed distribution - N(0, 1)
Dns = []
for i in range(10000):  # increase this for better precision
    x = distrib.rvs(size=n, random_state=i+1)  # really follows distrib
    Dns.append(compute_Dn(x, distrib.cdf))
Dns = np.array(Dns)
```

*Now let us compute the proportion of cases which lead to $\hat{D}_n$ greater than the critical value $K_n$:*

```
len(Dns[Dns >= scipy.stats.kstwo.ppf(1-0.001, n)]) / len(Dns)
## 0.0016
```

*In theory, this should be equal to 0.001. But our values are necessarily approximate, because we rely on randomness. Increasing the number of trials from 10,000 to, say, 1,000,000 will make the above estimate more precise.*

*It is also worth checking out that the density histogram of* `Dns` *resembles the Kolmogorov distribution that we can compute via* ***scipy.stats.kstwo.pdf***.

**Exercise 6.13** *(\*) It might also be interesting to check out the test's* power*, i.e., the probability that when the null hypothesis is false, it will actually be rejected. Modify the above code in such a way that x in the* ***for*** *loop is not generated from N(0, 1), but N(0.1, 1), N(0.2, 1), etc., and check the proportion of cases where we deem the sample distribution different from N(0, 1). Small differences in the location parameter μ are usually ignored, and this improves with sample size n.*

### 6.4.5 Adding Jitter (White Noise)

We mentioned that measurements might be subject to observational error. Rounding can also occur as early as the data collection phase. In particular, our `heights` dataset is precise up to 1 fractional digit. However, in statistics, when we say that data follow a continuous distribution, the probability of having two identical values in a sample is 0. Therefore, some data analysis methods might assume that there are no ties in the input vector, i.e., all values are unique.

The easiest way to deal with such numerical inconveniences is to add some white noise with the expected value of 0, either uniformly or normally distributed.

For example, for `heights` it makes sense to add some jitter from U[-0.05, 0.05]:

```
heights_jitter = heights + (np.random.rand(len(heights))*0.1-0.05)
heights_jitter[:6]  # preview
## array([160.21704623, 152.68870195, 161.24482407, 157.3675293 ,
##        154.61663465, 144.68964596])
```

Adding noise also might be performed for aesthetic reasons, e.g., when drawing scatterplots.

## 6.5 Further Reading

For an excellent general introductory course on probability and statistics; see [33, 35]. More advanced students are likely to enjoy the two classics [5, 15]. Topics in random number generation are covered in [32, 50, 70].

For a more comprehensive introduction to exploratory data analysis, see the classical books by Tukey [78, 79] and Tufte [77].

We took the logarithm of the log-normally distributed incomes and obtained a normally distributed sample. In statistical practice, it is not rare to apply different nonlinear transforms of the input vectors at the data preprocessing stage (see, e.g., Section 9.2.6). In particular, the Box–Cox (power) transform [8] is of the form $x \mapsto \frac{x^\lambda - 1}{\lambda}$ for some $\lambda$. Interestingly, in the limit as $\lambda \to 0$, this formula yields $x \mapsto \log x$ which is exactly what we were applying in this chapter.

[12, 60] give a nice overview of the power-law-like behaviour of some "rich" or otherwise extreme datasets. It is worth noting that the logarithm of a Paretian sample divided by the minimum follows an exponential distribution (which we discuss in Chapter 16). For a comprehensive catalogue of statistical distributions, their properties, and relationships between them, see [23].

## 6.6 Exercises

**Exercise 6.14** *Why is the notion of the mean income confusing the general public?*

**Exercise 6.15** *When manually setting the seed of a random number generator makes sense?*

**Exercise 6.16** *Given a log-normally distributed sample x, how can we turn it to a normally distributed one, i.e., y=f(x), with f being... what?*

**Exercise 6.17** *What is the 3σ rule for normally distributed data?*

**Exercise 6.18** *(\*) How can we verify graphically if a sample follows a hypothesised theoretical distribution?*

**Exercise 6.19** *(\*) Explain the meaning of type I error, significance level, and a test's power.*

# Part III

# Multidimensional Data

# 7

## *Multidimensional Numeric Data at a Glance*

From the perspective of structured datasets, a vector often represents $n$ independent measurements of the same quantitative property, e.g., heights of $n$ different patients, incomes in $n$ randomly chosen households, or ages of $n$ runners. More generally, these are all instances of a bag of $n$ points on the real line. By far[1] we should have become quite fluent with the methods for processing such one-dimensional arrays.

Let us increase the level of complexity by allowing each of the $n$ entities to be described by $m$ features, for any $m \geq 1$. In other words, we will be dealing with $n$ points in an $m$-dimensional space, $\mathbb{R}^m$.

We can arrange all the observations in a table with $n$ rows and $m$ columns (just like in spreadsheets). Such an object can be expressed with `numpy` as a two-dimensional array which we will refer to as *matrices*. Thanks to matrices, we can keep the $n$ tuples of length $m$ together in a single object and process them all at once (or $m$ tuples of length $n$, depending on how we want to look at them). Very convenient.

---

**Important**  Just like vectors, matrices were designed to store data of the same type. In Chapter 10, we will cover *data frames*, which further increase the degree of complexity (and freedom) by not only allowing for mixed data types (e.g., numerical and categorical; this will enable us to perform data analysis in subgroups more easily) but also for the rows and columns be named.

Many data analysis algorithms convert data frames to matrices automatically and deal with them as such. From the computational side, it is `numpy` that does most of the "mathematical" work. `pandas` implements many recipes for basic data wrangling tasks, but we want to go way beyond that. After all, we would like to be able to tackle *any* problem.

---

[1] Assuming we solved all the suggested exercises, which we did, didn't we? See Rule #3.

## 7.1   Creating Matrices

### 7.1.1   Reading CSV Files

Tabular data are often stored and distributed in a very portable plain-text format called CSV (comma-separated values) or variants thereof.

**numpy.loadtxt** supports them quite well as long as they do not feature column names (comment lines are accepted, though). As for most CSV files the opposite is the case, we suggest relying on a corresponding function from the **pandas** package:

```python
body = pd.read_csv("https://raw.githubusercontent.com/gagolews/" +
    "teaching-data/master/marek/nhanes_adult_female_bmx_2020.csv",
    comment="#")
body = np.array(body)  # convert to matrix
```

Notice that we converted the data frame to a matrix by calling the **numpy.array** function. Here is a preview of the first few rows:

```python
body[:6, :]  # six first rows, all columns
## array([[ 97.1, 160.2,  34.7,  40.8,  35.8, 126.1, 117.9],
##        [ 91.1, 152.7,  33.5,  33. ,  38.5, 125.5, 103.1],
##        [ 73. , 161.2,  37.4,  38. ,  31.8, 106.2,  92. ],
##        [ 61.7, 157.4,  38. ,  34.7,  29. , 101. ,  90.5],
##        [ 55.4, 154.6,  34.6,  34. ,  28.3,  92.5,  73.2],
##        [ 62. , 144.7,  32.5,  34.2,  29.8, 106.7,  84.8]])
```

This is an extended version of the National Health and Nutrition Examination Survey (NHANES[2]), where the consecutive columns give the following body measurements of adult females:

```python
body_columns = np.array([
    "weight (kg)",
    "standing height (cm)",
    "upper arm length (cm)",
    "upper leg length (cm)",
    "arm circumference (cm)",
    "hip circumference (cm)",
    "waist circumference (cm)"
])
```

**numpy** matrices do not support column naming. This is why we noted them down separately. It is only a minor inconvenience. **pandas** data frames will have this capability, but from the algebraic side, they are not as convenient as matrices for the purpose of scientific computing.

---

[2] https://wwwn.cdc.gov/nchs/nhanes/search/datapage.aspx

What we are dealing with is still a `numpy` array:

```python
type(body)  # class of this object
## <class 'numpy.ndarray'>
```

But this time it is a two-dimensional one:

```python
body.ndim  # number of dimensions
## 2
```

which means that the `shape` slot is now a tuple of length 2:

```python
body.shape
## (4221, 7)
```

The above gave the total number of rows and columns, respectively.

## 7.1.2 Enumerating Elements

`numpy.array` can create a two-dimensional array based on a list of lists or vector-like objects, all of the same lengths. Each of them will constitute a separate row of the resulting matrix.

For example:

```python
np.array([  # list of lists
    [ 1,  2,  3,  4 ],  # the 1st row
    [ 5,  6,  7,  8 ],  # the 2nd row
    [ 9, 10, 11, 12 ]   # the 3rd row
])
## array([[ 1,  2,  3,  4],
##        [ 5,  6,  7,  8],
##        [ 9, 10, 11, 12]])
```

gives a 3-by-4 (3×4) matrix,

```python
np.array([ [1], [2], [3] ])
## array([[1],
##        [2],
##        [3]])
```

yields a 3-by-1 one (we call it a *column vector*, but it is a special matrix — we will soon learn that shapes can make a significant difference), and

```
np.array([ [1, 2, 3, 4] ])
## array([[1, 2, 3, 4]])
```

produces a 1-by-4 array (a *row vector*).

---

**Note**   An ordinary vector (a unidimensional array) only uses a single pair of square brackets:

```
np.array([1, 2, 3, 4])
## array([1, 2, 3, 4])
```

---

### 7.1.3   Repeating Arrays

The previously mentioned `numpy.tile` and `numpy.repeat` can also generate some nice matrices. For instance,

```
np.repeat([[1, 2, 3, 4]], 3, axis=0)
## array([[1, 2, 3, 4],
##        [1, 2, 3, 4],
##        [1, 2, 3, 4]])
```

repeats a row vector rowwisely (i.e., over axis 0 – the first one).

Replicating a column vector columnwisely (i.e., over axis 1 – the second one) is possible as well:

```
np.repeat([[1], [2], [3]], 4, axis=1)
## array([[1, 1, 1, 1],
##        [2, 2, 2, 2],
##        [3, 3, 3, 3]])
```

**Exercise 7.1**   *How can we generate matrices of the following kinds?*

$$
\begin{bmatrix} 1 & 2 \\ 1 & 2 \\ 1 & 2 \\ 3 & 4 \\ 3 & 4 \\ 3 & 4 \\ 3 & 4 \end{bmatrix}, \quad
\begin{bmatrix} 1 & 2 & 1 & 2 & 1 & 2 \\ 1 & 2 & 1 & 2 & 1 & 2 \end{bmatrix}, \quad
\begin{bmatrix} 1 & 1 & 2 & 2 & 2 \\ 3 & 3 & 4 & 4 & 4 \end{bmatrix}.
$$

### 7.1.4   Stacking Arrays

`numpy.column_stack` and `numpy.row_stack` take a tuple of array-like objects and bind them column- or rowwisely to form a new matrix:

```python
np.column_stack(([10, 20], [30, 40], [50, 60]))  # a tuple of lists
## array([[10, 30, 50],
##        [20, 40, 60]])
np.row_stack(([10, 20], [30, 40], [50, 60]))
## array([[10, 20],
##        [30, 40],
##        [50, 60]])
np.column_stack((
    np.row_stack(([10, 20], [30, 40], [50, 60])),
    [70, 80, 90]
))
## array([[10, 20, 70],
##        [30, 40, 80],
##        [50, 60, 90]])
```

**Exercise 7.2** *Perform similar operations using* `numpy.append`*,* `numpy.vstack`*,* `numpy.hstack`*,* `numpy.concatenate`*, and* (\*) `numpy.c_`*.*

**Exercise 7.3** *Using* `numpy.insert`*, and a new row/column at the beginning, end, and in the middle of an array. Let us stress that this function returns a new array.*

### 7.1.5 Other Functions

Many built-in functions allow for generating arrays of arbitrary shapes (not only vectors). For example:

```python
np.random.seed(123)
np.random.rand(2, 5)  # not: rand((2, 5))
## array([[0.69646919, 0.28613933, 0.22685145, 0.55131477, 0.71946897],
##        [0.42310646, 0.9807642 , 0.68482974, 0.4809319 , 0.39211752]])
```

The same with **scipy**:

```python
scipy.stats.uniform.rvs(0, 1, size=(2, 5), random_state=123)
## array([[0.69646919, 0.28613933, 0.22685145, 0.55131477, 0.71946897],
##        [0.42310646, 0.9807642 , 0.68482974, 0.4809319 , 0.39211752]])
```

The way we specify the output shapes might differ across functions and packages. Consequently, as usual, it is always best to refer to their documentation.

**Exercise 7.4** *Check out the documentation of the following functions:* `numpy.eye`*,* `numpy.diag`*,* `numpy.zeros`*,* `numpy.ones`*, and* `numpy.empty`*.*

## 7.2    Reshaping Matrices

Let us take an example 3-by-4 matrix:

```
A = np.array([
    [ 1,  2,  3,  4 ],
    [ 5,  6,  7,  8 ],
    [ 9, 10, 11, 12 ]
])
```

Internally, a matrix is represented using a *long* flat vector where elements are stored in the row-major[3] order:

```
A.size  # total number of elements
## 12
A.ravel()  # the underlying array
## array([ 1,  2,  3,  4,  5,  6,  7,  8,  9, 10, 11, 12])
```

It is the `shape` slot that is causing the 12 elements to be treated as if they were arranged on a 3-by-4 grid, for example in different algebraic computations and during the printing thereof. This arrangement can be altered anytime without modifying the underlying array:

```
A.shape = (4, 3)
A
## array([[ 1,  2,  3],
##        [ 4,  5,  6],
##        [ 7,  8,  9],
##        [10, 11, 12]])
```

This way, we obtained a different *view* of the same data.

For convenience, there is also the `reshape` method that returns a modified version of the object it is applied on:

```
A.reshape(-1, 6)
## array([[ 1,  2,  3,  4,  5,  6],
##        [ 7,  8,  9, 10, 11, 12]])
```

Here, "-1" means that `numpy` must deduce by itself how many rows we want in the result. Twelve elements are supposed to be arranged in six columns, so the maths behind it is not rocket science.

Thanks to this, generating row or column vectors is straightforward:

---

[3] (*) Sometimes referred to as a C-style array, as opposed to Fortran-style which is used in, e.g., R.

```
np.linspace(0, 1, 5).reshape(1, -1)
## array([[0.  , 0.25, 0.5 , 0.75, 1.  ]])
np.array([9099, 2537, 1832]).reshape(-1, 1)
## array([[9099],
##        [2537],
##        [1832]])
```

Reshaping is not the same as matrix *transpose*, which also changes the order of elements in the underlying array:

```
A  # before
## array([[ 1,  2,  3],
##        [ 4,  5,  6],
##        [ 7,  8,  9],
##        [10, 11, 12]])
A.T  # transpose of A
## array([[ 1,  4,  7, 10],
##        [ 2,  5,  8, 11],
##        [ 3,  6,  9, 12]])
```

We see that the rows became columns and vice versa.

---

**Note** (*) Higher-dimensional arrays are also possible. For example,

```
np.arange(24).reshape(2, 4, 3)
## array([[[ 0,  1,  2],
##        [ 3,  4,  5],
##        [ 6,  7,  8],
##        [ 9, 10, 11]],
##
##        [[12, 13, 14],
##        [15, 16, 17],
##        [18, 19, 20],
##        [21, 22, 23]]])
```

Is an array of "depth" 2, "height" 4, and "width" 3; we can see it as two 4-by-3 matrices stacked together. Theoretically, they can be used for representing contingency tables for products of many factors. Still, in our application areas, we prefer to stick with long data frames instead; see Section 10.6.2. This is due to their more aesthetic display and better handling of sparse data.

---

## 7.3   Mathematical Notation

Here is some standalone mathematical notation that we shall be employing in this course. A matrix with $n$ rows and $m$ columns (an $n$-by-$m$ matrix) $\mathbf{X}$ can be written as:

$$\mathbf{X} = \left[ \begin{array}{cccc} x_{1,1} & x_{1,2} & \cdots & x_{1,m} \\ x_{2,1} & x_{2,2} & \cdots & x_{2,m} \\ \vdots & \vdots & \ddots & \vdots \\ x_{n,1} & x_{n,2} & \cdots & x_{n,m} \end{array} \right].$$

Mathematically, we denote this as $\mathbf{X} \in \mathbb{R}^{n \times m}$. Looking at the above, if this makes us think of how data are displayed in spreadsheets, we are correct, because the latter was inspired by the former.

We see that $x_{i,j} \in \mathbb{R}$ denotes the element in the $i$-th row (e.g., the $i$-th *observation*) and the $j$-th column (e.g., the $j$-th *feature* or *variable*), for every $i = 1, \dots, n, j = 1, \dots, m$.

In particular, if $\mathbf{X}$ denoted the `body` dataset, then $x_{1,2}$ would be the height of the 1st person.

---

**Important**   Matrices are a convenient means of representing many different kinds of data:

- $n$ points in an $m$-dimensional space (like $n$ observations for which there are $m$ measurements/features recorded, where each row describes a different object; exactly the case of the `body` dataset above) – this is the most common scenario;

- $m$ time series sampled at $n$ points in time (e.g., prices of $m$ different currencies on $n$ consecutive days; see Chapter 16);

- a single kind of measurement for data in $m$ groups, each consisting of $n$ subjects (e.g., heights of $n$ males and $n$ females); here, the order of elements in each column does not usually matter as observations are not *paired*; there is no relationship between $x_{i,j}$ and $x_{i,k}$ for $j \neq k$; a matrix is used merely as a convenient container for storing a few unrelated vectors of identical sizes; we will be dealing with a more generic case of possibly nonhomogeneous groups in Chapter 12;

- two-way contingency tables (see Section 11.2.2), where an element $x_{i,j}$ gives the number of occurrences of items at the $i$-th level of the first categorical variable and, at the same time, being at the $j$-th level of the second variable (e.g., blue-eyed *and* blonde-haired);

- graphs and other relationships between objects, e.g., $x_{i,j} = 0$ might denote that the $i$-th object is not connected[4] with the $j$-th one and $x_{k,l} = 0.2$ that there is a weak connection between $k$ and $l$ (e.g., who is a friend of whom, whether a user recommends a particular item);

---

[4] Such matrices are usually sparse, i.e., have many elements equal to 0. We have special, memory-efficient data structures for handling these data; see `scipy.sparse` for more details as this goes beyond the scope of our introductory course.

- images, where $x_{i,j}$ represents the intensity of a colour component (e.g., red, green, blue or shades of grey or hue, saturation, brightness; compare Section 16.4) of a pixel in the $(n - i + 1)$-th row and the $j$-th column.

---

**Note** In practice, more complex and less-structured data can quite often be mapped to a tabular form. For instance, a set of audio recordings can be described by measuring the overall loudness, timbre, and danceability of each song. Also, a collection of documents can be described by means of the degrees of belongingness to some automatically discovered topics (e.g., someone said that Joyce's *Ulysses* is 80% travel literature, 70% comedy, and 50% heroic fantasy, but let us not take it for granted).

---

### 7.3.1 Row and Column Vectors

Additionally, will sometimes use the following notation to emphasise that $\mathbf{X}$ consists of $n$ rows:

$$\mathbf{X} = \left[ \begin{array}{c} \mathbf{x}_{1,\cdot} \\ \mathbf{x}_{2,\cdot} \\ \vdots \\ \mathbf{x}_{n,\cdot} \end{array} \right].$$

Here, $\mathbf{x}_{i,\cdot}$ is a *row vector* of length $m$, i.e., a $(1 \times m)$-matrix:

$$\mathbf{x}_{i,\cdot} = \left[ \begin{array}{cccc} x_{i,1} & x_{i,2} & \cdots & x_{i,m} \end{array} \right].$$

Alternatively, we can specify the $m$ columns:

$$\mathbf{X} = \left[ \begin{array}{cccc} \mathbf{x}_{\cdot,1} & \mathbf{x}_{\cdot,2} & \cdots & \mathbf{x}_{\cdot,m} \end{array} \right],$$

where $\mathbf{x}_{\cdot,j}$ is a *column vector* of length $n$, i.e., an $(n \times 1)$-matrix:

$$\mathbf{x}_{\cdot,j} = \left[ \begin{array}{cccc} x_{1,j} & x_{2,j} & \cdots & x_{n,j} \end{array} \right]^{T} = \left[ \begin{array}{c} x_{1,j} \\ x_{2,j} \\ \vdots \\ x_{n,j} \end{array} \right],$$

where $\cdot^{T}$ denotes the transpose of a given matrix (thanks to which we can save some vertical space, we do not want this book to be 1000 pages long, do we?).

Also, recall that we are used to denoting *vectors* of length $m$ with $\boldsymbol{x} = (x_1, \dots, x_m)$. A vector is a one-dimensional array (not a two-dimensional one), hence a slightly different font in the case where ambiguity can be troublesome.

---

**Note**  To avoid notation clutter, we will often be implicitly promoting vectors like $\boldsymbol{x} = (x_1, \ldots, x_m)$ to row vectors $\mathbf{x} = [x_1 \cdots x_m]$, because this is the behaviour that `numpy`[5] uses; see Chapter 8.

---

### 7.3.2  Transpose

The *transpose* of a matrix $\mathbf{X} \in \mathbb{R}^{n \times m}$ is an $(m \times n)$-matrix $\mathbf{Y}$ given by:

$$\mathbf{Y} = \mathbf{X}^T = \left[ \begin{array}{cccc} x_{1,1} & x_{2,1} & \cdots & x_{m,1} \\ x_{1,2} & x_{2,2} & \cdots & x_{m,2} \\ \vdots & \vdots & \ddots & \vdots \\ x_{1,n} & x_{2,n} & \cdots & x_{m,n} \end{array} \right],$$

i.e., it enjoys $y_{i,j} = x_{j,i}$.

**Exercise 7.5**  *Compare the display of an example matrix A and its transpose A.T above.*

### 7.3.3  Identity and Other Diagonal Matrices

**I** denotes the *identity matrix*, being a square $n \times n$ matrix (with $n$ most often clear from the context) with 0s everywhere except on the main diagonal, where 1s lie.

```
np.eye(5)  # I
## array([[1., 0., 0., 0., 0.],
##        [0., 1., 0., 0., 0.],
##        [0., 0., 1., 0., 0.],
##        [0., 0., 0., 1., 0.],
##        [0., 0., 0., 0., 1.]])
```

The identity matrix is a neutral element of the matrix multiplication (Section 8.3).

More generally, any diagonal matrix, $\text{diag}(a_1, \ldots, a_n)$, can be constructed from a given sequence of elements by calling:

```
np.diag([1, 2, 3, 4])
## array([[1, 0, 0, 0],
##        [0, 2, 0, 0],
##        [0, 0, 3, 0],
##        [0, 0, 0, 4]])
```

---

[5] Some textbooks assume that all vectors are *column* vectors.

## 7.4 Visualising Multidimensional Data

Let us go back to our body dataset:

```
body[:6, :]  # preview
## array([[ 97.1, 160.2,  34.7,  40.8,  35.8, 126.1, 117.9],
##        [ 91.1, 152.7,  33.5,  33. ,  38.5, 125.5, 103.1],
##        [ 73. , 161.2,  37.4,  38. ,  31.8, 106.2,  92. ],
##        [ 61.7, 157.4,  38. ,  34.7,  29. , 101. ,  90.5],
##        [ 55.4, 154.6,  34.6,  34. ,  28.3,  92.5,  73.2],
##        [ 62. , 144.7,  32.5,  34.2,  29.8, 106.7,  84.8]])
body.shape
## (4221, 7)
```

This is an example of tabular ("structured") data. The important property is that the elements in each row describe the same person. We can freely reorder all the columns at the same time (change the order of participants). Still, sorting a single column and leaving the other ones unchanged will be semantically invalid.

Mathematically, we consider the above as a set of 4221 points in a seven-dimensional space, $\mathbb{R}^7$. Let us discuss how we can try visualising different natural *projections* thereof.

### 7.4.1 2D Data

A *scatterplot* can be used to visualise one variable against another one.

```
plt.plot(body[:, 1], body[:, 3], "o", c="#00000022")
plt.xlabel(body_columns[1])
plt.ylabel(body_columns[3])
plt.show()
```

Figure 7.1 depicts upper leg length (the y-axis) vs (versus; against; as a function of) standing height (the x-axis) in the form of a point cloud with $(x, y)$ coordinates like (body[i, 1], body[i, 3]).

**Example 7.6** *Here are the exact coordinates of the point corresponding to the person of the smallest height:*

```
body[np.argmin(body[:, 1]), [1, 3]]
## array([131.1,  30.8])
```

*and here is the one with the greatest upper leg length:*

```
body[np.argmax(body[:, 3]), [1, 3]]
## array([168.9,  49.1])
```

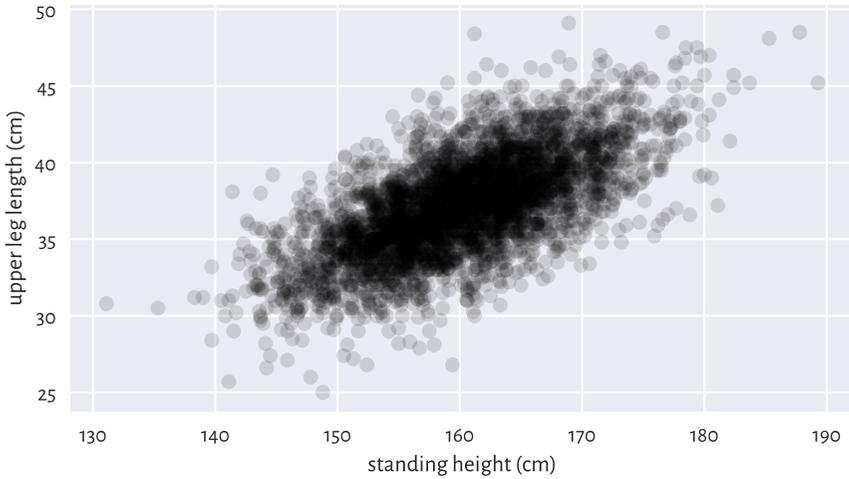

Figure 7.1: An example scatterplot

*Locate them in Figure 7.1.*

As the points are abundant, normally we cannot easily see *where* most of them are located. To remedy this, we applied the simple trick of plotting the points using a semi-transparent colour. Here, the colour specifier was of the form #rrggbbaa, giving the intensity of the red, green, blue, and alpha (opaqueness) channel in three series of two hexadecimal digits (between 00 = 0 and ff = 255).

Overall, the plot reveals that there is a *general tendency* for small heights and small upper leg lengths to occur frequently together. The same with larger pairs. In Chapter 9, we explore some measures of correlation that will enable us to quantify the degree of association between variable pairs.

### 7.4.2    3D Data and Beyond

If we have more than two variables to visualise, we might be tempted to use, e.g., a three-dimensional scatterplot like the one in Figure 7.2.

```
fig = plt.figure()
ax = fig.add_subplot(projection="3d", facecolor="#ffffff00")
ax.scatter(body[:, 1], body[:, 3], body[:, 0], color="#00000011")
ax.view_init(elev=30, azim=60, vertical_axis="y")
ax.set_xlabel(body_columns[1])
ax.set_ylabel(body_columns[3])
ax.set_zlabel(body_columns[0])
plt.show()
```

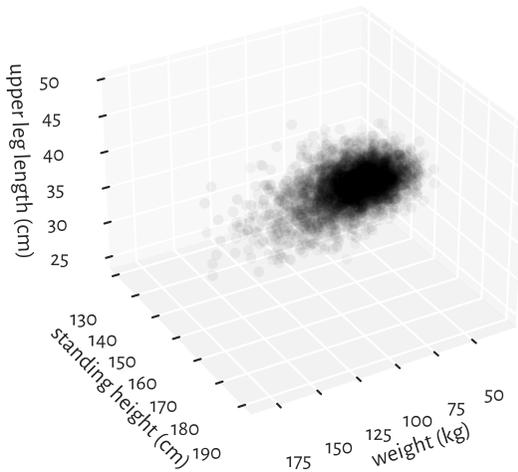

Figure 7.2: A three-dimensional scatterplot reveals almost nothing

Infrequently will such a 3D plot provide us with readable results, though. We are projecting a three-dimensional reality onto a two-dimensional screen or page. Some information must inherently be lost. Also, what we see is relative to the position of the virtual camera.

**Exercise 7.7**  *(\*) Try finding an* interesting *elevation and azimuth angle by playing with the arguments passed to the* `mpl_toolkits.mplot3d.axes3d.Axes3D.view_init` *function. Also, depict arm circumference, hip circumference, and weight on a 3D plot.*

---

**Note**   (\*) Sometimes there might be facilities available to create an interactive scatterplot (running the above from the Python's console enables this), where the virtual camera can be freely repositioned with a mouse/touchpad. This can give some more insight into our data. Also, there are means of creating animated sequences, where we can fly over the data scene. Some people find it cool, others find it annoying, but the biggest problem therewith is that they cannot be included in printed material. Yet, if we are only targeting the display for the Web (this includes mobile devices), we can try some Python libraries[6] that output HTML+CSS+JavaScript code to be rendered by a browser engine.

---

**Example 7.8**  *Instead of drawing a 3D plot, it might be better to play with different marker colours (or sometimes sizes: think of them as bubbles). Suitable colour maps[7] can be used to distinguish between low and high values of an additional variable, as in Figure 7.3.*

---

[6] https://wiki.python.org/moin/NumericAndScientific/Plotting
[7] https://matplotlib.org/stable/tutorials/colors/colormaps.html

```python
from matplotlib import cm
plt.scatter(
    body[:, 4],      # x
    body[:, 5],      # y
    c=body[:, 0],    # "z" - colours
    cmap=cm.get_cmap("copper"),    # colour map
    alpha=0.5   # opaqueness level between 0 and 1
)
plt.xlabel(body_columns[4])
plt.ylabel(body_columns[5])
plt.axis("equal")
plt.rcParams["axes.grid"] = False
cbar = plt.colorbar()
plt.rcParams["axes.grid"] = True
cbar.set_label(body_columns[0])
plt.show()
```

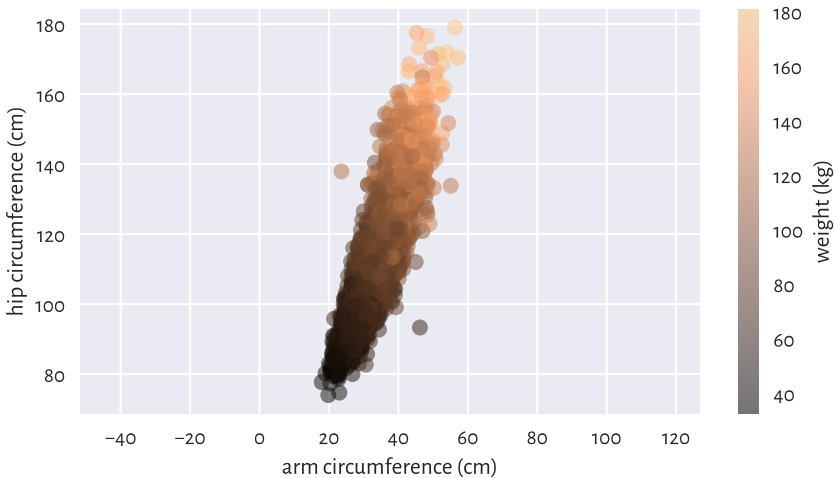

Figure 7.3: A two-dimensional scatter plot displaying three variables

We can see some tendency for the weight to be greater as both the arm and the hip circumferences increase.

**Exercise 7.9** *Play around with different colour palettes. However, be wary that ca. every 1 in 12 men (8%) and 1 in 200 women (0.5%) have colour vision deficiencies, especially in the red-green or blue-yellow spectrum. For this reason, some diverging colour maps might be worse than others.*

A piece of paper is two-dimensional. We only have height and width. Looking around us, we also understand the notion of depth. So far so good. But when the case of more-dimensional data is concerned, well, suffice it to say that we are three-dimensional creatures and any attempts towards visualising them will simply not work, don't even trip.

Luckily, this is where mathematics comes to our rescue. With some more knowledge and intuitions, and this book helps us develop them, it will be as easy[8] as imagining a generic $m$-dimensional space, and then assuming that, say, $m=7$ or 42.

This is exactly why data science relies on automated methods for knowledge/pattern discovery. Thanks to them, we can identify, describe, and analyse the structures that might be present in the data, but cannot be perceived with our imperfect senses.

---

**Note** Linear and nonlinear dimensionality reduction techniques can be applied to visualise some aspects of high-dimensional data in the form of 2D (or 3D) plots. In particular, the principal component analysis (PCA) finds an *interesting* angle from which looking at the data might be worth considering; see Section 9.3.

---

### 7.4.3 Scatterplot Matrix (Pairplot)

We may also try depicting all (or most – ones that we deem interesting) pairs of variables in the form of a scatterplot matrix; see Figure 7.4.

```
sns.pairplot(
    data=pd.DataFrame(  # sns.pairplot needs a DataFrame...
        body[:, [0, 1, 4, 5]],
        columns=body_columns[[0, 1, 4, 5]]
    ),
    plot_kws=dict(alpha=0.1)
)
# plt.show()  # not needed :/
```

Plotting variables against themselves is uninteresting (exercise: what would that be?). This is why we included histograms on the main diagonal to see how they are distributed (the *marginal distributions*).

A scatterplot matrix can be a valuable tool for identifying interesting combinations of columns in our datasets. We see that some pairs of variables are more "structured" than others, e.g., hip circumference and weight are more or less aligned on a straight line. This is why in Chapter 9 we will be interested in describing the possible relationships between the variables.

**Exercise 7.10** (*) *Use* `matplotlib.pyplot.subplot` *and other functions we learned in the previous part to create a scatterplot matrix manually. Draw weight, arm circumference, and hip circumference on a logarithmic scale.*

---

[8] This is an old funny joke that most funny mathematicians find funny. Ha.

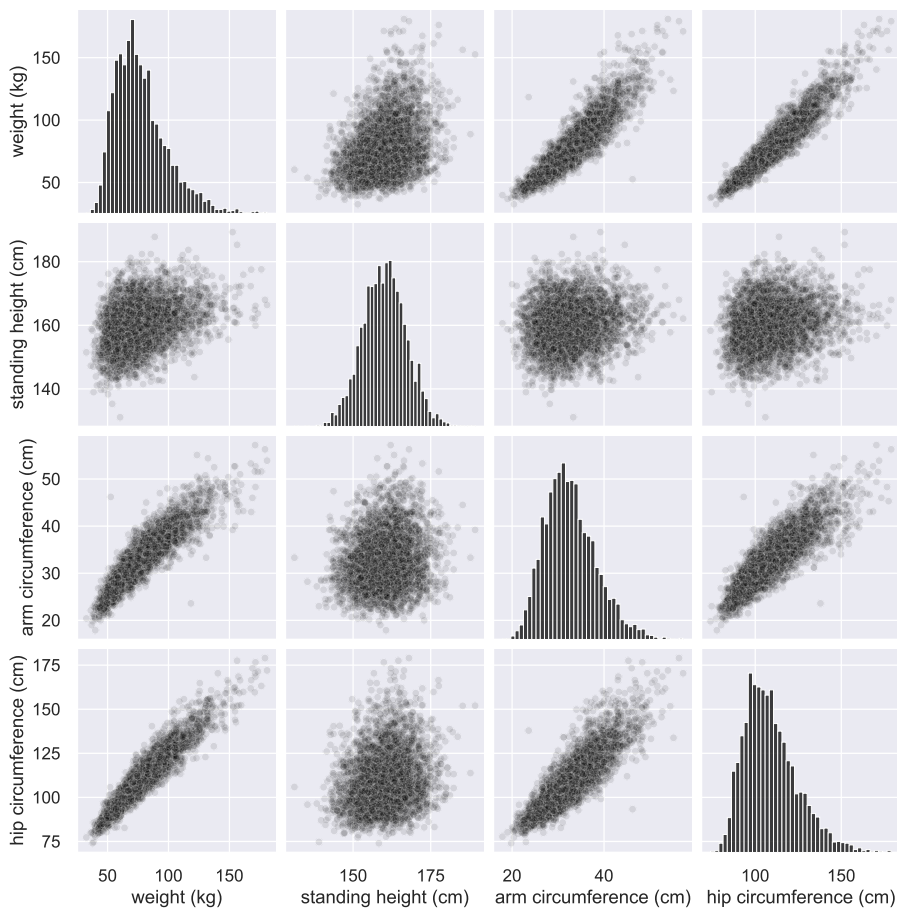

Figure 7.4: Scatterplot matrix for selected columns in the body dataset: scatterplots for all unique pairs of variables together with histograms on the main diagonal

## 7.5 Exercises

**Exercise 7.11** *What is the difference between* `[1, 2, 3]`, `[[1, 2, 3]]`, *and* `[[1], [2], [3]]` *in the context of array creation?*

**Exercise 7.12** *If* `A` *is a matrix with 5 rows and 6 columns, what is the difference between* `A.reshape(6, 5)` *and* `A.T`?

**Exercise 7.13** *If* `A` *is a matrix with 5 rows and 6 columns, what is the meaning of:* `A.reshape(-1)`, `A.reshape(3, -1)`, `A.reshape(-1, 3)`, `A.reshape(-1, -1)`, `A.shape = (3, 10)`, *and* `A.shape = (-1, 3)`?

**Exercise 7.14** *List some methods to add a new row and add a new column to an existing matrix.*

**Exercise 7.15** *Give some ways to visualise three-dimensional data.*

**Exercise 7.16** *How can we set point opaqueness/transparency when drawing a scatter plot? Why would we be interested in this?*

# 8

## *Processing Multidimensional Data*

## 8.1 From Vectors to Matrices

Let us study how the vector operations that we discussed in, amongst others, Chapter 5 can be extended to matrices. In many cases, we will end up applying the same transform either on every matrix element separately, or on each row or column. They are all brilliant examples of the *write less, do more* principle in practice.

### 8.1.1 Vectorised Mathematical Functions

Applying vectorised functions such as `numpy.round`, `numpy.log`, and `numpy.exp` returns an array of the same shape, with all elements transformed accordingly.

```
A = np.array([
    [0.2, 0.6, 0.4, 0.4],
    [0.0, 0.2, 0.4, 0.7],
    [0.8, 0.8, 0.2, 0.1]
]) # example matrix that we will be using below
```

For example:

```
np.square(A)
## array([[0.04, 0.36, 0.16, 0.16],
##        [0.  , 0.04, 0.16, 0.49],
##        [0.64, 0.64, 0.04, 0.01]])
```

takes the square of every element.

More generally, we will be denoting such operations with:

$$f(\mathbf{X}) = \begin{bmatrix} f(x_{1,1}) & f(x_{1,2}) & \cdots & f(x_{1,m}) \\ f(x_{2,1}) & f(x_{2,2}) & \cdots & f(x_{2,m}) \\ \vdots & \vdots & \ddots & \vdots \\ f(x_{n,1}) & f(x_{n,2}) & \cdots & f(x_{n,m}) \end{bmatrix}.$$

## 8.1.2 Componentwise Aggregation

Unidimensional aggregation functions (e.g., `numpy.mean`, `numpy.quantile`) can be applied to summarise:

- all data into a single number (`axis=None`, being the default),
- data in each column (`axis=0`), as well as
- data in each row (`axis=1`).

Here are the examples corresponding to the above cases:

```
np.mean(A)
## 0.39999999999999997
np.mean(A, axis=0)
## array([0.33333333, 0.53333333, 0.33333333, 0.4       ])
np.mean(A, axis=1)
## array([0.4  , 0.325, 0.475])
```

---

**Important** Let us repeat, `axis=1` does not mean that we get the column means (even though columns constitute the 2nd axis, and we count starting at 0). It denotes the axis *along* which the matrix is sliced. Sadly, even yours truly sometimes does not get it right on the first attempt.

---

**Exercise 8.1** *Given the* nhanes_adult_female_bmx_2020[1] *dataset, compute the mean, standard deviation, minimum, and maximum of each body measurement.*

We will get back to the topic of the aggregation of multidimensional data in Section 8.4.

## 8.1.3 Arithmetic, Logical, and Comparison Operations

Recall that for vectors, binary operators such as `+`, `*`, `==`, `<=`, and `&` as well as similar elementwise functions (e.g., `numpy.minimum`) can be applied if both inputs are of the same length, for example:

```
np.array([1, 10, 100, 1000]) * np.array([7, -6, 2, 8])  # elementwisely
## array([   7,  -60,  200, 8000])
```

Alternatively, one input can be a scalar:

```
np.array([1, 10, 100, 1000]) * -3
## array([   -3,   -30,  -300, -3000])
```

---

[1] https://github.com/gagolews/teaching-data/raw/master/marek/nhanes_adult_female_bmx_2020.csv

More generally, a set of rules referred to in the **numpy** manual as *broadcasting*[2] describes how this package handles arrays of different shapes.

---

**Important** Generally, for two matrices, their column/row numbers much match or be equal to 1. Also, if one operand is a one-dimensional array, it will be promoted to a row vector.

---

Let us explore all the possible scenarios.

**Matrix vs Scalar**

If one operand is a scalar, then it is going to be propagated over all matrix elements, for example:

```
(-1)*A
## array([[-0.2, -0.6, -0.4, -0.4],
##        [-0. , -0.2, -0.4, -0.7],
##        [-0.8, -0.8, -0.2, -0.1]])
```

changes the sign of every element, which is, mathematically, an instance of multiplying a matrix $\mathbf{X}$ by a scalar $c$:

$$c\mathbf{X} = \left[ \begin{array}{cccc} cx_{1,1} & cx_{1,2} & \cdots & cx_{1,m} \\ cx_{2,1} & cx_{2,2} & \cdots & cx_{2,m} \\ \vdots & \vdots & \ddots & \vdots \\ cx_{n,1} & cx_{n,2} & \cdots & cx_{n,m} \end{array} \right].$$

Furthermore:

```
A**2
## array([[0.04, 0.36, 0.16, 0.16],
##        [0.  , 0.04, 0.16, 0.49],
##        [0.64, 0.64, 0.04, 0.01]])
```

takes the square[3] of each element. Also:

```
A >= 0.25
## array([[False,  True,  True,  True],
##        [False, False,  True,  True],
##        [ True,  True, False, False]])
```

compares each element to 0.25.

**Matrix vs Matrix**

For two matrices of identical sizes, we act on the *corresponding* elements:

---

[2] https://numpy.org/devdocs/user/basics.broadcasting.html
[3] This is not the same as matrix-multiply by itself which we cover in Section 8.3.

```
B = np.tri(A.shape[0], A.shape[1])  # just an example
B  # a lower triangular 0-1 matrix
## array([[1., 0., 0., 0.],
##        [1., 1., 0., 0.],
##        [1., 1., 1., 0.]])
```

And now:

```
A * B
## array([[0.2, 0. , 0. , 0. ],
##        [0. , 0.2, 0. , 0. ],
##        [0.8, 0.8, 0.2, 0. ]])
```

multiplies each $a_{i,j}$ by the corresponding $b_{i,j}$.

This extends on the idea from algebra that given $\mathbf{A}$, $\mathbf{B}$ with $n$ rows and $m$ columns each, the result of $+$ (or $-$) would be for instance:

$$\mathbf{A} + \mathbf{B} = \begin{bmatrix} a_{1,1} + b_{1,1} & a_{1,2} + b_{1,2} & \cdots & a_{1,m} + b_{1,m} \\ a_{2,1} + b_{2,1} & a_{2,2} + b_{2,2} & \cdots & a_{2,m} + b_{2,m} \\ \vdots & \vdots & \ddots & \vdots \\ a_{n,1} + b_{n,1} & a_{n,2} + b_{n,2} & \cdots & a_{n,m} + b_{n,m} \end{bmatrix}.$$

Thanks to the matrix-matrix and matrix-scalar operations we can perform various tests on a per-element basis, e.g.,

```
(A >= 0.25) & (A <= 0.75)  # logical matrix & logical matrix
## array([[False,  True,  True,  True],
##        [False, False,  True,  True],
##        [False, False, False, False]])
```

**Example 8.2** (*) *Figure 8.1 depicts a (filled) contour plot of the Himmelblau's function,* $f(x, y) = (x^2 + y - 11)^2 + (x + y^2 - 7)^2$, *for* $x \in [-5, 5]$ *and* $y \in [-4, 4]$. *To draw it, we probed 250 points from the two said ranges and called* **numpy.meshgrid** *to generate two matrices, both of shape 250 by 250, giving the x- and y-coordinates of all the points on the corresponding two-dimensional grid. Thanks to this, we were able to use vectorised mathematical operations to compute the values of f thereon.*

```
x = np.linspace(-5, 5, 250)
y = np.linspace(-4, 4, 250)
xg, yg = np.meshgrid(x, y)
z = (xg**2 + yg - 11)**2 + (xg + yg**2 - 7)**2
plt.contourf(x, y, z, levels=20)
CS = plt.contour(x, y, z, levels=[1, 5, 10, 20, 50, 100, 150, 200, 250])
plt.clabel(CS, colors="black")
plt.show()
```

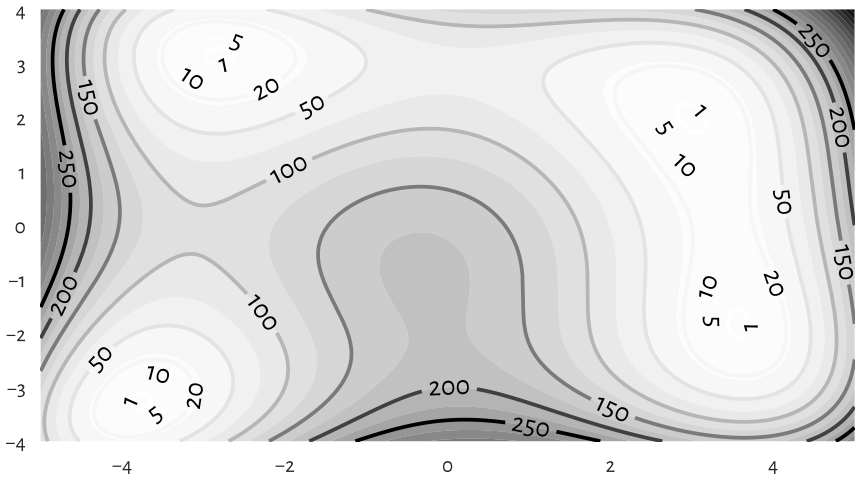

Figure 8.1: An example filled contour plot with additional labelled contour lines

To understand the result generated by **numpy.meshgrid**, here is its output for a smaller number of probe points:

```
x = np.linspace(-5, 5, 3)
y = np.linspace(-4, 4, 5)
xg, yg = np.meshgrid(x, y)
xg
## array([[-5.,  0.,  5.],
##        [-5.,  0.,  5.],
##        [-5.,  0.,  5.],
##        [-5.,  0.,  5.],
##        [-5.,  0.,  5.]])
```

Here, each column is the same.

```
yg
## array([[-4., -4., -4.],
##        [-2., -2., -2.],
##        [ 0.,  0.,  0.],
##        [ 2.,  2.,  2.],
##        [ 4.,  4.,  4.]])
```

In this case, each row is identical. Thanks to this, calling:

```
(xg**2 + yg - 11)**2 + (xg + yg**2 - 7)**2
## array([[116., 306., 296.],
```

```
##          [208., 178., 148.],
##          [340., 170., 200.],
##          [320.,  90., 260.],
##          [340., 130., 520.]])
```

gives a matrix **Z** such that $z_{i,j}$ is generated by considering the i-th element in y and the j-th item in x, which is exactly what we desired.

**Matrix vs Any Vector**

An *n×m* matrix can also be combined with an *n×1* column vector:

```
A * np.array([1, 10, 100]).reshape(-1, 1)
## array([[ 0.2,  0.6,  0.4,  0.4],
##        [ 0. ,  2. ,  4. ,  7. ],
##        [80. , 80. , 20. , 10. ]])
```

The above propagated the column vector over all columns (left to right).

Similarly, combining with a *1×m* row vector:

```
A + np.array([1, 2, 3, 4]).reshape(1, -1)
## array([[1.2, 2.6, 3.4, 4.4],
##        [1. , 2.2, 3.4, 4.7],
##        [1.8, 2.8, 3.2, 4.1]])
```

recycles the row vector over all rows (top to bottom).

If one operand is a one-dimensional array or a list of length *m*, it will be treated as a row vector. For example:

```
np.round(A - np.mean(A, axis=0), 3)  # matrix - vector
## array([[-0.133,  0.067,  0.067, -0.  ],
##        [-0.333, -0.333,  0.067,  0.3 ],
##        [ 0.467,  0.267, -0.133, -0.3 ]])
```

On a side note, this is an instance of *centring* of each column. An explicit `.reshape(1, -1)` was not necessary.

Mathematically, although it is not necessarily a standard notation, we will allow adding and subtracting row vectors from matrices of compatible sizes:

$$\mathbf{X} + \mathbf{t} = \mathbf{X} + [t_1\ t_2\ \cdots\ t_m] = \begin{bmatrix} x_{1,1} + t_1 & x_{1,2} + t_2 & \dots & x_{1,m} + t_m \\ x_{2,1} + t_1 & x_{2,2} + t_2 & \dots & x_{2,m} + t_m \\ \vdots & \vdots & \ddots & \vdots \\ x_{n,1} + t_1 & x_{n,2} + t_2 & \dots & x_{n,m} + t_m \end{bmatrix}.$$

This corresponds to shifting (translating) every row in the matrix.

**Exercise 8.3** *In the* `nhanes_adult_female_bmx_2020`[4] *dataset, standardise, normalise, and min-max scale every column. In every case, a single line of code is sufficient.*

**Row Vector vs Column Vector (\*)**

A row vector combined with a column vector results in an operation's being performed on each *combination* of *all* pairs of elements in the two arrays (i.e., the cross-product; not just the *corresponding* pairs).

```
np.arange(1, 8).reshape(1, -1) * np.array([1, 10, 100]).reshape(-1, 1)
## array([[  1,   2,   3,   4,   5,   6,   7],
##        [ 10,  20,  30,  40,  50,  60,  70],
##        [100, 200, 300, 400, 500, 600, 700]])
```

**Exercise 8.4** *Check out that* **`numpy.nonzero`** *relies on similar shape broadcasting rules as the binary operators we discussed here, but not with respect to all three arguments.*

**Example 8.5** *(\*) Himmelblau's function in Figure 8.1 is only defined by means of arithmetic operators, which all accept the kind of shape broadcasting that we discuss in this section. Consequently, calling* **`numpy.meshgrid`** *in that example to evaluate f on a grid of points was not really necessary:*

```
x = np.linspace(-5, 5, 3)
y = np.linspace(-4, 4, 5)
xg = x.reshape(1, -1)
yg = y.reshape(-1, 1)
(xg**2 + yg - 11)**2 + (xg + yg**2 - 7)**2
## array([[116., 306., 296.],
##        [208., 178., 148.],
##        [340., 170., 200.],
##        [320.,  90., 260.],
##        [340., 130., 520.]])
```

*See also the* `sparse` *parameter in* **`numpy.meshgrid`** *and Figure 12.9 where this function turns out useful after all.*

## 8.1.4 Other Row and Column Transforms (\*)

Some functions that we discussed in the previous part of this course are equipped with the `axis` argument, which allows to process each row or column independently, for example:

```
np.sort(A, axis=1)
## array([[0.2, 0.4, 0.4, 0.6],
```

*(continues on next page)*

```
##         [0. , 0.2, 0.4, 0.7],
##         [0.1, 0.2, 0.8, 0.8]])
```

sorts every row (separately). Moreover:

```
scipy.stats.rankdata(A, axis=0)
## array([[2. , 2. , 2.5, 2. ],
##         [1. , 1. , 2.5, 3. ],
##         [3. , 3. , 1. , 1. ]])
```

computes the ranks of elements in each column.

Some functions have the default argument `axis=-1`, which means that they are applied along the last (i.e., columns in the matrix case) axis:

```
np.diff(A)  # axis=1 here
## array([[ 0.4, -0.2,  0. ],
##         [ 0.2,  0.2,  0.3],
##         [ 0. , -0.6, -0.1]])
```

Still, the aforementioned `numpy.mean` is amongst the many exceptions to this rule.

Compare the above with:

```
np.diff(A, axis=0)
## array([[-0.2, -0.4,  0. ,  0.3],
##         [ 0.8,  0.6, -0.2, -0.6]])
```

which gives the iterated differences for each column separately (along the rows).

If a function (built-in or custom) in not equipped with the `axis` argument and – instead – it was designed to work with individual vectors, we can propagate it over all the rows or columns by calling `numpy.apply_along_axis`.

For instance, here is another (did you solve the suggested exercise?) way to compute the column z-scores:

```
def standardise(x):
    return (x-np.mean(x))/np.std(x)

np.round(np.apply_along_axis(standardise, 0, A), 2)
## array([[-0.39,  0.27,  0.71, -0.  ],
##         [-0.98, -1.34,  0.71,  1.22],
##         [ 1.37,  1.07, -1.41, -1.22]])
```

But of course we prefer `(x - np.mean(x, axis=0))/np.std(x, axis=0)`.

**Note** (*) Matrices are iterable (in the sense of Section 3.4), but in an interesting way. Namely, an iterator traverses through each row in a matrix. Writing:

```
r1, r2, r3 = A  # A has 3 rows
```

creates three variables, each representing a separate row in `A`, the second of which is:

```
r2
## array([0. , 0.2, 0.4, 0.7])
```

## 8.2 Indexing Matrices

Recall that for unidimensional arrays, we have four possible choices of indexers (i.e., where performing filtering like `x[i]`):

- scalar (extracts a single element),
- slice (selects a regular subsequence, e.g., every 2nd element or the first 6 items; returns a *view* on existing data – it does not make an independent copy of the subsetted elements),
- integer vector (selects elements at given indexes),
- logical vector (selects elements that correspond to `True` in the indexer).

Matrices are two-dimensional arrays. Subsetting thereof will require two indexes. We write `A[i, j]` to select rows given by `i` and columns given by `j`. Both `i` and `j` can be one of the four above types, so we have at least 10 different cases to consider (skipping the symmetric ones).

**Important** Generally:

- each scalar index reduces the dimensionality of the subsetted object by 1;
- slice-slice and slice-scalar indexing returns a view on the existing array, so we need to be careful when modifying the resulting object;
- usually, indexing returns a submatrix (subblock), which is a combination of elements at given rows and columns;
- indexing with two integer or logical vectors at the same time should be avoided.

Let us look at all the possible scenarios in greater detail.

## 8.2.1 Slice-Based Indexing

Our favourite example matrix again:

```
A = np.array([
    [0.2, 0.6, 0.4, 0.4],
    [0.0, 0.2, 0.4, 0.7],
    [0.8, 0.8, 0.2, 0.1]
])
```

Indexing based on two slices selects a submatrix:

```
A[::2, 3:]  # every second row, skip the first three columns
## array([[0.4],
##        [0.1]])
```

An empty slice selects all elements on the corresponding axis:

```
A[:, ::-1]  # all rows, reversed columns
## array([[0.4, 0.4, 0.6, 0.2],
##        [0.7, 0.4, 0.2, 0. ],
##        [0.1, 0.2, 0.8, 0.8]])
```

Let us stress that the result is *always* a matrix.

## 8.2.2 Scalar-Based Indexing

Indexing by a scalar selects a given row or column, reducing the dimensionality of the output object.

```
A[:, 3]
## array([0.4, 0.7, 0.1])
```

selects the 4th column and gives a flat vector (we can always use the **reshape** method to convert the resulting object back to a matrix).

Furthermore:

```
A[0, -1]
## 0.4
```

yields the element (scalar) in the first row and the last column.

## 8.2.3 Mixed Boolean/Integer Vector and Scalar/Slice Indexers

A logical and integer vector-like object can also be used for element selection. If the other indexer is a slice or a scalar, the result is quite predictable, for instance:

```
A[ [0, -1, 0], ::-1 ]
## array([[0.4, 0.4, 0.6, 0.2],
##        [0.1, 0.2, 0.8, 0.8],
##        [0.4, 0.4, 0.6, 0.2]])
```

selects the first, the last, and the first row again and reverses the order of columns.

```
A[ A[:, 0] > 0.1, : ]
## array([[0.2, 0.6, 0.4, 0.4],
##        [0.8, 0.8, 0.2, 0.1]])
```

selects the rows from `A` where the values in the first column of `A` are greater than 0.1.

```
A[np.mean(A, axis=1) > 0.35, : ]
## array([[0.2, 0.6, 0.4, 0.4],
##        [0.8, 0.8, 0.2, 0.1]])
```

selects the rows whose mean is greater than 0.35.

```
A[np.argsort(A[:, 0]), : ]
## array([[0. , 0.2, 0.4, 0.7],
##        [0.2, 0.6, 0.4, 0.4],
##        [0.8, 0.8, 0.2, 0.1]])
```

orders the matrix with respect to the values in the first column (all rows permuted in the same way, together).

**Exercise 8.6** *In the `nhanes_adult_female_bmx_2020`[5] dataset, select all the participants whose heights are within their mean ± 2 standard deviations.*

### 8.2.4   Two Vectors as Indexers (*)

With two vectors (logical or integer) things are a tad more horrible, as in this case not only some form of *shape broadcasting* comes into play but also all the headache-inducing exceptions listed in the perhaps not the most clearly written Advanced Indexing[6] section of the **numpy** manual. Cheer up, though: things in **pandas** are much worse; see Section 10.5.

For the sake of our maintaining sanity, in practice, it is best to be extra careful when using two vector indexers and stick only to the scenarios discussed below.

For two flat integer indexers, we pick elementwisely:

---

[5] https://github.com/gagolews/teaching-data/raw/master/marek/nhanes_adult_female_bmx_2020.csv

[6] https://numpy.org/doc/stable/user/basics.indexing.html

```
A[ [0, -1, 0, 2, 0], [1, 2, 0, 2, 1] ]
## array([0.6, 0.2, 0.2, 0.2, 0.6])
```

yields `A[0, 1]`, `A[-1, 2]`, `A[0, 0]`, `A[2, 2]`, and `A[0, 1]`.

To select a submatrix using integer indexes, it is best to make sure that the first indexer is a column vector, and the second one is a row vector (or some objects like these, e.g., compatible lists of lists).

```
A[ [[0], [-1]], [[1, 3]] ]  # column vector-like list, row vector-like list
## array([[0.6, 0.4],
##        [0.8, 0.1]])
```

Further, if indexing involves logical vectors, it is best to convert them to integer ones first (e.g., by calling **numpy.flatnonzero**).

```
A[ np.flatnonzero(np.mean(A, axis=1) > 0.35).reshape(-1, 1), [[0, 2, 3, 0]] ]
## array([[0.2, 0.4, 0.4, 0.2],
##        [0.8, 0.2, 0.1, 0.8]])
```

The necessary reshaping can be done automatically with the **numpy.ix_** function:

```
A[ np.ix_( np.mean(A, axis=1) > 0.35, [0, 2, 3, 0] ) ]  # np.ix_(rows, cols)
## array([[0.2, 0.4, 0.4, 0.2],
##        [0.8, 0.2, 0.1, 0.8]])
```

Alternatively, we can always apply indexing twice instead:

```
A[np.mean(A, axis=1) > 0.45, :][:, [0, 2, 3, 0]]
## array([[0.8, 0.2, 0.1, 0.8]])
```

This is only a mild inconvenience. We will be forced to apply such double indexing anyway in **pandas** whenever selecting rows *by position* and columns *by name* is required; see Section 10.5.

### 8.2.5  Views on Existing Arrays (*)

Only the indexing involving two slices or a slice and a scalar returns a view[7] on an existing array.

For example:

```
B = A[:, ::2]
B
## array([[0.2, 0.4],
```

---
[7] https://numpy.org/devdocs/user/basics.copies.html

```
##         [0. , 0.4],
##         [0.8, 0.2]])
```

Now `B` and `A` share memory. By modifying `B` in place, e.g.:

```
B *= -1
```

the changes will be visible in `A` as well:

```
A
## array([[-0.2,  0.6, -0.4,  0.4],
##         [-0. ,  0.2, -0.4,  0.7],
##         [-0.8,  0.8, -0.2,  0.1]])
```

This is time and memory efficient, but might lead to some unexpected results if we are being rather absent-minded. We have been warned.

### 8.2.6 Adding and Modifying Rows and Columns

With slice/scalar-based indexers, rows/columns/individual elements can be replaced by new content in a natural way:

```
A[:, 0] = A[:, 0]**2
```

With **numpy** arrays, however, brand new rows or columns cannot be added via the index operator. Instead, the whole array needs to be created from scratch using, e.g., one of the functions discussed in Section 7.1.4. For example:

```
A = np.column_stack((A, np.sqrt(A[:, 0])))
A
## array([[ 0.04,  0.6 , -0.4 ,  0.4 ,  0.2 ],
##        [ 0.  ,  0.2 , -0.4 ,  0.7 ,  0.  ],
##        [ 0.64,  0.8 , -0.2 ,  0.1 ,  0.8 ]])
```

## 8.3 Matrix Multiplication, Dot Products, and the Euclidean Norm

Matrix algebra is at the core of all the methods used in data analysis with the matrix multiply being the most fundamental operation therein (e.g., [18, 34]).

Given $\mathbf{A} \in \mathbb{R}^{n \times p}$ and $\mathbf{B} \in \mathbb{R}^{p \times m}$, their *multiply* is a matrix $\mathbf{C} = \mathbf{AB} \in \mathbb{R}^{n \times m}$ such that $c_{i,j}$ is the sum of the $i$-th row in $\mathbf{A}$ and the $j$-th column in $\mathbf{B}$ multiplied element-

wisely:

$$c_{i,j} = a_{i,1}b_{1,j} + a_{i,2}b_{2,j} + \cdots + a_{i,p}b_{p,j} = \sum_{k=1}^{p} a_{i,k}b_{k,j},$$

for $i = 1, \ldots, n$ and $j = 1, \ldots, m$.

For example:

```
A = np.array([
    [1, 0, 1],
    [2, 2, 1],
    [3, 2, 0],
    [1, 2, 3],
    [0, 0, 1],
])
B = np.array([
    [1, 0, 0, 0],
    [0, 4, 1, 3],
    [2, 0, 3, 1],
])
```

And now:

```
C = A @ B   # or: A.dot(B)
C
## array([[ 3,  0,  3,  1],
##        [ 4,  8,  5,  7],
##        [ 3,  8,  2,  6],
##        [ 7,  8, 11,  9],
##        [ 2,  0,  3,  1]])
```

Mathematically, we can write the above as:

$$\begin{bmatrix} 1 & 0 & 1 \\ 2 & 2 & 1 \\ 3 & 2 & 0 \\ \mathbf{1} & \mathbf{2} & \mathbf{3} \\ 0 & 0 & 1 \end{bmatrix} \begin{bmatrix} 1 & 0 & \mathbf{0} & 0 \\ 0 & 4 & \mathbf{1} & 3 \\ 2 & 0 & \mathbf{3} & 1 \end{bmatrix} = \begin{bmatrix} 3 & 0 & 3 & 1 \\ 4 & 8 & 5 & 7 \\ 3 & 8 & 2 & 6 \\ 7 & 8 & \mathbf{11} & 9 \\ 2 & 0 & 3 & 1 \end{bmatrix}.$$

For example, the element in the 4th row and 3rd column, $c_{4,3}$ takes the 4th row in the left matrix $\mathbf{a}_{4,\cdot} = [1\,2\,3]$ and the 3rd column in the right matrix $\mathbf{b}_{\cdot,3} = [0\,1\,3]^T$ (they are marked in bold), multiplies the corresponding elements and computes their sum, i.e., $c_{4,3} = 1 \cdot 0 + 2 \cdot 1 + 3 \cdot 3 = 11$.

---

**Important**  Matrix multiplication can only be performed on two matrices of *compatible*

*sizes* – the number of columns in the left matrix must match the number of rows in the right operand.

Another example:

```
A = np.array([
    [1, 2],
    [3, 4]
])
I = np.array([  # np.eye(2)
    [1, 0],
    [0, 1]
])
A @ I  # or A.dot(I)
## array([[1, 2],
##        [3, 4]])
```

We matrix-multiplied **A** by the identity matrix **I**, which is the neutral element of the said operation. This is why the result is identical to **A**.

---

**Important** In most textbooks, just like in this one, **AB** always denotes the *matrix* multiplication. This is a very different operation from the *elementwise* multiplication.

---

Compare the above to:

```
A * I  # elementwise multiplication
## array([[1, 0],
##        [0, 4]])
```

**Exercise 8.7** (*) *Show that* $(\mathbf{AB})^T = \mathbf{B}^T\mathbf{A}^T$. *Also notice that, typically, matrix multiplication is not commutative.*

---

**Note** By definition, matrix multiplication gives a convenient means for denoting sums of products of corresponding elements in many pairs of vectors, which we refer to as dot products.

Given two vectors $\boldsymbol{x}, \boldsymbol{y} \in \mathbb{R}^p$, their *dot (or scalar) product* is given by:

$$\boldsymbol{x} \cdot \boldsymbol{y} = \sum_{i=1}^{p} x_i y_i.$$

In matrix multiplication terms, if **x** is a row vector and $\mathbf{y}^T$ is a column vector, then the above can be written as $\mathbf{x}\mathbf{y}^T$. The result is a single number.

In particular, a dot product of a vector and itself:

$$x \cdot x = \sum_{i=1}^{p} x_i^2,$$

is the square of the Euclidean norm of $x$, which – as we said in Section 5.3.2 – is used to measure the *magnitude* of a vector:

$$\|x\| = \sqrt{\sum_{i=1}^{p} x_i^2} = \sqrt{x \cdot x} = \sqrt{xx^T}.$$

It is worth pointing out that the Euclidean norm fulfils (amongst others) the condition that $\|x\| = 0$ if and only if $x = \mathbf{0} = (0, 0, \dots, 0)$. The same of course holds for its square.

**Exercise 8.8**  *Show that* $A^T A$ *gives the matrix that consists of the dot products of all the pairs of columns in* $A$ *and* $AA^T$ *stores the dot products of all the pairs of rows.*

In Section 9.3.2, we will see that matrix multiplication can be used as a way to express certain geometrical transformations of points in a dataset, e.g., scaling and rotating.

Also, in Section 9.3.3, we briefly discuss the concept of the inverse of a matrix and in Section 9.3.4, we introduce its singular value decomposition.

## 8.4  Pairwise Distances and Related Methods

Many data analysis methods rely on the notion of *distances* between points, which quantify the extent to which two points (e.g., two rows in a matrix) are different from each other. Here we will be dealing with the most natural[8] distance called the Euclidean metric. We know it from school, where we measured how two points are far away from each other using a ruler.

### 8.4.1  The Euclidean Metric

Given two points in $\mathbb{R}^m$, $u = (u_1, \dots, u_m)$ and $v = (v_1, \dots, v_m)$, the *Euclidean metric* is defined in terms of the corresponding Euclidean norm:

$$\|u - v\| = \sqrt{(u_1 - v_1)^2 + (u_2 - v_2)^2 + \cdots + (u_m - v_m)^2} = \sqrt{\sum_{i=1}^{m}(u_i - v_i)^2},$$

that is, it is the square root of the sum of squared differences between the corresponding coordinates.

---

[8] There are many possible distances, allowing to measure the similarity of points not only in $\mathbb{R}^m$, but also character strings (e.g., the Levenshtein metric), ratings (e.g., cosine dissimilarity), etc.; there is even an encyclopedia of distances [21].

**Important**  Given two vectors of equal lengths $x, y \in \mathbb{R}^m$, the dot product of their difference:

$$(\boldsymbol{x} - \boldsymbol{y}) \cdot (\boldsymbol{x} - \boldsymbol{y}) = (\mathbf{x} - \mathbf{y})(\mathbf{x} - \mathbf{y})^T = \sum_{i=1}^{m} (x_i - y_i)^2,$$

is nothing else than the square of the Euclidean distance between them.

In particular, for unidimensional data ($m = 1$), we have $\|\boldsymbol{u} - \boldsymbol{v}\| = |u_1 - v_1|$, i.e., the absolute value of the difference.

**Exercise 8.9**  *Consider the following matrix* $\mathbf{X} \in \mathbb{R}^{4 \times 2}$:

$$\mathbf{X} = \begin{bmatrix} 0 & 0 \\ 1 & 0 \\ -\frac{3}{2} & 1 \\ 1 & 1 \end{bmatrix}.$$

*Calculate (by hand):* $\|\mathbf{x}_{1,\cdot} - \mathbf{x}_{2,\cdot}\|$, $\|\mathbf{x}_{1,\cdot} - \mathbf{x}_{3,\cdot}\|$, $\|\mathbf{x}_{1,\cdot} - \mathbf{x}_{4,\cdot}\|$, $\|\mathbf{x}_{2,\cdot} - \mathbf{x}_{4,\cdot}\|$, $\|\mathbf{x}_{2,\cdot} - \mathbf{x}_{3,\cdot}\|$, $\|\mathbf{x}_{1,\cdot} - \mathbf{x}_{1,\cdot}\|$, *and* $\|\mathbf{x}_{2,\cdot} - \mathbf{x}_{1,\cdot}\|$.

The distances between all the possible pairs of rows in two matrices $\mathbf{X} \in \mathbb{R}^{n \times m}$ and $\mathbf{Y} \in \mathbb{R}^{k \times m}$ can be computed by calling `scipy.spatial.distance.cdist`. We need to be careful, though, because they result in a distance matrix of size $n \times k$, which can become quite large (e.g., for $n = k = 100{,}000$ we would need ca. 80 GB of RAM to store it).

Here are the distances between all the pairs of points in the same dataset.

```
X = np.array([
    [0,    0],
    [1,    0],
    [-1.5, 1],
    [1,    1]
])
import scipy.spatial.distance
D = scipy.spatial.distance.cdist(X, X)
D
## array([[0.        , 1.        , 1.80277564, 1.41421356],
##        [1.        , 0.        , 2.6925824 , 1.        ],
##        [1.80277564, 2.6925824 , 0.        , 2.5        ],
##        [1.41421356, 1.        , 2.5        , 0.        ]])
```

Hence, $d_{i,j} = \|\mathbf{x}_{i,\cdot} - \mathbf{x}_{j,\cdot}\|$. That we have zeros on the diagonal is due to the fact that $\|\boldsymbol{u} - \boldsymbol{v}\| = 0$ if and only if $\boldsymbol{u} = \boldsymbol{v}$. Furthermore, $\|\boldsymbol{u} - \boldsymbol{v}\| = \|\boldsymbol{v} - \boldsymbol{u}\|$, which implies the symmetry of $\mathbf{D}$, i.e., it holds $\mathbf{D}^T = \mathbf{D}$

Figure 8.2 illustrates all six non-trivial pairwise distances. Let us emphasise that our perception of distance is disturbed because the aspect ratio (the ratio between the range of the x-axis to the range of the y-axis) is not 1:1. This is why it is very important, when judging spatial relationships between the points, to call `matplotlib.pyplot.axis("equal")` or set the axis limits manually (which is left as an exercise).

```
plt.plot(X[:, 0], X[:, 1], "ko")
for i in range(X.shape[0]-1):
    for j in range(i+1, X.shape[0]):
        plt.plot(X[[i,j], 0], X[[i,j], 1], "k-", alpha=0.2)
        plt.text(
            np.mean(X[[i,j], 0]),
            np.mean(X[[i,j], 1]),
            np.round(D[i, j], 2)
        )
plt.show()
```

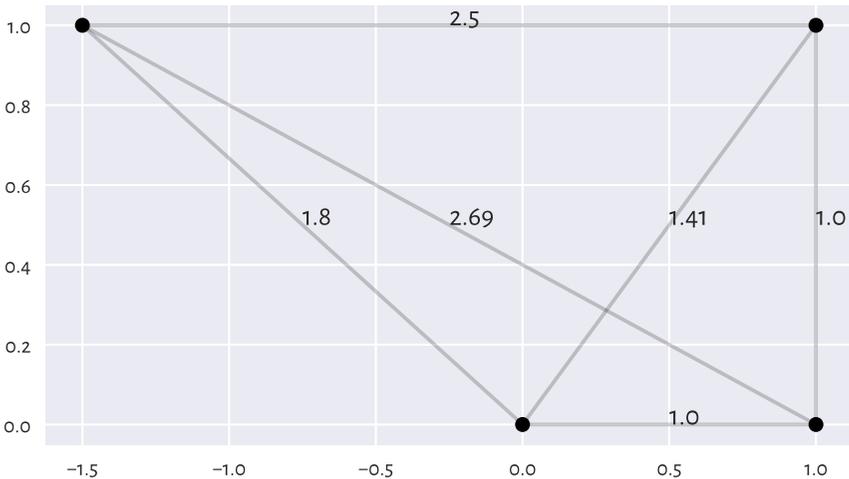

Figure 8.2: Distances between four example points; their perception is disturbed because the aspect ratio is not 1:1

---

**Important**  Some popular techniques in data science rely on computing pairwise distances, including:

- multidimensional data aggregation (see below),

- $k$-means clustering (Section 12.4),

- $k$-nearest neighbour regression (Section 9.2.1) and classification (Section 12.3.1),

- missing value imputation (Section 15.1),

- density estimation (which we can use outlier detection, see Section 15.4).

In the sequel, whenever we apply them, we will be assuming that data have been appropriately preprocessed: in particular, that columns are on the same scale (e.g., are standardised). Otherwise, computing sums of their squared differences might not make sense at all.

---

## 8.4.2 Centroids

So far we have been only discussing ways to aggregate unidimensional data (for instance, each matrix column separately). It turns out that some summaries can be generalised to the multidimensional case.

For instance, it can be shown that the arithmetic mean of a vector $(x_1, \dots, x_n)$ is a point $c$ that minimises the sum of the *squared* unidimensional distances between itself and all the $x_i$s, i.e., $\sum_{i=1}^{n} \|x_i - c\|^2 = \sum_{i=1}^{n} (x_i - c)^2$.

We can define the *centroid* of a dataset $\mathbf{X} \in \mathbb{R}^{n \times m}$ as the point $c \in \mathbb{R}^m$ to which the overall *squared* distance is the smallest:

$$\text{minimise} \sum_{i=1}^{n} \|\mathbf{x}_{i,\cdot} - c\|^2 \qquad \text{w.r.t. } c.$$

It can be shown that the solution to the above is:

$$c = \frac{1}{n} \left( \mathbf{x}_{1,\cdot} + \mathbf{x}_{2,\cdot} + \cdots + \mathbf{x}_{n,\cdot} \right) = \frac{1}{n} \sum_{i=1}^{n} \mathbf{x}_{i,\cdot},$$

which is the componentwise arithmetic mean, i.e., its $j$-th component is:

$$c_j = \frac{1}{n} \sum_{i=1}^{n} x_{i,j}.$$

For instance, the centroid of the dataset depicted in Figure 8.2 is:

```
c = np.mean(X, axis=0)
c
## array([0.125, 0.5  ])
```

Centroids are, amongst others, a basis for the $k$-means clustering method that we discuss in Section 12.4.

## 8.4.3 Multidimensional Dispersion and Other Aggregates

Furthermore, as a measure of multidimensional dispersion, we can consider the natural generalisation of the standard deviation:

$$s = \sqrt{\frac{1}{n} \sum_{i=1}^{n} \|\mathbf{x}_{i,\cdot} - c\|^2},$$

being the square root of the average squared distance to the centroid. Notice that $s$ is a single number.

```
np.sqrt(np.mean(scipy.spatial.distance.cdist(X, c.reshape(1, -1))**2))
## 1.1388041973930374
```

---

**Note**    (**) Generalising other aggregation functions is not a trivial task, because, amongst others, there is no natural linear ordering relation in the multidimensional space (see, e.g., [67]). For instance, any point on the convex hull of a dataset could serve as an analogue of the minimal and maximal observation.

Furthermore, the componentwise median does not behave nicely (it may, for example, fall outside the convex hull). Instead, we usually consider a different generalisation of the median: the point $\boldsymbol{m}$ which minimises the sum of distances (not squared), $\sum_{i=1}^{n} \|\mathbf{x}_{i,\cdot} - \boldsymbol{m}\|$. Sadly, it does not have an analytic solution, but it can be determined algorithmically.

---

**Note** (**) A bag plot [71] is one of the possible multidimensional generalisations of the box-and-whisker plot. Unfortunately, its use is quite limited due to its low popularity amongst practitioners.

---

### 8.4.4    Fixed-Radius and $K$-Nearest Neighbour Search

Several data analysis techniques rely upon aggregating information about what is happening in the *local neighbourhoods* of the points. Let $\mathbf{X} \in \mathbb{R}^{n \times m}$ be a dataset and $\boldsymbol{x}' \in \mathbb{R}^m$ be some point, not necessarily from $\mathbf{X}$. We have two options:

- *fixed-radius search*: for some radius $r > 0$, we seek the indexes of all the points in $\mathbf{X}$ whose distance to $\boldsymbol{x}'$ is not greater than $r$:

$$B_r(\boldsymbol{x}') = \left\{ i : \|\mathbf{x}_{i,\cdot} - \boldsymbol{x}'\| \leq r \right\};$$

- *few nearest neighbour search*: for some (usually small) integer $k \geq 1$, we seek the indexes of the $k$ points in $\mathbf{X}$ which are the closest to $\boldsymbol{x}'$:

$$N_k(\boldsymbol{x}') = \{i_1, i_2, \ldots, i_k\},$$

such that for all $j \notin \{i_1, \ldots, i_k\}$:

$$\|\mathbf{x}_{i_1,\cdot} - \boldsymbol{x}'\| \leq \|\mathbf{x}_{i_2,\cdot} - \boldsymbol{x}'\| \leq \ldots \leq \|\mathbf{x}_{i_k,\cdot} - \boldsymbol{x}'\| \leq \|\mathbf{x}_{j,\cdot} - \boldsymbol{x}'\|.$$

---

**Important**  In $\mathbb{R}^1$, $B_r(\boldsymbol{x}')$ is an interval of length $2r$ centred at $\boldsymbol{x}'$, i.e., $[x_1' - r, x_1' + r]$.

In $\mathbb{R}^2$, $B_r(\boldsymbol{x}')$ is a circle of radius $r$ centred at $(x_1', x_2')$. More generally, we call $B_r(\boldsymbol{x})$ an $m$-dimensional (Euclidean) ball or a solid hypersphere.

---

Here is an example dataset, consisting of some randomly generated points (see Figure 8.3).

```
np.random.seed(777)
X = np.random.randn(25, 2)
x_test = np.array([0, 0])
```

Local neighbourhoods can of course be determined by computing the distances between each point in $\mathbf{X}$ and $\boldsymbol{x}'$.

```
import scipy.spatial.distance
D = scipy.spatial.distance.cdist(X, x_test.reshape(1, -1))
```

For instance, here are the indexes of the points in $B_{0.75}(\boldsymbol{x}')$:

```
r = 0.75
B = np.flatnonzero(D <= r)
B
## array([ 1, 11, 14, 16, 24])
```

And here are the 11 nearest neighbours, $N_{11}(\boldsymbol{x}')$:

```
k = 11
N = np.argsort(D.reshape(-1))[:k]
N
## array([14, 24, 16, 11,  1, 22,  7, 19,  0,  9, 15])
```

See Figure 8.3 for an illustration (observe that the aspect ratio is set to 1:1 as otherwise the circle would look like an ellipse).

```
fig, ax = plt.subplots()
ax.add_patch(plt.Circle(x_test, r, color="red", alpha=0.1))
for i in range(k):
    plt.plot(
        [x_test[0], X[N[i], 0]],
        [x_test[1], X[N[i], 1]],
        "r:", alpha=0.4
    )
plt.plot(X[:, 0], X[:, 1], "bo", alpha=0.1)
for i in range(X.shape[0]):
    plt.text(X[i, 0], X[i, 1], str(i), va="center", ha="center")
plt.plot(x_test[0], x_test[1], "rX")
plt.text(x_test[0], x_test[1], "$\\mathbf{x}'$", va="center", ha="center")
```

*(continued from previous page)*

```
plt.axis("equal")
plt.show()
```

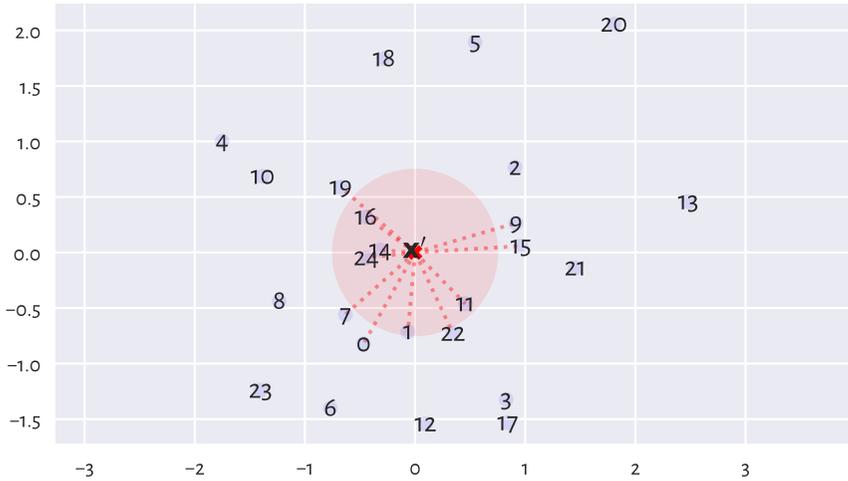

Figure 8.3: Fixed-radius vs few nearest neighbour search

### 8.4.5   Spatial Search with *K*-d Trees

For efficiency reasons, it is better to rely on dedicated spatial search data structures, especially if we have a large number of neighbourhood-related queries. **scipy** implements such a search algorithm based on the so-called *K*-dimensional trees (*K*-d trees[9]).

---

**Note** (*) In *K*-d trees, the data space is partitioned into hyperrectangles along the axes of the Cartesian coordinate system (standard basis). Thanks to such a representation, all subareas which are too far from the point of interest can be pruned to speed up the search.

---

Let us create the data structure for searching within the above **X** matrix.

```
import scipy.spatial
T = scipy.spatial.KDTree(X)
```

Assume we would like to make queries with regard to the 3 following pivot points.

---

[9] In our context, we should prefer referring to them as *m*-d trees, but let us stick with the traditional name.

```
X_test = np.array([
    [0,  0],
    [2,  2],
    [2, -2]
])
```

Here are the results for the fixed radius searches ($r = 0.75$):

```
T.query_ball_point(X_test, 0.75)
## array([list([1, 11, 14, 16, 24]), list([20]), list([])], dtype=object)
```

We see that the search was nicely vectorised: we made a query about three points at the same time. As a result, we received a list-like object storing three lists representing the indexes of interest. Note that in the case of the 3rd point, there are no elements in **X** within the range (ball) of interest, hence the empty index list.

And here are the 5 nearest neighbours:

```
distances, indexes = T.query(X_test, 5)  # returns a tuple of length 2
```

We obtained both the distances to the nearest neighbours:

```
distances
## array([[0.31457701, 0.44600012, 0.54848109, 0.64875661, 0.71635172],
##        [0.20356263, 1.45896222, 1.61587605, 1.64870864, 2.04640408],
##        [1.2494805 , 1.35482619, 1.93984334, 1.95938464, 2.08926502]])
```

as well as the indexes:

```
indexes
## array([[14, 24, 16, 11,  1],
##        [20,  5, 13,  2,  9],
##        [17,  3, 21, 12, 22]])
```

Each of them is a matrix with 3 rows (corresponding to the number of pivot points) and 5 columns (the number of neighbours sought).

---

**Note** (*) We expect the $K$-d trees to be much faster than the brute-force approach (where we compute all pairwise distances) in low-dimensional spaces. Nonetheless, due to the phenomenon called the *curse of dimensionality*, sometimes already for $m \geq 5$ the speed gains might be very small; see, e.g., [7].

---

## 8.5 Exercises

**Exercise 8.10** *Does* `numpy.mean(A, axis=0)` *compute rowwise or columnwise means?*

**Exercise 8.11** *How does shape broadcasting work? List the most common pairs of shape cases when performing arithmetic operations like addition or multiplication.*

**Exercise 8.12** *What are the possible matrix indexing schemes and how do they behave?*

**Exercise 8.13** *Which kinds of matrix indexers return a view on an existing array?*

**Exercise 8.14** *(\*) How can we select a submatrix comprised of the first and the last row and the first and the last column?*

**Exercise 8.15** *Why appropriate data preprocessing is required when computing the Euclidean distance between points?*

**Exercise 8.16** *What is the relationship between the dot product, the Euclidean norm, and the Euclidean distance?*

**Exercise 8.17** *What is a centroid? How is it defined by means of the Euclidean distance between the points in a dataset?*

**Exercise 8.18** *What is the difference between the fixed-radius and few nearest-neighbours search?*

**Exercise 8.19** *(\*) When K-d trees or other spatial search data structures might be better than a brute-force search with* `scipy.spatial.distance.cdist`?

# *Exploring Relationships Between Variables*

Let us go back to National Health and Nutrition Examination Survey (NHANES study) excerpt that we were playing with in Section 7.4:

```python
body = pd.read_csv("https://raw.githubusercontent.com/gagolews/" +
    "teaching-data/master/marek/nhanes_adult_female_bmx_2020.csv",
    comment="#")
body = np.array(body)  # convert to matrix
body[:6, :]  # preview: 6 first rows, all columns
## array([[ 97.1, 160.2,  34.7,  40.8,  35.8, 126.1, 117.9],
##        [ 91.1, 152.7,  33.5,  33. ,  38.5, 125.5, 103.1],
##        [ 73. , 161.2,  37.4,  38. ,  31.8, 106.2,  92. ],
##        [ 61.7, 157.4,  38. ,  34.7,  29. , 101. ,  90.5],
##        [ 55.4, 154.6,  34.6,  34. ,  28.3,  92.5,  73.2],
##        [ 62. , 144.7,  32.5,  34.2,  29.8, 106.7,  84.8]])
body.shape
## (4221, 7)
```

We thus have $n = 4221$ participants and 7 different features describing them, in this order:

```python
body_columns = np.array([
    "weight",
    "height",
    "arm len",
    "leg len",
    "arm circ",
    "hip circ",
    "waist circ"
])
```

We expect the data in different columns to be *related* to each other (e.g., a taller person *usually tends to* weight more). This is why we will now be interested in quantifying the degree of association between the variables, modelling the possible functional relationships, and finding new interesting combinations of columns.

## 9.1 Measuring Correlation

Scatterplots let us identify some simple patterns or structure in data. Figure 7.4 indicates that higher hip circumferences *tend to* occur more often together with higher arm circumferences and that the latter does not really tell us much about height.

Let us explore some basic means for measuring (expressing as a single number) the degree of association between a set of pairs of points.

### 9.1.1 Pearson's Linear Correlation Coefficient

The Pearson *linear correlation* coefficient is given by:

$$r(\boldsymbol{x}, \boldsymbol{y}) = \frac{1}{n} \sum_{i=1}^{n} \frac{x_i - \bar{x}}{s_x} \frac{y_i - \bar{y}}{s_y},$$

with $s_x, s_y$ denoting the standard deviations and $\bar{x}, \bar{y}$ being the means of $\boldsymbol{x} = (x_1, \ldots, x_n)$ and $\boldsymbol{y} = (y_1, \ldots, y_n)$, respectively.

---

**Note** Look carefully: we are computing the mean of the pairwise products of standardised versions of the two vectors. It is a normalised measure of how they *vary together* (co-variance).

(\*) Furthermore, in Section 9.3.1, we mention that $r$ is the cosine of the angle between centred and normalised versions of the vectors.

---

Here is how we can compute it manually:

```
x = body[:, 4]  # arm circumference
y = body[:, 5]  # hip circumference
x_std = (x-np.mean(x))/np.std(x)  # z-scores for x
y_std = (y-np.mean(y))/np.std(y)  # z-scores for y
np.mean(x_std*y_std)
## 0.8680627457873239
```

And here is a built-in function (for the lazy, in a good sense) that implements the same formula:

```
scipy.stats.pearsonr(x, y)[0]
## 0.8680627457873241
```

Note the [0] part: the function returns more than we actually need.

---

**Important** Basic properties of Pearson's $r$ include:

1. $r(\boldsymbol{x}, \boldsymbol{y}) = r(\boldsymbol{y}, \boldsymbol{x})$ (symmetric);

2. $|r(\boldsymbol{x}, \boldsymbol{y})| \le 1$ (bounded from below by -1 and from above by 1);

3. $r(\boldsymbol{x}, \boldsymbol{y}) = 1$ if and only if $\boldsymbol{y} = a\boldsymbol{x} + b$ for some $a > 0$ and any $b$, (reaches the maximum when one variable is an increasing linear function of the other one);

4. $r(\boldsymbol{x}, -\boldsymbol{y}) = -r(\boldsymbol{x}, \boldsymbol{y})$ (negative scaling (reflection) of one variable changes the sign of the coefficient);

5. $r(\boldsymbol{x}, a\boldsymbol{y} + b) = r(\boldsymbol{x}, \boldsymbol{y})$ for any $a > 0$ and $b$ (invariant to translation and scaling of inputs that does not change the sign of elements).

---

To get more insight, below we shall illustrate some interesting *correlations* using the following function that draws a scatter plot and prints out Pearson's *r* (and Spearman's $\rho$ which we discuss in Section 9.1.4; let us ignore it by then):

```python
def plot_corr(x, y, axes_eq=False):
    r = scipy.stats.pearsonr(x, y)[0]
    rho = scipy.stats.spearmanr(x, y)[0]
    plt.plot(x, y, "o")
    plt.title(f"$r = {r:.3}$, $\\rho = {rho:.3}$",
        fontdict=dict(fontsize=10))
    if axes_eq: plt.axis("equal")
```

**Perfect Linear Correlation**

The aforementioned properties imply that $r(\boldsymbol{x}, \boldsymbol{y}) = -1$ if and only if $\boldsymbol{y} = a\boldsymbol{x} + b$ for some $a < 0$ and any $b$ (reaches the minimum when one variable is a decreasing linear function of the other one) Furthermore, a variable is trivially perfectly correlated with itself, $r(\boldsymbol{x}, \boldsymbol{x}) = 1$.

Consequently, we get perfect *linear correlation* (-1 or 1) when one variable is a scaled and shifted version (linear function) of the other variable; see Figure 9.1.

```python
x = np.random.rand(100)
plt.subplot(1, 2, 1); plot_corr(x, -0.5*x+3,  axes_eq=True) # negative slope
plt.subplot(1, 2, 2); plot_corr(x,    3*x+10, axes_eq=True) # positive slope
plt.show()
```

A negative correlation means that when one variable increases, the other one decreases (like: a car's braking distance vs velocity).

**Strong Linear Correlation**

Next, if two variables are *more or less* linear functions of themselves, the correlations will be close to -1 or 1, with the degree of association diminishing as the linear relationship becomes less and less evident; see Figure 9.2.

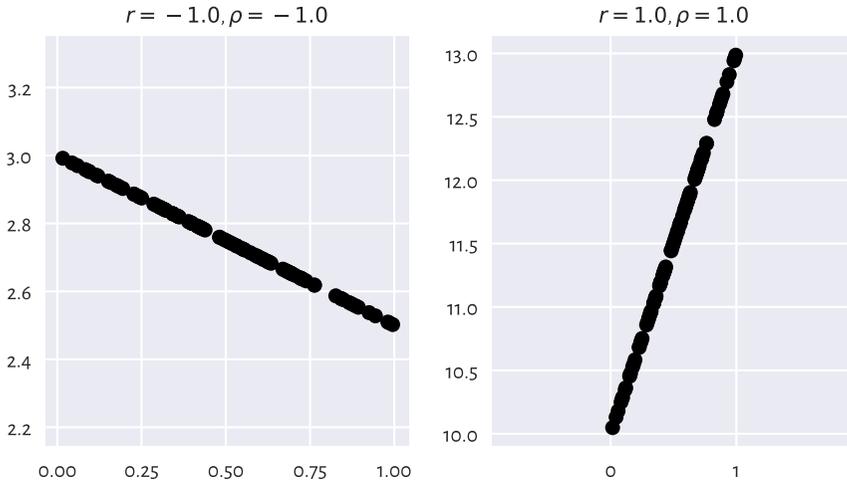

Figure 9.1: Perfect linear correlation (negative and positive)

```
x = np.random.rand(100)      # random x (whatever)
y = 0.5*x                     # y is a linear function of x
e = np.random.randn(len(x))  # random Gaussian noise (expected value 0)
plt.subplot(2, 2, 1); plot_corr(x, y)          # original y
plt.subplot(2, 2, 2); plot_corr(x, y+0.05*e)   # some noise added to y
plt.subplot(2, 2, 3); plot_corr(x, y+0.1*e)    # more noise
plt.subplot(2, 2, 4); plot_corr(x, y+0.25*e)   # even more noise
plt.show()
```

Notice again that the arm and hip circumferences enjoy quite high positive degree of linear correlation. Their scatterplot (Figure 7.4) looks somewhat similar to one of the cases presented here.

**Exercise 9.1** *Draw a series of similar plots but for the case of negatively correlated point pairs, e.g., $y = -2x + 5$.*

---

**Important**   As a rule of thumb, linear correlation degree of 0.9 or greater (or -0.9 or smaller) is quite decent. Between -0.8 and 0.8 we probably should not be talking about two variables being linearly correlated at all. Some textbooks are more lenient, but we have higher standards. In particular, it is not uncommon in social sciences to consider 0.6 a decent degree of correlation, but this is like building on sand. If a dataset at hand does not provide us with strong evidence, it is our ethical duty to refrain ourselves from making unjustified statements.

---

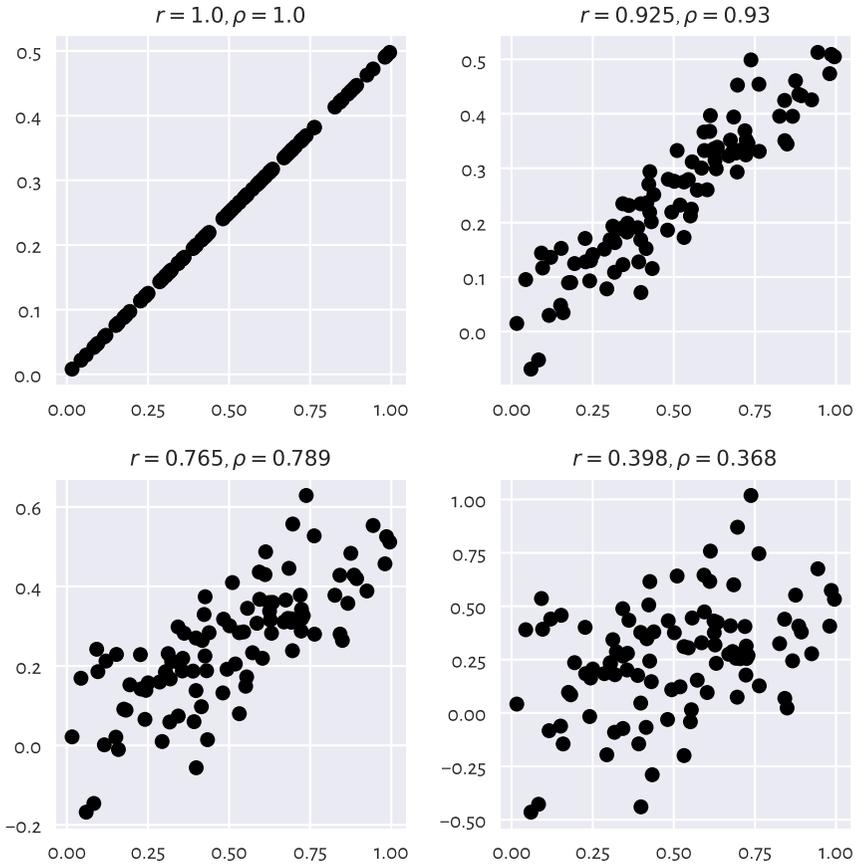

Figure 9.2: Linear correlation coefficients for data with different amounts of noise

**No Linear Correlation Does Not Imply Independence**

For two independent variables, we expect the correlation coefficient be approximately equal to 0. Nevertheless, we should stress that correlation close to 0 does not necessarily mean that two variables are unrelated to each other. Pearson's $r$ is a *linear* correlation coefficient, so we are quantifying only[1] these types of relationships. See Figure 9.3 for an illustration of this fact.

```
plt.subplot(2, 2, 1)
plot_corr(x, np.random.rand(100))  # independent (not correlated)
plt.subplot(2, 2, 2)
```

*(continues on next page)*

---

[1] Note that in Section 6.2.3, we were also testing one concrete hypothesis: whether a distribution was normal or whether it was anything else. We only know that if the data really follow that distribution, the null hypothesis will not be rejected in 0.1% of the cases. The rest is silence.

```
plot_corr(x, (2*x-1)**2-1)        # quadratic dependence
plt.subplot(2, 2, 3)
plot_corr(x, np.abs(2*x-1))       # absolute value
plt.subplot(2, 2, 4)
plot_corr(x, np.sin(10*np.pi*x))  # sine
plt.show()
```

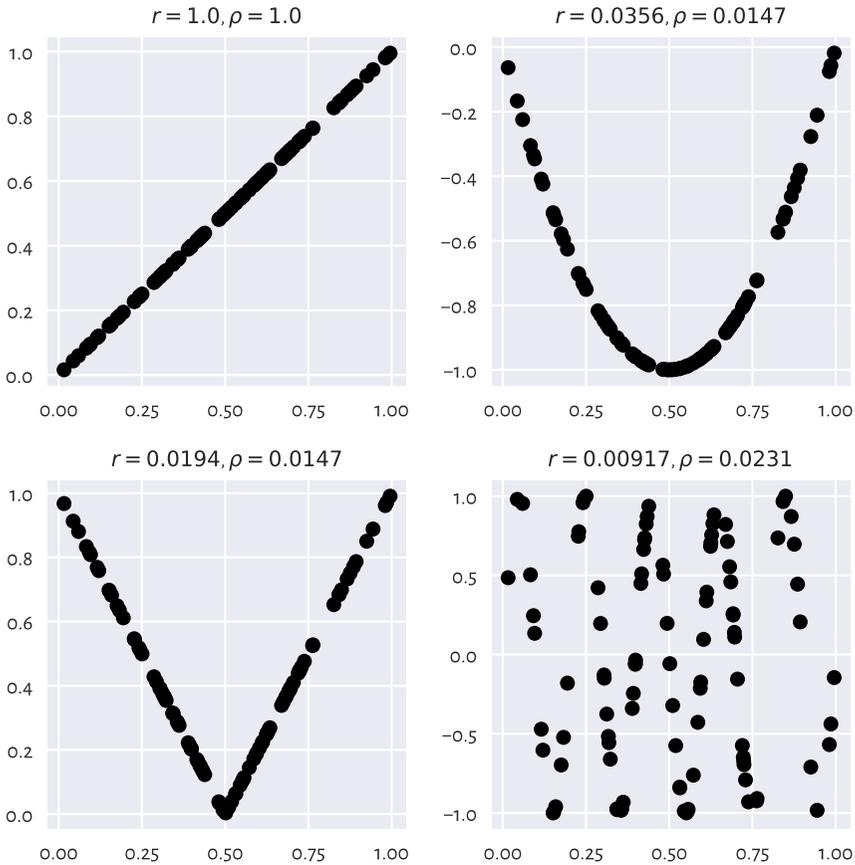

Figure 9.3: Are all of these really uncorrelated?

## False Linear Correlations

What is more, sometimes we can detect *false* correlations – when data are functionally dependent, the relationship is not linear, but it kind of looks like linear. Refer to Figure 9.4 for some examples.

```python
plt.subplot(2, 2, 1)
plot_corr(x, np.sin(0.6*np.pi*x))   # sine
plt.subplot(2, 2, 2)
plot_corr(x, np.log(x+1))           # logarithm
plt.subplot(2, 2, 3);
plot_corr(x, np.exp(x**2))          # exponential of square
plt.subplot(2, 2, 4)
plot_corr(x, 1/(x/2+0.2))           # reciprocal
plt.show()
```

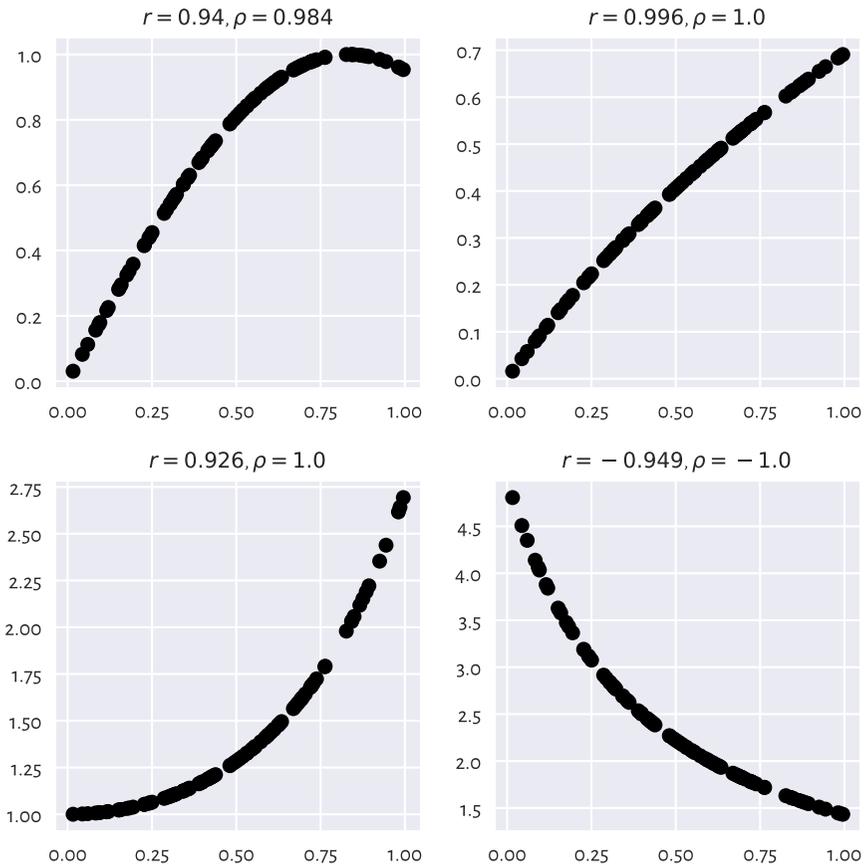

Figure 9.4: Example non-linear relationships that look like linear, at least to Pearson's $r$

No single measure is perfect – we are trying to compress $2n$ data points into a single number — it is obvious that there will be many different datasets, sometimes remarkably diverse, that will yield the same correlation value.

**Correlation Is Not Causation**

A high correlation degree (either positive or negative) does not mean that there is any *causal* relationship between the two variables. We cannot say that having large arm circumference affects hip size or the other way around. There might be some *latent* variable that influences these two (e.g., maybe also related to weight?).

**Exercise 9.2** *Quite often, medical advice is formulated based on correlations and similar association-measuring tools. We should know how to interpret them, as it is never a true cause-effect relationship; rather, it is all about detecting common patterns in larger populations. For instance, in "obesity increases the likelihood of lower back pain and diabetes" we do not say that one necessarily* implies *another or that if you are not overweight, there is no risk of getting the two said conditions. It might also work the other way around, as lower back pain may lead to less exercise and then weight gain. Reality is complex. Find similar patterns in sets of health conditions.*

---

**Note**   Measuring correlations can aid in constructing regression models, where we would like to identify the transformation that expresses a variable as a function of one or more other ones. When we say that *y* can be modelled approximately by $ax + b$, regression analysis can identify the best matching *a* and *b* coefficients; see Section 9.2.3 for more details.

---

### 9.1.2   Correlation Heatmap

Calling **numpy.corrcoef**(body.T) (note the matrix transpose) allows for determining the linear correlation coefficients between all pairs of variables.

We can depict them nicely on a heatmap; see Figure 9.5.

```python
from matplotlib import cm
order = [4, 5, 6, 0, 2, 1, 3]
R = np.corrcoef(body.T)
sns.heatmap(
    R[np.ix_(order, order)],
    xticklabels=body_columns[order],
    yticklabels=body_columns[order],
    annot=True, fmt=".2f", cmap=cm.get_cmap("copper")
)
plt.show()
```

Notice that we ordered[2] the columns to reveal some naturally occurring variable *clusters*: for instance, arm, hip, waist circumference and weight are all quite strongly correlated.

Of course, we have 1.0s on the main diagonal because a variable is trivially correl-ated with itself. Interestingly, this heatmap is symmetric which is due to the property $r(\boldsymbol{x}, \boldsymbol{y}) = r(\boldsymbol{y}, \boldsymbol{x})$.

---

[2] (**) This can be done automatically via some hierarchical clustering algorithm applied onto the transformed correlation matrix, $1 - |\mathbf{R}|$ or $1 - \mathbf{R}^2$.

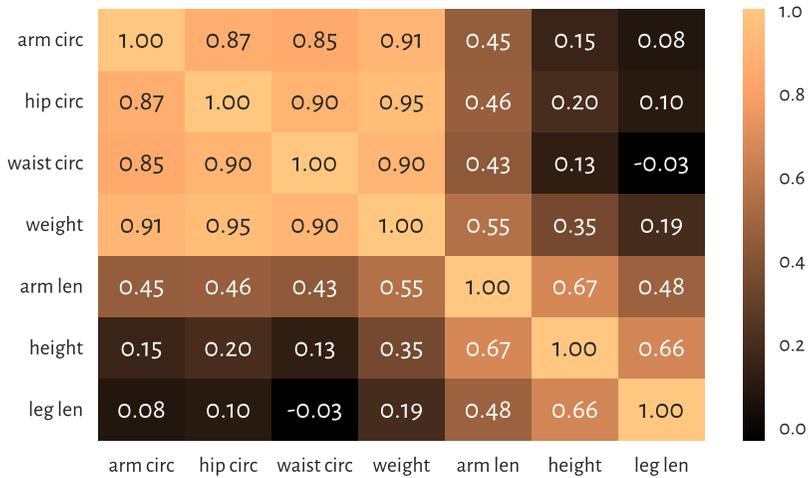

Figure 9.5: A correlation heatmap

**Example 9.3** *(\*) To fetch the row and column index of the most correlated pair of variables (either positively or negatively), we should first take the upper (or lower) triangle of the correlation matrix (see* **numpy.triu** *or* **numpy.tril**) *to ignore the irrelevant and repeating items:*

```
Ru = np.triu(np.abs(R), 1)
np.round(Ru, 2)
## array([[0.  , 0.35, 0.55, 0.19, 0.91, 0.95, 0.9 ],
##        [0.  , 0.  , 0.67, 0.66, 0.15, 0.2 , 0.13],
##        [0.  , 0.  , 0.  , 0.48, 0.45, 0.46, 0.43],
##        [0.  , 0.  , 0.  , 0.  , 0.08, 0.1 , 0.03],
##        [0.  , 0.  , 0.  , 0.  , 0.  , 0.87, 0.85],
##        [0.  , 0.  , 0.  , 0.  , 0.  , 0.  , 0.9 ],
##        [0.  , 0.  , 0.  , 0.  , 0.  , 0.  , 0.  ]])
```

*and then find the location of the maximum:*

```
pos = np.unravel_index(np.argmax(Ru), Ru.shape)
pos  # (row, column)
## (0, 5)
body_columns[ list(pos) ]  # indexing by a tuple has a different meaning
## array(['weight', 'hip circ'], dtype='<U10')
```

*Weight and hip circumference is the most strongly correlated pair.*

*Note that **numpy.argmax** returns an index in the flattened (unidimensional) array. We had to use **numpy.unravel_index** to convert it to a two-dimensional one.*

### 9.1.3    Linear Correlation Coefficients on Transformed Data

Pearson's coefficient can of course also be applied on nonlinearly transformed versions of variables, e.g., logarithms (remember incomes?), squares, square roots, etc.

Let us consider an excerpt from the 2020 CIA World Factbook[3], where we have data on gross domestic product per capita (based on purchasing power parity) and life expectancy at birth in many countries.

```
world = pd.read_csv("https://raw.githubusercontent.com/gagolews/" +
    "teaching-data/master/marek/world_factbook_2020_subset1.csv",
    comment="#")
world = np.array(world)  # convert to matrix
world[:6, :]  # preview
## array([[ 2000. ,    52.8],
##        [12500. ,    79. ],
##        [15200. ,    77.5],
##        [11200. ,    74.8],
##        [49900. ,    83. ],
##        [ 6800. ,    61.3]])
```

Figure 9.6 depicts these data on a scatterplot.

```
plt.subplot(1, 2, 1)
plot_corr(world[:, 0], world[:, 1])
plt.xlabel("per capita GDP PPP")
plt.ylabel("life expectancy (years)")
plt.subplot(1, 2, 2)
plot_corr(np.log(world[:, 0]), world[:, 1])
plt.xlabel("log(per capita GDP PPP)")
plt.yticks()
plt.show()
```

If we compute Pearson's *r* between these two, we will note a quite weak linear correlation:

```
scipy.stats.pearsonr(world[:, 0], world[:, 1])[0]
## 0.656471945486374
```

Anyhow, already the logarithm of GDP is quite strongly linearly correlated with life expectancy:

---

[3] https://www.cia.gov/the-world-factbook/

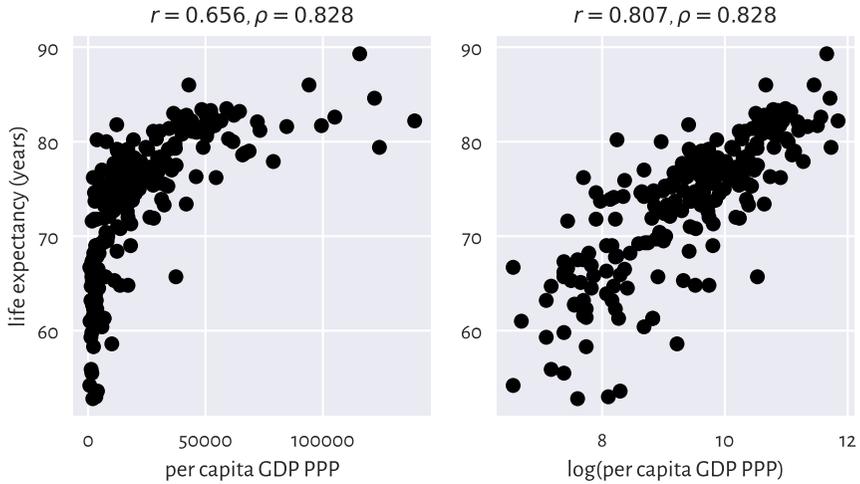

Figure 9.6: Scatterplots for life expectancy vs gross domestic product (purchasing power parity) on linear (left-) and log-scale (righthand side)

```
scipy.stats.pearsonr(np.log(world[:, 0]), world[:, 1])[0]
## 0.8066505089380016
```

which means that modelling our data via $y = a \log x + b$ could be an idea worth considering.

### 9.1.4 Spearman's Rank Correlation Coefficient

Sometimes we might be interested in measuring the degree of any kind of *monotonic* correlation – to what extent one variable is an increasing or decreasing function of another one (linear, logarithmic, quadratic over the positive domain, etc.).

Spearman's rank correlation coefficient is frequently used in such a scenario:

$$\rho(\boldsymbol{x}, \boldsymbol{y}) = r(R(\boldsymbol{x}), R(\boldsymbol{y})),$$

which is[4] the Pearson coefficient computed over vectors of the corresponding ranks of all the elements in $\boldsymbol{x}$ and $\boldsymbol{y}$ (denoted with $R(\boldsymbol{x})$ and $R(\boldsymbol{y})$, respectively).

Hence, the two following calls are equivalent:

```
scipy.stats.spearmanr(world[:, 0], world[:, 1])[0]
## 0.8275220380818622
```

---

[4] If a method Y is nothing else than X on transformed data, we should not consider it a totally new method.

```
scipy.stats.pearsonr(
    scipy.stats.rankdata(world[:, 0]),
    scipy.stats.rankdata(world[:, 1])
)[0]
## 0.8275220380818621
```

Let us point out that this measure is invariant with respect to monotone transformations of the input variables (up to the sign):

```
scipy.stats.spearmanr(np.log(world[:, 0]), -np.sqrt(world[:, 1]))[0]
## -0.8275220380818622
```

This is because such transformations do not change the observations' ranks (or only reverse them).

**Exercise 9.4** *We included the ρs in all the outputs generated by our* **plot_corr** *function. Review all the above figures.*

**Exercise 9.5** *Apply* **numpy.corrcoef** *and* **scipy.stats.rankdata** *(with the appropriate axis argument) to compute the Spearman correlation matrix for all the variable pairs in body. Draw it on a heatmap.*

**Exercise 9.6** *(\*) Draw the scatterplots of the ranks of each column in the* world *and* body *datasets.*

## 9.2 Regression Tasks

Let us assume that we are given a *training/reference* set of $n$ points in an $m$-dimensional space represented as a matrix $\mathbf{X} \in \mathbb{R}^{n \times m}$ and a set of $n$ corresponding numeric outcomes $\mathbf{y} \in \mathbb{R}^n$. *Regression* aims to find a function between the $m$ *independent/explanatory/predictor* variables and a chosen *dependent/response/predicted* variable that can be applied on any test point $\boldsymbol{x}' \in \mathbb{R}^m$:

$$\hat{y}' = f(x'_1, x'_2, \dots, x'_m),$$

and which *approximates* the reference outcomes in a *usable* way.

### 9.2.1 *K*-Nearest Neighbour Regression

A quite straightforward approach to regression relies on aggregating the reference outputs that are associated with a few nearest neighbours of the point $\boldsymbol{x}'$ tested; compare Section 8.4.4.

In $k$-nearest neighbour regression, for a fixed $k \geq 1$ and any given $\boldsymbol{x}' \in \mathbb{R}^m$, $\hat{y} = f(\boldsymbol{x}')$ is computed as follows.

1. Find the indices $N_k(\boldsymbol{x}') = \{i_1, \ldots, i_k\}$ of the $k$ points from $\mathbf{X}$ closest to $\boldsymbol{x}'$, i.e., ones that fulfil for all $j \notin \{i_1, \ldots, i_k\}$:

$$\|\mathbf{x}_{i_1,\cdot} - \boldsymbol{x}'\| \leq \ldots \leq \|\mathbf{x}_{i_k,\cdot} - \boldsymbol{x}'\| \leq \|\mathbf{x}_{j,\cdot} - \boldsymbol{x}'\|.$$

2. Return the arithmetic mean of $(y_{i_1}, \ldots, y_{i_k})$ as the result.

Here is a straightforward implementation that generates the predictions for each point in X_test:

```python
def knn_regress(X_test, X_train, y_train, k):
    t = scipy.spatial.KDTree(X_train.reshape(-1, 1))
    i = t.query(X_test.reshape(-1, 1), k)[1]  # indices of NNs
    y_nn_pred = y_train[i]  # corresponding reference outputs
    return np.mean(y_nn_pred, axis=1)
```

For example, let us try expressing weight (the 1st column) as a function of hip circumference (the 6th column) in the body dataset:

$$\text{weight} = f_1(\text{hip circumference}) \qquad (+\text{some error}).$$

We can also model the life expectancy at birth in different countries (world dataset) as a function of their GDP per capita (PPP):

$$\text{life expectancy} = f_2(\text{GDP per capita}) \qquad (+\text{some error}).$$

Both are instances of the *simple* regression problem, i.e., where there is only one independent variable ($m = 1$). We can easily create an appealing visualisation thereof by means of the following function:

```python
def knn_regress_plot(x, y, K, num_test_points=1001):
    """
    x - 1D vector - reference inputs
    y - 1D vector - corresponding outputs
    K - numbers of near neighbours to test
    num_test_points - number of points to test at
    """
    plt.plot(x, y, "o", alpha=0.1)
    _x = np.linspace(x.min(), x.max(), num_test_points)
    for k in K:
        _y = knn_regress(_x, x, y, k)  # see above
        plt.plot(_x, _y, label=f"$k={k}$")
    plt.legend()
```

Figure 9.7 depicts the fitted functions for a few different $k$s.

```
plt.subplot(1, 2, 1)
knn_regress_plot(body[:, 5], body[:, 0], [5, 25, 100])
plt.xlabel("hip circumference")
plt.ylabel("weight")

plt.subplot(1, 2, 2)
knn_regress_plot(world[:, 0], world[:, 1], [5, 25, 100])
plt.xlabel("per capita GDP PPP")
plt.ylabel("life expectancy (years)")

plt.show()
```

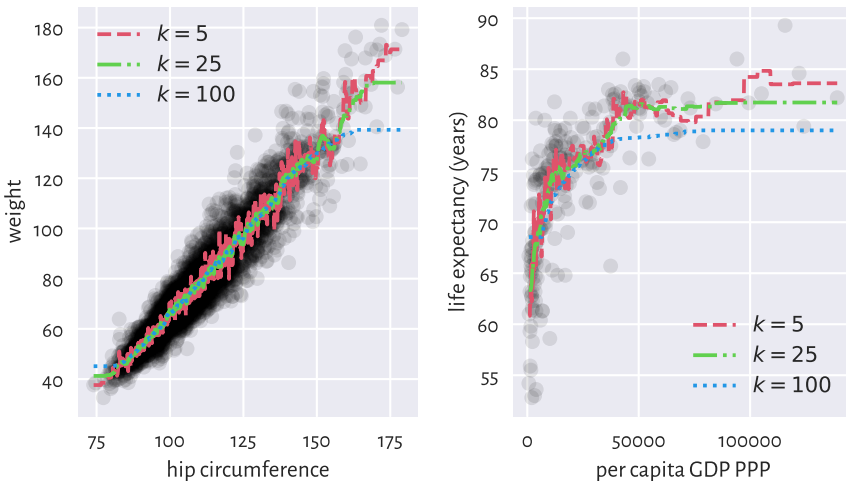

Figure 9.7: *K*-nearest neighbour regression curves for example datasets; the greater the *k*, the more coarse-grained the approximation

We obtained a *smoothened* version of the original dataset. The fact that we do not reproduce the reference data points in an exact manner is reflected by the (figurative) error term in the above equations. Its role is to emphasise the existence of some natural data variability; after all, one's weight is not purely determined by their hip size and life is not all about money.

For small *k* we adapt to the data points better. This can be a good thing unless data are very noisy. The greater the *k*, the smoother the approximation at the cost of losing fine detail and restricted usability at the domain boundaries (here: in the left and right part of the plots).

Usually, the number of neighbours is chosen by trial and error (just like the number of bins in a histogram; compare Section 4.3.3).

**Note** (\*\*) Some methods use weighted arithmetic means for aggregating the $k$ reference outputs, with weights inversely proportional to the distances to the neighbours (closer inputs are considered more important).

Also, instead of few nearest neighbours, we can easily implement some form of fixed-radius search regression, by simply replacing $N_k(x')$ with $B_r(x')$; compare Section 8.4.4. Yet, note that this way we might make the function undefined in sparsely populated regions of the domain.

### 9.2.2 From Data to (Linear) Models

Unfortunately, to generate predictions for new data points, $k$-nearest neighbours regression requires that the training sample is available at all times. It does not *synthesise* or *simplify* the inputs; instead, it works as a kind of a black box. If we were to provide a mathematical equation for the generated prediction, it would be disgustingly long and obscure.

In such cases, to emphasise that $f$ is dependent on the training sample, we sometimes use the more explicit notation $f(x'|\mathbf{X}, y)$ or $f_{\mathbf{X},y}(x')$.

In many contexts we might prefer creating a data *model* instead, in the form of an easily interpretable mathematical function. A simple yet still quite flexible choice tackles regression problems via affine maps of the form:

$$y = f(x_1, x_2, \ldots, x_m) = c_1 x_1 + c_2 x_2 + \cdots + c_m x_m + c_{m+1},$$

or, in matrix multiplication terms:

$$y = \mathbf{c}\mathbf{x}^T + c_{m+1},$$

where $\mathbf{c} = [c_1\ c_2\ \cdots\ c_m]$ and $\mathbf{x} = [x_1\ x_2\ \cdots\ x_m]$.

For $m = 1$, the above simply defines a straight line, which we traditionally denote with:

$$y = ax + b,$$

i.e., where we mapped $x_1 \mapsto x$, $c_1 \mapsto a$ (slope), and $c_2 \mapsto b$ (intercept).

For $m > 1$, we obtain different hyperplanes (high-dimensional generalisations of the notion of a plane).

**Important** A separate intercept "$+c_{m+1}$" term in the defining equation can be quite inconvenient, notationwisely. We will thus restrict ourselves to linear maps like:

$$y = \mathbf{c}\mathbf{x}^T,$$

but where we can possibly have an explicit constant-1 component somewhere *inside* $\mathbf{x}$, for instance:

$$\mathbf{x} = [x_1 \ x_2 \ \cdots \ x_m \ 1].$$

Together with $\mathbf{c} = [c_1 \ c_2 \ \cdots \ c_m \ c_{m+1}]$, as trivially $c_{m+1} \cdot 1 = c_{m+1}$, this new setting is equivalent to the original one.

Without loss in generality, from now on we assume that $\mathbf{x}$ is $m$-dimensional, regardless of its having a constant-1 inside or not.

---

### 9.2.3  Least Squares Method

A linear model is uniquely[5] encoded using only the coefficients $c_1, \ldots, c_m$. To find them, for each point $\mathbf{x}_{i,\cdot}$ from the input (training) set, we typically desire the *predicted* value:

$$\hat{y}_i = f(x_{i,1}, x_{i,2}, \ldots, x_{i,m}) = f(\mathbf{x}_{i,\cdot}|\mathbf{c}) = \mathbf{c}\mathbf{x}_{i,\cdot}^T,$$

to be as *close* to the corresponding reference $y_i$ as possible.

There are many measures of *closeness*, but the most popular one[6] uses the notion of the *sum of squared residuals* (true minus predicted outputs):

$$\mathrm{SSR}(\boldsymbol{c}|\mathbf{X}, \mathbf{y}) = \sum_{i=1}^{n} \left(y_i - \hat{y}_i\right)^2 = \sum_{i=1}^{n} \left(y_i - (c_1 x_{i,1} + c_2 x_{i,2} + \cdots + c_m x_{i,m})\right)^2,$$

which is a function of $\boldsymbol{c} = (c_1, \ldots, c_m)$ (for fixed $\mathbf{X}, \mathbf{y}$).

The *least squares* solution to the stated linear regression problem will be defined by the coefficient vector $\boldsymbol{c}$ that minimises the SSR. Based on what we said about matrix multiplication, this is equivalent to solving the optimisation task:

$$\text{minimise } \left(\mathbf{y} - \mathbf{c}\mathbf{X}^T\right)\left(\mathbf{y} - \mathbf{c}\mathbf{X}^T\right)^T \qquad \text{w.r.t. } (c_1, \ldots, c_m) \in \mathbb{R}^m,$$

because $\hat{\mathbf{y}} = \mathbf{c}\mathbf{X}^T$ gives the predicted values as a row vector (the kind reader is encouraged to check that on a piece of paper now), $\mathbf{r} = \mathbf{y} - \hat{\mathbf{y}}$ computes all the $n$ residuals, and $\mathbf{r}\mathbf{r}^T$ gives their sum of squares.

The method of least squares is one of the simplest and most natural approaches to regression analysis (curve fitting). Its theoretical foundations (calculus…) were developed more than 200 years ago by Gauss and then were polished by Legendre.

---

**Note**  (*) Had the points lain on a hyperplane exactly (the interpolation problem),

---

[5] To memorise the model for further reference, we only need to serialise its $m$ coefficients, e.g., in a JSON or CSV file.

[6] Due to computability and mathematical analysability, which we usually explore in more advanced courses on statistical data analysis such as [6, 20, 42].

$\mathbf{y} = \mathbf{cX}^T$ would have an exact solution, equivalent to solving the linear system of equations $\mathbf{y} - \mathbf{cX}^T = \mathbf{0}$. However, in our setting we assume that there might be some measurement errors or other discrepancies between the reality and the theoretical model. To account for this, we are trying to solve a more general problem of finding a hyperplane for which $\|\mathbf{y} - \mathbf{cX}^T\|^2$ is as small as possible.

This optimisation task can be solved analytically (compute the partial derivatives of SSR with respect to each $c_1, \dots, c_m$, equate them to 0, and solve a simple system of linear equations). This results in $\mathbf{c} = \mathbf{yX}(\mathbf{X}^T\mathbf{X})^{-1}$, where $\mathbf{A}^{-1}$ is the inverse of a matrix $\mathbf{A}$, i.e., the matrix such that $\mathbf{AA}^{-1} = \mathbf{A}^{-1}\mathbf{A} = \mathbf{I}$; compare `numpy.linalg.inv`. As inverting larger matrices directly is not too robust, numerically speaking, we prefer relying upon some more specialised algorithms to determine the solution.

The `scipy.linalg.lstsq` function that we use below provides a quite numerically stable (yet, see Section 9.2.9) procedure that is based on the singular value decomposition of the model matrix.

---

Let us go back to the NHANES study excerpt and express weight (the 1st column) as function of hip circumference (the 6th column) again, but this time using an affine map of the form[7]:

$$\text{weight} = a \cdot \text{hip circumference} + b \qquad (+\text{some error}).$$

The *design (model) matrix* $\mathbf{X}$ and the reference *y*s are:

```
x_original = body[:, [5]]      # a column vector
X_train = x_original**[1, 0]   # hip circumference, 1s
y_train = body[:, 0]           # weight
```

We used the vectorised exponentiation operator to convert each $x_i$ (the $i$-th hip circumference) to a pair $\mathbf{x}_{i,.} = (x_i^1, x_i^0) = (x_i, 1)$, which is a nice trick to append a column of 1s to a matrix. This way, we included the intercept term in the model (as discussed in Section 9.2.2). Here is a preview:

```
preview_indices = [4, 5, 6, 8, 12, 13]
X_train[preview_indices, :]
## array([[ 92.5,    1. ],
##        [106.7,    1. ],
##        [ 96.3,    1. ],
##        [102. ,    1. ],
##        [ 94.8,    1. ],
##        [ 97.5,    1. ]])
```

*(continues on next page)*

---

[7] We sometimes explicitly list the error term that corresponds to the residuals. This is to assure the reader that we are not naïve and that we know what we are doing. We see from the scatterplot of the involved variables that the data do not lie on a straight line perfectly. Each model is merely an idealisation/simplification of the described reality. It is wise to remind ourselves about that every so often.

```
y_train[preview_indices]
## array([55.4, 62. , 66.2, 77.2, 64.2, 56.8])
```

Let us determine the least squares solution to our regression problem:

```
import scipy.linalg
res = scipy.linalg.lstsq(X_train, y_train)
```

That's it. The optimal coefficients vector (the one that minimises the SSR) is:

```
c = res[0]
c
## array([  1.3052463 , -65.10087248])
```

The estimated model is:

$$\text{weight} = 1.305 \cdot \text{hip circumference} - 65.1 \qquad (+\text{some error}).$$

Let us contemplate the fact that the model is nicely interpretable. For instance, as hip circumference increases, we expect the weights to be greater and greater. As we said before, it does not mean that there is some *causal* relationship between the two (for instance, there can be some latent variables that affect both of them). Instead, there is some general tendency regarding how the data align in the sample space. For instance, that the "best guess" (according to the current model – there can be many; see below) weight for a person with hip circumference of 100 cm is 65.4 kg. Thanks to such models, we might understand certain phenomena better or find some proxies for different variables (especially if measuring them directly is tedious, costly, dangerous, etc.).

Let us determine the predicted weights for all of the participants:

```
y_pred = c @ X_train.T
np.round(y_pred[preview_indices], 2)  # preview
## array([55.63, 74.17, 60.59, 68.03, 58.64, 62.16])
```

The scatterplot and the fitted regression line in Figure 9.8 indicates a quite good fit, but of course there is some natural variability.

```
plt.plot(x_original, y_train, "o", alpha=0.1)  # scatterplot
_x = np.array([x_original.min(), x_original.max()]).reshape(-1, 1)
_y = c @ (_x**[1, 0]).T
plt.plot(_x, _y, "r-")  # a line that goes through the two extreme points
plt.xlabel("hip circumference")
plt.ylabel("weight")
plt.show()
```

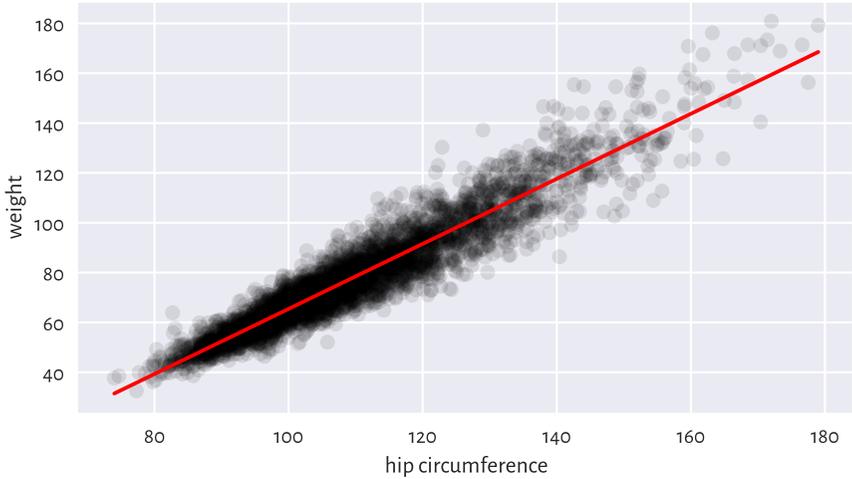

Figure 9.8: The least squares line for weight vs hip circumference

**Exercise 9.7** *The Anscombe quartet[8] is a famous example dataset, where we have four pairs of variables that have almost identical means, variances, and linear correlation coefficients. Even though they can be approximated by the same straight line, their scatter plots are vastly different. Reflect upon this toy example.*

### 9.2.4   Analysis of Residuals

The residuals (i.e., the estimation errors – what we expected vs what we got), for the chosen 6 observations are visualised in Figure 9.9.

```
r = y_train - y_pred  # residuals
np.round(r[preview_indices], 2)  # preview
## array([ -0.23, -12.17,   5.61,   9.17,   5.56,  -5.36])
```

We wanted the squared residuals (on average – across all the points) to be as small as possible. The least squares method assures that this is the case *relative to the chosen model*, i.e., a linear one. Nonetheless, it still does not mean that what we obtained constitutes a good fit to the training data. Thus, we need to perform the *analysis of residuals*.

Interestingly, the average of residuals is always zero:

$$\frac{1}{n} \sum_{i=1}^{n} (y_i - \hat{y}_i) = 0.$$

Therefore, if we want to summarise the residuals into a single number, we should in-

---

[8] https://github.com/gagolews/teaching-data/raw/master/r/anscombe.csv

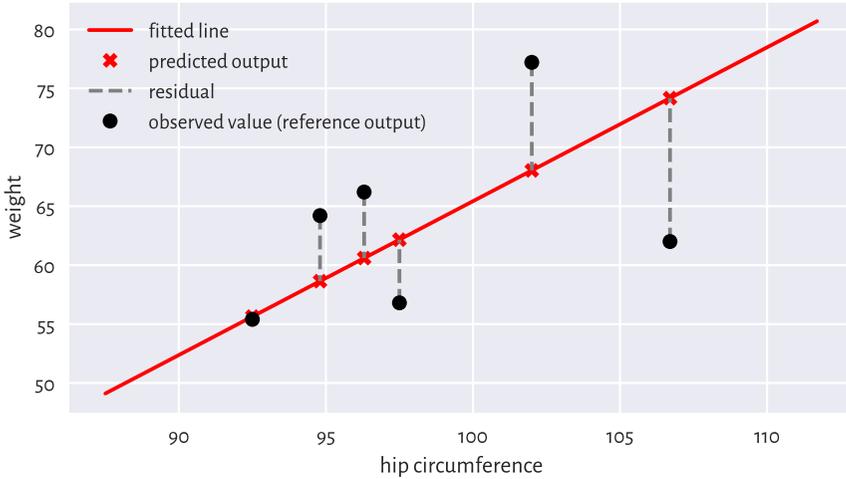

Figure 9.9: Example residuals in a simple linear regression task

stead use, for example, the root mean squared error:

$$\mathrm{RMSE}(\mathbf{y}, \hat{\mathbf{y}}) = \sqrt{\frac{1}{n} \sum_{i=1}^{n} (y_i - \hat{y}_i)^2}.$$

```
np.sqrt(np.mean(r**2))
## 6.948470091176111
```

Hopefully we can see that RMSE is a function of SSR that we sought to minimise above.

Alternatively, we can compute the mean absolute error:

$$\mathrm{MAE}(\mathbf{y}, \hat{\mathbf{y}}) = \frac{1}{n} \sum_{i=1}^{n} |y_i - \hat{y}_i|.$$

```
np.mean(np.abs(r))
## 5.207073583769202
```

MAE is nicely interpretable, because it measures by how many kilograms we err *on average*. Not bad.

**Exercise 9.8** *Fit a regression line explaining weight as a function of the waist circumference and compute the corresponding RMSE and MAE. Are they better than when hip circumference is used as an explanatory variable?*

**Note**  Generally, fitting simple (involving one independent variable) linear models can only make sense for highly linearly correlated variables. Interestingly, if $y$ and $x$ are both standardised, and $r$ is their Pearson's coefficient, then the least squares solution is given by $y = rx$.

To verify whether a fitted model is not extremely wrong (e.g., when we fit a linear model to data that clearly follows a different functional relationship), a plot of residuals against the fitted values can be of help; see Figure 9.10. Ideally, the points should be aligned totally at random therein, without any dependence structure (homoscedasticity).

```python
plt.plot(y_pred, r, "o", alpha=0.1)
plt.axhline(0, ls="--", color="red")  # horizontal line at y=0
plt.xlabel("fitted values")
plt.ylabel("residuals")
plt.show()
```

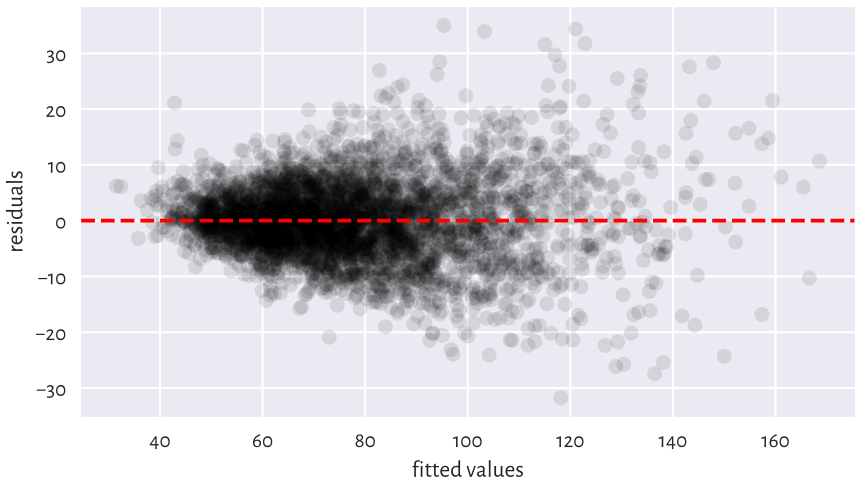

Figure 9.10: Residuals vs fitted values for the linear model explaining weight as a function of hip circumference; the variance of residuals slightly increases as $\hat{y}_i$ increases, which is not ideal, but it could be much worse than this

**Exercise 9.9** *Compare[9] the RMSE and MAE for the k-nearest neighbour regression curves depicted in the lefthand side of Figure 9.7. Also, draw the residuals vs fitted plot.*

---

[9] In $k$-nearest neighbour regression, we are not aiming to minimise anything in particular. If the model is good with respect to some metrics such as RMSE or MAE, we can consider ourselves lucky. Nevertheless, some asymptotic results guarantee the optimality of the outcomes generated for large sample sizes (e.g., consistency); see, e.g., [20].

For linear models fitted using the least squares method, it can be shown that it holds:

$$\frac{1}{n} \sum_{i=1}^{n} (y_i - \bar{y})^2 = \frac{1}{n} \sum_{i=1}^{n} (\hat{y}_i - \bar{y})^2 + \frac{1}{n} \sum_{i=1}^{n} (y_i - \hat{y}_i)^2 .$$

In other words, the variance of the dependent variable (left) can be decomposed into the sum of the variance of the predictions and the averaged squared residuals. Multiplying the above by $n$, we have that the *total* sum of squares is equal to the *explained* sum of squares plus the *residual* sum of squares:

$$\text{TSS} = \text{ESS} + \text{RSS}.$$

We of course yearn for ESS to be as close to TSS as possible. Equivalently, it would be jolly nice to have RSS equal to 0.

The *coefficient of determination* (unadjusted R-Squared, sometimes referred to as simply the *score*) is a popular normalised, unitless measure that is easier to interpret than raw ESS or RSS when we have no domain-specific knowledge of the modelled problem. It is given by:

$$R^2(\mathbf{y}, \hat{\mathbf{y}}) = \frac{\text{ESS}}{\text{TSS}} = 1 - \frac{\text{RSS}}{\text{TSS}} = 1 - \frac{s_r^2}{s_y^2}.$$

```
1 - np.var(y_train-y_pred)/np.var(y_train)
## 0.8959634726270759
```

The coefficient of determination in the current context[10] is thus the proportion of variance of the dependent variable explained by the independent variables in the model. The closer it is to 1, the better. A dummy model that always returns the mean of $\mathbf{y}$ gives R-squared of 0.

In our case, $R^2 \simeq 0.9$ is quite high, which indicates a rather good fit.

---

**Note**    (*) There are certain statistical results that can be relied upon provided that the residuals are independent random variables with expectation zero and the same variance (e.g., the Gauss–Markov theorem). Further, if they are normally distributed, then we have several hypothesis tests available (e.g., for the significance of coefficients). This is why in various textbooks such assumptions are additionally verified. But we do not go that far in this introductory course.

---

### 9.2.5 Multiple Regression

As another example, let us fit a model involving two independent variables, arm and hip circumference:

---

[10] For a model that is *not* generated via least squares, the coefficient of determination can also be negative, particularly when the fit is extremely bad. Also, note that this measure is dataset-dependent. Therefore, it should not be used for comparing models explaining different dependent variables.

```
X_train = np.insert(body[:, [4, 5]], 2, 1, axis=1)  # append a column of 1s
res = scipy.linalg.lstsq(X_train, y_train)
c = res[0]
np.round(c, 2)
## array([  1.3 ,    0.9 , -63.38])
```

We fitted the plane:

$$\text{weight} = 1.3\,\text{arm circumference} + 0.9\,\text{hip circumference} - 63.38.$$

We skip the visualisation part, because we do not expect it to result in a readable plot: these are multidimensional data. The coefficient of determination is:

```
y_pred = c @ X_train.T
r = y_train - y_pred
1-np.var(r)/np.var(y_train)
## 0.9243996585518783
```

Root mean squared error:

```
np.sqrt(np.mean(r**2))
## 5.923223870044694
```

Mean absolute error:

```
np.mean(np.abs(r))
## 4.431548244333893
```

It is a slightly better model than the previous one. We can predict the participants' weights with better precision, at the cost of an increased model's complexity.

### 9.2.6 Variable Transformation and Linearisable Models (*)

We are not restricted merely to linear functions of the input variables. By applying arbitrary transformations upon the columns of the design matrix, we can cover many diverse scenarios.

For instance, a polynomial model involving two variables:

$$g(v_1, v_2) = \beta_0 + \beta_1 v_1 + \beta_2 v_1^2 + \beta_3 v_1 v_2 + \beta_4 v_2 + \beta_5 v_2^2,$$

can be obtained by substituting $x_1 = 1$, $x_2 = v_1$, $x_3 = v_1^2$, $x_4 = v_1 v_2$, $x_5 = v_2$, $x_6 = v_2^2$, and then fitting a linear model involving six variables:

$$f(x_1, x_2, \dots, x_6) = c_1 x_1 + c_2 x_2 + \cdots + x_6 x_6.$$

The design matrix is made of rubber, it can handle almost anything. If we have a linear model, but with respect to transformed data, the algorithm does not care. This is the beauty of the underlying mathematics; see also [8].

A creative modeller can also turn models such as $u = ce^{av}$ into $y = ax + b$ by replacing $y = \log u$, $x = v$, and $b = \log c$. There are numerous possibilities based on the properties of the log and exp functions that we listed in Section 5.2. We call them *linearisable models*.

As an example, let us model the life expectancy at birth in different countries as a function of their GDP per capita (PPP).

We will consider four different models:

1. $y = c_1 + c_2 x$ (linear),

2. $y = c_1 + c_2 x + c_3 x^2$ (quadratic),

3. $y = c_1 + c_2 x + c_3 x^2 + c_4 x^3$ (cubic),

4. $y = c_1 + c_2 \log x$ (logarithmic).

Here are the helper functions that create the model matrices:

```python
def make_model_matrix1(x):
    return x.reshape(-1, 1)**[0, 1]

def make_model_matrix2(x):
    return x.reshape(-1, 1)**[0, 1, 2]

def make_model_matrix3(x):
    return x.reshape(-1, 1)**[0, 1, 2, 3]

def make_model_matrix4(x):
    return (np.log(x)).reshape(-1, 1)**[0, 1]

make_model_matrix1.__name__ = "linear model"
make_model_matrix2.__name__ = "quadratic model"
make_model_matrix3.__name__ = "cubic model"
make_model_matrix4.__name__ = "logarithmic model"

model_matrix_makers = [
    make_model_matrix1,
    make_model_matrix2,
    make_model_matrix3,
    make_model_matrix4
]
x_original = world[:, 0]
Xs_train = [ make_model_matrix(x_original)
    for make_model_matrix in model_matrix_makers ]
```

Fitting the models:

```
y_train = world[:, 1]
cs = [ scipy.linalg.lstsq(X_train, y_train)[0]
    for X_train in Xs_train ]
```

Their coefficients of determination are equal to:

```
for i in range(len(Xs_train)):
    R2 = 1 - np.var(y_train - cs[i] @ Xs_train[i].T)/np.var(y_train)
    print(f"{model_matrix_makers[i].__name__:20} R2={R2:.3f}")
## linear model         R2=0.431
## quadratic model      R2=0.567
## cubic model          R2=0.607
## logarithmic model    R2=0.651
```

The logarithmic model is thus the best (out of the models we considered). The four models are depicted in Figure 9.11.

```
plt.plot(x_original, y_train, "o", alpha=0.1)
_x = np.linspace(x_original.min(), x_original.max(), 101).reshape(-1, 1)
for i in range(len(model_matrix_makers)):
    _y = cs[i] @ model_matrix_makers[i](_x).T
    plt.plot(_x, _y, label=model_matrix_makers[i].__name__)
plt.legend()
plt.xlabel("per capita GDP PPP")
plt.ylabel("life expectancy (years)")
plt.show()
```

**Exercise 9.10** *Draw box plots and histograms of residuals for each model as well as the scatterplots of residuals vs fitted values.*

### 9.2.7 Descriptive vs Predictive Power (*)

We *approximated* the life vs GDP relationship using a few different functions. Nevertheless, we see that the above quadratic and cubic models possibly do not make much sense, semantically speaking. Sure, as far as individual points *in the training set* are concerned, they do fit the data better than the linear model. After all, they have smaller mean squared errors (again: at these given points). Looking at the way they behave, one does not need a university degree in economics/social policy to conclude that they are not the best *description* of how the reality behaves (on average).

---

**Important** Naturally, a model's goodness of fit to observed data tends to improve as the model's complexity increases. The Razor principle (by William of Ockham et al.) advises that if some phenomenon can be explained in many different ways, the simplest explanation should be chosen (*do not multiply entities* [here: introduce independent variables] *without necessity*).

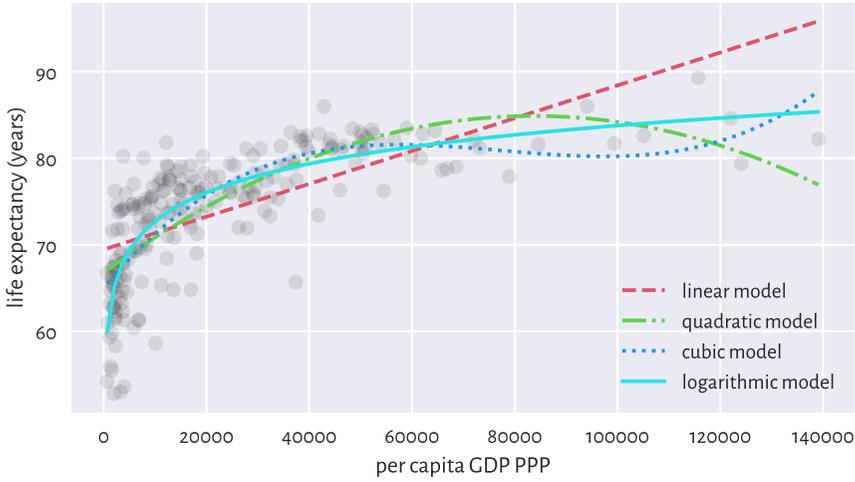

Figure 9.11: Different models for life expectancy vs GDP

In particular, the more independent variables we have in the model, the greater the $R^2$ coefficient will be. We can try correcting for this phenomenon by considering the *adjusted $R^2$*:

$$\bar{R}^2(\mathbf{y}, \hat{\mathbf{y}}) = 1 - (1 - R^2(\mathbf{y}, \hat{\mathbf{y}})) \frac{n-1}{n-m-1},$$

which, to some extent, penalises more complex models.

**Note**  (**) Model quality measures adjusted for the number of model parameters, $m$, can also be useful in automated variable selection. For example, the Akaike Information Criterion is a popular measure given by:

$$\text{AIC} = 2m + n \log(\text{SSR}) - n \log n.$$

Furthermore, the Bayes Information Criterion is defined via:

$$\text{BIC} = m \log n + n \log(\text{SSR}) - n \log n.$$

Unfortunately, they are both dependent on the scale of $\mathbf{y}$.

We should also be interested in a model's *predictive* power – how well does it generalise to data points that we do not have now (or pretend we do not have), but might face in the future. As we observe the modelled reality only at a few different points, the question is how the model performs when filling the gaps between the dots it connects.

In particular, we should definitely be careful when *extrapolating* the data, i.e., making predictions outside of its usual domain. For example, the linear model predicts the following life expectancy for an imaginary country with $500,000 per capita GDP:

```
cs[0] @ model_matrix_makers[0](np.array([500000])).T
## array([164.3593753])
```

and the quadratic one gives:

```
cs[1] @ model_matrix_makers[1](np.array([500000])).T
## array([-364.10630779])
```

Nonsense.

**Example 9.11** *Let us consider the following theoretical illustration. Assume that a true model of some reality is $y = 5 + 3x^3$.*

```
def true_model(x):
    return 5 + 3*(x**3)
```

*Still, for some reason we are only able to gather a small ($n = 25$) sample from this model. What is even worse, it is subject to some measurement error:*

```
np.random.seed(42)
x = np.random.rand(25)                          # random xs on [0, 1]
y = true_model(x) + 0.2*np.random.randn(len(x))  # true_model(x) + noise
```

*The least-squares fitting of $y = c_1 + c_2 x^3$ to the above gives:*

```
X03 = x.reshape(-1, 1)**[0, 3]
c03 = scipy.linalg.lstsq(X03, y)[0]
ssr03 = np.sum((y-c03 @ X03.T)**2)
np.round(c03, 2)
## array([5.01, 3.13])
```

*which is not too far, but still somewhat[11] distant from the true coefficients, 5 and 3.*

*We can also fit a more flexible cubic polynomial, $y = c_1 + c_2 x + c_3 x^2 + c_4 x_3$:*

```
X0123 = x.reshape(-1, 1)**[0, 1, 2, 3]
c0123 = scipy.linalg.lstsq(X0123, y)[0]
ssr0123 = np.sum((y-c0123 @ X0123.T)**2)
```

---

[11] For large $n$, we expect the pinpoint the true coefficients exactly. This is because, in our scenario (independent, normally distributed errors with the expectation of 0), the least squares method is the maximum likelihood estimator of the model parameters. As a consequence, it is consistent.

```
np.round(c0123, 2)
## array([4.89, 0.32, 0.57, 2.23])
```

*In terms of the SSR, this more complex model of course explains* the training data *better:*

```
ssr03, ssr0123
## (1.0612111154029558, 0.9619488226837544)
```

*Yet, it is farther away from the* truth *(which, whilst performing the fitting task based only on given* $x$ *and* $y$*, is unknown). We may thus say that the first model* generalises *better on yet-to-be-observed data; see* Figure 9.12 *for an illustration.*

```
_x = np.linspace(0, 1, 101)
plt.plot(x, y, "o")
plt.plot(_x, true_model(_x), "--", label="true model")
plt.plot(_x, c0123 @ (_x.reshape(-1, 1)**[0, 1, 2, 3]).T,
    label="fitted model y=x**[0, 1, 2, 3]")
plt.plot(_x, c03 @ (_x.reshape(-1, 1)**[0, 3]).T,
    label="fitted model y=x**[0, 3]")
plt.legend()
plt.show()
```

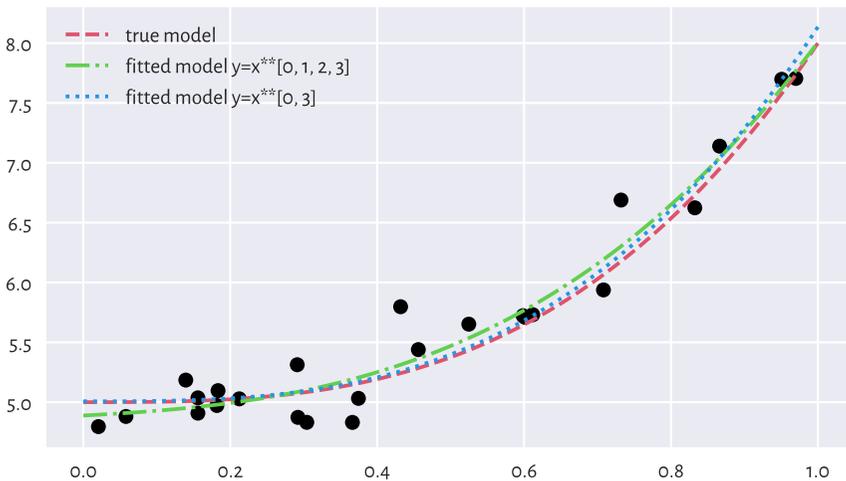

Figure 9.12: The true (theoretical) model vs some guesstimates (fitted based on noisy data); more degrees of freedom is not always better

**Example 9.12** (**) *We defined the sum of squared residuals (and its function, the root mean squared error) by means of the averaged deviation from the reference values, which in fact are* *themselves subject to error. Even though they are our best-shot approximation of the truth, they should be taken with a degree of scepticism.*

*In the above example, given the true (reference) model $f$ defined over the domain $D$ (in our case, $f(x) = 5 + 3x^3$ and $D = [0, 1]$) and an empirically fitted model $\hat{f}$, we can compute the square root of the integrated squared error over the whole $D$:*

$$\text{RMSE}(f, \hat{f}) = \sqrt{\int_D (f(x) - \hat{f}(x))^2 \, dx}.$$

*For polynomials and other simple functions, RMSE can be computed analytically. More generally, we can approximate it numerically by sampling the above at sufficiently many points and applying the trapezoidal rule (e.g., [66]). As this can be an educative programming exercise, below we consider a range of polynomial models of different degrees.*

```python
cs, rmse_train, rmse_test = [], [], []  # result containers
ps = np.arange(1, 10)  # polynomial degrees
for p in ps:              # for each polynomial degree:
    c = scipy.linalg.lstsq(x.reshape(-1, 1)**np.arange(p+1), y)[0]  # fit
    cs.append(c)

    y_pred = c @ (x.reshape(-1, 1)**np.arange(p+1)).T   # predictions
    rmse_train.append(np.sqrt(np.mean((y-y_pred)**2)))  # RMSE

    _x = np.linspace(0, 1, 101)                    # many _xs
    _y = c @ (_x.reshape(-1, 1)**np.arange(p+1)).T       # f(_x)
    _r = (true_model(_x) - _y)**2                  # residuals
    rmse_test.append(np.sqrt(0.5*np.sum(
        np.diff(_x)*(_r[1:]+_r[:-1])  # trapezoidal rule for integration
    )))

plt.plot(ps, rmse_train, label="RMSE (training set)")
plt.plot(ps, rmse_test, label="RMSE (theoretical)")
plt.legend()
plt.yscale("log")
plt.xlabel("model complexity (polynomial degree)")
plt.show()
```

*In Figure 9.13, we see that a model's ability to make correct generalisations onto unseen data, with the increased complexity initially improves, but then becomes worse. It is quite a typical behaviour. In fact, the model with the smallest RMSE on the training set, overfits to the input sample, see Figure 9.14.*

```python
plt.plot(x, y, "o")
plt.plot(_x, true_model(_x), "--", label="true model")
for i in [0, 1, 8]:
    plt.plot(_x, cs[i] @ (_x.reshape(-1, 1)**np.arange(ps[i]+1)).T,
```

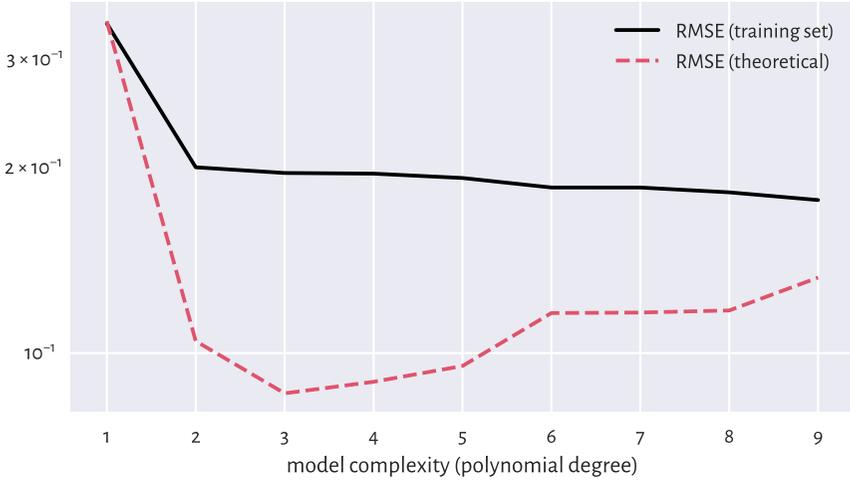

Figure 9.13: Small RMSE on training data does not necessarily imply good generalisation abilities

```
        label=f"fitted degree-{ps[i]} polynomial")
plt.legend()
plt.show()
```

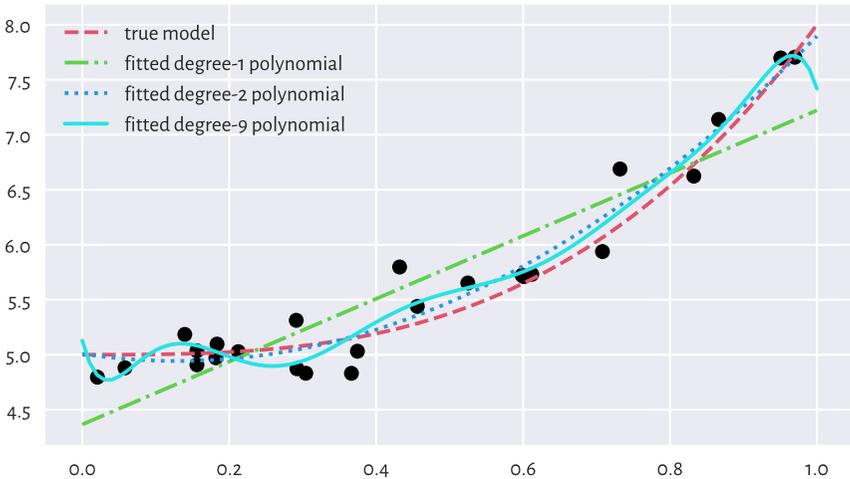

Figure 9.14: Under- and overfitting to training data

**Important**  When evaluating a model's quality in terms of predictive power on unseen data, we should go beyond inspecting its behaviour merely on the points from the training sample. As the *truth* is usually not known (if it were, we would not need any guessing), a common approach in case where we have a dataset of a considerable size is to divide it (randomly; see Section 10.5.3) into two parts:

- training sample (say, 60%) – used to fit a model,

- test sample (the remaining 40%) – used to assess its quality (e.g., by means of RMSE).

This might *emulate* an environment where some new data arrives later, see Section 12.3.3 for more details.

Furthermore, if model selection is required, we may apply a training/validation/test split (say, 60/20/20%; see Section 12.3.4). Here, many models are constructed on the training set, the validation set is used to compute the metrics and choose the best model, and then the test set gives the final model's valuation to assure its usefulness/uselessness (because we do not want it to overfit to the test set).

Overall, models should never be blindly trusted – common sense must always be applied. The fact that we fitted something using a sophisticated procedure on a dataset that was hard to obtain does not justify its use. Mediocre models must be discarded, and we should move on, regardless of how much time/resources we have invested whilst developing them. Too many bad models go into production and make our daily lives harder. We should end this madness.

### 9.2.8  Fitting Regression Models with `scikit-learn` (*)

`scikit-learn`[12] (`sklearn`; [64]) is a huge Python package built on top of `numpy`, `scipy`, and `matplotlib`. It has a consistent API and implements or provides wrappers for many regression, classification, clustering, and dimensionality reduction algorithms (amongst others).

**Important**  `sklearn` is very convenient but allows for fitting models even if we do not understand the mathematics behind them. This is dangerous – it is like driving a sports car without the necessary skills and, at the same time, wearing a blindfold. Advanced students and practitioners will appreciate it, but if used by beginners, it needs to be handled with care; we should not mistake something's being easily accessible with its being safe to use. Remember that if we are given a function implementing some procedure for which we are not able to provide its definition/mathematical properties/explain its idealised version using pseudocode, we should refrain from using it (see Rule#7).

---

[12] https://scikit-learn.org/stable/index.html

Because of the above, we shall only present a quick demo of **scikit-learn**'s API. Let us do that by fitting a multiple linear regression model for, again, weight as a function of the arm and the hip circumference:

```
X_train = body[:, [4, 5]]
y_train = body[:, 0]
```

In **scikit-learn**, once we construct an object representing the model to be fitted, the **fit** method determines the optimal parameters.

```
import sklearn.linear_model
lm = sklearn.linear_model.LinearRegression(fit_intercept=True)
lm.fit(X_train, y_train)
lm.intercept_, lm.coef_
## (-63.383425410947524, array([1.30457807, 0.8986582 ]))
```

We of course obtained the same solution as with **scipy.linalg.lstsq**.

Computing the predicted values can be done via the **predict** method. For example, we can calculate the coefficient of determination:

```
y_pred = lm.predict(X_train)
import sklearn.metrics
sklearn.metrics.r2_score(y_train, y_pred)
## 0.9243996585518783
```

The above function is convenient, but can we really recall the formula for the score and what it measures? We should always be able to do that.

### 9.2.9  Ill-Conditioned Model Matrices (*)

Our approach to regression analysis relies on solving an optimisation problem (the method least squares). Nevertheless, sometimes the "optimal" solution that the algorithm returns might have nothing to do with the *true* minimum. And this is despite the fact that we have the theoretical results stating that the solution is unique[13] (the objective is convex). The problem stems from our using the computer's finite-precision floating point arithmetic; compare Section 5.5.6.

Let us fit a degree-4 polynomial to the life expectancy vs per capita GDP dataset.

```
x_original = world[:, 0]
X_train = (x_original.reshape(-1, 1))**[0, 1, 2, 3, 4]
y_train = world[:, 1]
cs = dict()
```

---

[13] There are methods in statistical learning where there might be multiple local minima – this is even more difficult; see Section 12.4.4.

We store the estimated model coefficients in a dictionary, because many methods will follow next. First, **scipy**:

```
res = scipy.linalg.lstsq(X_train, y_train)
cs["scipy_X"] = res[0]
cs["scipy_X"]
## array([ 2.33103950e-16,  6.42872371e-12,  1.34162021e-07,
##        -2.33980973e-12,  1.03490968e-17])
```

If we drew the fitted polynomial now (see Figure 9.15), we would see that the fit is unbelievably bad. The result returned by **scipy.linalg.lstsq** is now not at all optimal. All coefficients are approximately equal to 0.

It turns out that the fitting problem is extremely *ill-conditioned* (and it is not the algorithm's fault): GDPs range from very small to very large ones. Furthermore, taking the powers of 4 thereof results in numbers of ever greater range. Finding the least squares solution involves some form of matrix inverse (not necessarily directly) and our model matrix may be close to singular (one that is not invertible).

As a measure of the model matrix's ill-conditioning, we often use the so-called *condition number*, denoted $\kappa(\mathbf{X}^T)$, being the ratio of the largest to the smallest so-called *singular values*[14] of $\mathbf{X}^T$. They are in fact returned by the **scipy.linalg.lstsq** method itself:

```
s = res[3]    # singular values of X_train.T
s
## array([5.63097211e+20, 7.90771769e+14, 4.48366565e+09, 6.77575417e+04,
##        5.76116463e+00])
```

Note that they are already sorted nonincreasingly. The condition number $\kappa(\mathbf{X}^T)$ is equal to:

```
s[0] / s[-1]  # condition number (largest/smallest singular value)
## 9.774017018467434e+19
```

As a rule of thumb, if the condition number is $10^k$, we are losing $k$ digits of numerical precision when performing the underlying computations. We are thus currently faced with a very ill-conditioned problem, because the above number is exceptionally large. We expect that if the values in $\mathbf{X}$ or $\mathbf{y}$ are perturbed even slightly, it can result in very large changes in the computed regression coefficients.

---

**Note** (\*\*) The least squares regression problem can be solved by means of the singular value decomposition of the model matrix, see Section 9.3.4. Let **USQ** be the SVD of

---

[14] (\*\*) Being themselves the square roots of eigenvalues of $\mathbf{X}^T\mathbf{X}$. Equivalently, $\kappa(\mathbf{X}^T) = \|(\mathbf{X}^T)^{-1}\| \|\mathbf{X}^T\|$ with respect to the spectral norm. Seriously, we really need linear algebra when we even remotely think about practising data science. Let us add it to our life skills bucket list.

$\mathbf{X}^T$. Then $\mathbf{c} = \mathbf{U}\mathbf{S}^{-1}\mathbf{Q}\mathbf{y}$, with $\mathbf{S}^{-1} = \mathrm{diag}(1/s_{1,1}, \dots, 1/s_{m,m})$. As $s_{1,1} \geq \dots \geq s_{m,m}$ gives the singular values of $\mathbf{X}^T$, the aforementioned condition number can simply be computed as $s_{1,1}/s_{m,m}$.

---

Let us verify the method used by `scikit-learn`. As it fits the intercept separately, we expect it to be slightly better-behaving. Nevertheless, let us keep in mind that it is merely a wrapper around `scipy.linalg.lstsq` with a different API.

```
import sklearn.linear_model
lm = sklearn.linear_model.LinearRegression(fit_intercept=True)
lm.fit(X_train[:, 1:], y_train)
cs["sklearn"] = np.r_[lm.intercept_, lm.coef_]
cs["sklearn"]
## array([ 6.92257708e+01,  5.05752755e-13,  1.38835643e-08,
##        -2.18869346e-13,  9.09347772e-19])
```

Here is the condition number of the underlying model matrix:

```
lm.singular_[0] / lm.singular_[-1]
## 1.4026032298428496e+16
```

The condition number is also enormous. Still, `scikit-learn` did not warn us about this being the case (insert frowning face emoji here). Had we trusted the solution returned by it, we would end up with conclusions from our data analysis built on sand. As we said in Section 9.2.8, the package design assumes that its users know what they are doing. This is okay, we are all adults here, although some of us are still learning.

Overall, if the model matrix is close to singular, the computation of its inverse is prone to enormous numerical errors. One way of dealing with this is to remove highly correlated variables (the multicollinearity problem). Interestingly, standardisation can *sometimes* make the fitting more numerically stable.

Let $\mathbf{Z}$ be a standardised version of the model matrix $\mathbf{X}$ with the intercept part (the column of 1s) not included, i.e., with $\mathbf{z}_{\cdot,j} = (\mathbf{x}_{\cdot,j} - \bar{x}_j)/s_j$ where $\bar{x}_j$ and $s_j$ denotes the arithmetic mean and standard deviation of the $j$-th column in $\mathbf{X}$. If $(d_1, \dots, d_{m-1})$ is the least squares solution for $\mathbf{Z}$, then the least squares solution to the underlying original regression problem is:

$$\mathbf{c} = \left( \bar{y} - \sum_{j=1}^{m-1} \frac{d_j}{s_j} \bar{x}_j, \frac{d_1}{s_1}, \frac{d_2}{s_2}, \dots, \frac{d_{m-1}}{s_{m-1}} \right),$$

with the first term corresponding to the intercept.

Let us test this approach with `scipy.linalg.lstsq`:

```
means = np.mean(X_train[:, 1:], axis=0)
stds = np.std(X_train[:, 1:], axis=0)
Z_train = (X_train[:, 1:]-means)/stds
resZ = scipy.linalg.lstsq(Z_train, y_train)
c_scipyZ = resZ[0]/stds
cs["scipy_Z"] = np.r_[np.mean(y_train) - (c_scipyZ @ means.T), c_scipyZ]
cs["scipy_Z"]
## array([ 6.35946784e+01,  1.04541932e-03, -2.41992445e-08,
##         2.39133533e-13, -8.13307828e-19])
```

The condition number is:

```
s = resZ[3]
s[0] / s[-1]
## 139.42792257372338
```

This is still far from perfect (we would prefer a value close to 1) but nevertheless way better.

Figure 9.15 depicts the three fitted models, each claiming to be *the* solution to the original regression problem. Note that, luckily, we know that in our case the logarithmic model is better than the polynomial one.

```
plt.plot(x_original, y_train, "o", alpha=0.1)
_x = np.linspace(x_original.min(), x_original.max(), 101).reshape(-1, 1)
_X = _x**[0, 1, 2, 3, 4]
for lab, c in cs.items():
    ssr = np.sum((y_train - c @ X_train.T)**2)
    plt.plot(_x, c @ _X.T, label=f"{lab:10} SSR={ssr:.2f}")
plt.legend()
plt.ylim(20, 120)
plt.xlabel("per capita GDP PPP")
plt.ylabel("life expectancy (years)")
plt.show()
```

---

**Important** Always check the model matrix's condition number.

---

**Exercise 9.13** *Check the condition numbers of all the models fitted so far in this chapter via the least squares method.*

To be strict, if we read a paper in, say, social or medical sciences (amongst others) where the researchers fit a regression model but do not provide the model matrix's condition number, we should doubt the conclusions they make.

On a final note, we might wonder why the standardisation is not done automatically by the least squares solver. As usual with most numerical methods, there is no one-fits-all solution: e.g., when there are columns of extremely small variance or there are outliers in data. This is why we need to study all the topics deeply: to be able to respond flexibly to many different scenarios ourselves.

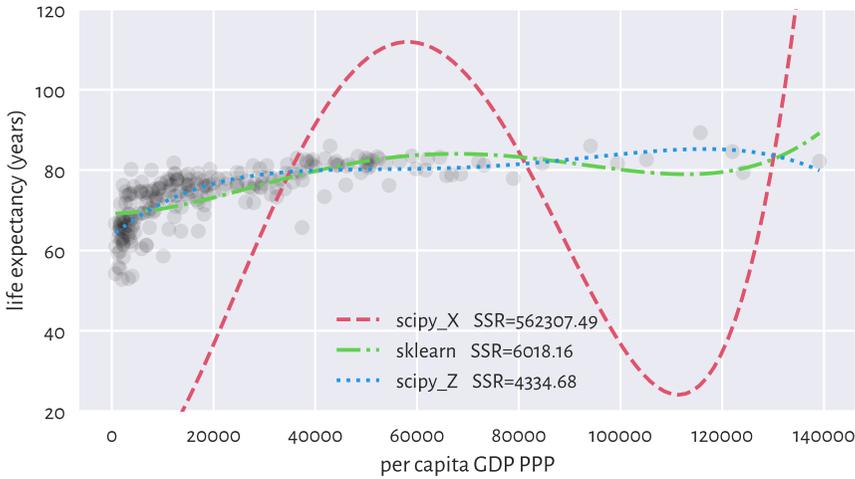

Figure 9.15: Ill-conditioned model matrix can result in the resulting models' being very wrong

## 9.3   Finding Interesting Combinations of Variables (*)

### 9.3.1   Dot Products, Angles, Collinearity, and Orthogonality

It turns out that the dot product (Section 8.3) has a nice geometrical interpretation:

$$\boldsymbol{x} \cdot \boldsymbol{y} = \|\boldsymbol{x}\| \, \|\boldsymbol{y}\| \, \cos \alpha,$$

where $\alpha$ is the angle between two given vectors $\boldsymbol{x}, \boldsymbol{y} \in \mathbb{R}^n$. In plain English, it is the product of the magnitudes of the two vectors and the cosine of the angle between them.

We can obtain the cosine part by computing the dot product of the *normalised* vectors, i.e., such that their magnitudes are equal to 1:

$$\cos \alpha = \frac{\boldsymbol{x}}{\|\boldsymbol{x}\|} \cdot \frac{\boldsymbol{y}}{\|\boldsymbol{y}\|}.$$

For example, consider two vectors in $\mathbb{R}^2$, $\boldsymbol{u} = (1/2, 0)$ and $\boldsymbol{v} = (\sqrt{2}/2, \sqrt{2}/2)$, which are depicted in Figure 9.16.

```
u = np.array([0.5, 0])
v = np.array([np.sqrt(2)/2, np.sqrt(2)/2])
```

Their dot product is equal to:

```
np.sum(u*v)
## 0.3535533905932738
```

The dot product of their normalised versions, i.e., the cosine of the angle between them is:

```
u_norm = u/np.sqrt(np.sum(u*u))
v_norm = v/np.sqrt(np.sum(v*v))   # BTW: this vector is already normalised
np.sum(u_norm*v_norm)
## 0.7071067811865476
```

The angle itself can be determined by referring to the inverse of the cosine function, i.e., arccosine.

```
np.arccos(np.sum(u_norm*v_norm)) * 180/np.pi
## 45.0
```

Notice that we converted the angle from radians to degrees.

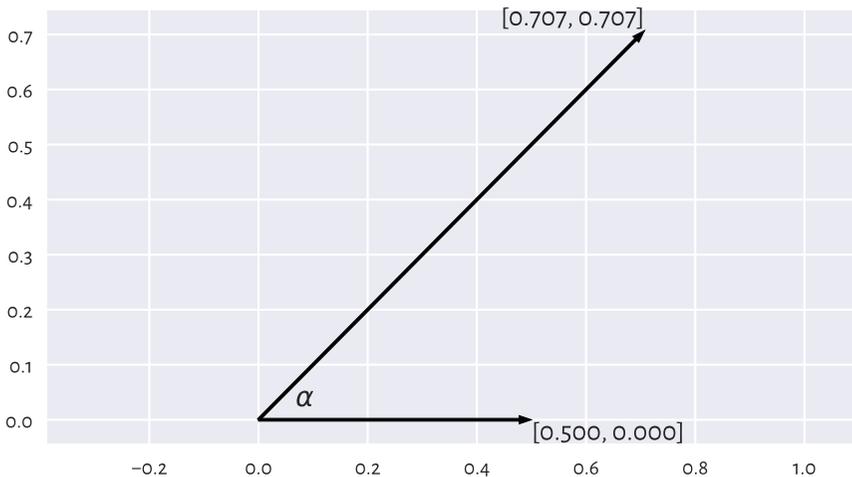

Figure 9.16: Example vectors and the angle between them

**Important**  If two vectors are collinear (*codirectional*, one is a scaled version of another, angle 0), then $\cos 0 = 1$. If they point in opposite directions ($\pm\pi = \pm 180°$ angle), then $\cos \pm \pi = -1$. For vectors that are *orthogonal* (perpendicular, $\pm \frac{\pi}{2} = \pm 90^c irc$ angle), we get $\cos \pm \frac{\pi}{2} = 0$.

---

**Note** (\*\*) The standard deviation $s$ of a vector $x \in \mathbb{R}^n$ that has already been centred (whose components' mean is 0) is a scaled version of its magnitude, i.e., $s = \|x\|/\sqrt{n}$. Looking at the definition of the Pearson linear correlation coefficient in Section 9.1.1, we see that it is the dot product of the standardised versions of two vectors $x$ and $y$ divided by the number of elements therein. If the vectors are centred, we can rewrite the formula equivalently as $r(x, y) = \frac{x}{\|x\|} \cdot \frac{y}{\|y\|}$ and thus $r(x, y) = \cos \alpha$. It is not easy to imagine vectors in high-dimensional spaces, but from this observation we can at least imply the fact that $r$ is bounded between -1 and 1. In this context, being not linearly correlated corresponds to the vectors' orthogonality.

## 9.3.2 Geometric Transformations of Points

For certain square matrices of size $m$-by-$m$, matrix multiplication can be thought of as an application of the corresponding geometrical transformation of points in $\mathbb{R}^m$

Let $\mathbf{X}$ be a matrix of shape $n$-by-$m$, which we treat as representing the coordinates of $n$ points in an $m$-dimensional space. For instance, if we are given a diagonal matrix:

$$\mathbf{S} = \text{diag}(s_1, s_2, \ldots, s_m) = \begin{bmatrix} s_1 & 0 & \ldots & 0 \\ 0 & s_2 & \ldots & 0 \\ \vdots & \vdots & \ddots & \vdots \\ 0 & 0 & \ldots & s_m \end{bmatrix},$$

then $\mathbf{XS}$ represents *scaling* (stretching) with respect to the individual axes of the coordinate system, because:

$$\mathbf{XS} = \begin{bmatrix} s_1 x_{1,1} & s_2 x_{1,2} & \ldots & s_m x_{1,m} \\ s_1 x_{2,1} & s_2 x_{2,2} & \ldots & s_m x_{2,m} \\ \vdots & \vdots & \ddots & \vdots \\ s_1 x_{n-1,1} & s_2 x_{n-1,2} & \ldots & s_m x_{n-1,m} \\ s_1 x_{n,1} & s_2 x_{n,2} & \ldots & s_m x_{n,m} \end{bmatrix}.$$

In `numpy`, this can be implemented without referring to the matrix multiplication. A notation like X * np.array([s1, s2, ..., sm]).reshape(1, -1) will suffice (elementwise multiplication and proper shape broadcasting).

Furthermore, let $\mathbf{Q}$ is an *orthonormal*[15] matrix, i.e., a square matrix whose columns and rows are unit vectors (normalised), all orthogonal to each other:

- $\|\mathbf{q}_{i,\cdot}\| = 1$ for all $i$,

---

[15] Orthonormal matrices are sometimes simply referred to as orthogonal ones.

- $\mathbf{q}_{i,\cdot} \cdot \mathbf{q}_{k,\cdot} = 0$ for all $i, k$,
- $\|\mathbf{q}_{\cdot,j}\| = 1$ for all $j$,
- $\mathbf{q}_{\cdot,j} \cdot \mathbf{q}_{\cdot,k} = 0$ for all $j, k$.

In such a case, $\mathbf{XQ}$ represents a combination of rotations and reflections.

---

**Important** By definition, a matrix $\mathbf{Q}$ is *orthonormal* if and only if $\mathbf{Q}^T \mathbf{Q} = \mathbf{Q}\mathbf{Q}^T = \mathbf{I}$. It is due to the $\cos \pm \frac{\pi}{2} = 0$ interpretation of the dot products of normalised orthogonal vectors.

---

In particular, for any angle $\alpha$, the matrix representing the corresponding rotation in $\mathbb{R}^2$:

$$\mathbf{R}(\alpha) = \left[ \begin{array}{cc} \cos \alpha & \sin \alpha \\ -\sin \alpha & \cos \alpha \end{array} \right],$$

is orthonormal (which can be easily verified using the basic trigonometric equalities).

Furthermore:

$$\left[ \begin{array}{cc} 1 & 0 \\ 0 & -1 \end{array} \right] \quad \text{and} \quad \left[ \begin{array}{cc} -1 & 0 \\ 0 & 1 \end{array} \right],$$

represent the two reflections, one against the x- and the other against the y-axis, respectively. Both are orthonormal matrices as well.

Consider a dataset $\mathbf{X}'$ in $\mathbb{R}^2$:

```
np.random.seed(12345)
Xp = np.random.randn(10000, 2) * 0.25
```

and its scaled, rotated, and translated (shifted) version:

$$\mathbf{X} = \mathbf{X}' \left[ \begin{array}{cc} 2 & 0 \\ 0 & 0.5 \end{array} \right] \left[ \begin{array}{cc} \cos \frac{\pi}{6} & \sin \frac{\pi}{6} \\ -\sin \frac{\pi}{6} & \cos \frac{\pi}{6} \end{array} \right] + \left[ \begin{array}{cc} 3 & 2 \end{array} \right].$$

```
t = np.array([3, 2])
S = np.diag([2, 0.5])
S
## array([[2. , 0. ],
##        [0. , 0.5]])
alpha = np.pi/6
Q = np.array([
    [ np.cos(alpha), np.sin(alpha)],
    [-np.sin(alpha), np.cos(alpha)]
```

```
])
Q
## array([[ 0.8660254,  0.5      ],
##        [-0.5      ,  0.8660254]])
X = Xp @ S @ Q + t
```

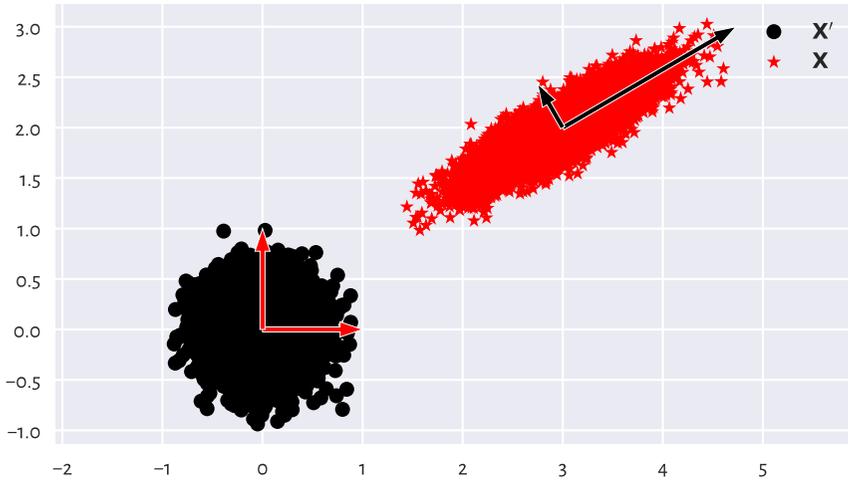

Figure 9.17: A dataset and its scaled, rotated, and shifted version

We can consider $\mathbf{X} = \mathbf{X'SQ} + \mathbf{t}$ a version of $\mathbf{X'}$ in a new coordinate system (basis), see Figure 9.17. Each column in the transformed matrix is a shifted linear combination of the columns in the original matrix:

$$\mathbf{x}_{\cdot,j} = t_j + \sum_{k=1}^{m} (s_{k,k} q_{k,j}) \mathbf{x'}_{\cdot,k}.$$

The computing of such linear combinations of columns is not rare during a dataset's preprocessing step, especially if they are on the same scale or are unitless. As a matter of fact, the standardisation itself is a form of scaling and translation.

**Exercise 9.14**  *Assume that we have a dataset with two columns, giving the number of apples and the number of oranges in clients' baskets. What orthonormal and scaling transforms should be applied to obtain a matrix bearing the total number of fruits and surplus apples (e.g., a row $(4, 7)$ should be converted to $(11, -3)$)?*

### 9.3.3   Matrix Inverse

The *inverse* of a square matrix $\mathbf{A}$ (if it exists) is denoted with $\mathbf{A}^{-1}$ – it is the matrix fulfilling the identity:

$$\mathbf{A}^{-1}\mathbf{A} = \mathbf{A}\mathbf{A}^{-1} = \mathbf{I}.$$

Noting that the identity matrix $\mathbf{I}$ is the neutral element of the matrix multiplication, the above is thus the analogue of the inverse of a scalar: something like $3 \cdot 3^{-1} = 3 \cdot \frac{1}{3} = \frac{1}{3} \cdot 3 = 1$.

---

**Important**   For any invertible matrices of admissible shapes, it might be shown that the following noteworthy properties hold:

- $(\mathbf{A}^{-1})^T = (\mathbf{A}^T)^{-1}$,

- $(\mathbf{AB})^{-1} = \mathbf{B}^{-1}\mathbf{A}^{-1}$,

- a matrix equality $\mathbf{A} = \mathbf{BC}$ holds if and only if $\mathbf{AC}^{-1} = \mathbf{BCC}^{-1} = \mathbf{B}$; this is also equivalent to $\mathbf{B}^{-1}\mathbf{A} = \mathbf{B}^{-1}\mathbf{BC} = \mathbf{C}$.

---

Matrix inverse allows us to identify the inverses of geometrical transformations. Knowing that $\mathbf{X} = \mathbf{X}'\mathbf{SQ} + \mathbf{t}$, we can recreate the original matrix by applying:

$$\mathbf{X}' = (\mathbf{X} - \mathbf{t})(\mathbf{SQ})^{-1} = (\mathbf{X} - \mathbf{t})\mathbf{Q}^{-1}\mathbf{S}^{-1}.$$

It is worth knowing that if $\mathbf{S} = \mathrm{diag}(s_1, s_2, \ldots, s_m)$ is a diagonal matrix, then its inverse is $\mathbf{S}^{-1} = \mathrm{diag}(1/s_1, 1/s_2, \ldots, 1/s_m)$, which we can denote as $(1/\mathbf{S})$. In addition, the inverse of an orthonormal matrix $\mathbf{Q}$ is always equal to its transpose, $\mathbf{Q}^{-1} = \mathbf{Q}^T$. Luckily, we will not be inverting other matrices in this introductory course.

As a consequence:

$$\mathbf{X}' = (\mathbf{X} - \mathbf{t})\mathbf{Q}^T(1/\mathbf{S}).$$

Let us verify this numerically (testing equality up to some inherent round-off error):

```
np.allclose(Xp, (X-t) @ Q.T @ np.diag(1/np.diag(S)))
## True
```

### 9.3.4   Singular Value Decomposition

It turns out that given any real $n$-by-$m$ matrix $\mathbf{X}$ with $n \geq m$, we can find an interesting scaling and orthonormal transform that, when applied on a dataset whose columns are already normalised, yields exactly $\mathbf{X}$.

Namely, the singular value decomposition (SVD in the so-called compact form) is a factorisation:

$$\mathbf{X} = \mathbf{USQ},$$

where:

- **U** is an $n$-by-$m$ semi-orthonormal matrix (its columns are orthonormal vectors; it holds $\mathbf{U}^T \mathbf{U} = \mathbf{I}$),

- **S** is an $m$-by-$m$ diagonal matrix such that $s_{1,1} \geq s_{2,2} \geq ... \geq s_{m,m} \geq 0$,

- **Q** is an $m$-by-$m$ orthonormal matrix.

---

**Important**   In data analysis, we usually apply the SVD on matrices that have already been centred (so that their column means are all 0).

---

For example:

```python
import scipy.linalg
n = X.shape[0]
X_centred = X - np.mean(X, axis=0)
U, s, Q = scipy.linalg.svd(X_centred, full_matrices=False)
```

And now:

```python
U[:6, :]  # preview first few rows
## array([[-0.00195072,  0.00474569],
##        [-0.00510625, -0.00563582],
##        [ 0.01986719,  0.01419324],
##        [ 0.00104386,  0.00281853],
##        [ 0.00783406,  0.01255288],
##        [ 0.01025205, -0.0128136 ]])
```

The norms of all the columns in **U** are all equal to 1 (and hence standard deviations are $1/\sqrt{n}$). Consequently, they are on the same scale:

```python
np.std(U, axis=0), 1/np.sqrt(n)  # compare
## (array([0.01, 0.01]), 0.01)
```

What is more, they are orthogonal: their dot products are all equal to 0. Regarding what we said about Pearson's linear correlation coefficient and its relation to dot products of normalised vectors, we imply that the columns in **U** are not linearly correlated. In some sense, they form *independent* dimensions.

Now, it holds $\mathbf{S} = \mathrm{diag}(s_1, ..., s_m)$, with the elements on the diagonal being:

```python
s
## array([49.72180455, 12.5126241 ])
```

The elements on the main diagonal of **S** are used to scale the corresponding columns in **U**. The fact that they are ordered decreasingly means that the first column in **US** has the greatest standard deviation, the second column has the second greatest variability, and so forth.

```
S = np.diag(s)
US = U @ S
np.std(US, axis=0)  # equal to s/np.sqrt(n)
## array([0.49721805, 0.12512624])
```

Multiplying **US** by **Q** simply rotates and/or reflects the dataset. This brings **US** to a new coordinate system where, by construction, the dataset projected onto the direction determined by the first row in **Q**, i.e., $\mathbf{q}_{1,\cdot}$ has the largest variance, projection onto $\mathbf{q}_{2,\cdot}$ has the second largest variance, and so on.

```
Q
## array([[ 0.86781968,  0.49687926],
##        [-0.49687926,  0.86781968]])
```

This is why we refer to the rows in **Q** as *principal directions* (or *components*). Their scaled versions (proportional to the standard deviations along them) are depicted in Figure 9.18. Note that we have more or less recreated the steps needed to construct **X** from **X′** above (by the way we generated **X′**, we expect it to have linearly uncorrelated columns; yet, **X′** and **U** have different column variances).

```
plt.plot(X_centred[:, 0], X_centred[:, 1], "o", alpha=0.1)
plt.arrow(
    0, 0, Q[0, 0]*s[0]/np.sqrt(n), Q[0, 1]*s[0]/np.sqrt(n), width=0.02,
    facecolor="red", edgecolor="white", length_includes_head=True, zorder=2)
plt.arrow(
    0, 0, Q[1, 0]*s[1]/np.sqrt(n), Q[1, 1]*s[1]/np.sqrt(n), width=0.02,
    facecolor="red", edgecolor="white", length_includes_head=True, zorder=2)
plt.show()
```

### 9.3.5 Dimensionality Reduction with SVD

Let us consider the following example three-dimensional dataset.

```
chainlink = np.loadtxt("https://raw.githubusercontent.com/gagolews/" +
    "teaching-data/master/clustering/fcps_chainlink.csv")
```

As we said in Section 7.4, the plotting is always done on a two-dimensional surface (be it the computer screen or book page). We can look at the dataset only from one *angle* at a time.

In particular, a scatterplot matrix only depicts the dataset from the perspective of the axes of the Cartesian coordinate system (standard basis); see Figure 9.19.

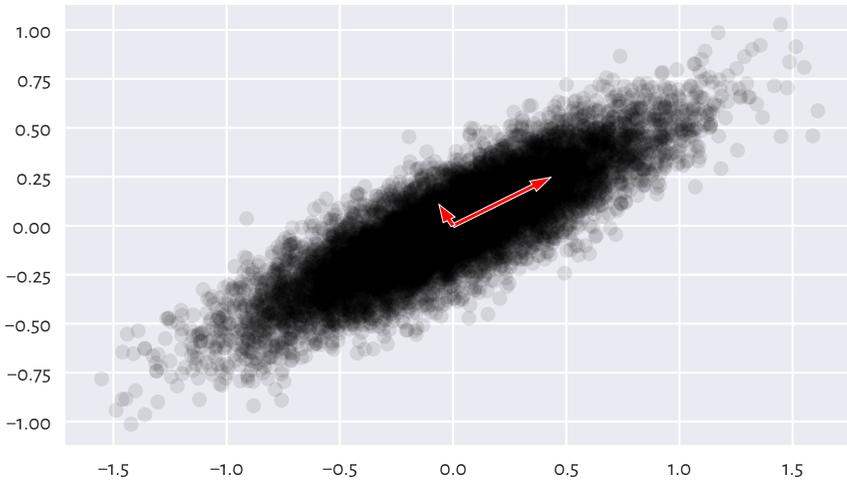

Figure 9.18: Principal directions of an example dataset (scaled so that they are proportional to the standard deviations along them)

```
sns.pairplot(data=pd.DataFrame(chainlink))
# plt.show()  # not needed :/
```

These viewpoints by no means must reveal the true geometric structure of the dataset. However, we know that we can rotate the virtual camera and find some more *interesting* angle. It turns out that our dataset represents two nonintersecting rings, hopefully visible Figure 9.20.

```
fig = plt.figure()
ax = fig.add_subplot(1, 3, 1, projection="3d", facecolor="#ffffff00")
ax.scatter(chainlink[:, 0], chainlink[:, 1], chainlink[:, 2])
ax.view_init(elev=45, azim=45, vertical_axis="z")
ax = fig.add_subplot(1, 3, 2, projection="3d", facecolor="#ffffff00")
ax.scatter(chainlink[:, 0], chainlink[:, 1], chainlink[:, 2])
ax.view_init(elev=37, azim=0, vertical_axis="z")
ax = fig.add_subplot(1, 3, 3, projection="3d", facecolor="#ffffff00")
ax.scatter(chainlink[:, 0], chainlink[:, 1], chainlink[:, 2])
ax.view_init(elev=10, azim=150, vertical_axis="z")
plt.show()
```

It turns out that we may find a noteworthy viewpoint using the SVD. Namely, we can perform the decomposition of a centred dataset which we denote with $\mathbf{X}$:

$$\mathbf{X} = \mathbf{USQ}.$$

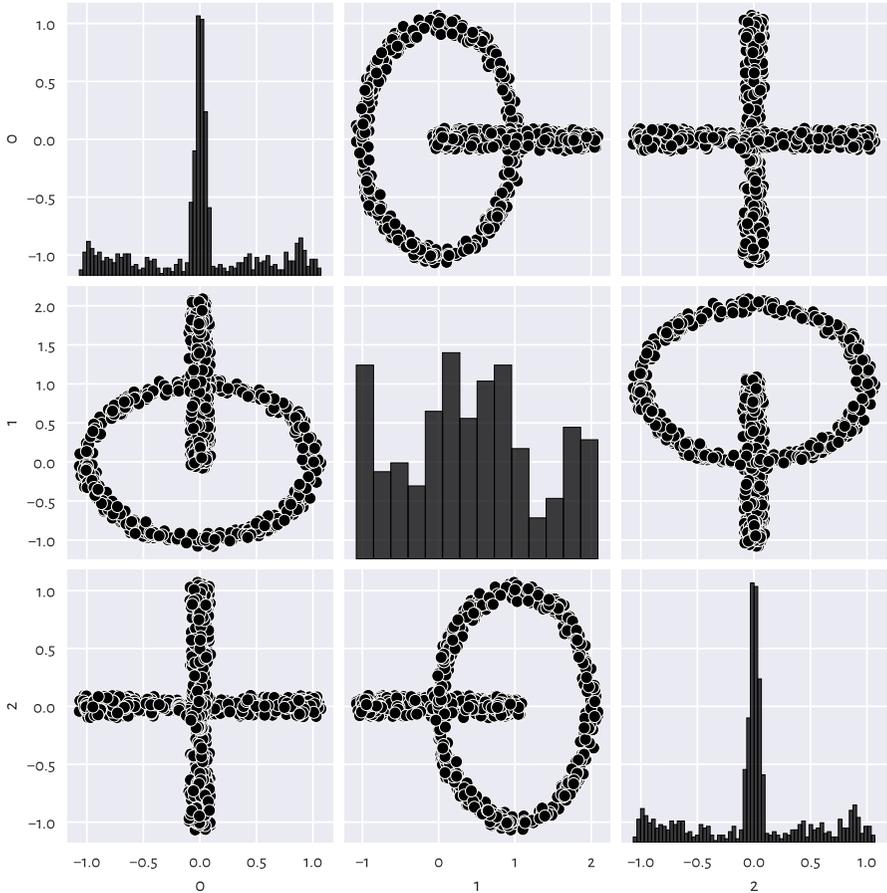

Figure 9.19: Views from the perspective of the main axes

```python
import scipy.linalg
X_centered = chainlink-np.mean(chainlink, axis=0)
U, s, Q = scipy.linalg.svd(X_centered, full_matrices=False)
```

Then, considering its rotated/reflected version:

$$\mathbf{P} = \mathbf{XQ}^{-1} = \mathbf{US},$$

we know that its first column has the highest variance, the second column has the second highest variability, and so on. It might indeed be worth looking at that dataset from that *most informative* perspective.

Figure 9.21 gives the scatter plot for $\mathbf{p}_{\cdot,1}$ and $\mathbf{p}_{\cdot,2}$. Maybe this does not reveal the true geometric structure of the dataset (no single two-dimensional projection can do that), but at least it is better than the initial ones (from the pairplot).

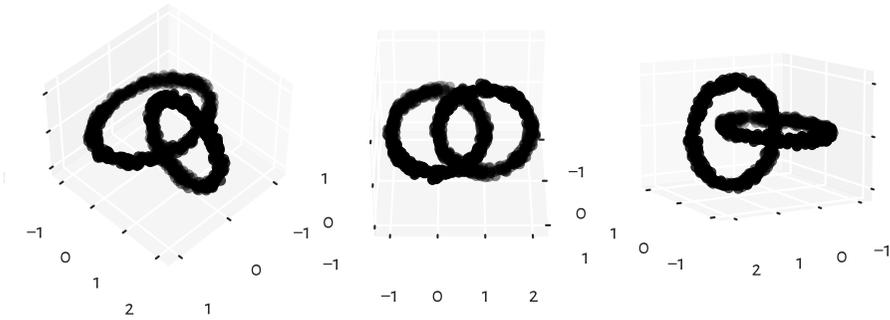

Figure 9.20: Different views on the same dataset

```
P2 = U[:, :2] @ np.diag(s[:2])  # the same as (U@np.diag(s))[:, :2]
plt.plot(P2[:, 0], P2[:, 1], "o")
plt.axis("equal")
plt.show()
```

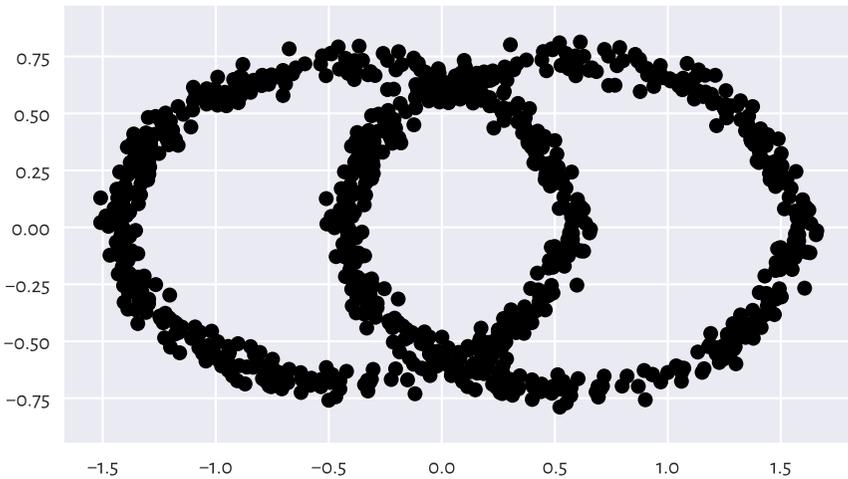

Figure 9.21: The view from the two principal axes

What we just did is a kind of *dimensionality reduction*. We found a viewpoint (in the form of an orthonormal matrix, being a mixture of rotations and reflections) on **X** such that its orthonormal projection onto the first two axes of the Cartesian coordinate system is the most informative[16] (in terms of having the highest variance along these axes).

### 9.3.6    Principal Component Analysis

*Principal component analysis* (PCA) is a fancy name for the entire process involving our brainstorming upon what happens along the projections onto the most variable dimensions. It can be used not only for data visualisation and deduplication, but also for feature engineering (as in fact it creates new columns that are linear combinations of existing ones).

Let us consider a few chosen countrywise 2016 Sustainable Society Indices[17].

```
ssi = pd.read_csv("https://raw.githubusercontent.com/gagolews/" +
    "teaching-data/master/marek/ssi_2016_indicators.csv",
    comment="#")
X = np.array(ssi.iloc[:, [3, 5, 13, 15, 19] ])  # select columns, make matrix
n = X.shape[0]
X[:6, :]  # preview
## array([[ 9.32      , 8.13333333, 8.386     , 8.5757    , 5.46249573],
##        [ 8.74      , 7.71666667, 7.346     , 6.8426    , 6.2929302 ],
##        [ 5.11      , 4.31666667, 8.788     , 9.2035    , 3.91062849],
##        [ 9.61      , 7.93333333, 5.97      , 5.5232    , 7.75361284],
##        [ 8.95      , 7.81666667, 8.032     , 8.2639    , 4.42350654],
##        [10.        , 8.65      , 1.        , 1.        , 9.66401848]])
```

Each index is on the scale from 0 to 10. These are, in this order:

1. Safe Sanitation,

2. Healthy Life,

3. Energy Use,

4. Greenhouse Gases,

5. Gross Domestic Product.

Above we displayed the data corresponding to the 6 following countries:

```
countries = list(ssi.iloc[:, 0])  # select the 1st column from the data frame
countries[:6]  # preview
## ['Albania', 'Algeria', 'Angola', 'Argentina', 'Armenia', 'Australia']
```

This is a five-dimensional dataset. We cannot easily visualise it. That the pairplot does not reveal much is left as an exercise. Let us thus perform the SVD decomposition of a standardised version of this dataset, **Z** (recall that the centring is necessary, at the very least).

---

[16] (**) The Eckart–Young–Mirsky theorem states that $\mathbf{U}_{\cdot,:k}\mathbf{S}_{:k,:k}\mathbf{Q}_{:k,\cdot}$ (where ":$k$" denotes "first $k$ rows or columns") is the best rank-$k$ approximation of **X** with respect to both the Frobenius and spectral norms.
[17] https://ssi.wi.th-koeln.de/

```
Z = (X - np.mean(X, axis=0))/np.std(X, axis=0)
U, s, Q = scipy.linalg.svd(Z, full_matrices=False)
```

The standard deviations of the data projected onto the consecutive principal components (columns in **US**) are:

```
s/np.sqrt(n)
## array([2.02953531, 0.7529221 , 0.3943008 , 0.31897889, 0.23848286])
```

It is customary to check the ratios of the cumulative variances explained by the consecutive principal components, which is a normalised measure of their importances. We can compute them by calling:

```
np.cumsum(s**2)/np.sum(s**2)
## array([0.82380272, 0.93718105, 0.96827568, 0.98862519, 1.        ])
```

As in some sense the variability within the first two components covers ca. 94% of the variability of the whole dataset, we can restrict ourselves only to a two-dimensional projection of this dataset (actually, we are quite lucky here – or someone has selected these countrywise indices for us in a very clever fashion).

The rows in **Q** feature the so-called *loadings*. They give the coefficients defining the linear combinations of the rows in **Z** that correspond to the principal components.

Let us try to interpret them.

```
np.round(Q[0, :], 2)  # loadings – the 1st principal axis
## array([-0.43, -0.43,  0.44,  0.45, -0.47])
```

The first row in **Q** consists of similar values, but with different signs. We can consider them a scaled version of the average Energy Use (column 3), Greenhouse Gases (4), and MINUS Safe Sanitation (1), MINUS Healthy Life (2), MINUS Gross Domestic Product (5). We could call this a measure of a country's overall eco-unfriendliness(?), because countries with low Healthy Life and high Greenhouse Gasses will score highly on this scale.

```
np.round(Q[1, :], 2)  # loadings – the 2nd principal axis
## array([ 0.52,  0.5 ,  0.52,  0.45, -0.02])
```

The second row in **Q** defines a scaled version of the average of Safe Sanitation (1), Healthy Life (2), Energy Use (3), and Greenhouse Gases (4), almost completely ignoring the GDP (5). Should we call it a measure of industrialisation? Something like this. But this naming is just for fun[18].

---

[18] Although someone might take these results seriously and write, for example, a research thesis about

Figure 9.22 is a scatter plot of the countries projected onto the said two principal directions. For readability, we only display a few chosen labels. This is merely a projection/approximation, but it might be an interesting one for some practitioners.

```python
P2 = U[:, :2] @ np.diag(s[:2])  # == Y @ Q[:2, :].T
plt.plot(P2[:, 0], P2[:, 1], "o", alpha=0.1)
which = [   # hand-crafted/artisan
    141, 117, 69, 123, 35, 80, 93, 45, 15, 2, 60, 56, 14,
    104, 122, 8, 134, 128, 0, 94, 114, 50, 34, 41, 33, 77,
    64, 67, 152, 135, 148, 99, 149, 126, 111, 57, 20, 63
]
for i in which:
    plt.text(P2[i, 0], P2[i, 1], countries[i], ha="center")
plt.axis("equal")
plt.xlabel("1st principal component (eco-unfriendliness?)")
plt.ylabel("2nd principal component (industrialisation?)")
plt.show()
```

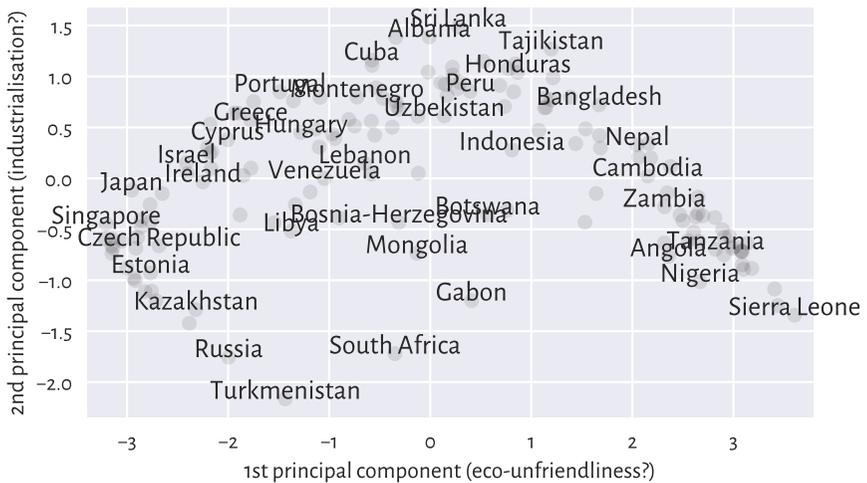

Figure 9.22: An example principal component analysis of countries

---

it. Mathematics – unlike the brains of ordinary mortals – does not need our imperfect interpretations/fairy tales to function properly. We need more maths in our lives.

## 9.4 Further Reading

Other approaches to regression via linear models include ridge and lasso, the latter having the nice property of automatically getting rid of noninformative variables from the model. Furthermore, instead of minimising squared residuals, we can also consider, e.g., least absolute deviation.

There are of course many other approaches to dimensionality reduction, also nonlinear ones, including kernel PCA, feature agglomeration via hierarchical clustering, autoencoders, t-SNE, etc.

A popular introductory text in statistical learning is [42]. We recommend [6, 7, 20] for more advanced students.

## 9.5 Exercises

**Exercise 9.15** *Why correlation is not causation?*

**Exercise 9.16** *What does the linear correlation of 0.9 mean? How about the rank correlation of 0.9? And the linear correlation of 0.0?*

**Exercise 9.17** *How is Spearman's coefficient related to Pearson's one?*

**Exercise 9.18** *State the optimisation problem behind the least squares fitting of linear models.*

**Exercise 9.19** *What are the different ways of the numerical summarising of residuals?*

**Exercise 9.20** *Why is it important for the residuals to be homoscedastic?*

**Exercise 9.21** *Is a more complex model always better?*

**Exercise 9.22** *Why should extrapolation be handled with care?*

**Exercise 9.23** *Why did we say that novice users should refrain from using* `scikit-learn`*?*

**Exercise 9.24** *What is the condition number of a model matrix and why should we always check it?*

**Exercise 9.25** *What is the geometrical interpretation of the dot product of two normalised vectors?*

**Exercise 9.26** *How can we verify if two vectors are orthonormal? What is an orthonormal projection? What is the inverse of an orthonormal matrix?*

**Exercise 9.27** *What is the inverse of a diagonal matrix?*

**Exercise 9.28** *Characterise the general properties of the three matrices obtained by performing the singular value decomposition of a given matrix of shape* n-by-m.

**Exercise 9.29** *How can we obtain the first principal component of a given centred matrix?*

**Exercise 9.30** *How can we compute the ratios of the variances explained by the consecutive principal components?*

# Part IV

# Heterogeneous Data

## *Introducing Data Frames*

**numpy** arrays are an extremely versatile tool for performing data analysis exercises and other numerical computations of various kinds. Although theoretically possible otherwise, in practice we only store elements of the same type therein, most often numbers.

**pandas**[1] [56] is amongst over one hundred thousand[2] open-source packages and repositories that use **numpy** to provide additional data wrangling functionality. It was originally written by Wes McKinney but was heavily inspired by the `data.frame`[3] objects in S and R as well as tables in relational (think: SQL) databases and spreadsheets.

**pandas** delivers a few classes, of which the most important are:

- `DataFrame` – for representing tabular data (matrix-like) with columns of possibly different types, in particular a mix of numerical and categorical variables,

- `Series` – vector-like objects for representing individual columns,

- `Index` – vector-like (usually) objects for labelling individual rows and columns in `DataFrames` and items in `Series` objects,

together with many methods for:

- transforming/aggregating/processing data, also in groups determined by categorical variables or products thereof,

- reshaping (e.g., from wide to long format) and joining datasets,

- importing/exporting data from/to various sources and formats, e.g., CSV and HDF5 files or relational databases,

- handling missing data,

all of which we introduce in this part.

Before we delve into the world of **pandas**, let us point out that it is customary to load this package under the following alias:

---

[1] https://pandas.pydata.org/
[2] https://libraries.io/pypi/numpy
[3] Data frames were first introduced in the 1991 version of the S language [11].

```python
import pandas as pd
```

---

**Important**  Let us repeat: `pandas` is built on top of **numpy** and most objects therein can be processed by **numpy** functions as well. Many other functions (e.g., in **sklearn**) accept both `DataFrame` and `ndarray` objects, but often convert the former to the latter internally to enable data processing using fast C/C++/Fortran routines.

What we have learned so far[4] still applies. But there is of course more, hence this part.

---

## 10.1  Creating Data Frames

Data frames can be created, amongst others, using the `DataFrame` class constructor, which can be fed, for example, with a **numpy** matrix:

```python
np.random.seed(123)
pd.DataFrame(
    np.random.rand(4, 3),
    columns=["a", "b", "c"]
)
##           a         b         c
## 0  0.696469  0.286139  0.226851
## 1  0.551315  0.719469  0.423106
## 2  0.980764  0.684830  0.480932
## 3  0.392118  0.343178  0.729050
```

Notice that rows and columns are labelled (and how readable that is).

A dictionary of vector-like objects of equal lengths is another common option:

```python
np.random.seed(123)
pd.DataFrame(dict(
    u=np.round(np.random.rand(5), 2),
    v=[True, True, False, False, True],
    w=["A", "B", "C", "D", "E"],
    x=["spam", "spam", "bacon", "spam", "eggs"]
))
##       u      v  w      x
## 0  0.70   True  A   spam
## 1  0.29   True  B   spam
```

*(continues on next page)*

---

[4] If by any chance the kind reader frivolously decided to start their journey with this superb book at this chapter, it is now the time to go back to the *Preface* and learn everything in the right order. See you later.

```
## 2   0.23   False   C   bacon
## 3   0.55   False   D    spam
## 4   0.72    True   E    eggs
```

This illustrates the possibility of having columns of different types.

**Exercise 10.1**  *Check out **pandas.DataFrame.from_dict** and **from_records** in the document-ation[5]. Use them to create some example data frames.*

Further, data frames can be read from files in different formats, for instance, CSV:

```
body = pd.read_csv("https://raw.githubusercontent.com/gagolews/" +
    "teaching-data/master/marek/nhanes_adult_female_bmx_2020.csv",
    comment="#")
body.head()   # display first few rows (5 by default)
##     BMXWT  BMXHT  BMXARML  BMXLEG  BMXARMC  BMXHIP  BMXWAIST
## 0   97.1  160.2     34.7    40.8     35.8   126.1     117.9
## 1   91.1  152.7     33.5    33.0     38.5   125.5     103.1
## 2   73.0  161.2     37.4    38.0     31.8   106.2      92.0
## 3   61.7  157.4     38.0    34.7     29.0   101.0      90.5
## 4   55.4  154.6     34.6    34.0     28.3    92.5      73.2
```

Reading from URLs and local files is of course supported; compare Section 13.6.1.

**Exercise 10.2**  *Check out the other **pandas.read_*** functions in the **pandas** documentation – some of which we shall be discussing later.*

## 10.1.1  Data Frames Are Matrix-Like

Data frames are modelled through **numpy** matrices. We can thus already feel quite at home with them.

For example, given:

```
np.random.seed(123)
df = pd.DataFrame(dict(
    u=np.round(np.random.rand(5), 2),
    v=[True, True, False, False, True],
    w=["A", "B", "C", "D", "E"],
    x=["spam", "spam", "bacon", "spam", "eggs"]
))
df
##       u      v   w      x
## 0   0.70   True  A   spam
## 1   0.29   True  B   spam
```

---

[5] https://pandas.pydata.org/docs/

```
## 2  0.23  False  C  bacon
## 3  0.55  False  D   spam
## 4  0.72   True  E   eggs
```

it is easy to fetch the number of rows and columns:

```
df.shape
## (5, 4)
```

or the type of each column:

```
df.dtypes  # returns a Series object; see below
## u    float64
## v       bool
## w     object
## x     object
## dtype: object
```

Recall that arrays are equipped with the `dtype` slot.

### 10.1.2  Series

There is a separate class for storing individual data frame columns: it is called `Series`.

```
s = df.loc[:, "u"]  # extract the `u` column; alternatively: d.u
s
## 0    0.70
## 1    0.29
## 2    0.23
## 3    0.55
## 4    0.72
## Name: u, dtype: float64
```

Data frames with one column are printed out slightly differently. We get the column name at the top, but do not have the `dtype` information at the bottom.

```
s.to_frame()  # or: pd.DataFrame(s)
##       u
## 0  0.70
## 1  0.29
## 2  0.23
## 3  0.55
## 4  0.72
```

Indexing will of course be discussed later.

**Important**  It is crucial to know when we are dealing with a `Series` and when with a `DataFrame` object, because each of them defines a slightly different set of methods.

We will now be relying upon object-oriented syntax (compare Section 2.2.3) much more frequently than before.

**Example 10.3**  *By calling:*

```
s.mean()
## 0.49800000000000005
```

*we refer to* **pandas.Series.mean** *(which returns a scalar), whereas:*

```
df.mean(numeric_only=True)
## u    0.498
## v    0.600
## dtype: float64
```

*uses* **pandas.DataFrame.mean** *(which yields a* `Series`*).*

*Look up these two methods in the* **pandas** *manual. Note that their argument list is slightly different.*

`Series` are vector-like objects:

```
s.shape
## (5,)
s.dtype
## dtype('float64')
```

They are wrappers around **numpy** arrays.

```
s.values
## array([0.7 , 0.29, 0.23, 0.55, 0.72])
```

Most importantly, **numpy** functions can be called directly on them:

```
np.mean(s)
## 0.49800000000000005
```

As a consequence, what we covered in the part of this book that dealt with vector processing still holds for data frame columns (but there will be more).

`Series` can also be *named*.

```
s.name
## 'u'
```

This is convenient when we convert them to a data frame, because the `name` sets the label of the newly created column:

```
s.rename("spam").to_frame()
##    spam
## 0  0.70
## 1  0.29
## 2  0.23
## 3  0.55
## 4  0.72
```

### 10.1.3  Index

Another important class is called `Index`[6]. The `index` (lowercase) *slot* of a data frame stores an `Index` object that gives the row labels:

```
df.index  # row labels
## RangeIndex(start=0, stop=5, step=1)
```

The above represents a sequence (0, 1, 2, 3, 4).

Furthermore, the `column` slot gives:

```
df.columns  # column labels
## Index(['u', 'v', 'w', 'x'], dtype='object')
```

Also, we can label the individual elements in `Series` objects:

```
s.index
## RangeIndex(start=0, stop=5, step=1)
```

The **set_index** method is a quite frequent operation. We will be applying it to make a data frame column act as a vector of row labels:

```
df2 = df.set_index("x")
df2
##             u      v  w
## x
## spam   0.70   True  A
## spam   0.29   True  B
## bacon  0.23  False  C
## spam   0.55  False  D
## eggs   0.72   True  E
```

This `Index` object is named:

---

[6] The name `Index` is confusing not only because it clashes with the *index* operator (square brackets), but also the concept of an *index* in relational databases. In **pandas**, we can have non-unique row names.

```
df2.index.name
## 'x'
```

We can also rename the axes on the fly:

```
df2.rename_axis(index="ROWS", columns="COLS")
## COLS        u      v  w
## ROWS
## spam   0.70   True  A
## spam   0.29   True  B
## bacon  0.23  False  C
## spam   0.55  False  D
## eggs   0.72   True  E
```

Having a named `index` slot is handy when we decide that we want to convert the vector of row labels back to a standalone column:

```
df2.rename_axis(index="NAME_OF_THE_NEW_COLUMN").reset_index()
##    NAME_OF_THE_NEW_COLUMN      u      v  w
## 0                    spam   0.70   True  A
## 1                    spam   0.29   True  B
## 2                   bacon   0.23  False  C
## 3                    spam   0.55  False  D
## 4                    eggs   0.72   True  E
```

There is also an option to get rid of the current `index` and to replace it with the default label sequence, i.e., 0, 1, 2, …:

```
df2.reset_index(drop=True)
##       u      v  w
## 0  0.70   True  A
## 1  0.29   True  B
## 2  0.23  False  C
## 3  0.55  False  D
## 4  0.72   True  E
```

Take note of the fact that **reset_index** and many other methods that we have used so far do not modify the data frame in place.

---

**Important**  We will soon get used to calling **reset_index**(drop=True) quite frequently, sometimes more than once in a single series of commands.

---

**Exercise 10.4**  *Use the **pandas.DataFrame.rename** method to change the name of the u column in df to spam.*

Also, a *hierarchical* index – one that is comprised of more than one level – is possible.

For example, here is a sorted (see Section 10.6.1) version of `df` with a new index based on two columns at the same time:

```
df.sort_values("x", ascending=False).set_index(["x", "v"])
##                 u   w
## x     v
## spam  True   0.70  A
##       True   0.29  B
##       False  0.55  D
## eggs  True   0.72  E
## bacon False  0.23  C
```

Note that a group of three consecutive `spam`s, when printed, does not have the series of identical labels repeated. This is for increased readability.

**Example 10.5** *Hierarchical indexes might arise after aggregating data in groups. We will discuss that in more detail in Chapter 12.*

```
nhanes = pd.read_csv("https://raw.githubusercontent.com/gagolews/" +
    "teaching-data/master/marek/nhanes_p_demo_bmx_2020.csv",
    comment="#")
res = nhanes.groupby(["RIAGENDR", "DMDBORN4"])["BMXBMI"].mean()
res  # BMI by gender and US born-ness
## RIAGENDR  DMDBORN4
## 1         1            25.734110
##           2            27.405251
## 2         1            27.120261
##           2            27.579448
##           77           28.725000
##           99           32.600000
## Name: BMXBMI, dtype: float64
```

*This returned a Series object with a hierarchical index. Let us fret not, though: **reset_index** can always come to our rescue:*

```
res.reset_index()
##     RIAGENDR  DMDBORN4      BMXBMI
## 0          1         1   25.734110
## 1          1         2   27.405251
## 2          2         1   27.120261
## 3          2         2   27.579448
## 4          2        77   28.725000
## 5          2        99   32.600000
```

## 10.2 Aggregating Data Frames

Here is an example data frame:

```python
np.random.seed(123)
d = pd.DataFrame(dict(
    u = np.round(np.random.rand(5), 2),
    v = np.round(np.random.randn(5), 2),
    w = ["spam", "bacon", "spam", "eggs", "sausage"]
), index=["a", "b", "c", "d", "e"])
d
##       u     v        w
## a  0.70  0.32     spam
## b  0.29 -0.05    bacon
## c  0.23 -0.20     spam
## d  0.55  1.98     eggs
## e  0.72 -1.62  sausage
```

All **numpy** functions can be applied directly on individual columns, i.e., objects of type Series, because they are vector-like.

```python
u = d.loc[:, "u"]  # extract the column named `u` (gives a Series; see below)
np.quantile(u, [0, 0.5, 1])
## array([0.23, 0.55, 0.72])
```

Most **numpy** functions also work if they are fed with data frames, but we will need to extract the numeric columns manually.

```python
uv = d.loc[:, ["u", "v"]]  # select two columns (a DataFrame; see below)
np.quantile(uv, [0, 0.5, 1], axis=0)
## array([[ 0.23, -1.62],
##        [ 0.55, -0.05],
##        [ 0.72,  1.98]])
```

Sometimes the results will automatically be coerced to a Series object with the index slot set appropriately:

```python
np.mean(uv, axis=0)
## u    0.498
## v    0.086
## dtype: float64
```

Many operations, for convenience, were also implemented as methods for the Series and DataFrame classes, e.g., **mean**, **median**, **min**, **max**, **quantile**, **var**, **std**, and **skew**.

```
d.mean(numeric_only=True)
## u    0.498
## v    0.086
## dtype: float64
d.quantile([0, 0.5, 1], numeric_only=True)
##          u      v
## 0.0   0.23  -1.62
## 0.5   0.55  -0.05
## 1.0   0.72   1.98
```

Also note the **describe** method, which returns a few statistics at the same time.

```
d.describe()
##              u           v
## count  5.000000   5.000000
## mean   0.498000   0.086000
## std    0.227969   1.289643
## min    0.230000  -1.620000
## 25%    0.290000  -0.200000
## 50%    0.550000  -0.050000
## 75%    0.700000   0.320000
## max    0.720000   1.980000
```

**Exercise 10.6** *Check out the **pandas.DataFrame.agg** method that can apply all aggregates given by a list of functions. Write a call equivalent to **d.describe()**.*

---

**Note** (*) Let us stress that above we see the corrected for bias (but still only asymptotically unbiased) version of standard deviation, given by $\sqrt{\frac{1}{n-1} \sum_{i=1}^{n} (x_i - \bar{x})^2}$; compare Section 5.1. In **pandas**, **std** methods assume ddof=1 by default, whereas we recall that **numpy** uses ddof=0.

```
np.round([u.std(), np.std(u), np.std(np.array(u)), u.std(ddof=0)], 3)
## array([0.228, 0.204, 0.204, 0.204])
```

This is an unfortunate inconsistency between the two packages, but please do not blame the messenger.

---

## 10.3 Transforming Data Frames

By applying the already well-known vectorised mathematical functions from **numpy**, we can transform each data cell and return an object of the same type as the input one.

```
np.exp(d.loc[:, "u"])
## a    2.013753
## b    1.336427
## c    1.258600
## d    1.733253
## e    2.054433
## Name: u, dtype: float64
np.exp(d.loc[:, ["u", "v"]])
##           u          v
## a  2.013753  1.377128
## b  1.336427  0.951229
## c  1.258600  0.818731
## d  1.733253  7.242743
## e  2.054433  0.197899
```

When applying the binary arithmetic, comparison, and logical operators on an object of class `Series` and a scalar or a **numpy** vector, the operations are performed element-wisely – a style with which we are already familiar.

For instance, here is a standardised version of the `u` column:

```
u = d.loc[:, "u"]
(u - np.mean(u)) / np.std(u)
## a     0.990672
## b    -1.020098
## c    -1.314357
## d     0.255025
## e     1.088759
## Name: u, dtype: float64
```

Binary operators act on the elements with corresponding labels. For two objects having identical `index` slots (this is the most common scenario), this is the same as elementwise vectorisation. For instance:

```
d.loc[:, "u"] > d.loc[:, "v"]   # here: elementwise comparison
## a      True
## b      True
## c      True
## d     False
```

```
## e      True
## dtype: bool
```

For transforming many numerical columns at once, it is a good idea either to convert them to a numeric matrix explicitly and then use the basic **numpy** functions:

```
uv = np.array(d.loc[:, ["u", "v"]])
uv2 = (uv-np.mean(uv, axis=0))/np.std(uv, axis=0)
uv2
## array([[ 0.99067229,  0.20826225],
##        [-1.0200982 , -0.11790285],
##        [-1.3143573 , -0.24794275],
##        [ 0.25502455,  1.64197052],
##        [ 1.08875866, -1.47898717]])
```

or to use the **pandas.DataFrame.apply** method which invokes a given function on each column separately:

```
uv2 = d.loc[:, ["u", "v"]].apply(lambda x: (x-np.mean(x))/np.std(x))
uv2
##            u         v
## a  0.990672  0.202862
## b -1.020098 -0.117903
## c -1.314357 -0.247943
## d  0.255025  1.641971
## e  1.088759 -1.478987
```

Anticipating what we cover in the next section, in both cases, we can write d.loc[:, ["u", "v"]] = uv2 to replace the old content. Also, new columns can be added based on the transformed versions of the existing ones, for instance:

```
d.loc[:, "uv_squared"] = (d.loc[:, "u"] * d.loc[:, "v"])**2
d
##       u     v        w  uv_squared
## a  0.70  0.32     spam    0.050176
## b  0.29 -0.05    bacon    0.000210
## c  0.23 -0.20     spam    0.002116
## d  0.55  1.98     eggs    1.185921
## e  0.72 -1.62  sausage    1.360489
```

**Example 10.7** *(\*) Binary operations on objects with different* index *slots are in fact vectorised labelwise:*

```
x = pd.Series([1, 10, 1000, 10000, 100000], index=["a", "b", "a", "a", "c"])
```

```
x
## a         1
## b        10
## a      1000
## a     10000
## c    100000
## dtype: int64
y = pd.Series([1, 2, 3, 4, 5], index=["b", "b", "a", "d", "c"])
y
## b    1
## b    2
## a    3
## d    4
## c    5
## dtype: int64
```

*And now:*

```
x * y
## a         3.0
## a      3000.0
## a     30000.0
## b        10.0
## b        20.0
## c    500000.0
## d         NaN
## dtype: float64
```

Here, each element in the first `Series` named *a* was multiplied by each (there was only one) element labelled *a* in the second `Series`. For *d*, there were no matches, hence the result's being marked as missing; compare *Chapter 15*. Thus, this behaves like a full outer join-type operation; see *Section 10.6.4*.

The above is different from elementwise *vectorisation in* **numpy**:

```
np.array(x) * np.array(y)
## array([     1,     20,   3000,  40000, 500000])
```

Labelwise vectorisation can be useful in certain contexts, but we should be aware of this (yet another) incompatibility between the two packages.

## 10.4   Indexing `Series` Objects

Recall that each `DataFrame` and `Series` object is equipped with a slot called `index`, which is an object of class `Index` (or subclass thereof), giving the row and element labels, respectively. It turns out that we may apply the *index* operator, `[...]`, to subset these objects not only through the *indexers* known from the **numpy** part (e.g., numerical ones, i.e., by position) but also ones that pinpoint the items via their labels. That is quite a lot of index-ing.

Let us study different forms thereof in very detail. For illustration, we will be playing with the two following objects of class `Series`:

```python
np.random.seed(123)
b = pd.Series(np.round(np.random.rand(10), 2))
b.index = np.random.permutation(np.arange(10))
b
## 2    0.70
## 1    0.29
## 8    0.23
## 7    0.55
## 9    0.72
## 4    0.42
## 5    0.98
## 6    0.68
## 3    0.48
## 0    0.39
## dtype: float64
```

and:

```python
c = b.copy()
c.index = list("abcdefghij")
c
## a    0.70
## b    0.29
## c    0.23
## d    0.55
## e    0.72
## f    0.42
## g    0.98
## h    0.68
## i    0.48
## j    0.39
## dtype: float64
```

They consist of the same values, in the same order, but have different labels (`index` slots). In particular, `b`'s labels are integers that *do not* match the physical element positions (where 0 would denote the first element, etc.).

---

**Important**   For `numpy` vectors, we had four different indexing schemes: via a scalar (extracts an element at a given position), a slice, an integer vector, and a logical vector. `Series` objects are *additionally* labelled. Therefore, they can also be accessed through the contents of the `index` slot.

---

### 10.4.1   Do Not Use [ . . . ] Directly

Applying the index operator, [ . . . ], directly on `Series` is generally not a good idea:

```
b[0]
## 0.39
b[ [0] ]
## 0    0.39
## dtype: float64
```

both do not select the first item, but the item labelled 0.

However:

```
b[0:1]  # slice - 0 only
## 2    0.7
## dtype: float64
```

and

```
c[0]  # there is no label `0`
## 0.7
```

both fall back to position-based indexing.

Confusing? Well, with some self-discipline, the solution is easy:

---

**Important**   We should never apply [ . . . ] directly on `Series` nor `DataFrame` objects.

To avoid ambiguity, we should be referring to the **loc**[ . . . ] and **iloc**[ . . . ] accessors for the label- and position-based filtering, respectively.

---

### 10.4.2   `loc[...]`

`Series.loc`[ . . . ] implements label-based indexing.

```
b.loc[0]
## 0.39
```

This returned the element labelled `0`. On the other hand, `c.loc[0]` will raise a `KeyError`, because `c` consists of string labels only. But in this case, we can write:

```
c.loc["j"]
## 0.39
```

Next, we can use lists of labels to select a *subset*.

```
b.loc[ [0, 1, 0] ]
## 0    0.39
## 1    0.29
## 0    0.39
## dtype: float64
c.loc[ ["j", "b", "j"] ]
## j    0.39
## b    0.29
## j    0.39
## dtype: float64
```

The result is always of type `Series`.

Slicing behaves differently as the range is *inclusive* (sic![7]) at both sides:

```
b.loc[1:7]
## 1    0.29
## 8    0.23
## 7    0.55
## dtype: float64
b.loc[0:4:-1]
## 0    0.39
## 3    0.48
## 6    0.68
## 5    0.98
## 4    0.42
## dtype: float64
c.loc["d":"g"]
## d    0.55
## e    0.72
## f    0.42
## g    0.98
## dtype: float64
```

---

[7] Inconsistency; but makes sense when selecting column ranges.

return all elements between the two indicated labels.

Be careful that if there are repeated labels, then we will be returning *all* (sic![8]) the matching items:

```
d = pd.Series([1, 2, 3, 4], index=["a", "b", "a", "c"])
d.loc["a"]
## a    1
## a    3
## dtype: int64
```

The result is not a scalar but a `Series` object.

### 10.4.3  `iloc[...]`

Here are some examples of position-based indexing with the **`iloc`**`[...]` accessor. It is worth stressing that, fortunately, its behaviour is consistent with its **`numpy`** counterpart, i.e., the ordinary square brackets applied on objects of class `ndarray`.

For example:

```
b.iloc[0]  # the same: c.iloc[0]
## 0.7
```

returns the first element.

```
b.iloc[1:7]  # the same: b.iloc[1:7]
## 1    0.29
## 8    0.23
## 7    0.55
## 9    0.72
## 4    0.42
## 5    0.98
## dtype: float64
```

returns the 2nd, 3rd, ..., 7th element (not including `b.iloc[7]`, i.e., the 8th one).

### 10.4.4  Logical Indexing

Indexing using a logical vector-like object is also available. For this purpose, we will usually be using **`loc`**`[...]` with either a logical `Series` object of identical `index` slot as the subsetted object, or a Boolean **`numpy`** vector.

```
b.loc[(b > 0.4) & (b < 0.6)]
## 7    0.55
## 4    0.42
```

---

[8] Inconsistency; but makes sense for hierarchical indexes with repeated.

```
## 3    0.48
## dtype: float64
```

For **iloc**[...], the indexer must be unlabelled, e.g., be an ordinary **numpy** vector.
`

## 10.5   Indexing Data Frames

### 10.5.1   loc[...] and iloc[...]

For data frames, `iloc` and `loc` can be used too, but they now require two arguments,
serving as row and column selectors.

For example:

```
np.random.seed(123)
d = pd.DataFrame(dict(
    u = np.round(np.random.rand(5), 2),
    v = np.round(np.random.randn(5), 2),
    w = ["spam", "bacon", "spam", "eggs", "sausage"],
    x = [True, False, True, False, True]
), index=["a", "b", "c", "d", "e"])
```

And now:

```
d.loc[ d.loc[:, "u"] > 0.5, "u":"w" ]
##        u     v        w
## a   0.70  0.32     spam
## d   0.55  1.98     eggs
## e   0.72 -1.62  sausage
```

selects the rows where the values in the u column are greater than 0.5 and then returns
all columns between u and w (inclusive!).

Furthermore,

```
d.iloc[:3, :].loc[:, ["u", "w"]]
##        u     w
## a   0.70  spam
## b   0.29  bacon
## c   0.23  spam
```

fetches the first 3 rows (by position – `iloc` is necessary) and then selects two indicated
columns.

**Important**  We can write `d.u` as a shorter version of `d.loc[:, "u"]`. This improves the readability in contexts such as:

```
d.loc[(d.u >= 0.5) & (d.u <= 0.7), ["u", "w"]]
##        u     w
## a  0.70  spam
## d  0.55  eggs
```

This accessor is, sadly, not universal. We can verify this by considering a data frame featuring a column named, e.g., `mean` (which clashes with a built-in method).

**Exercise 10.8**  *Use **pandas.DataFrame.drop** to select all columns except v in d.*

**Exercise 10.9**  *Use **pandas.Series.isin** (amongst others) to select all rows with spam and bacon on the d's menu.*

**Exercise 10.10**  *In the `tips`[9] dataset, select data on male customers where the total bills were in the* $[10, 20]$ *interval. Also, select Saturday and Sunday records where the tips were greater than $5.*

## 10.5.2  Adding Rows and Columns

`loc[...]` can also be used to add new columns to an existing data frame:

```
d.loc[:, "y"] = d.loc[:, "u"]**2  # or d.loc[:, "y"] = d.u**2
d
##        u     v        w      x       y
## a  0.70  0.32     spam   True  0.4900
## b  0.29 -0.05    bacon  False  0.0841
## c  0.23 -0.20     spam   True  0.0529
## d  0.55  1.98     eggs  False  0.3025
## e  0.72 -1.62  sausage   True  0.5184
```

**Important**  Notation like "`d.new_column = ...`" does not work. As we said, `loc` and `iloc` are universal. For other accessors, this is not necessarily the case.

**Exercise 10.11**  *Use **pandas.DataFrame.insert** to add a new column not necessarily at the end of d.*

**Exercise 10.12**  *Use **pandas.DataFrame.append** to add a few more rows to d.*

---

[9] https://github.com/gagolews/teaching-data/raw/master/other/tips.csv

### 10.5.3    Pseudorandom Sampling and Splitting

As a simple application of what we have covered so far, let us consider the ways to sample several rows from an existing data frame.

We can use the **pandas.DataFrame.sample** method in the most basic scenarios, such as:

- select five rows, without replacement,

- select 20% rows, with replacement,

- rearrange all the rows.

For example:

```
body = pd.read_csv("https://raw.githubusercontent.com/gagolews/" +
    "teaching-data/master/marek/nhanes_adult_female_bmx_2020.csv",
    comment="#")
body.sample(5, random_state=123)  # 5 rows without replacement
##         BMXWT  BMXHT  BMXARML  BMXLEG  BMXARMC  BMXHIP  BMXWAIST
## 4214    58.4  156.2     35.2    34.7     27.2    99.5      77.5
## 3361    73.7  161.0     36.5    34.5     29.0   107.6      98.2
## 3759    61.4  164.6     37.5    40.4     26.9    93.5      84.4
## 3733   120.4  158.8     33.5    34.6     40.5   147.2     129.3
## 1121   123.5  157.5     35.5    29.0     50.5   143.0     136.4
```

Notice the `random_state` argument which controls the seed of the pseudorandom number generator so that we get reproducible results. Alternatively, we could call **numpy.random.seed**.

**Exercise 10.13** *Show how the three aforementioned scenarios can be implemented manually using* **iloc[...]** *and* **numpy.random.permutation** *or* **numpy.random.choice***.*

In machine learning practice, we are used to training and evaluating machine learning models on different (mutually disjoint) subsets of the whole data frame.

For instance, in Section 12.3.3, we mention that we may be interested in performing the so-called *training/test split* (partitioning), where 80% (or 60% or 70%) of the randomly selected rows would constitute the first new data frame and the remaining 20% (or 40% or 30%, respectively) would go to the second one.

Given a data frame like:

```
x = body.head(10)  # this is just an example
x
##     BMXWT  BMXHT  BMXARML  BMXLEG  BMXARMC  BMXHIP  BMXWAIST
## 0    97.1  160.2     34.7    40.8     35.8   126.1     117.9
## 1    91.1  152.7     33.5    33.0     38.5   125.5     103.1
## 2    73.0  161.2     37.4    38.0     31.8   106.2      92.0
## 3    61.7  157.4     38.0    34.7     29.0   101.0      90.5
## 4    55.4  154.6     34.6    34.0     28.3    92.5      73.2
```

```
## 5   62.0   144.7     32.5     34.2     29.8    106.7      84.8
## 6   66.2   166.5     37.5     37.6     32.0     96.3      95.7
## 7   75.9   154.5     35.4     37.6     32.7    107.7      98.7
## 8   77.2   159.2     38.5     40.5     35.7    102.0      97.5
## 9   91.6   174.5     36.1     45.9     35.2    121.3     100.3
```

one way to perform the aforementioned split is to generate a random permutation of the set of row indexes:

```
np.random.seed(123)  # reproducibility matters
idx = np.random.permutation(x.shape[0])
idx
## array([4, 0, 7, 5, 8, 3, 1, 6, 9, 2])
```

And then to pick the first 80% of them to construct the data frame number one:

```
k = int(x.shape[0]*0.8)
x.iloc[idx[:k], :]
##     BMXWT   BMXHT   BMXARML   BMXLEG   BMXARMC   BMXHIP   BMXWAIST
## 4   55.4   154.6      34.6     34.0      28.3     92.5       73.2
## 0   97.1   160.2      34.7     40.8      35.8    126.1      117.9
## 7   75.9   154.5      35.4     37.6      32.7    107.7       98.7
## 5   62.0   144.7      32.5     34.2      29.8    106.7       84.8
## 8   77.2   159.2      38.5     40.5      35.7    102.0       97.5
## 3   61.7   157.4      38.0     34.7      29.0    101.0       90.5
## 1   91.1   152.7      33.5     33.0      38.5    125.5      103.1
## 6   66.2   166.5      37.5     37.6      32.0     96.3       95.7
```

and the remaining ones to generate the second dataset:

```
x.iloc[idx[k:], :]
##     BMXWT   BMXHT   BMXARML   BMXLEG   BMXARMC   BMXHIP   BMXWAIST
## 9   91.6   174.5      36.1     45.9      35.2    121.3      100.3
## 2   73.0   161.2      37.4     38.0      31.8    106.2       92.0
```

**Exercise 10.14** *In the* `wine_quality_all`[10] *dataset, leave out all but the white wines. Partition the resulting data frame randomly into three data frames:* `wines_train` *(60% of the rows),* `wines_validate` *(another 20% of the rows), and* `wines_test` *(the remaining 20%).*

**Exercise 10.15** *Write a function* ***kfold*** *which takes a data frame* ***x*** *and an integer* $k > 1$ *as arguments. Return a list of data frames resulting in stemming from randomly partitioning* ***x*** *into* $k$ *disjoint chunks of equal (or almost equal if that is not possible) sizes.*

---

[10] https://github.com/gagolews/teaching-data/raw/master/other/wine_quality_all.csv

## 10.5.4 Hierarchical Indexes (*)

Consider the following `DataFrame` object with a hierarchical index:

```python
np.random.seed(123)
d = pd.DataFrame(dict(
    year = np.repeat([2023, 2024, 2025], 4),
    quarter = np.tile(["Q1", "Q2", "Q3", "Q4"], 3),
    data = np.round(np.random.rand(12), 2)
)).set_index(["year", "quarter"])
d
##              data
## year quarter
## 2023 Q1      0.70
##      Q2      0.29
##      Q3      0.23
##      Q4      0.55
## 2024 Q1      0.72
##      Q2      0.42
##      Q3      0.98
##      Q4      0.68
## 2025 Q1      0.48
##      Q2      0.39
##      Q3      0.34
##      Q4      0.73
```

The index has both levels named, but this is purely for aesthetic reasons.

Indexing using **loc**[. . . ] by default relates to the first level of the hierarchy:

```python
d.loc[2023, :]
##         data
## quarter
## Q1      0.70
## Q2      0.29
## Q3      0.23
## Q4      0.55
```

Note that we selected *all* rows corresponding to a given label and dropped (!) this level of the hierarchy.

Another example:

```python
d.loc[ [2023, 2025], : ]
##              data
## year quarter
## 2023 Q1      0.70
##      Q2      0.29
```

```
##      Q3      0.23
##      Q4      0.55
## 2025 Q1      0.48
##      Q2      0.39
##      Q3      0.34
##      Q4      0.73
```

To access deeper levels, we can use tuples as indexers:

```
d.loc[ (2023, "Q1"), : ]
## data    0.7
## Name: (2023, Q1), dtype: float64
d.loc[ [(2023, "Q1"), (2024, "Q3")], : ]
##              data
## year quarter
## 2023 Q1      0.70
## 2024 Q3      0.98
```

In certain scenarios, though, it will probably be much easier to subset a hierarchical index by using **reset_index** and **set_index** creatively (together with **loc**[...] and **pandas.Series.isin**, etc.).

Let us stress again that the `:` operator can only be used *directly* within the square brackets. Nonetheless, we can always use the **slice** constructor to create a slice in any context:

```
d.loc[ (slice(None), ["Q1", "Q3"]), : ]  # :, ["Q1", "Q3"]
##              data
## year quarter
## 2023 Q1      0.70
##      Q3      0.23
## 2024 Q1      0.72
##      Q3      0.98
## 2025 Q1      0.48
##      Q3      0.34
d.loc[ (slice(None, None, -1), slice("Q2", "Q3")), : ]  # ::-1, "Q2":"Q3"
##              data
## year quarter
## 2025 Q3      0.34
##      Q2      0.39
## 2024 Q3      0.98
##      Q2      0.42
## 2023 Q3      0.23
##      Q2      0.29
```

## 10.6    Further Operations on Data Frames

One of the many roles of data frames is to represent tables of values for their nice presentation, e.g., in reports from data analysis or research papers. Here are some functions that can aid in their formatting.

### 10.6.1    Sorting

Let us consider another example dataset. Here are the yearly (for 2018) average air quality data[11] in the Australian state of Victoria.

```
air = pd.read_csv("https://raw.githubusercontent.com/gagolews/" +
    "teaching-data/master/marek/air_quality_2018_means.csv",
    comment="#")
air = (
    air.
    loc[air.param_id.isin(["BPM2.5", "NO2"]), :].
    reset_index(drop=True)
)
```

We chose two air quality parameters using **pandas.Series.isin**, which determines whether each element in a Series is enlisted in a given sequence. We could also have used **set_index** and **loc[...]** for that.

Notice that the above code spans many lines. We needed to enclose it in round brackets to avoid a syntax error. Alternatively, we could have used backslashes at the end of each line.

Anyway, here is the data frame:

```
air
##           sp_name param_id      value
## 0       Alphington   BPM2.5   7.848758
## 1       Alphington      NO2   9.558120
## 2      Altona North     NO2   9.467912
## 3        Churchill   BPM2.5   6.391230
## 4        Dandenong      NO2   9.800705
## 5        Footscray   BPM2.5   7.640948
## 6        Footscray      NO2  10.274531
## 7     Geelong South   BPM2.5   6.502762
## 8     Geelong South     NO2   5.681722
## 9      Melbourne CBD   BPM2.5   8.072998
## 10             Moe   BPM2.5   6.427079
```

*(continues on next page)*

---

[11] https://discover.data.vic.gov.au/dataset/epa-air-watch-all-sites-air-quality-hourly-averages-yearly

```
## 11   Morwell East   BPM2.5   6.784596
## 12  Morwell South   BPM2.5   6.512849
## 13  Morwell South      NO2   5.124430
## 14      Traralgon   BPM2.5   8.024735
## 15      Traralgon      NO2   5.776333
```

**sort_values** is a convenient means to order the rows with respect to one criterion, be it numeric or categorical.

```
air.sort_values("value", ascending=False)
##               sp_name param_id       value
## 6            Footscray      NO2   10.274531
## 4            Dandenong      NO2    9.800705
## 1           Alphington      NO2    9.558120
## 2         Altona North      NO2    9.467912
## 9        Melbourne CBD   BPM2.5    8.072998
## 14           Traralgon   BPM2.5    8.024735
## 0           Alphington   BPM2.5    7.848758
## 5            Footscray   BPM2.5    7.640948
## 11         Morwell East  BPM2.5    6.784596
## 12        Morwell South  BPM2.5    6.512849
## 7         Geelong South  BPM2.5    6.502762
## 10                 Moe   BPM2.5    6.427079
## 3            Churchill   BPM2.5    6.391230
## 15           Traralgon      NO2    5.776333
## 8         Geelong South     NO2    5.681722
## 13        Morwell South     NO2    5.124430
```

It is also possible to take into account more sorting criteria:

```
air.sort_values(["param_id", "value"], ascending=[True, False])
##               sp_name param_id       value
## 9        Melbourne CBD   BPM2.5    8.072998
## 14           Traralgon   BPM2.5    8.024735
## 0           Alphington   BPM2.5    7.848758
## 5            Footscray   BPM2.5    7.640948
## 11         Morwell East  BPM2.5    6.784596
## 12        Morwell South  BPM2.5    6.512849
## 7         Geelong South  BPM2.5    6.502762
## 10                 Moe   BPM2.5    6.427079
## 3            Churchill   BPM2.5    6.391230
## 6            Footscray      NO2   10.274531
## 4            Dandenong      NO2    9.800705
## 1           Alphington      NO2    9.558120
## 2         Altona North      NO2    9.467912
```

```
## 15       Traralgon      NO2    5.776333
## 8    Geelong South      NO2    5.681722
## 13  Morwell South      NO2    5.124430
```

Here, in each group of identical parameters, we get a decreasing order with respect to the value.

**Exercise 10.16** *Compare the ordering with respect to* `param_id` *and* `value` *vs* `value` *and then* `param_id`.

---

**Note** (*) `DataFrame.sort_values` by default lamentably (!) uses a non-stable algorithm (a modified quicksort). If a data frame is sorted with respect to one criterion, and then we reorder it with respect to another one, tied observations are not guaranteed to be listed in the original order:

```python
(pd.read_csv("https://raw.githubusercontent.com/gagolews/" +
    "teaching-data/master/marek/air_quality_2018_means.csv",
    comment="#")
    .sort_values("sp_name")
    .sort_values("param_id")
    .set_index("param_id")
    .loc[["BPM2.5", "NO2"], :]
    .reset_index())
##    param_id        sp_name        value
## 0    BPM2.5   Melbourne CBD    8.072998
## 1    BPM2.5             Moe    6.427079
## 2    BPM2.5       Footscray    7.640948
## 3    BPM2.5    Morwell East    6.784596
## 4    BPM2.5       Churchill    6.391230
## 5    BPM2.5   Morwell South    6.512849
## 6    BPM2.5       Traralgon    8.024735
## 7    BPM2.5      Alphington    7.848758
## 8    BPM2.5   Geelong South    6.502762
## 9       NO2   Morwell South    5.124430
## 10      NO2       Traralgon    5.776333
## 11      NO2   Geelong South    5.681722
## 12      NO2     Altona North    9.467912
## 13      NO2      Alphington    9.558120
## 14      NO2       Dandenong    9.800705
## 15      NO2       Footscray   10.274531
```

We lost the ordering based on station names in the two subgroups. To switch to a mergesort-like method (timsort), we should pass `kind="stable"`.

```
(pd.read_csv("https://raw.githubusercontent.com/gagolews/" +
    "teaching-data/master/marek/air_quality_2018_means.csv",
    comment="#")
    .sort_values("sp_name")
    .sort_values("param_id", kind="stable")  # !
    .set_index("param_id")
    .loc[["BPM2.5", "NO2"], :]
    .reset_index())
##    param_id        sp_name      value
## 0    BPM2.5      Alphington   7.848758
## 1    BPM2.5       Churchill   6.391230
## 2    BPM2.5       Footscray   7.640948
## 3    BPM2.5   Geelong South   6.502762
## 4    BPM2.5   Melbourne CBD   8.072998
## 5    BPM2.5             Moe   6.427079
## 6    BPM2.5    Morwell East   6.784596
## 7    BPM2.5   Morwell South   6.512849
## 8    BPM2.5       Traralgon   8.024735
## 9       NO2      Alphington   9.558120
## 10      NO2    Altona North   9.467912
## 11      NO2       Dandenong   9.800705
## 12      NO2       Footscray  10.274531
## 13      NO2   Geelong South   5.681722
## 14      NO2   Morwell South   5.124430
## 15      NO2       Traralgon   5.776333
```

**Exercise 10.17** (*) *Perform identical reorderings but using only* **loc[...]**, **iloc[...]**, *and* **numpy.argsort**.

## 10.6.2 Stacking and Unstacking (Long and Wide Forms)

Let us discuss some further ways to transform data frames, that benefit from, make sense because of, or are possible due to their being able to store data of different types.

The above `air` dataset is in the so-called *long* format, where all measurements are *stacked* one after/below another. Such a form is quite convenient for the storing of data, especially when there are few recorded values, but many possible combinations of levels (sparse data).

The long format might not necessarily be optimal in all data processing activities, though; compare [85]. In the matrix processing part of this book, it was much more natural for us to have a single *observation* (e.g., data for each measurement station) in each row.

We can *unstack* the `air` data frame (convert to the *wide* format) quite easily:

```
air_wide = air.set_index(["sp_name", "param_id"]).unstack().loc[:, "value"]
air_wide
## param_id       BPM2.5          NO2
## sp_name
## Alphington    7.848758     9.558120
## Altona North       NaN     9.467912
## Churchill     6.391230          NaN
## Dandenong          NaN     9.800705
## Footscray     7.640948    10.274531
## Geelong South 6.502762     5.681722
## Melbourne CBD 8.072998          NaN
## Moe           6.427079          NaN
## Morwell East  6.784596          NaN
## Morwell South 6.512849     5.124430
## Traralgon     8.024735     5.776333
```

The missing values are denoted with `NaN`s (not-a-number); see Section 15.1 for more details. Interestingly, we got a hierarchical index in the columns (sic!) slot, hence the **loc**[...] part to drop the last level of the hierarchy. Also notice that the `index` and `columns` slots are named.

The other way around, we can use the **stack** method:

```
air_wide.T.rename_axis(index="location", columns="param").\
    stack().rename("value").reset_index()
##     location        param        value
## 0     BPM2.5     Alphington     7.848758
## 1     BPM2.5       Churchill     6.391230
## 2     BPM2.5       Footscray     7.640948
## 3     BPM2.5   Geelong South     6.502762
## 4     BPM2.5   Melbourne CBD     8.072998
## 5     BPM2.5             Moe     6.427079
## 6     BPM2.5   Morwell East     6.784596
## 7     BPM2.5   Morwell South     6.512849
## 8     BPM2.5       Traralgon     8.024735
## 9        NO2     Alphington     9.558120
## 10       NO2   Altona North     9.467912
## 11       NO2       Dandenong     9.800705
## 12       NO2       Footscray    10.274531
## 13       NO2   Geelong South     5.681722
## 14       NO2   Morwell South     5.124430
## 15       NO2       Traralgon     5.776333
```

We used the data frame transpose (**T**) to get a location-major order (less boring an outcome in this context). Missing values are gone now. We do not need them anymore. Nevertheless, passing `dropna=False` would help us identify the combinations of `location` and `param` for which the readings are not provided.

### 10.6.3 Set-Theoretic Operations and Removing Duplicates

Here are two not at all disjoint sets of imaginary persons:

```python
A = pd.read_csv("https://raw.githubusercontent.com/gagolews/" +
    "teaching-data/master/marek/some_birth_dates1.csv",
    comment="#")
A
##                   Name    BirthDate
## 0    Paitoon Ornwimol   26.06.1958
## 1        Antónia Lata   20.05.1935
## 2    Bertoldo Mallozzi   17.08.1972
## 3        Nedeljko Bukv   19.12.1921
## 4        Micha Kitchen   17.09.1930
## 5      Mefodiy Shachar   01.10.1914
## 6          Paul Meckler   29.09.1968
## 7       Katarzyna Lasko   20.10.1971
## 8          Åge Trelstad   07.03.1935
## 9  Duchanee Panomyaong   19.06.1952
```

and:

```python
B = pd.read_csv("https://raw.githubusercontent.com/gagolews/" +
    "teaching-data/master/marek/some_birth_dates2.csv",
    comment="#")
B
##                    Name    BirthDate
## 0   Hushang Naigamwala   25.08.1991
## 1             Zhen Wei   16.11.1975
## 2        Micha Kitchen   17.09.1930
## 3          Jodoc Alwin   16.11.1969
## 4           Igor Mazał   14.05.2004
## 5      Katarzyna Lasko   20.10.1971
## 6  Duchanee Panomyaong   19.06.1952
## 7      Mefodiy Shachar   01.10.1914
## 8         Paul Meckler   29.09.1968
## 9        Noe Tae-Woong   11.07.1970
## 10        Åge Trelstad   07.03.1935
```

In both datasets, there is a single categorical column whose elements uniquely identify each record (i.e., `Name`). In the language of relational databases, we would call it the *primary key*. In such a case, implementing the set-theoretic operations is relatively easy, as we can refer to the **pandas.Series.isin** method.

First, $A \cap B$ (intersection), includes only the rows that are *both* in $A$ and in $B$:

```python
A.loc[A.Name.isin(B.Name), :]
```

```
##                     Name    BirthDate
## 4          Micha Kitchen   17.09.1930
## 5        Mefodiy Shachar   01.10.1914
## 6           Paul Meckler   29.09.1968
## 7         Katarzyna Lasko   20.10.1971
## 8            Åge Trelstad   07.03.1935
## 9    Duchanee Panomyaong   19.06.1952
```

Second, $A \setminus B$ (difference), gives all the rows that are in $A$ *but not* in $B$:

```
A.loc[~A.Name.isin(B.Name), :]
##                     Name    BirthDate
## 0      Paitoon Ornwimol   26.06.1958
## 1          Antónia Lata   20.05.1935
## 2      Bertoldo Mallozzi   17.08.1972
## 3          Nedeljko Bukv   19.12.1921
```

Third, $A \cup B$ (union), returns the rows that exist in $A$ *or* are in $B$:

```
pd.concat((A, B.loc[~B.Name.isin(A.Name), :]))
##                     Name    BirthDate
## 0      Paitoon Ornwimol   26.06.1958
## 1          Antónia Lata   20.05.1935
## 2      Bertoldo Mallozzi   17.08.1972
## 3          Nedeljko Bukv   19.12.1921
## 4          Micha Kitchen   17.09.1930
## 5        Mefodiy Shachar   01.10.1914
## 6           Paul Meckler   29.09.1968
## 7         Katarzyna Lasko   20.10.1971
## 8            Åge Trelstad   07.03.1935
## 9    Duchanee Panomyaong   19.06.1952
## 0    Hushang Naigamwala   25.08.1991
## 1               Zhen Wei   16.11.1975
## 3            Jodoc Alwin   16.11.1969
## 4             Igor Mazał   14.05.2004
## 9          Noe Tae-Woong   11.07.1970
```

There are no duplicate rows in any of the above outputs.

**Exercise 10.18** *Determine* $(A \cup B) \setminus (A \cap B) = (A \setminus B) \cup (B \setminus A)$ *(symmetric difference).*

**Exercise 10.19** *(\*) Determine the union, intersection, and difference of the* `wine_sample1`[12] *and* `wine_sample2`[13] *datasets, where there is no column uniquely identifying the observations. Hint:*

---

[12] https://github.com/gagolews/teaching-data/raw/master/other/wine_sample1.csv
[13] https://github.com/gagolews/teaching-data/raw/master/other/wine_sample2.csv

*consider using* **pandas.concat** *and* **pandas.DataFrame.duplicated** *or* **pandas.DataFrame. drop_duplicates**.

### 10.6.4  Joining (Merging)

In database design, it is common to normalise the datasets. We do this to avoid the duplication of information and pathologies stemming from them (e.g., [17]).

**Example 10.20** *The above air quality parameters are separately described in another data frame:*

```
param = pd.read_csv("https://raw.githubusercontent.com/gagolews/" +
    "teaching-data/master/marek/air_quality_2018_param.csv",
    comment="#")
param.rename(dict(param_std_unit_of_measure="unit"), axis=1)
##    param_id              param_name   unit       param_short_name
## 0       API   Airborne particle index   none   Visibility Reduction
## 1   BPM2.5 BAM  Particles < 2.5 micron   ug/m3                 PM2.5
## 2        CO          Carbon Monoxide    ppm                    CO
## 3     HPM10            Hivol PM10    ug/m3                   NaN
## 4      NO2         Nitrogen Dioxide     ppb                   NO2
## 5       O3                  Ozone     ppb                    O3
## 6     PM10   TEOM Particles <10micron   ug/m3                  PM10
## 7    PPM2.5         Partisol PM2.5    ug/m3                   NaN
## 8      SO2           Sulfur Dioxide     ppb                   SO2
```

*We could have stored them alongside the* `air` *data frame, but that would be a waste of space. Also, if we wanted to modify some datum (note, e.g., the annoying double space in* `param_name` *for* `BPM2.5`*), we would have to update all the relevant records.*

*Instead, we can always match the records in both data frames that have the same* `param_ids`*, and join (merge) these datasets only when we really need this.*

Let us discuss the possible join operations by studying the two following toy data sets:

```
A = pd.DataFrame({
    "x": ["a0", "a1", "a2", "a3"],
    "y": ["b0", "b1", "b2", "b3"]
})
A
##     x    y
## 0  a0   b0
## 1  a1   b1
## 2  a2   b2
## 3  a3   b3
```

and:

```
B = pd.DataFrame({
    "x": ["a0", "a2", "a2", "a4"],
    "z": ["c0", "c1", "c2", "c3"]
})
B
##     x   z
## 0  a0  c0
## 1  a2  c1
## 2  a2  c2
## 3  a4  c3
```

They both have one column somewhat *in common*, x.

The *inner (natural) join* returns the records that have a match in both datasets:

```
pd.merge(A, B, on="x")
##     x   y   z
## 0  a0  b0  c0
## 1  a2  b2  c1
## 2  a2  b2  c2
```

The *left join* of *A* with *B* guarantees to return all the records from *A*, even those which are not matched by anything in *B*.

```
pd.merge(A, B, how="left", on="x")
##     x   y    z
## 0  a0  b0   c0
## 1  a1  b1  NaN
## 2  a2  b2   c1
## 3  a2  b2   c2
## 4  a3  b3  NaN
```

The *right join* of *A* with *B* is the same as the left join of *B* with *A*:

```
pd.merge(A, B, how="right", on="x")
##     x    y   z
## 0  a0   b0  c0
## 1  a2   b2  c1
## 2  a2   b2  c2
## 3  a4  NaN  c3
```

Finally, the *full outer join* is the set-theoretic union of the left and the right join:

```
pd.merge(A, B, how="outer", on="x")
##     x    y    z
## 0  a0   b0   c0
```

```
## 1   a1   b1   NaN
## 2   a2   b2   c1
## 3   a2   b2   c2
## 4   a3   b3   NaN
## 5   a4   NaN  c3
```

**Exercise 10.21** *Join* `air_quality_2018_value`[14] *with* `air_quality_2018_point`[15], *and then with* `air_quality_2018_param`[16].

**Exercise 10.22** *Normalise* `air_quality_2018`[17] *so that you get the three separate data frames mentioned in the previous exercise (*`value`*,* `point`*, and* `param`*).*

**Exercise 10.23** *(\*) In the National Health and Nutrition Examination Survey, each participant is uniquely identified by their sequence number (*`SEQN`*). This token is mentioned in numerous datasets, including:*

- *demographic variables*[18],
- *body measures*[19],
- *audiometry*[20],
- *and many more*[21].

*Join a few chosen datasets that you find interesting.*

### 10.6.5   ...And (Too) Many More

Looking at the list of methods for the `DataFrame` and `Series` classes in the **pandas** package's documentation[22], we can see that they are abundant. Together with the object-oriented syntax, we will often find ourselves appreciating the high readability of even quite complex operation chains such as `data.drop_duplicates().groupby(["year", "month"]).mean().reset_index()`; see Chapter 12.

Nevertheless, the methods are probably too plentiful to our taste. Their developers were overgenerous. They wanted to include a list of all the possible *verbs* related to data analysis, even if they can be trivially expressed by a composition of 2-3 simpler operations from **numpy** or **scipy** instead.

As strong advocates of minimalism, we would rather save ourselves from being overloaded with too much new information. This is why our focus in this book is on devel-

---

oping the most *transferable*[23] skills. Our approach is also slightly more hygienic. We do not want the reader to develop a hopeless mindset, the habit of looking everything up on the internet when faced with even the simplest kinds of problems. We have brains for a reason.

## 10.7 Exercises

**Exercise 10.24** *How are data frames different from matrices?*

**Exercise 10.25** *What are the use cases of the `name` slot in `Series` and `Index` objects?*

**Exercise 10.26** *What is the purpose of `set_index` and `reset_index`?*

**Exercise 10.27** *Why learning `numpy` is crucial when someone wants to become a proficient user of `pandas`?*

**Exercise 10.28** *What is the difference between `iloc[...]` and `loc[...]`?*

**Exercise 10.29** *Why applying the index operator `[...]` directly on a `Series` or `DataFrame` object is not necessarily a good idea?*

**Exercise 10.30** *What is the difference between `index`, `Index`, and `columns`?*

**Exercise 10.31** *How can we compute the arithmetic mean and median of all the numeric columns in a data frame, using a single line of code?*

**Exercise 10.32** *What is a training/test split and how to perform it using `numpy` and `pandas`?*

**Exercise 10.33** *What is the difference between stacking and unstacking? Which one yields a wide (as opposed to long) format?*

**Exercise 10.34** *Name different data frame join (merge) operations and explain how they work.*

**Exercise 10.35** *How does sorting with respect to more than one criterion work?*

**Exercise 10.36** *Name the basic set-theoretic operations on data frames.*

---

[23] This is also in line with the observation that Python with `pandas` is not the only environment where we can work with data frames (e.g., base R or Julia with `DataFrame.jl` allows that too).

# 11

## Handling Categorical Data

So far, we have been mostly dealing with *quantitative* (numeric) data, on which we were able to apply various mathematical operations, such as computing the arithmetic mean or taking the square thereof. Naturally, not every transformation must always make sense in every context (e.g., multiplying temperatures – what does it mean when we say that it is twice as hot today as compared to yesterday?), but still, the possibilities were plenty.

*Qualitative* data (also known as categorical data, factors, or enumerated types) such as eye colour, blood type, or a flag whether a patient is ill, on the other hand, take a small number of unique values. They support an extremely limited set of admissible operations. Namely, we can only determine whether two entities are equal or not.

In datasets involving many features, which we shall cover in Chapter 12, categorical variables are often used for observation *grouping* (e.g., so that we can compute the best and average time for marathoners in each age category or draw box plots for finish times of men and women separately). Also, they may serve as target variables in statistical classification tasks (e.g., so that we can determine if an email is "spam" or "not spam").

## 11.1 Representing and Generating Categorical Data

Common ways to represent a categorical variable with $l$ distinct levels $\{L_1, L_2, \dots, L_l\}$ is by storing it as:

- a vector of strings,

- a vector of integers between 0 (inclusive) and $l$ (exclusive[1]).

These two are easily interchangeable.

For $l = 2$ (binary data), another convenient representation is by means of logical vectors. This can be extended to a so-called one-hot encoded representation using a logical vector of length $l$.

---

[1] This coincides with the possible indexes into an array of length $l$. In some other languages, e.g., R, we would use integers between 1 and $l$ (inclusive). Nevertheless, a dataset creator is free to encode the labels however they want. For example, `DMDBORN4` in NHANES has: 1 (born in 50 US states or Washington, DC), 2 (others), 77 (refused to answer), and 99 (do not know).

Let us consider the data on the original whereabouts of the top 16 marathoners (the 37th PZU native Marathon dataset):

```python
marathon = pd.read_csv("https://raw.githubusercontent.com/gagolews/" +
    "teaching-data/master/marek/37_pzu_warsaw_marathon_simplified.csv",
    comment="#")
cntrs = np.array(marathon.country, dtype="str")
cntrs16 = cntrs[:16]
cntrs16
## array(['KE', 'KE', 'KE', 'ET', 'KE', 'KE', 'ET', 'MA', 'PL', 'PL', 'IL',
##        'PL', 'KE', 'KE', 'PL', 'PL'], dtype='<U2')
```

These are two-letter ISO 3166 country codes, encoded of course as strings (notice the dtype="str" argument).

Calling **pandas.unique** allows us to determine the set of distinct categories:

```python
cat_cntrs16 = pd.unique(cntrs16)
cat_cntrs16
## array(['KE', 'ET', 'MA', 'PL', 'IL'], dtype='<U2')
```

Hence, cntrs16 is a categorical vector of length $n = 16$ (**len**(cntrs16)) with data assuming one of $l = 5$ different levels (**len**(cat_cntrs16)).

---

**Note**  We could have also used **numpy.unique** discussed in Section 5.5.3, but it would sort the distinct values lexicographically. In other words, they would *not* be listed in the order of appearance.

---

### 11.1.1  Encoding and Decoding Factors

To *encode* a label vector using a set of consecutive nonnegative integers, we can call **pandas.factorize**:

```python
codes_cntrs16, cat_cntrs16 = pd.factorize(cntrs16)  # sort=False
cat_cntrs16
## array(['KE', 'ET', 'MA', 'PL', 'IL'], dtype='<U2')
codes_cntrs16
## array([0, 0, 0, 1, 0, 0, 1, 2, 3, 3, 4, 3, 0, 0, 3, 3])
```

The code sequence 0, 0, 0, 1, ... corresponds to the 1st, 1st, 1st, 2nd, ... level in cat_cntrs16, i.e., Kenya, Kenya, Kenya, Ethiopia, ....

---

**Important**  When we represent categorical data using numeric codes, it is possible to introduce non-occurring levels. Such information can be useful, e.g., we could explicitly indicate that there were no runners from Australia in the top 16.

Even though we can *represent* categorical variables using a set of integers, it does not mean that they become instances of a quantitative type. Arithmetic operations thereon do not really make sense.

---

The values between 0 (inclusive) and 5 (exclusive) can be used to index a given array of length $l = 5$. As a consequence, to *decode* our factor, we can simply write:

```
cat_cntrs16[codes_cntrs16]
## array(['KE', 'KE', 'KE', 'ET', 'KE', 'KE', 'ET', 'MA', 'PL', 'PL', 'IL',
##        'PL', 'KE', 'KE', 'PL', 'PL'], dtype='<U2')
```

We can use any other set of labels now:

```
np.array(["Kenya", "Ethiopia", "Morocco", "Poland", "Israel"])[codes_cntrs16]
## array(['Kenya', 'Kenya', 'Kenya', 'Ethiopia', 'Kenya', 'Kenya',
##        'Ethiopia', 'Morocco', 'Poland', 'Poland', 'Israel', 'Poland',
##        'Kenya', 'Kenya', 'Poland', 'Poland'], dtype='<U8')
```

This is an instance of the *relabelling* of a categorical variable.

**Exercise 11.1** *(\*\*) Here is a way of* recoding *a variable, i.e., changing the order of labels and permuting the numeric codes:*

```
new_codes = np.array([3, 0, 2, 4, 1])  # an example permutation of labels
new_cat_cntrs16 = cat_cntrs16[new_codes]
new_cat_cntrs16
## array(['PL', 'KE', 'MA', 'IL', 'ET'], dtype='<U2')
```

*Then we make use of the fact that* **numpy.argsort** *applied on a vector representing a permutation, determines its very inverse:*

```
new_codes_cntrs16 = np.argsort(new_codes)[codes_cntrs16]
new_codes_cntrs16
## array([1, 1, 1, 4, 1, 1, 4, 2, 0, 0, 3, 0, 1, 1, 0, 0])
```

*Verification:*

```
np.all(cntrs16 == new_cat_cntrs16[new_codes_cntrs16])
## True
```

**Exercise 11.2** *(\*\*) Determine the set of unique values in* `cntrs16` *in the order of appearance (and not sorted lexicographically), but without using* **pandas.unique** *nor* **pandas.factorize**. *Then, encode* `cntrs16` *using this level set.*

Hint: check out the `return_index` argument to **numpy.unique** and **numpy.searchsorted**.

Furthermore, **pandas** includes[2] a special `dtype` for storing categorical data. Namely, we can write:

```
cntrs16_series = pd.Series(cntrs16, dtype="category")
```

or, equivalently:

```
cntrs16_series = pd.Series(cntrs16).astype("category")
```

These two yield a `Series` object displayed as if it was represented using string labels:

```
cntrs16_series.head()  # preview
## 0    KE
## 1    KE
## 2    KE
## 3    ET
## 4    KE
## dtype: category
## Categories (5, object): ['ET', 'IL', 'KE', 'MA', 'PL']
```

Instead, it is encoded using the aforementioned numeric representation:

```
np.array(cntrs16_series.cat.codes)
## array([2, 2, 2, 0, 2, 2, 0, 3, 4, 4, 1, 4, 2, 2, 4, 4], dtype=int8)
cntrs16_series.cat.categories
## Index(['ET', 'IL', 'KE', 'MA', 'PL'], dtype='object')
```

This time the labels are sorted lexicographically.

Most often we will be storing categorical data in data frames as ordinary strings, unless a relabelling on-the-fly is required:

```
(marathon.iloc[:16, :].country.astype("category")
    .cat.reorder_categories(
        ["KE",     "IL",     "MA",     "ET",     "PL"]
    )
    .cat.rename_categories(
        ["Kenya", "Israel", "Morocco", "Ethiopia", "Poland"]
    ).astype("str")
).head()
## 0      Kenya
## 1      Kenya
## 2      Kenya
## 3    Ethiopia
## 4      Kenya
## Name: country, dtype: object
```

---

[2] https://pandas.pydata.org/pandas-docs/stable/user_guide/categorical.html

### 11.1.2   Binary Data as Logical and Probability Vectors

*Binary data* is a special case of the qualitative setting, where we only have $l = 2$ categories. For example:

- 0, e.g., healthy/fail/off/non-spam/absent/...),

- 1, e.g., ill/success/on/spam/present/...).

Usually, the *interesting* or *noteworthy* category is denoted with 1.

---

**Important**   When converting logical to numeric, `False` becomes 0 and `True` becomes 1. Conversely, 0 is converted to `False` and anything else (including -0.326) to `True`.

---

Hence, instead of working with vectors of 0s and 1s, we might equivalently be playing with logical arrays. For example:

```
np.array([True, False, True, True, False]).astype(int)
## array([1, 0, 1, 1, 0])
```

The other way around:

```
np.array([-2, -0.326, -0.000001, 0.0, 0.1, 1, 7643]).astype(bool)
## array([ True,  True,  True, False,  True,  True,  True])
```

or, equivalently:

```
np.array([-2, -0.326, -0.000001, 0.0, 0.1, 1, 7643]) != 0
## array([ True,  True,  True, False,  True,  True,  True])
```

---

**Important**   It is not rare to work with vectors of probabilities, where the $i$-th element therein, say `p[i]`, denotes the likelihood of an observation's belonging to class 1. Consequently, the probability of being a member of class 0 is `1-p[i]`. In the case where we would rather work with *crisp* classes, we can simply apply the conversion (`p>=0.5`) to get a logical vector.

---

**Exercise 11.3**   *Given a numeric vector x, create a vector of the same length as x whose i-th element is equal to "yes" if x[i] is in the unit interval and to "no" otherwise. Use **numpy.where**, which can act as a vectorised version of the `if` statement.*

### 11.1.3   One-Hot Encoding (*)

Let $\boldsymbol{x}$ be a vector of $n$ integer labels in $\{0, ..., l - 1\}$. Its *one-hot* encoded version is a 0/1 (or, equivalently, logical) matrix $\mathbf{R}$ of shape $n$-by-$l$ such that $r_{i,j} = 1$ if and only if $x_i = j$.

For example, if $x = (0, 1, 2, 1)$ and $l = 4$, then:

$$\mathbf{R} = \begin{bmatrix} 1 & 0 & 0 & 0 \\ 0 & 1 & 0 & 0 \\ 0 & 0 & 1 & 0 \\ 0 & 1 & 0 & 0 \end{bmatrix}.$$

One can easily verify that each row consists of one and only one 1 (the number of 1s per one row is 1). Such a representation is adequate when solving a multiclass classification problem by means of $l$ binary classifiers. For example, if *spam*, *bacon*, and *hot dogs* are on the menu, then *spam* is encoded as $(1, 0, 0)$, i.e., yeah-spam, nah-bacon, and nah-hot dog. We can build three binary classifiers, each narrowly specialising in sniffing one particular type of food.

**Example 11.4** *Write a function to one-hot encode a given categorical vector represented using character strings.*

**Example 11.5** *Write a function to decode a one-hot encoded matrix.*

**Example 11.6** *(\*) We can also work with matrices like $\mathbf{P} \in [0, 1]^{n \times l}$, where $p_{i,j}$ denotes the probability of the i-th object's belonging to the j-th class. Given an example matrix of this kind, verify that in each row the probabilities sum to 1 (up to a small numeric error). Then, decode such a matrix by choosing the greatest element in each row.*

### 11.1.4  Binning Numeric Data (Revisited)

Numerical data can be converted to categorical via binning (quantisation). Even though this results in information loss, it may open some new possibilities. In fact, we needed binning to draw all the histograms in Chapter 4. Also, reporting observation counts in each bin instead of raw data enables us to include them in printed reports (in the form of tables).

---

**Note**  As strong proponents of openness and transparency, we always encourage all entities (governments, universities, non-profits, corporations, etc.) to share raw, unabridged versions of their datasets under the terms of some open data license. This is to enable public scrutiny and to get the most out of the possibilities they can bring for benefit of the community.

Of course, sometimes the sharing of unprocessed information can violate the privacy of the subjects. In such a case, it might be a good idea to communicate them in a binned form.

---

---

**Note**  Rounding is a kind of binning. In particular, `numpy.round` allows for rounding to the nearest tenths, hundredths, ..., as well as tens, hundreds, .... It is useful if data are naturally imprecise, and we do not want to give the impression that it is otherwise.

Nonetheless, rounding can easily introduce tied observations, which are problematic on their own; see Section 5.5.3.

---

Consider the 16 best marathon finish times (in minutes):

```
mins = np.array(marathon.mins)
mins16 = mins[:16]
mins16
## array([129.32, 130.75, 130.97, 134.17, 134.68, 135.97, 139.88, 143.2 ,
##         145.22, 145.92, 146.83, 147.8 , 149.65, 149.88, 152.65, 152.88])
```

**numpy.searchsorted** can be used to determine the interval where each value in `mins` falls.

```
bins = [130, 140, 150]
codes_mins16 = np.searchsorted(bins, mins16)
codes_mins16
## array([0, 1, 1, 1, 1, 1, 1, 2, 2, 2, 2, 2, 2, 2, 3, 3])
```

By default, the intervals are of the form $(a, b]$ (not including $a$, including $b$). Code 0 corresponds to values less than the first bin bound, whereas code 3 – greater than or equal to the last bound:

**pandas.cut** gives us another interface to the same binning method. It returns a vector-like object with `dtype="category"`, with very readable labels generated automatically (and ordered; see Section 11.4.7):

```
cut_mins16 = pd.Series(pd.cut(mins16, [-np.inf, 130, 140, 150, np.inf]))
cut_mins16.iloc[ [0, 1, 6, 7, 13, 14, 15] ].astype("str")  # preview
## 0       (-inf, 130.0]
## 1      (130.0, 140.0]
## 6      (130.0, 140.0]
## 7      (140.0, 150.0]
## 13     (140.0, 150.0]
## 14       (150.0, inf]
## 15       (150.0, inf]
## dtype: object
cut_mins16.cat.categories.astype("str")
## Index(['(-inf, 130.0]', '(130.0, 140.0]', '(140.0, 150.0]',
##        '(150.0, inf]'],
##        dtype='object')
```

**Example 11.7** *(\*) We can create a set of the corresponding categories manually, for example, as follows:*

```
bins2 = np.r_[-np.inf, bins, np.inf]
```

```
np.array(
    [f"({bins2[i]}, {bins2[i+1]}]" for i in range(len(bins2)-1)]
)
## array(['(-inf, 130.0]', '(130.0, 140.0]', '(140.0, 150.0]',
##        '(150.0, inf]'], dtype='<U14')
```

**Exercise 11.8** (*) *Check out the* **numpy.histogram_bin_edges** *function which tries to determine some informative interval bounds based on a few simple heuristics. Recall that* **numpy.linspace** *and* **numpy.geomspace** *can be used for generating equidistant bounds on linear and logarithmic scales, respectively.*

## 11.1.5   Generating Pseudorandom Labels

**numpy.random.choice** allows us to create a pseudorandom sample with categories picked with given probabilities:

```
np.random.seed(123)
np.random.choice(
    a=["spam", "bacon", "eggs", "tempeh"],
    p=[   0.7,     0.1,    0.15,      0.05],
    replace=True,
    size=16
)
## array(['spam', 'spam', 'spam', 'spam', 'bacon', 'spam', 'tempeh', 'spam',
##        'spam', 'spam', 'spam', 'bacon', 'spam', 'spam', 'spam', 'bacon'],
##        dtype='<U6')
```

If we generate a sufficiently long vector, we will expect `"spam"` to occur ca. 70% times, and `"tempeh"` to be drawn in 5% of the cases, etc.

## 11.2   Frequency Distributions

### 11.2.1   Counting

**pandas.Series.value_counts** creates a frequency table in the form of a `Series` object equipped with a readable index (element labels):

```
pd.Series(cntrs16).value_counts()  # sort=True, ascending=False
## KE     7
## PL     5
## ET     2
## MA     1
```

```
## IL      1
## dtype: int64
```

By default, data are ordered with respect to the counts, decreasingly.

If we *already* have an array of integer codes between 0 and $l - 1$, **numpy.bincount** will return the number of times each code appears therein.

```
counts_cntrs16 = np.bincount(codes_cntrs16)
counts_cntrs16
## array([7, 2, 1, 5, 1])
```

A vector of counts can easily be turned into a vector of proportions (fractions) or percentages (if we multiply them by 100):

```
counts_cntrs16 / np.sum(counts_cntrs16) * 100
## array([43.75, 12.5 ,  6.25, 31.25,  6.25])
```

Almost 31.25% of the top runners were from Poland (this marathon is held in Warsaw after all...).

**Exercise 11.9** *Using **numpy.argsort**, sort* `counts_cntrs16` *increasingly together with the corresponding items in* `cat_cntrs16`.

## 11.2.2    Two-Way Contingency Tables: Factor Combinations

Some datasets may feature many categorical columns, each having possibly different levels. Let us now consider all the runners in the `marathon` dataset:

```
marathon.loc[:, "age"] = marathon.category.str.slice(1)  # first two chars
marathon.loc[marathon.age >= "60", "age"] = "60+"  # too few runners aged 70+
marathon = marathon.loc[:, ["sex", "age", "country"]]
marathon.head()
##   sex age country
## 0   M  20      KE
## 1   M  20      KE
## 2   M  20      KE
## 3   M  20      ET
## 4   M  30      KE
```

The three columns are: sex, age (in 10-year brackets), and country. We can of course analyse the data distribution in each column individually, but this we leave as an exercise. Instead, we note that some interesting patterns might also arise when we study the *combinations* of levels of different variables.

Here are the levels of the sex and age variables:

```
pd.unique(marathon.sex)
## array(['M', 'F'], dtype=object)
pd.unique(marathon.age)
## array(['20', '30', '50', '40', '60+'], dtype=object)
```

We have $2 \cdot 5 = 10$ combinations thereof. We can use **pandas.DataFrame.value_counts** to determine the number of observations at each two-dimensional level:

```
counts2 = marathon.loc[:, ["sex", "age"]].value_counts()
counts2
## sex  age
## M    30     2200
##      40     1708
##      20      879
##      50      541
## F    30      449
##      40      262
##      20      240
## M    60+     170
## F    50       43
##      60+      19
## dtype: int64
```

These can be converted to a two-way *contingency table*, which is a matrix that gives the number of occurrences of each pair of values from the two factors:

```
V = counts2.unstack(fill_value=0)
V
## age   20    30    40   50   60+
## sex
## F     240   449   262   43    19
## M     879  2200  1708  541   170
```

For example, there were 19 women aged at least 60 amongst the marathoners. Jolly good.

The *marginal* (one-dimensional) frequency distributions can be recreated by computing the rowwise and columnwise sums of V:

```
np.sum(V, axis=1)
## sex
## F    1013
## M    5498
## dtype: int64
np.sum(V, axis=0)
## age
```

```
## 20      1119
## 30      2649
## 40      1970
## 50       584
## 60+      189
## dtype: int64
```

### 11.2.3 Combinations of Even More Factors

`pandas.DataFrame.value_counts` can also be used with a combination of more than two categorical variables:

```
counts3 = (marathon
    .loc[
        marathon.country.isin(["PL", "UA", "SK"]),
        ["country", "sex", "age"]
    ]
    .value_counts()
    .rename("count")
    .reset_index()
)
counts3
##     country sex  age  count
## 0        PL   M   30   2081
## 1        PL   M   40   1593
## 2        PL   M   20    824
## 3        PL   M   50    475
## 4        PL   F   30    422
## 5        PL   F   40    248
## 6        PL   F   20    222
## 7        PL   M  60+    134
## 8        PL   F   50     26
## 9        PL   F  60+      8
## 10       UA   M   20      8
## 11       UA   M   30      8
## 12       UA   M   50      3
## 13       UA   F   30      2
## 14       UA   M   40      2
## 15       SK   F   50      1
## 16       SK   M   40      1
## 17       SK   M  60+      1
```

The display is in the *long* format (compare Section 10.6.2), because we cannot nicely print a three-dimensional array. Yet, we can always partially unstack the dataset, for aesthetic reasons:

```
counts3.set_index(["country", "sex", "age"]).unstack()
##                 count
## age                20        30        40       50      60+
## country sex
## PL      F       222.0     422.0     248.0     26.0      8.0
##         M       824.0    2081.0    1593.0    475.0    134.0
## SK      F         NaN       NaN       NaN      1.0      NaN
##         M         NaN       NaN       1.0      NaN      1.0
## UA      F         NaN       2.0       NaN      NaN      NaN
##         M         8.0       8.0       2.0      3.0      NaN
```

Let us again appreciate how versatile is the concept of data frames. Not only can we represent data to be investigated (one row per observation, variables possibly of mixed types), but also we can store the results of such analyses (neatly formatted tables).

## 11.3 Visualising Factors

Methods for visualising categorical data are by no means fascinating (unless we use them as grouping variables in more complex datasets, but this is a topic that we cover in Chapter 12).

### 11.3.1 Bar Plots

Example data:

```
x = (marathon.age.astype("category")
    .cat.reorder_categories(["20", "30", "40", "50", "60+"])
    .value_counts(sort=False)
)
x
## 20      1119
## 30      2649
## 40      1970
## 50       584
## 60+      189
## Name: age, dtype: int64
```

Bar plots are self-explanatory and hence will do the trick most of the time; see Figure 11.1.

```
ind = np.arange(len(x))  # 0, 1, 2, 3, 4
plt.bar(ind, height=x, color="lightgray", edgecolor="black", alpha=0.8)
```

*(continues on next page)*

```
plt.xticks(ind, x.index)
plt.show()
```

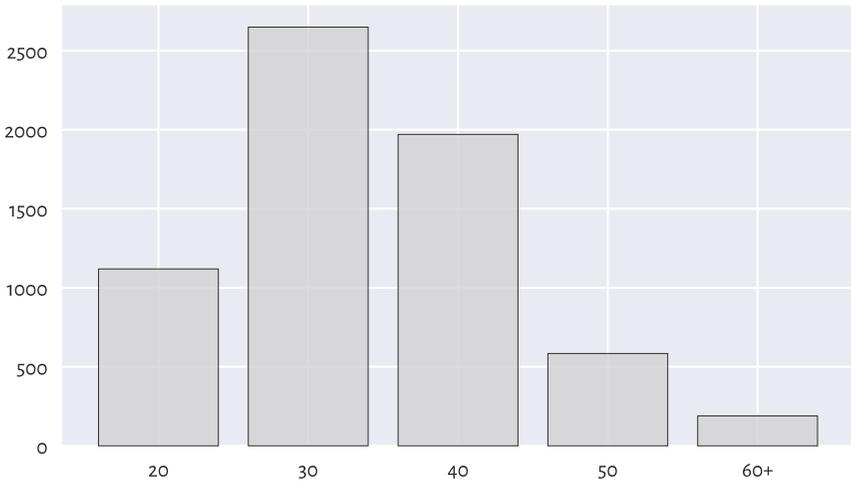

Figure 11.1: Runners' age categories

The `ind` vector gives the x-coordinates of the bars; here: consecutive integers. By calling `matplotlib.pyplot.xticks` we assign them readable labels.

**Exercise 11.10** *Draw a bar plot for the five most prevalent foreign (i.e., excluding Polish) marathoners' original whereabouts which features an additional bar that represents "all other" countries. Depict percentages instead of counts, so that the total bar height is 100%. Assign a different colour to each bar.*

A bar plot is a versatile tool for visualising the counts also in the two-variable case; see Figure 11.2. Let us use `seaborn.barplot` now, which is a pleasant wrapper around `matplotlib.pyplot.bar` (but, as usual, gives less control over the details):

```
v = (marathon.loc[:, ["sex", "age"]].value_counts(sort=False)
    .rename("count").reset_index()
)
sns.barplot(x="age", hue="sex", y="count", data=v)
plt.show()
```

**Exercise 11.11** *(\*) Draw a similar chart using `matplotlib.pyplot.bar`.*

**Exercise 11.12** *(\*\*) Create a stacked bar plot similar to the one in Figure 11.3, where we have horizontal bars for data that have been normalised so that, for each sex, their sum is 100%.*

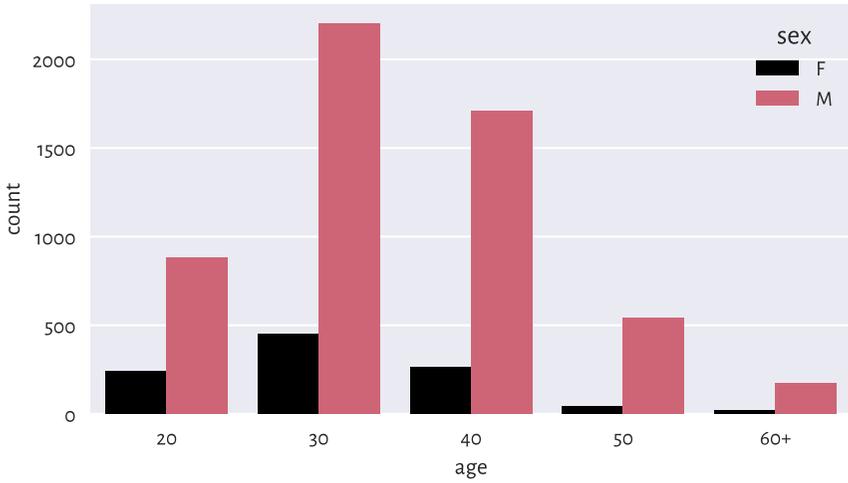

Figure 11.2: Number of runners by age category and sex

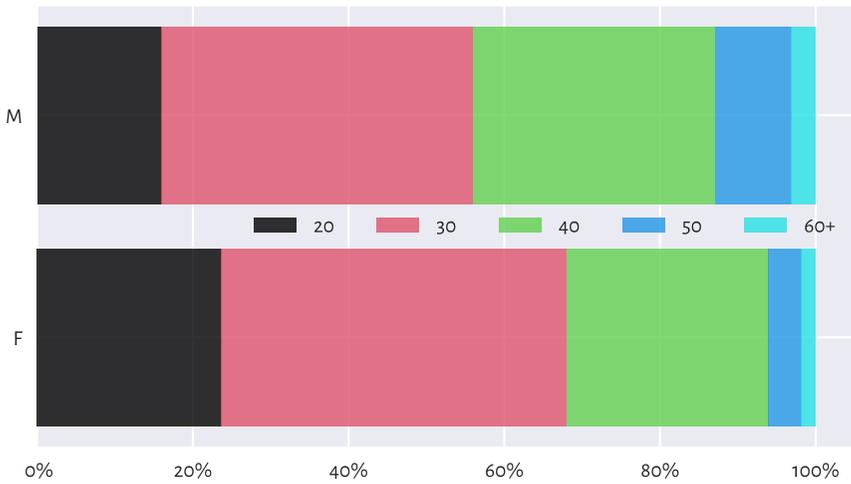

Figure 11.3: An example stacked bar plot: Age distribution for different sexes amongst all the runners

## 11.3.2    Political Marketing and Statistics

Even such a simple chart as bar plot can be manipulated. In the second round of the presidential elections that were held in Poland in 2020, Andrzej Duda won against Rafał Trzaskowski. In Figure 11.4, we have the official results that might be presented by the pro-government media outlets:

```python
plt.bar([1, 2], height=[51.03, 48.97], width=0.25,
    color="lightgray", edgecolor="black", alpha=0.8)
plt.xticks([1, 2], ["Duda", "Trzaskowski"])
plt.ylabel("%")
plt.xlim(0, 3)
plt.ylim(48.9, 51.1)
plt.show()
```

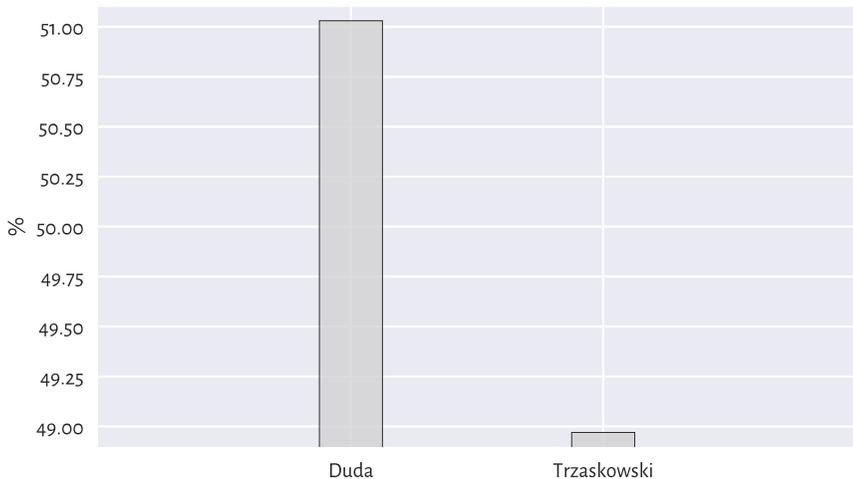

Figure 11.4: Such a great victory! Wait... Just look at the y-axis tick marks.

Another media conglomerate could have reported it like in Figure 11.5:

```python
plt.bar([1, 2], height=[51.03, 48.97], width=0.25,
    color="lightgray", edgecolor="black", alpha=0.8)
plt.xticks([1, 2], ["Duda", "Trzaskowski"])
plt.ylabel("%")
plt.xlim(0, 3)
plt.ylim(0, 250)
plt.yticks([0, 100])
plt.show()
```

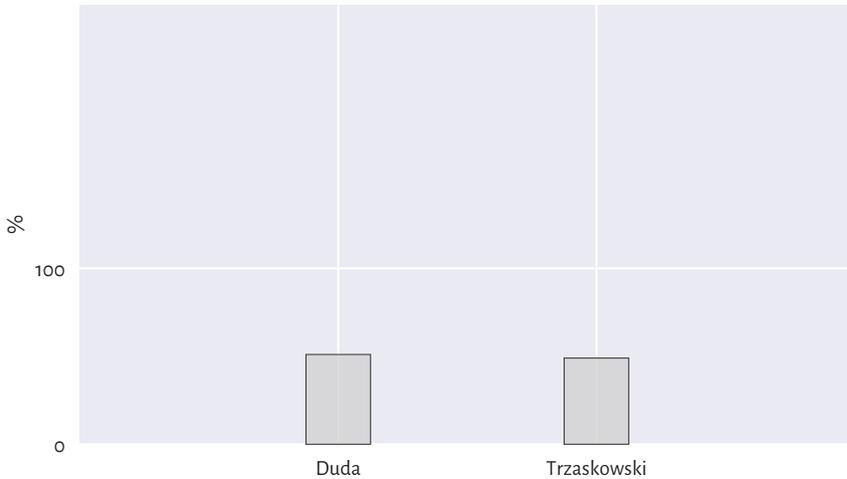

Figure 11.5: It was a draw! So close!

---

**Important** We should always read the axis tick marks. And when drawing our own bar plots, we must never trick the reader; this is unethical; compare Rule#9.

---

### 11.3.3 Pie Cha... Don't Even Trip

We are definitely not going to discuss the infamous pie charts, because their use in data analysis has been widely criticised for a long time (it is difficult to judge the ratios of areas of their slices). Do not draw them. Ever. Good morning.

### 11.3.4 Pareto Charts (*)

As a general (empirical) rule, it is usually the case that most instances of something's happening (usually 70–90%) are due to only a few causes (10–30%). This is known as the *Pareto principle* (with 80% vs 20% being an often cited rule of thumb).

**Example 11.13** *In Chapter 6, we modelled the US cities' population dataset using the Pareto distribution (the very same Pareto, but a different, yet related mathematical object). We discovered that only ca. 14% of the settlements (those with 10,000 or more inhabitants) are home to as much as 84% of the population. Hence, we may say that this data domain follows the Pareto rule.*

Here is a dataset[3] fabricated by the Clinical Excellence Commission in New South Wales, Australia, listing the most frequent causes of medication errors:

---

[3] https://www.cec.health.nsw.gov.au/CEC-Academy/quality-improvement-tools/pareto-charts

```
cat_med = np.array([
    "Unauthorised drug", "Wrong IV rate", "Wrong patient", "Dose missed",
    "Underdose", "Wrong calculation","Wrong route", "Wrong drug",
    "Wrong time", "Technique error", "Duplicated drugs", "Overdose"
])
counts_med = np.array([1, 4, 53, 92, 7, 16, 27, 76, 83, 3, 9, 59])
np.sum(counts_med)  # total number of medication errors
## 430
```

Let us display the dataset ordered with respect to the counts, decreasingly:

```
med = pd.Series(counts_med, index=cat_med).sort_values(ascending=False)
med
## Dose missed         92
## Wrong time          83
## Wrong drug          76
## Overdose            59
## Wrong patient       53
## Wrong route         27
## Wrong calculation   16
## Duplicated drugs     9
## Underdose            7
## Wrong IV rate        4
## Technique error      3
## Unauthorised drug    1
## dtype: int64
```

Pareto charts may aid in visualising the datasets where the Pareto principle is likely to hold, at least approximately. They include bar plots with some extras:

- bars are listed in decreasing order,

- the cumulative percentage curve is added.

The plotting of the Pareto chart is a little tricky, because it involves using two different Y axes (as usual, fine-tuning the figure and studying the manual of the **matplotlib** package is left as an exercise.)

```
x = np.arange(len(med))  # 0, 1, 2, ...
p = 100.0*med/np.sum(med)  # percentages

fig, ax1 = plt.subplots()
ax1.set_xticks(x-0.5, med.index, rotation=60)
ax1.set_ylabel("%")
ax1.bar(x, height=p, color="lightgray", edgecolor="black")
ax2 = ax1.twinx()  # creates a new coordinate system with a shared x-axis
ax2.plot(x, np.cumsum(p), "ro-")
```

```
ax2.grid(visible=False)
ax2.set_ylabel("cumulative %")

fig.tight_layout()
plt.show()
```

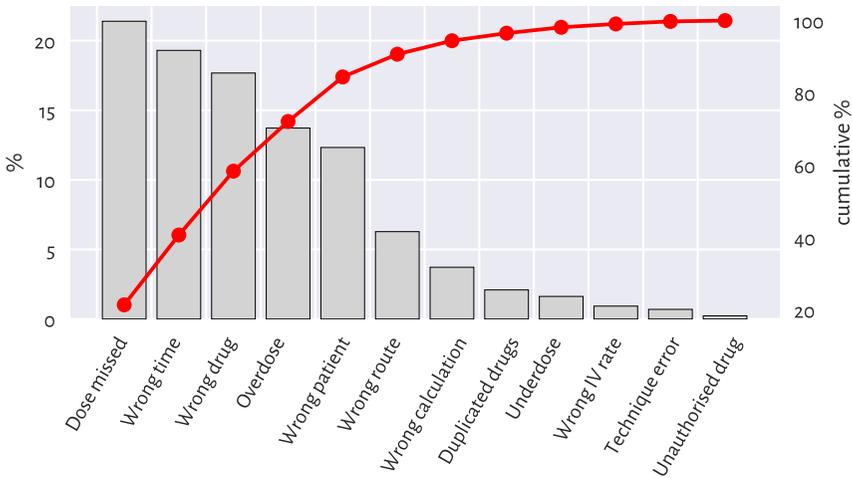

Figure 11.6: An example Pareto chart: the most frequent causes for medication errors

In Figure 11.6, we can read that the first five causes (less than 40%) correspond to ca. 85% of the medication errors. More precisely, the cumulative probabilities are:

```
med.cumsum()/np.sum(med)
## Dose missed          0.213953
## Wrong time           0.406977
## Wrong drug           0.583721
## Overdose             0.720930
## Wrong patient        0.844186
## Wrong route          0.906977
## Wrong calculation    0.944186
## Duplicated drugs     0.965116
## Underdose            0.981395
## Wrong IV rate        0.990698
## Technique error      0.997674
## Unauthorised drug    1.000000
## dtype: float64
```

Note that there is an explicit assumption here that a single error is only due to a single cause. Also, we presume that each medication error has a similar degree of severity.

Policymakers and quality controllers often rely on similar simplifications. They most probably are going to be addressing only the top causes. If we ever wondered why some processes (mal)function the way they do, there is a hint above. Inventing something more effective yet so simple at the same time requires much more effort.

It would be also nice to report the number of cases where no mistakes are made and the cases where errors are insignificant. Healthcare workers are doing a wonderful job for our communities, especially in the public system. Why add to their stress?

### 11.3.5 Heat Maps

Two-way contingency tables can be depicted by means of a heatmap, where each count influences the corresponding cell's colour intensity; see Figure 11.7.

```
from matplotlib import cm
V = marathon.loc[:, ["sex", "age"]].value_counts().unstack(fill_value=0)
sns.heatmap(V, annot=True, fmt="d", cmap=cm.get_cmap("copper"))
plt.show()
```

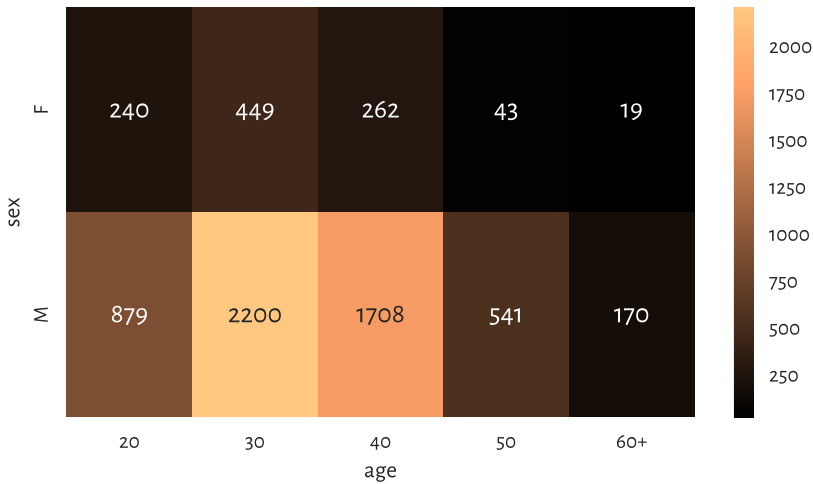

Figure 11.7: A heatmap for the marathoners' sex and age category

## 11.4 Aggregating and Comparing Factors

### 11.4.1 A Mode

The only operation on categorical data on which we can rely is counting.

```
counts = marathon.country.value_counts()
counts.head()
## PL    6033
## GB      71
## DE      38
## FR      33
## SE      30
## Name: country, dtype: int64
```

Therefore, as far as qualitative data aggregation is concerned, what we are left with is the *mode*, i.e., the most frequently occurring value.

```
counts.index[0]   # counts are sorted
## 'PL'
```

---

**Important** A mode might be ambiguous.

---

It turns out that amongst the fastest 22 runners (a nicely round number), there is a tie between Kenya and Poland – both meet our definition of a mode:

```
counts = marathon.country.iloc[:22].value_counts()
counts
## KE    7
## PL    7
## ET    3
## IL    3
## MA    1
## MD    1
## Name: country, dtype: int64
```

To avoid any bias, it is always best to report all the potential mode candidates:

```
counts.loc[counts == counts.max()].index
## Index(['KE', 'PL'], dtype='object')
```

If one value is required, though, we can pick one at random (calling `numpy.random.choice`).

### 11.4.2   Binary Data as Logical Vectors

Recall that we are used to representing binary data as logical vectors or, equivalently, vectors of 0s and 1s.

Perhaps the most useful arithmetic operation on logical vectors is the sum. For example:

```
np.sum(marathon.country == "PL")
## 6033
```

This gave the number of elements that are equal to `"PL"` (because the sum of `0`s and `1`s is equal to the number of `1`s in the sequence). Note that (`country == "PL"`) is a logical vector that represents a binary categorical variable with levels: not-Poland (`False`) and Poland (`True`).

If we divide the above result by the length of the vector, we will get the proportion:

```
np.mean(marathon.country == "PL")
## 0.9265857779142989
```

About 93% of the runners were from Poland. As this is greater than 0.5, `"PL"` is definitely the mode.

**Exercise 11.14** *What is the meaning of* ***numpy.all***, ***numpy.any***, ***numpy.min***, ***numpy.max***, ***numpy.cumsum***, *and* ***numpy.cumprod*** *applied on logical vectors?*

---

**Note** (\*\*) Having the 0/1 (or zero/nonzero) vs `False`/`True` correspondence allows us to perform some logical operations using integer arithmetic. In mathematics, 0 is the annihilator of multiplication and the neutral element of addition, whereas 1 is the neutral element of multiplication. In particular, assuming that `p` and `q` are logical values and `a` and `b` are numeric ones, we have, what follows:

- `p+q != 0` means that at least one value is `True` and `p+q == 0` if and only if both are `False`;

- more generally, `p+q == 2` if both elements are `True`, `p+q == 1` if only one is `True` (we call it exclusive-or, XOR), and `p+q == 0` if both are `False`;

- `p*q != 0` means that both values are `True` and `p*q == 0` holds whenever at least one is `False`;

- `1-p` corresponds to the negation of `p` (changes 1 to 0 and 0 to 1);

- `p*a + (1-p)*b` is equal to `a` if `p` is `True` and equal to `b` otherwise.

---

### 11.4.3 Pearson's Chi-Squared Test (\*)

The Kolmogorov–Smirnov test that we described in Section 6.2.3 verifies whether a given sample differs significantly from a hypothesised *continuous*[4] distribution, i.e., it works for *numeric* data.

---

[4] (\*) There exists a discrete version of the Kolmogorov–Smirnov test, but it is defined in a different way than in Section 6.2.3; compare [2, 14].

For binned/categorical data, we can use a classical and easy-to-understand test developed by Karl Pearson in 1900. It is supposed to judge whether the differences between the observed proportions $\hat{p}_1, \ldots, \hat{p}_l$ and the theoretical ones $p_1, \ldots, p_l$ are significantly large or not:

$$\begin{cases} H_0: & \hat{p}_i = p_i \text{ for all } i = 1, \ldots, l & \text{(null hypothesis)} \\ H_1: & \hat{p}_i \neq p_i \text{ for some } i = 1, \ldots, l & \text{(alternative hypothesis)} \end{cases}$$

Having such a test is beneficial, e.g., when the data we have at hand are based on small surveys that are supposed to serve as estimates of what might be happening in a larger population.

Going back to our political example from Section 11.3.2, it turns out that one of the pre-election polls indicated that $c = 516$ out of $n = 1017$ people would vote for the first candidate. We have $\hat{p}_1 = 50.74\%$ (Duda) and $\hat{p}_2 = 49.26\%$ (Trzaskowski). If we would like to test whether the observed proportions are significantly different from each other, we could test them against the theoretical distribution $p_1 = 50\%$ and $p_2 = 50\%$, which states that there is a tie between the competitors (up to a sampling error).

A natural test statistic is based on the relative squared differences:

$$\hat{T} = n \sum_{i=1}^{l} \frac{(\hat{p}_i - p_i)^2}{p_i}.$$

```
c, n = 516, 1017
p_observed = np.array([c, n-c]) / n
p_expected = np.array([0.5, 0.5])
T = n * np.sum( (p_observed-p_expected)**2 / p_expected )
T
## 0.2212389380530986
```

Similarly to the continuous case in Section 6.2.3, we should reject the null hypothesis, if:

$$\hat{T} \geq K.$$

The critical value $K$ is computed based on the fact that, if the null hypothesis is true, $\hat{T}$ follows the $\chi^2$ (chi-squared, hence the name of the test) distribution with $l-1$ degrees of freedom, see **scipy.stats.chi2**.

We thus need to query the theoretical quantile function to determine the test statistic that is not exceeded in 99.9% of the trials (under the null hypothesis):

```
alpha = 0.001  # significance level
scipy.stats.chi2.ppf(1-alpha, len(p_observed)-1)
## 10.827566170662733
```

As $\hat{T} < K$ (because $0.22 < 10.83$), we cannot deem the two proportions significantly different. In other words, this poll did not indicate (at the significance level 0.1%) any of the candidates as a clear winner.

**Exercise 11.15** *Assuming $n = 1017$, determine the smallest $c$, i.e., the number of respondents claiming they would vote for Duda, that leads to the rejection of the null hypothesis.*

### 11.4.4 Two-Sample Pearson's Chi-Squared Test (*)

Let us consider the data depicted in Figure 11.3 and test whether the runners' age distributions differ significantly between men and women.

```
V = marathon.loc[:, ["sex", "age"]].value_counts().unstack(fill_value=0)
c1, c2 = np.array(V)  # first row, second row
c1  # women
## array([240, 449, 262,  43,  19])
c2  # men
## array([ 879, 2200, 1708,  541,  170])
```

There are $l = 5$ age categories. First, denote the total number of observations in both groups with $n'$ and $n''$.

```
n1 = c1.sum()
n2 = c2.sum()
n1, n2
## (1013, 5498)
```

The observed proportions in the first group (females), denoted as $\hat{p}'_1, \ldots, \hat{p}'_l$, are, respectively:

```
p1 = c1/n1
p1
## array([0.23692004, 0.44323791, 0.25863771, 0.04244817, 0.01875617])
```

Here are the proportions in the second group (males), $\hat{p}''_1, \ldots, \hat{p}''_l$:

```
p2 = c2/n2
p2
## array([0.15987632, 0.40014551, 0.31065842, 0.09839942, 0.03092033])
```

We would like to verify whether the corresponding proportions are equal (up to some sampling error):

$$\begin{cases} H_0: & \hat{p}'_i = \hat{p}''_i \text{ for all } i = 1, \ldots, l & \text{(null hypothesis)} \\ H_1: & \hat{p}'_i \neq \hat{p}''_i \text{ for some } i = 1, \ldots, l & \text{(alternative hypothesis)} \end{cases}$$

In other words, we are interested whether the categorical data in the two groups come from the same discrete probability distribution.

Taking the estimated expected proportions:

$$\bar{p}_i = \frac{n_i' \hat{p}_i' + n_i'' \hat{p}_i''}{n' + n''},$$

for all $i = 1, \dots, l$, the test statistic this time is equal to:

$$\hat{T} = n' \sum_{i=1}^{l} \frac{\left(\hat{p}_i' - \bar{p}_i\right)^2}{\bar{p}_i} + n'' \sum_{i=1}^{l} \frac{\left(\hat{p}_i'' - \bar{p}_i\right)^2}{\bar{p}_i},$$

which is a variation on the one-sample theme presented in the previous subsection.

```
pp = (n1*p1+n2*p2)/(n1+n2)
T = n1 * np.sum( (p1-pp)**2 / pp ) + n2 * np.sum( (p2-pp)**2 / pp )
T
## 75.31373854741857
```

It can be shown that, if the null hypothesis is true, the test statistic *approximately* follows the $\chi^2$ distribution with $l - 1$ degrees of freedom[5]. The critical value $K$ is equal to:

```
alpha = 0.001  # significance level
scipy.stats.chi2.ppf(1-alpha, len(p1)-1)
## 18.46682695290317
```

As $\hat{T} \geq K$ (because $75.31 \geq 18.47$), we reject the null hypothesis. And so, the runners' age distribution differs across sexes (at significance level 0.1%).

### 11.4.5  Measuring Association (*)

Let us consider the Australian Bureau of Statistics National Health Survey 2018[6] data on the prevalence of certain medical conditions as a function of age. Here is the extracted contingency table:

```
l = [
    ["Arthritis", "Asthma", "Back problems", "Cancer (malignant neoplasms)",
     "Chronic obstructive pulmonary disease", "Diabetes mellitus",
     "Heart, stroke and vascular disease", "Kidney disease",
     "Mental and behavioural conditions", "Osteoporosis"],
    ["15-44", "45-64", "65+"]
]
C = 1000*np.array([
    [ 360.2,       1489.0,        1772.2],
```

*(continues on next page)*

---

[5] Notice that [66] in Section 14.3 recommends $l$ degrees of freedom, but we do not agree with this rather informal reasoning. Also, simple Monte Carlo simulations suggest that $l - 1$ is a better candidate.

[6] https://www.abs.gov.au/statistics/health/health-conditions-and-risks/
national-health-survey-first-results/2017-18

```
    [1069.7,      741.9,      433.7],
    [1469.6,     1513.3,      955.3],
    [  28.1,      162.7,      237.5],
    [ 103.8,      207.0,      251.9],
    [ 135.4,      427.3,      607.7],
    [  94.0,      344.4,      716.0],
    [  29.6,       67.7,      123.3],
    [2218.9,     1390.6,      725.0],
    [  36.1,      312.3,      564.7],
]).astype(int)
pd.DataFrame(C, index=l[0], columns=l[1])
```

```
##                                        15-44    45-64      65+
## Arthritis                             360000  1489000  1772000
## Asthma                               1069000   741000   433000
## Back problems                        1469000  1513000   955000
## Cancer (malignant neoplasms)           28000   162000   237000
## Chronic obstructive pulmonary disease  103000   207000   251000
## Diabetes mellitus                      135000   427000   607000
## Heart, stroke and vascular disease      94000   344000   716000
## Kidney disease                          29000    67000   123000
## Mental and behavioural conditions     2218000  1390000   725000
## Osteoporosis                           36000   312000   564000
```

Cramér's $V$ is one of a few ways to measure the degree of association between two categorical variables. It is equal to 0 (the lowest possible value) if the two variables are independent (there is no association between them) and 1 (the highest possible value) if they are tied.

Given a two-way contingency table $C$ with $n$ rows and $m$ columns and assuming that:

$$T = \sum_{i=1}^{n} \sum_{j=1}^{m} \frac{(c_{i,j} - e_{i,j})^2}{e_{i,j}},$$

where:

$$e_{i,j} = \frac{\left(\sum_{k=1}^{m} c_{i,k}\right)\left(\sum_{k=1}^{n} c_{k,j}\right)}{\sum_{i=1}^{n} \sum_{j=1}^{m} c_{i,j}},$$

the Cramér coefficient is given by:

$$V = \sqrt{\frac{T}{\min\{n-1, m-1\} \sum_{i=1}^{n} \sum_{j=1}^{m} c_{i,j}}}.$$

Here, $c_{i,j}$ gives the actually observed counts and $e_{i,j}$ denotes the number that we would expect to see if the two variables were really independent.

```
scipy.stats.contingency.association(C)
## 0.316237999724298
```

The above means that there might be a small association between age and the prevalence of certain conditions. In other words, it might be the case that some conditions are more prevalent in different age groups than others.

**Exercise 11.16**  *Compute the Cramér V using only* **numpy** *functions.*

**Example 11.17**  *(\*\*) We can easily verify the hypothesis whether V does not differ significantly from 0, i.e., whether the variables are independent. Looking at T, we see that this is essentially the test statistic in Pearson's chi-squared goodness-of-fit test.*

```
E = C.sum(axis=1).reshape(-1, 1) * C.sum(axis=0).reshape(1, -1) / C.sum()
T = np.sum((C-E)**2 / E)
T
## 3715440.465191512
```

*If the data are really independent, T follows the chi-squared distribution $n + m - 1$. As a consequence, the critical value K is equal to:*

```
alpha = 0.001  # significance level
scipy.stats.chi2.ppf(1-alpha, C.shape[0] + C.shape[1] - 1)
## 32.90949040736021
```

*As T is much greater than K, we conclude (at significance level 0.1%) that the health conditions are not independent of age.*

**Exercise 11.18**  *(\*) Take a look at* Table 19: Comorbidity of selected chronic conditions *in the* National Health Survey 2018[7], *where we clearly see that many disorders co-occur. Visualise them on some heatmaps and bar plots (including data grouped by sex and age).*

### 11.4.6  Binned Numeric Data

Generally, modes are meaningless for continuous data, where repeated values are – at least theoretically – highly unlikely. It might make sense to compute them on binned data, though.

Looking at a histogram, e.g., in Figure 4.2, the mode is the interval corresponding to the highest bar (hopefully assuming there is only one). If we would like to obtain a single number, we can choose for example the middle of this interval as the mode.

For numeric data, the mode will heavily depend on the coarseness and type of binning; compare Figure 4.4 and Figure 6.7. Thus, the question "what is the most popular income" is overall a quite difficult one to answer.

---

[7] https://www.abs.gov.au/statistics/health/health-conditions-and-risks/
national-health-survey-first-results/2017-18

**Exercise 11.19** *Compute some potentially informative modes for the* `uk_income_simulated_2020`[8] *dataset. Play around with different numbers of bins on linear and logarithmic scales and see how they affect the mode.*

### 11.4.7    Ordinal Data (*)

The case where the categories can be linearly ordered, is called *ordinal data*. For instance, Australian university grades are: F (fail) < P (pass) < C (credit) < D (distinction) < HD (high distinction), some questionnaires use Likert-type scales such as "strongly disagree" < "disagree" < "neutral" < "agree" < "strongly agree", etc.

A linear ordering allows us to go beyond the mode. This is because, due to the existence of order statistics and observation ranks, we can also easily define sample quantiles. Nevertheless, the standard methods for resolving ties will not work: we need to be careful.

For example, the median of a sample of student grades (P, P, C, D, HD) is C, but (P, P, C, D, HD, HD) is either C or D - we can choose one at random or just report that the solution is ambiguous (C+? D-? C/D?).

Another option, of course, is to treat ordinal data as numbers (e.g., F = 0, P = 1, ..., HD = 4). In the latter example, the median would be equal to 2.5.

There are some cases, though, where the conversion of labels to consecutive integers is far from optimal – because it gives the impression that the "distance" between different levels is always equal (linear).

**Exercise 11.20** (**) *The* `grades_results`[9] *dataset represents the grades (F, P, C, D, HD) of 100 students attending an imaginary course in an Australian university. You can load it in the form of an* ordered *categorical* `Series` *by calling:*

```
grades = np.loadtxt("https://raw.githubusercontent.com/gagolews/" +
    "teaching-data/master/marek/grades_results.txt", dtype="str")
grades = pd.Series(pd.Categorical(grades,
    categories=["F", "P", "C", "D", "HD"], ordered=True))
grades.value_counts()  # note the order of labels
## F     30
## P     29
## C     23
## HD    22
## D     19
## dtype: int64
```

*How would you determine the* average *grade represented as a number between 0 and 100, taking into account that for a P you need at least 50%, C is given for ≥ 60%, D for ≥ 70%, and HD for only*

---

[8] https://github.com/gagolews/teaching-data/raw/master/marek/uk_income_simulated_2020.txt
[9] https://github.com/gagolews/teaching-data/raw/master/marek/grades_results.txt

*(!) 80% of the points. Come up with a pessimistic, optimistic, and best-shot estimate, and then compare your result to the true corresponding scores listed in the grades_scores[10] dataset.*

---

## 11.5  Exercises

**Exercise 11.21**  *Does it make sense to compute the arithmetic mean of a categorical variable?*

**Exercise 11.22**  *Name the basic use cases for categorical data.*

**Exercise 11.23**  *(\*) What is a Pareto chart?*

**Exercise 11.24**  *How can we deal with the case of the mode being nonunique?*

**Exercise 11.25**  *What is the meaning of the sum and mean for binary data (logical vectors)?*

**Exercise 11.26**  *What is the meaning of `numpy.mean((x > 0) & (x < 1))`, where x is a numeric vector?*

**Exercise 11.27**  *List some ways to visualise multidimensional categorical data (combinations of two or more factors).*

**Exercise 11.28**  *(\*) State the null hypotheses verified by the one- and two-sample chi-squared goodness-of-fit tests.*

**Exercise 11.29**  *(\*) How is Cramér's V defined and what values does it take?*

---

# *Processing Data in Groups*

Let us consider another subset of the US Centres for Disease Control and Prevention National Health and Nutrition Examination Survey, this time carrying some body measures (P_BMX[1]) together with demographics (P_DEMO[2]).

```python
nhanes = pd.read_csv("https://raw.githubusercontent.com/gagolews/" +
    "teaching-data/master/marek/nhanes_p_demo_bmx_2020.csv",
    comment="#")
nhanes = (
    nhanes
    .loc[
        (nhanes.DMDBORN4 <= 2) & (nhanes.RIDAGEYR >= 18),
        ["RIDAGEYR", "BMXWT", "BMXHT", "BMXBMI", "RIAGENDR", "DMDBORN4"]
    ]  # age >= 18 and only US and non-US born
    .rename({
        "RIDAGEYR": "age",
        "BMXWT": "weight",
        "BMXHT": "height",
        "BMXBMI": "bmival",
        "RIAGENDR": "gender",
        "DMDBORN4": "usborn"
    }, axis=1)  # rename columns
    .dropna()   # remove missing values
    .reset_index(drop=True)
)
```

We consider only the adult (at least 18 years old) participants, whose country of birth (the US or not) is well-defined. Let us recode the usborn and gender variables (for readability) and introduce the BMI categories:

```python
nhanes.loc[:, "usborn"] = (
    nhanes.usborn.astype("category")
    .cat.rename_categories(["yes", "no"]).astype("str")      # recode usborn
)
nhanes.loc[:, "gender"] = (
```

---

[1] https://wwwn.cdc.gov/Nchs/Nhanes/2017-2018/P_BMX.htm
[2] https://wwwn.cdc.gov/Nchs/Nhanes/2017-2018/P_DEMO.htm

```
    nhanes.gender.astype("category")
    .cat.rename_categories(["male", "female"]).astype("str")  # recode gender
)
nhanes.loc[:, "bmicat"] = pd.cut(                             # new column
    nhanes.bmival,
    bins= [ 0,              18.5,          25,          30,        np.inf ],
    labels=[    "underweight",    "normal",  "overweight",  "obese"       ]
)
```

Here is a preview of this data frame:

```
nhanes.head()
##     age   weight   height   bmival   gender usborn      bmicat
## 0    29    97.1    160.2    37.8   female     no       obese
## 1    49    98.8    182.3    29.7     male    yes  overweight
## 2    36    74.3    184.2    21.9     male    yes      normal
## 3    68   103.7    185.3    30.2     male    yes       obese
## 4    76    83.3    177.1    26.6     male    yes  overweight
```

We have a mix of categorical (gender, US born-ness, BMI category) and numerical (age, weight, height, BMI) variables. Unless we had encoded qualitative variables as integers, this would not be possible with plain matrices, at least not with a single one.

In this section, we will treat the categorical columns as grouping variables, so that we can, e.g., summarise or visualise the data *in each group* separately, because it is likely that data distributions vary across different factor levels. This is much like having many data frames stored in one object, e.g., the heights of women and men separately.

`nhanes` is thus an example of heterogeneous data at their best.

## 12.1 Basic Methods

`DataFrame` and `Series` objects are equipped with the **groupby** methods, which assist in performing a wide range of operations in data groups defined by one or more data frame columns (compare [84]).

They return objects of class `DataFrameGroupBy` and `SeriesGroupby`:

```
type(nhanes.groupby("gender"))
## <class 'pandas.core.groupby.generic.DataFrameGroupBy'>
type(nhanes.groupby("gender").height)  # or (...)["height"]
## <class 'pandas.core.groupby.generic.SeriesGroupBy'>
```

**Important**   When we wish to browse the list of available attributes in the **pandas** manual, it is worth knowing that DataFrameGroupBy and SeriesGroupBy are separate types. Still, they have many methods and slots in common, because they both inherit from (extend) the GroupBy class.

**Exercise 12.1** *Skim through the documentation[3] of the said classes.*

For example, the **pandas.DataFrameGroupBy.size** method determines the number of observations in each group:

```python
nhanes.groupby("gender").size()
## gender
## female    4514
## male      4271
## dtype: int64
```

It returns an object of type Series. We can also perform the grouping with respect to a combination of levels in two qualitative columns:

```python
nhanes.groupby(["gender", "bmicat"]).size()
## gender  bmicat
## female  underweight      93
##         normal         1161
##         overweight     1245
##         obese          2015
## male    underweight      65
##         normal         1074
##         overweight     1513
##         obese          1619
## dtype: int64
```

This yields a Series with a hierarchical index (as discussed in Section 10.1.3). Nevertheless, we can always call **reset_index** to convert it to standalone columns:

```python
nhanes.groupby(["gender", "bmicat"]).size().rename("counts").reset_index()
##     gender       bmicat  counts
## 0   female  underweight      93
## 1   female       normal    1161
## 2   female   overweight    1245
## 3   female        obese    2015
## 4     male  underweight      65
## 5     male       normal    1074
## 6     male   overweight    1513
## 7     male        obese    1619
```

---

[3] https://pandas.pydata.org/pandas-docs/stable/reference/groupby.html

Take note of the **rename** part. It gave us some readable column names.

Furthermore, it is possible to group rows in a data frame using a list of any `Series` objects, i.e., not just column names in a given data frame; see Section 16.2.3 for an example.

**Exercise 12.2** *(\*) Note the difference between **pandas.GroupBy.count** and **pandas.GroupBy. size** methods (by reading their documentation).*

### 12.1.1    Aggregating Data in Groups

The `DataFrameGroupBy` and `SeriesGroupBy` classes are equipped with several well-known aggregation functions. For example:

```
nhanes.groupby("gender").mean().reset_index()
##    gender        age     weight      height     bmival
## 0  female  48.956580  78.351839  160.089189  30.489189
## 1    male  49.653477  88.589932  173.759541  29.243620
```

The arithmetic mean was computed only on numeric columns[4]. Further, a few common aggregates are generated by **describe**:

```
nhanes.groupby("gender").height.describe().reset_index()
##    gender   count        mean       std  ...    25%    50%    75%    max
## 0  female  4514.0  160.089189  7.035483  ...  155.3  160.0  164.8  189.3
## 1    male  4271.0  173.759541  7.702224  ...  168.5  173.8  178.9  199.6
##
## [2 rows x 9 columns]
```

But we can always apply a custom list of functions by using **aggregate**:

```
(nhanes.
    loc[:, ["gender", "height", "weight"]].
    groupby("gender").
    aggregate([np.mean, np.median, len, lambda x: (np.max(x)-np.min(x))/2]).
    reset_index()
)
##    gender      height                  ...       weight
##                  mean median   len ...       mean median   len <lambda_0>
## 0  female  160.089189  160.0  4514 ...  78.351839   74.1  4514      110.85
## 1    male  173.759541  173.8  4271 ...  88.589932   85.0  4271      102.90
##
## [2 rows x 9 columns]
```

---

[4] (\*) In this example, we called **pandas.GroupBy.mean**. Note that it has slightly different functionality from **pandas.DataFrame.mean** and **pandas.Series.mean**, which all needed to be implemented separately so that we can use them in complex operation chains. Still, they all call the underlying **numpy.mean** function. Object-oriented programming has its pros (more expressive syntax) and cons (sometimes more redundancy in the API design).

The result's `columns` slot features a hierarchical index.

---

**Note** The column names in the output object are generated by reading the applied functions' `__name__` slots, see, e.g., `print`(np.mean.`__name__`).

```python
mr = lambda x: (np.max(x)-np.min(x))/2
mr.__name__ = "midrange"
(nhanes.
    loc[:, ["gender", "height", "weight"]].
    groupby("gender").
    aggregate([np.mean, mr]).
    reset_index()
)
##     gender       height              weight
##               mean midrange       mean midrange
## 0   female  160.089189     29.1  78.351839   110.85
## 1     male  173.759541     27.5  88.589932   102.90
```

---

## 12.1.2 Transforming Data in Groups

We can easily transform individual columns relative to different data groups by means of the **transform** method for `GroupBy` objects.

```python
def std0(x, axis=None):
    return np.std(x, axis=axis, ddof=0)
std0.__name__ = "std0"

def standardise(x):
    return (x-np.mean(x, axis=0))/std0(x, axis=0)

nhanes["height_std"] = (
    nhanes.
    loc[:, ["height", "gender"]].
    groupby("gender").
    transform(standardise)
)

nhanes.head()
##     age  weight  height  bmival  gender  usborn       bmicat  height_std
## 0   29    97.1   160.2    37.8  female      no        obese    0.015752
## 1   49    98.8   182.3    29.7    male     yes  overweight    1.108960
## 2   36    74.3   184.2    21.9    male     yes      normal    1.355671
## 3   68   103.7   185.3    30.2    male     yes        obese    1.498504
## 4   76    83.3   177.1    26.6    male     yes  overweight    0.433751
```

The new column gives the *relative* z-scores: a woman with a relative z-score of 0 has height of 160.1 cm, whereas a man with the same z-score has height of 173.8 cm.

We can check that the means and standard deviations in both groups are equal to 0 and 1:

```
(nhanes.
    loc[:, ["gender", "height", "height_std"]].
    groupby("gender").
    aggregate([np.mean, std0])
)
##               height                 height_std
##                 mean        std0          mean  std0
## gender
## female   160.089189   7.034703   -1.351747e-15   1.0
## male     173.759541   7.701323    3.145329e-16   1.0
```

Interestingly, it is likely a bug in **pandas** that `groupby("gender").aggregate([np.std])` somewhat passes ddof=1 to **numpy.std**, hence our using a custom function.

**Exercise 12.3**  *Create a data frame comprised of the five tallest men and the five tallest women.*

## 12.1.3   Manual Splitting into Subgroups (*)

It turns out that `GroupBy` objects and their derivatives are *iterable*; compare Section 3.4. As a consequence, the grouped data frames and series can be easily processed manually in case where the built-in methods are insufficient (i.e., not so rarely).

Let us consider a small sample of our data frame.

```
grouped = (nhanes.head()
    .loc[:, ["gender", "weight", "height"]].groupby("gender")
)
list(grouped)
## [('female',    gender  weight  height
## 0   female    97.1   160.2), ('male',    gender   weight   height
## 1   male     98.8    182.3
## 2   male     74.3    184.2
## 3   male    103.7    185.3
## 4   male     83.3    177.1)]
```

The way Python formatted the above output is imperfect, so we need to contemplate it for a tick. We see that when iterating through a `GroupBy` object, we get access to pairs giving all the levels of the grouping variable and the subsets of the input data frame corresponding to these categories.

Here is a simple example where we make use of the above fact:

```
for level, df in grouped:
    # level is a string label
    # df is a data frame - we can do whatever we want
    print(f"There are {df.shape[0]} subject(s) with gender=`{level}`.")
## There are 1 subject(s) with gender=`female`.
## There are 4 subject(s) with gender=`male`.
```

We see that splitting followed by manual processing of the chunks in a loop is quite tedious in the case where we would merely like to compute some basic aggregates. These scenarios are extremely common. No wonder why the **pandas** developers introduced a convenient interface in the form of the **pandas.DataFrame.groupby** and **pandas.Series. groupby** methods and the DataFrameGroupBy and SeriesGroupBy classes. Still, for more ambitious tasks, the low-level way to perform the splitting will come in handy.

**Exercise 12.4** *(\*\*) Using the manual splitting and **matplotlib.pyplot.boxplot**, draw a box-and-whisker plot of heights grouped by BMI category (four boxes side by side).*

**Exercise 12.5** *(\*\*) Using the manual splitting, compute the relative z-scores of the height column separately for each BMI category.*

**Example 12.6** *Let us also demonstrate that the splitting can be done manually without the use of **pandas**. Namely, calling **numpy.split(a, ind)** returns a list with a (being an array-like object, e.g., a vector, a matrix, or a data frame) partitioned rowwisely into **len(ind)+1** chunks at indexes given by ind. For example:*

```
a = ["one", "two", "three", "four", "five", "six", "seven", "eight", "nine"]
for e in np.split(a, [2, 6]):
    print(repr(e))
## array(['one', 'two'], dtype='<U5')
## array(['three', 'four', 'five', 'six'], dtype='<U5')
## array(['seven', 'eight', 'nine'], dtype='<U5')
```

*To split a data frame into groups defined by a categorical column, we can first sort it with respect to the criterion of interest, for instance, the gender data:*

```
nhanes_srt = nhanes.sort_values("gender", kind="stable")
```

*Then, we can use **numpy.unique** to fetch the indexes of first occurrences of each series of identical labels:*

```
levels, where = np.unique(nhanes_srt.gender, return_index=True)
levels, where
## (array(['female', 'male'], dtype=object), array([   0, 4514]))
```

*This can now be used for dividing the sorted data frame into chunks:*

```
nhanes_grp = np.split(nhanes_srt, where[1:])  # where[0] is not interesting
```

*We obtained a list of data frames split at rows specified by* `where[1:]`. *Here is a preview of the first and the last row in each chunk:*

```python
for i in range(len(levels)):
    # process (levels[i], nhanes_grp[i])
    print(f"level='{levels[i]}'; preview:")
    print(nhanes_grp[i].iloc[ [0, -1], : ], end="\n\n")
## level='female'; preview:
##       age  weight  height  bmival  gender usborn  bmicat  height_std
## 0      29    97.1   160.2    37.8  female     no   obese    0.015752
## 8781   67    82.8   147.8    37.9  female     no   obese   -1.746938
##
## level='male'; preview:
##       age  weight  height  bmival gender usborn       bmicat  height_std
## 1      49    98.8   182.3    29.7   male    yes  overweight    1.108960
## 8784   74    59.7   167.5    21.3   male     no      normal   -0.812788
```

*Within each subgroup, we can apply any operation we have learned so far: our imagination is the only major limiting factor. For instance, we can aggregate some columns:*

```python
nhanes_agg = [
    dict(
        level=t.gender.iloc[0],  # they are all the same here – take first
        height_mean=np.round(np.mean(t.height), 2),
        weight_mean=np.round(np.mean(t.weight), 2)
    )
    for t in nhanes_grp
]
print(nhanes_agg[0])
## {'level': 'female', 'height_mean': 160.09, 'weight_mean': 78.35}
print(nhanes_agg[1])
## {'level': 'male', 'height_mean': 173.76, 'weight_mean': 88.59}
```

*The resulting list of dictionaries can be combined to form a data frame:*

```python
pd.DataFrame(nhanes_agg)
##     level  height_mean  weight_mean
## 0  female       160.09        78.35
## 1    male       173.76        88.59
```

*Furthermore, a simple trick to allow grouping with respect to more than one column is to apply* **numpy.unique** *on a string vector that combines the levels of the grouping variables, e.g., by concatenating them like* `nhanes_srt.gender + "___" + nhanes_srt.bmicat` *(assuming that* `nhanes_srt` *is ordered with respect to these two criteria).*

## 12.2    Plotting Data in Groups

The **seaborn** package is particularly convenient for plotting grouped data – it is highly interoperable with **pandas**.

### 12.2.1    Series of Box Plots

Figure 12.1 depicts a box plot with four boxes side by side:

```
sns.boxplot(x="bmival", y="gender", hue="usborn",
    data=nhanes, palette="Paired")
plt.show()
```

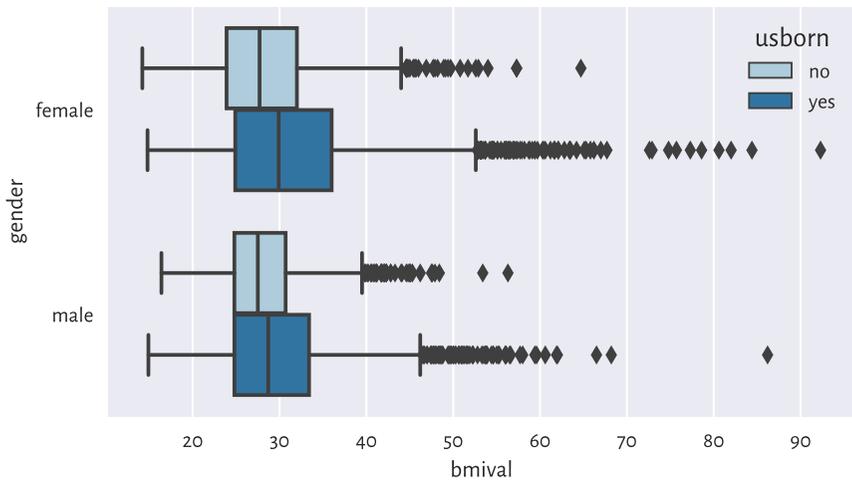

Figure 12.1: The distribution of BMIs for different genders and countries of birth

Let us contemplate for a while how easy it is now to compare the BMI distribution in different groups. Here, we have two grouping variables, as specified by the y and hue arguments.

**Exercise 12.7**    *Create a similar series of violin plots.*

**Exercise 12.8**    *(\*) Add the average BMIs in each group to the above box plot using* **matplotlib. pyplot.plot**. *Check* y l i m *to determine the range on the y-axis.*

### 12.2.2    Series of Bar Plots

In Figure 12.2, on the other hand, we have a bar plot representing a contingency table (obtained in a different way than in Chapter 11):

```
sns.barplot(
    y="counts", x="gender", hue="bmicat", palette="Paired",
    data=(
        nhanes.
        groupby(["gender", "bmicat"]).
        size().
        rename("counts").
        reset_index()
    )
)
plt.show()
```

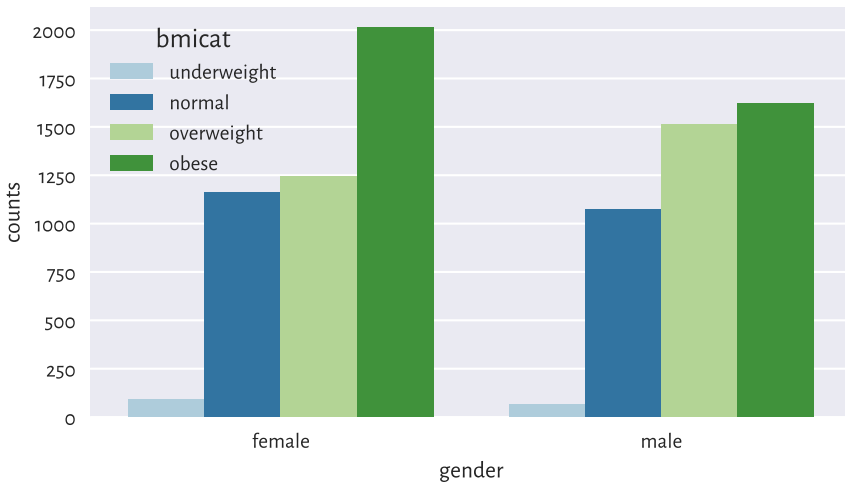

Figure 12.2: Number of persons for each gender and BMI category

**Exercise 12.9** *Draw a similar bar plot where the bar heights sum to 100% for each gender.*

**Exercise 12.10** *Using the two-sample chi-squared test, verify whether the BMI category distributions for men and women differ significantly from each other.*

### 12.2.3 Semitransparent Histograms

Figure 12.3 illustrates that playing with semitransparent objects can make comparisons easy. By passing `common_norm=False`, we scaled each density histogram separately, which is the behaviour we desire if samples are of different lengths.

```
sns.histplot(data=nhanes, x="weight", hue="usborn",
    element="step", stat="density", common_norm=False)
plt.show()
```

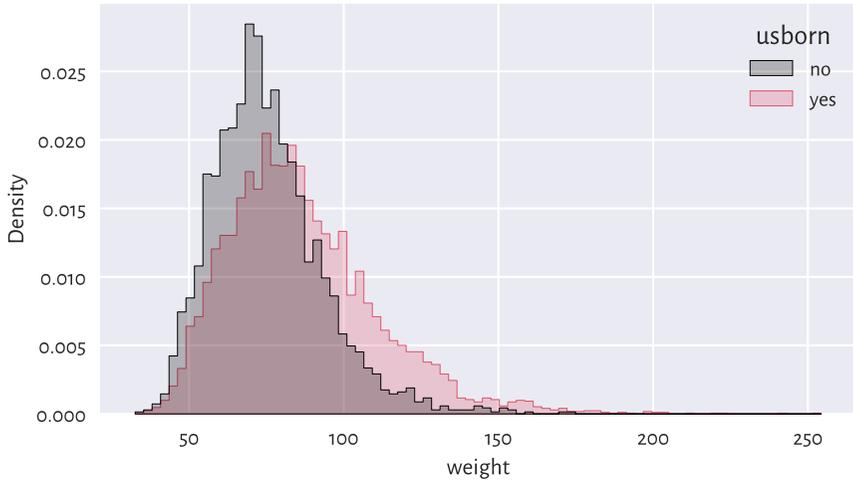

Figure 12.3: The weight distribution of the US-born participants has a higher mean and variance

### 12.2.4 Scatterplots with Group Information

Scatterplots for grouped data can display category information using points of different colours and/or styles, compare Figure 12.4.

```
sns.scatterplot(x="height", y="weight", hue="gender", style="gender",
    data=nhanes, alpha=0.2, markers=["o", "v"])
plt.show()
```

### 12.2.5 Grid (Trellis) Plots

Grid plot (also known as trellis, panel, or lattice plots) are a way to visualise data separately for each factor level. All the plots share the same coordinate ranges which makes them easily comparable. For instance, Figure 12.5 depicts a series of histograms of weights grouped by a combination of two categorical variables.

```
grid = sns.FacetGrid(nhanes, col="gender", row="usborn")
grid.map(sns.histplot, "weight", stat="density", color="lightgray")
# plt.show()  # not required...
```

**Exercise 12.11** *Pass hue="bmicat" additionally to **seaborn.FacetGrid**.*

---

**Important** Grid plots can bear any kind of data visualisation we have discussed so far (e.g., histograms, bar plots, scatterplots).

---

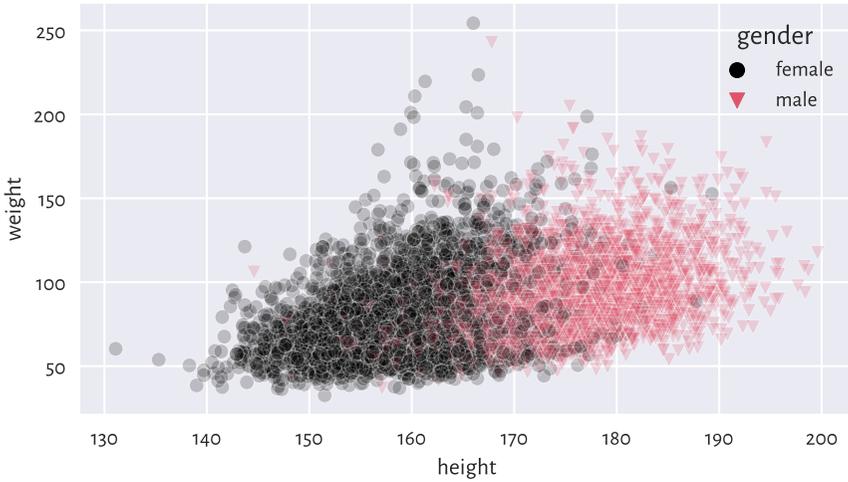

Figure 12.4: Weight vs height grouped by gender

**Exercise 12.12** *Draw a trellis plot with scatterplots of weight vs height grouped by BMI category and gender.*

### 12.2.6   Comparing ECDFs with the Kolmogorov–Smirnov Test (*)

Figure 12.6 compares the empirical cumulative distribution functions of the weight distributions for US and non-US born participants.

```
for usborn, weight in nhanes.groupby("usborn").weight:
    sns.ecdfplot(data=weight, legend=False, label=usborn)
plt.legend(title="usborn")
plt.show()
```

We have used manual splitting of the `weight` column into subgroups and then plotted the two ECDFs separately, because a call to **seaborn.ecdfplot**(data=nhanes, x="weight", hue="usborn") does not honour our wish to use alternating lines styles (most likely due to a bug).

A two-sample Kolmogorov–Smirnov test can be used to check whether two ECDFs $\hat{F}'_n$ (e.g., the weight of the US-born participants) and $\hat{F}''_m$ (e.g., the weight of non-US-born persons) are significantly different from each other:

$$\left\{ \begin{array}{ll} H_0: & \hat{F}'_n = \hat{F}''_n \quad \text{(null hypothesis)} \\ H_1: & \hat{F}'_n \neq \hat{F}''_n \quad \text{(two-sided alternative)} \end{array} \right.$$

The test statistic will be a variation of the one-sample setting discussed in Sec-

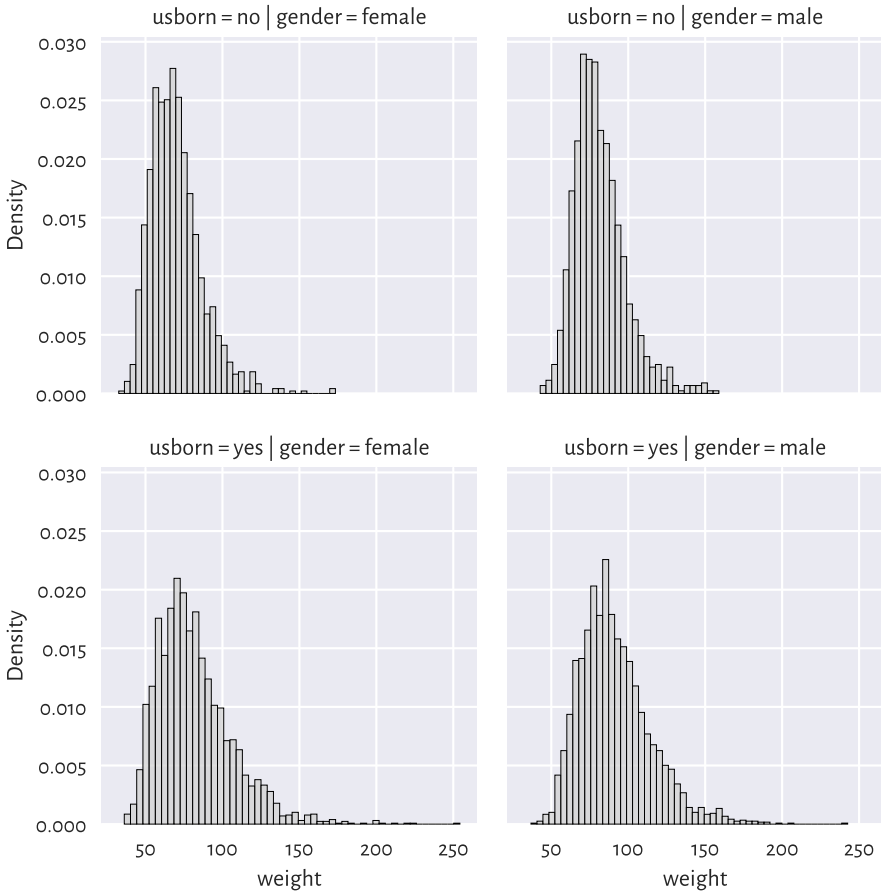

Figure 12.5: Distribution of weights for different genders and countries of birth

tion 6.2.3. Namely, let:

$$\hat{D}_{n,m} = \sup_{t \in \mathbb{R}} |\hat{F}'_n(t) - \hat{F}''_m(t)|.$$

Computing the above is slightly trickier than in the previous case[5], but luckily an appropriate procedure is already implemented in **scipy.stats**:

```python
x12 = nhanes.set_index("usborn").weight
x1 = x12.loc["yes"]  # first sample
x2 = x12.loc["no"]   # second sample
```

*(continues on next page)*

---

[5] Remember that this is an introductory course, and we are still being very generous here. We encourage the readers to upskill themselves (later, of course) not only in mathematics, but also in programming (e.g., algorithms and data structures).

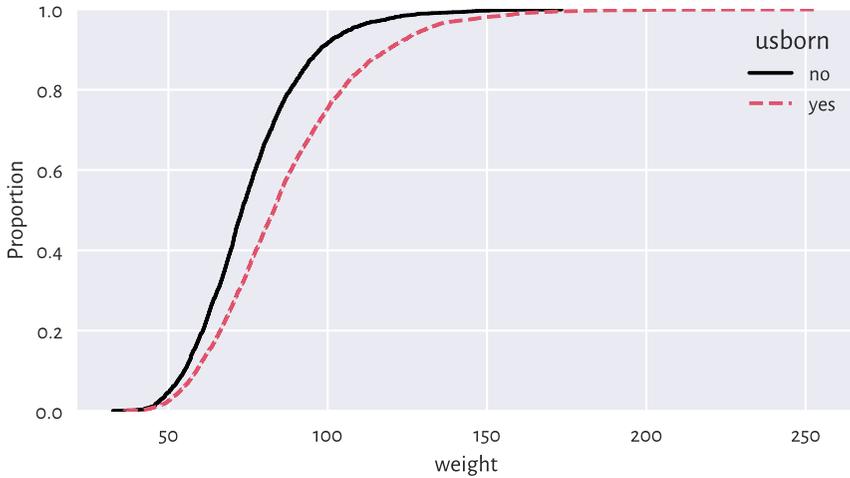

Figure 12.6: Empirical cumulative distribution functions of weight distributions for different birthplaces

*(continued from previous page)*

```
Dnm = scipy.stats.ks_2samp(x1, x2)[0]
Dnm
## 0.22068075889911914
```

Assuming significance level $\alpha = 0.001$, the critical value is approximately (for larger $n$ and $m$) equal to:

$$K_{n,m} = \sqrt{-\frac{\log(\alpha/2)(n+m)}{2nm}}.$$

```
alpha = 0.001
np.sqrt(-np.log(alpha/2) * (len(x1)+len(x2)) / (2*len(x1)*len(x2)))
## 0.04607410479813944
```

As usual, we reject the null hypothesis when $\hat{D}_{n,m} \geq K_{n,m}$, which is exactly the case here (at significance level 0.1%). In other words, weights of US- and non-US-born participants differ significantly.

---

**Important**   Frequentist hypothesis testing only takes into account the deviation between distributions that is explainable due to sampling effects (the assumed randomness of the data generation process). For large sample sizes, even very small deviations[6] will be deemed *statistically significant*, but it does not mean that we should consider them as *practically significant*. For instance, if a very costly, environmentally unfriendly, and generally inconvenient for everyone upgrade leads to a process' improvement such that we reject the null hypothesis stating that two distributions are equal, but it turns out that the gains are ca. 0.5%, the good old common sense should be applied.

---

[6] Including those that are merely due to round-off errors.

**Exercise 12.13** *Compare between the ECDFs of weights of men and women who are between 18 and 25 years old. Determine whether they are significantly different.*

**Important**    Some statistical textbooks and many research papers in the social sciences (amongst many others) employ the significance level of $\alpha = 5\%$, which is often criticised as too high[7]. Many stakeholders aggressively push towards constant improvements in terms of inventing bigger, better, faster, more efficient things. In this context, larger $\alpha$ allows for generating more *sensational* discoveries. This is because it considers smaller differences as already significant. This all adds to what we call the reproducibility crisis in the empirical sciences.

We, on the other hand, claim that it is better to err on the side of being cautious. This, in the long run, is more sustainable.

### 12.2.7    Comparing Quantiles

Plotting quantiles in two samples against each other can also give us some further (informal) insight with regard to the possible distributional differences. Figure 12.7 depicts an example Q-Q plot (see also the one-sample version in Section 6.2.2), where we see that the distributions have similar shapes (points more or less lie on a straight line), but they are shifted and/or scaled (if they were, they would be on the identity line).

```python
x = nhanes.weight.loc[nhanes.usborn == "yes"]
y = nhanes.weight.loc[nhanes.usborn == "no"]
xd = np.sort(x)
yd = np.sort(y)
if len(xd) > len(yd):  # interpolate between quantiles in a longer sample
    xd = np.quantile(xd, np.arange(1, len(yd)+1)/(len(yd)+1))
else:
    yd = np.quantile(yd, np.arange(1, len(xd)+1)/(len(xd)+1))
plt.plot(xd, yd, "o")
plt.axline((xd[len(xd)//2], xd[len(xd)//2]), slope=1,
    linestyle=":", color="gray")  # identity line
plt.xlabel(f"Sample quantiles (weight; usborn=yes)")
```

*(continues on next page)*

---

[7] For similar reasons, we do not introduce the notion of p-values. Most practitioners tend to misunderstand them anyway.

```
plt.ylabel(f"Sample quantiles (weight; usborn=no)")
plt.show()
```

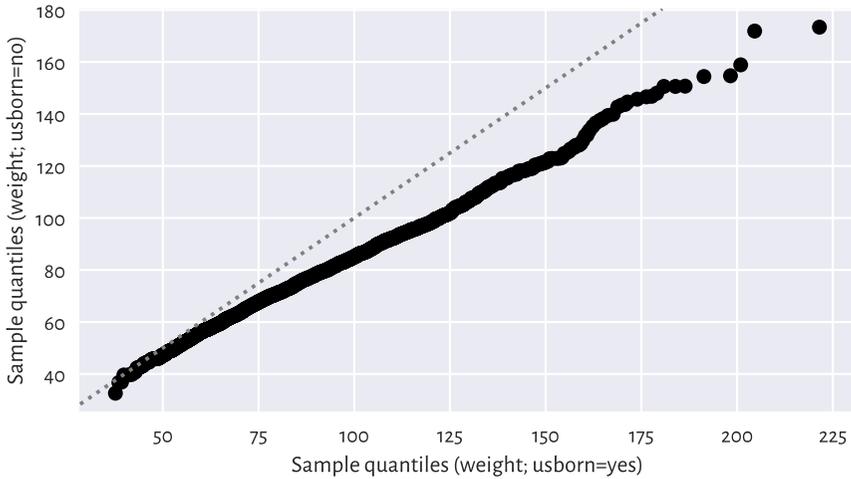

Figure 12.7: A two-sample Q-Q plot

Notice that we interpolated between the quantiles in a larger sample to match the length of the shorter vector.

## 12.3  Classification Tasks

Let us consider a small sample of white, rather sweet wines from a much larger wine quality[8] dataset.

```
wine_train = pd.read_csv("https://raw.githubusercontent.com/gagolews/" +
        "teaching-data/master/other/sweetwhitewine_train2.csv",
        comment="#")
wine_train.head()
##       alcohol       sugar   bad
## 0   10.625271   10.340159     0
## 1    9.066111   18.593274     1
## 2   10.806395    6.206685     0
```

---

[8] http://archive.ics.uci.edu/ml/datasets/Wine+Quality

```
## 3   13.432876    2.739529    0
## 4    9.578162    3.053025    0
```

We are given each wine's alcohol and residual sugar content, as well as a binary categorical variable stating whether a group of sommeliers deem a given beverage quite bad (1) or not (0). Figure 12.8 reveals that subpar wines are rather low in... alcohol and, to some extent, sugar.

```
sns.scatterplot(x="alcohol", y="sugar", data=wine_train,
    hue="bad", style="bad", markers=["o", "v"], alpha=0.5)
plt.xlabel("alcohol")
plt.ylabel("sugar")
plt.legend(title="bad")
plt.show()
```

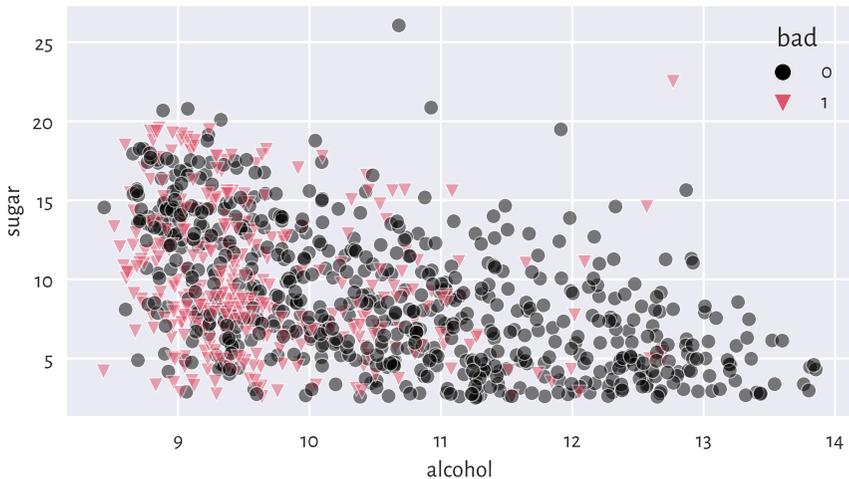

Figure 12.8: Scatterplot for sugar vs alcohol content for white, rather sweet wines, and whether they are considered bad (1) or drinkable (0) by some experts

Someone answer the door! We have a delivery: quite a few new wine bottles whose alcohol and sugar contents have, luckily, been given on their respective labels.

```
wine_test = pd.read_csv("https://raw.githubusercontent.com/gagolews/" +
        "teaching-data/master/other/sweetwhitewine_train2.csv",
        comment="#").iloc[:, :-1]
wine_test.head()
##      alcohol      sugar
## 0   10.625271   10.340159
```

```
## 1    9.066111   18.593274
## 2   10.806395    6.206685
## 3   13.432876    2.739529
## 4    9.578162    3.053025
```

We would like to determine which of the wines from the test set might be not-bad without asking an expert for their opinion. In other words, we would like to exercise a *classification* task (see, e.g., [6, 42]). More formally:

---

**Important**   Assume we are given a set of training points $\mathbf{X} \in \mathbb{R}^{n \times m}$ and the corresponding reference outputs $\boldsymbol{y} \in \{L_1, L_2, \ldots, L_l\}^n$ in the form of a categorical variable with $l$ distinct levels. The aim of a *classification* algorithm is to predict what the outputs for each point from a possibly different dataset $\mathbf{X}' \in \mathbb{R}^{n' \times m}$, i.e., $\hat{\boldsymbol{y}}' \in \{L_1, L_2, \ldots, L_l\}^{n'}$, might be.

---

In other words, we are asked to fill the gaps in a categorical variable. Recall that in a regression problem (Section 9.2), the reference outputs were numerical.

**Exercise 12.14** *Which of the following are instances of classification problems and which are regression tasks?*

- *Detect email spam.*
- *Predict a market stock price (good luck with that).*
- *Assess credit risk.*
- *Detect tumour tissues in medical images.*
- *Predict time-to-recovery of cancer patients.*
- *Recognise smiling faces on photographs (kind of creepy).*
- *Detect unattended luggage in airport security camera footage.*

*What kind of data should you gather to tackle them?*

## 12.3.1   *K*-Nearest Neighbour Classification

One of the simplest approaches to classification – good enough for such an introductory course – is based on the information about a test point's nearest neighbours living in the training sample; compare Section 8.4.4.

Fix $k \geq 1$. Namely, to classify some $\boldsymbol{x}' \in \mathbb{R}^m$:

1. Find the indexes $N_k(\boldsymbol{x}') = \{i_1, \ldots, i_k\}$ of the $k$ points from $\mathbf{X}$ closest to $\boldsymbol{x}'$, i.e., ones that fulfil for all $j \notin \{i_1, \ldots, i_k\}$:

$$\|\mathbf{x}_{i_1,\cdot} - \boldsymbol{x}'\| \leq \ldots \leq \|\mathbf{x}_{i_k,\cdot} - \boldsymbol{x}'\| \leq \|\mathbf{x}_{j,\cdot} - \boldsymbol{x}'\|.$$

2. Classify $x'$ as $\hat{y}' = \text{mode}(y_{i_1}, \ldots, y_{i_k})$, i.e., assign it the label that most frequently occurs amongst its $k$ nearest neighbours. If a mode is nonunique, resolve the ties, for example, at random.

It is thus a similar algorithm to $k$-nearest neighbour regression (Section 9.2.1). We only replaced the *quantitative* mean with the *qualitative* mode.

This is a variation on the theme: if you don't know what to do in a given situation, try to mimic what most of the other people around you are doing. Or, if you don't know what to think about a particular wine, but amongst the 5 similar ones (in terms of alcohol and sugar content) three were said to be awful, say that you don't like it because it's not sweet enough. Thanks to this, others will take you for a very refined wine taster.

Let us apply a 5-nearest neighbour classifier on the standardised version of the dataset: as we are about to use a technique based on pairwise distances, it would be best if the variables were on the same scale. Thus, we first compute the z-scores for the training set:

```
X_train = np.array(wine_train.loc[:, ["alcohol", "sugar"]])
means = np.mean(X_train, axis=0)
sds = np.std(X_train, axis=0)
Z_train = (X_train-means)/sds
```

Then, we determine the z-scores for the test set:

```
Z_test = (np.array(wine_test.loc[:, ["alcohol", "sugar"]])-means)/sds
```

Let us stress that we referred to the aggregates computed for the training set. This is a good example of a situation where we cannot simply use a built-in method from **pandas**. Instead, we apply what we have learned about **numpy**.

To make the predictions, we will use the following function:

```
def knn_class(X_test, X_train, y_train, k):
    nnis = scipy.spatial.KDTree(X_train).query(X_test, k)[1]
    nnls = y_train[nnis]  # same as: y_train[nnis.reshape(-1)].reshape(-1, k)
    return scipy.stats.mode(nnls.reshape(-1, k), axis=1)[0].reshape(-1)
```

First, we fetched the indexes of each test point's nearest neighbours (amongst the points in the training set). Then, we read their corresponding labels; they are stored in a matrix with $k$ columns. Finally, we computed the modes in each row. As a consequence, we have each point in the test set classified.

And now:

```
k = 5
y_train = np.array(wine_train.bad)
y_pred = knn_class(Z_test, Z_train, y_train, k)
```

*(continues on next page)*

```
y_pred[:10]  # preview
## array([0, 1, 0, 0, 1, 0, 0, 0, 0, 0])
```

---

**Note** Unfortunately, `scipy.stats.mode` does not resolve the possible ties at random. Nevertheless, in our case, $k$ is odd and the number of possible classes is $l = 2$. In this setting, the mode is always unique.

---

Figure 12.9 shows how nearest neighbour classification categorises different regions of a section of the two-dimensional plane. The greater the $k$, the smoother the decision boundaries. Naturally, in regions corresponding to few training points, we do not expect the classification accuracy to be good enough[9].

```
x1 = np.linspace(Z_train[:, 0].min(), Z_train[:, 0].max(), 100)
x2 = np.linspace(Z_train[:, 1].min(), Z_train[:, 1].max(), 100)
xg1, xg2 = np.meshgrid(x1, x2)
Xg12 = np.column_stack((xg1.reshape(-1), xg2.reshape(-1)))
ks = [5, 25]
for i in range(len(ks)):
    plt.subplot(1, len(ks), i+1)
    yg12 = knn_class(Xg12, Z_train, y_train, ks[i])
    plt.scatter(Z_train[y_train == 0, 0], Z_train[y_train == 0, 1],
        c="black", marker="o", alpha=0.5)
    plt.scatter(Z_train[y_train == 1, 0], Z_train[y_train == 1, 1],
        c="#DF536B", marker="v", alpha=0.5)
    plt.contourf(x1, x2, yg12.reshape(len(x2), len(x1)),
        cmap="gist_heat", alpha=0.5)
    plt.title(f"$k={ks[i]}$", fontdict=dict(fontsize=10))
    plt.xlabel("alcohol")
    if i == 0: plt.ylabel("sugar")
plt.show()
```

**Example 12.15** (*) *The same with the* `scikit-learn` *package:*

```
import sklearn.neighbors
knn = sklearn.neighbors.KNeighborsClassifier(k)
knn.fit(Z_train, y_train)
y_pred2 = knn.predict(Z_test)
```

*We can verify that the results are identical to the ones above by calling:*

---

[9] (*) As an exercise, we could implement a fixed-radius classifier; compare Section 8.4.4. In sparsely populated regions, the decision might be "unknown".

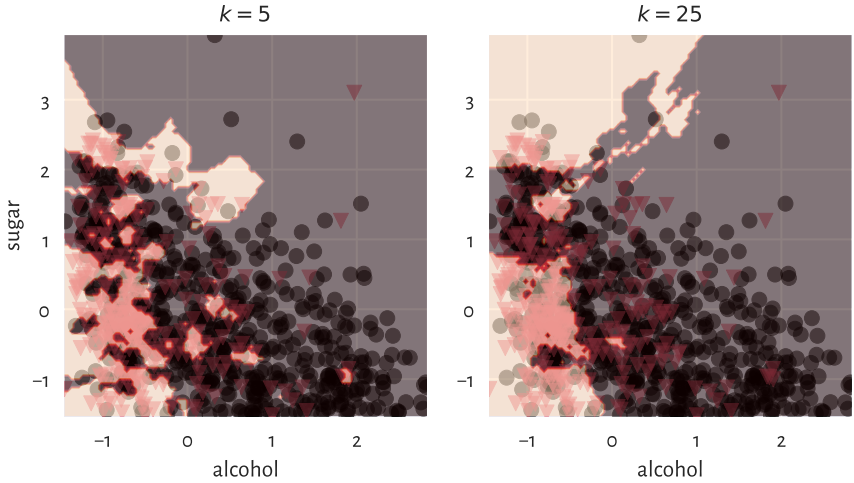

Figure 12.9: *k*-nearest neighbour classification of a whole, dense, two-dimensional grid of points for different *k*

```
np.all(y_pred2 == y_pred)
## True
```

## 12.3.2 Assessing the Quality of Predictions

It is time to reveal the truth: our test wines, it turns out, have already been assessed by some experts.

```
y_test = pd.read_csv("https://raw.githubusercontent.com/gagolews/" +
    "teaching-data/master/other/sweetwhitewine_train2.csv",
    comment="#")
y_test = np.array(y_test.bad)
y_test[:10]  # preview
## array([0, 1, 0, 0, 0, 1, 0, 0, 0, 1])
```

The *accuracy* score is the most straightforward measure of the similarity between these true labels (denoted $\mathbf{y'} = (y'_1, \dots, y'_{n'})$) and the ones predicted by the classifier (denoted $\hat{\mathbf{y}}' = (\hat{y}'_1, \dots, \hat{y}'_{n'})$). It is defined as a ratio between the correctly classified instances and all the instances:

$$\text{Accuracy}(\mathbf{y'}, \hat{\mathbf{y}}') = \frac{\sum_{i=1}^{n'} \mathbf{1}(y'_i = \hat{y}'_i)}{n'},$$

where the *indicator function* $\mathbf{1}(y'_i = \hat{y}'_i) = 1$ if and only if $y'_i = \hat{y}'_i$ and 0 otherwise.

Computing the above for our test sample gives:

```
np.mean(y_test == y_pred)
## 0.788
```

Thus, 79% of the wines were correctly classified with regard to their true quality. Before we get too enthusiastic, let us note that our dataset is slightly *imbalanced* in terms of the distribution of label counts:

```
pd.Series(y_test).value_counts()  # contingency table
## 0    639
## 1    361
## dtype: int64
```

It turns out that the majority of the wines (639 out of 1,000) in our sample are *truly* good. Notice that a dummy classifier which labels *all* the wines as great would have accuracy of ca. 64%. Our *k*-nearest neighbour approach to wine quality assessment is not that usable after all.

It is therefore always beneficial to analyse the corresponding *confusion matrix*, which is a two-way contingency table summarising the correct decisions and errors we make.

```
C = pd.DataFrame(
    dict(y_pred=y_pred, y_test=y_test)
).value_counts().unstack(fill_value=0)
C
## y_test    0    1
## y_pred
## 0       548  121
## 1        91  240
```

In the binary classification case ($l = 2$) such as this one, its entries are usually referred to as (see also the table below):

- TN – the number of cases where the true $y_i' = 0$ and the predicted $\hat{y}_i' = 0$ (true negative),
- TP – the number of instances such that the true $y_i' = 1$ and the predicted $\hat{y}_i' = 1$ (true positive),
- FN – how many times the true $y_i' = 1$ but the predicted $\hat{y}_i' = 0$ (false negative),
- FN – how many times the true $y_i' = 0$ but the predicted $\hat{y}_i' = 1$ (false positive).

The terms *positive* and *negative* refer to the output predicted by a classifier, i.e., they indicate whether some $\hat{y}_i'$ is equal to 1 and 0, respectively.

Table 12.1: The different cases of true vs predicted labels in a binary classification task ($l = 2$)

|  | $y_i' = 0$ | $y_i' = 1$ |
|---|---|---|
| $\hat{y}_i' = 0$ | **True Negative** | False Negative (Type II error) |
| $\hat{y}_i' = 1$ | False Positive (Type I error) | **True Positive** |

Ideally, the number of false positives and false negatives should be as low as possible. The accuracy score only takes the raw number of true negatives (TN) and true positives (TP) into account:

$$\text{Accuracy}(\boldsymbol{y}', \hat{\boldsymbol{y}}') = \frac{\text{TN} + \text{TP}}{\text{TN} + \text{TP} + \text{FN} + \text{FP}}.$$

Consequently, it might not be a good metric in imbalanced classification problems.

There are, fortunately, some more meaningful measures in the case where class 1 is less prevalent and where mispredicting it is considered more hazardous than making an inaccurate prediction with respect to class 0. This is because most will agree that it is better to be surprised by a vino mislabelled as bad, than be disappointed with a highly recommended product where we have already built some expectations around it. Further, not getting diagnosed as having COVID-19 where we are genuinely sick can be more dangerous for the people around us than being asked to stay at home with nothing but a headache.

*Precision* answers the question: If the classifier outputs 1, what is the probability that this is indeed true?

$$\text{Precision}(\boldsymbol{y}', \hat{\boldsymbol{y}}') = \frac{\text{TP}}{\text{TP} + \text{FP}} = \frac{\sum_{i=1}^{n'} y_i' \hat{y}_i'}{\sum_{i=1}^{n'} \hat{y}_i'}.$$

```
C = np.array(C)  # convert to matrix
C[1,1]/(C[1,1]+C[1,0])  # precision
## 0.7250755287009063
np.sum(y_test*y_pred)/np.sum(y_pred)  # equivalently
## 0.7250755287009063
```

When a classifier labels a vino as bad, in 73% of cases it is veritably undrinkable.

*Recall* (sensitivity, hit rate, or true positive rate) addresses the question: If the true class is 1, what is the probability that the classifier will detect it?

$$\text{Recall}(\boldsymbol{y}', \hat{\boldsymbol{y}}') = \frac{\text{TP}}{\text{TP} + \text{FN}} = \frac{\sum_{i=1}^{n'} y_i' \hat{y}_i'}{\sum_{i=1}^{n'} y_i'}.$$

```
C[1,1]/(C[1,1]+C[0,1])  # recall
## 0.6648199445983379
np.sum(y_test*y_pred)/np.sum(y_test)  # equivalently
## 0.6648199445983379
```

Only 66% of the really bad wines will be filtered out by the classifier.

*F-measure* (or $F_1$-measure), is the harmonic[10] mean of precision and recall in the case where we would rather have them aggregated into a single number:

$$
\mathrm{F}(\boldsymbol{y}', \hat{\boldsymbol{y}}') = \frac{1}{\frac{\frac{1}{\mathrm{Precision}} + \frac{1}{\mathrm{Recall}}}{2}} = \left( \frac{1}{2} \left( \mathrm{Precision}^{-1} + \mathrm{Recall}^{-1} \right) \right)^{-1} = \frac{\mathrm{TP}}{\mathrm{TP} + \frac{\mathrm{FP}+\mathrm{FN}}{2}}.
$$

```
C[1,1]/(C[1,1]+0.5*C[0,1]+0.5*C[1,0])  # F
## 0.6936416184971098
```

Overall, we can conclude that our classifier is rather weak.

**Exercise 12.16** *Would you use precision or recall in each of the following settings?*

- *Medical diagnosis,*
- *medical screening,*
- *suggestions of potential matches in a dating app,*
- *plagiarism detection,*
- *wine recommendation.*

### 12.3.3 Splitting into Training and Test Sets

The training set was used as a source of knowledge about our problem domain. The *k*-nearest neighbour classifier is technically *model-free*. As a consequence, to generate a new prediction, we need to be able to query all the points in the database every time.

Nonetheless, most statistical/machine learning algorithms, by construction, generalise the patterns discovered in the dataset in the form of mathematical functions (oftentimes, very complicated ones), that are fitted by minimising some error metric. Linear regression analysis by means of the least squares approximation uses exactly this kind of approach. Logistic regression for a binary response variable would be a conceptually similar classifier, but it is beyond our introductory course.

Either way, we used a separate *test set* to verify the quality of our classifier on so-far *unobserved* data, i.e., its *predictive* capabilities. We do not want our model to fit to the training data too closely. This could lead to its being completely useless when filling the gaps between the points it was exposed to. This is like being a student who can only repeat what the teacher says, and when faced with a slightly different real-world problem, they panic and say complete gibberish.

---

[10] (*) For any vector of nonnegative values, its minimum ≤ its harmonic mean ≤ its arithmetic mean.

In the above example, the training and test sets were created by yours truly. Still, normally, it is the data scientist who splits a single data frame into two parts themself; see Section 10.5.3. This way, they can *mimic* the situation where some *test* observations become available after the learning phase is complete.

### 12.3.4 Validating Many Models (Parameter Selection) (*)

In statistical modelling, there usually are many *hyperparameters* that should be tweaked. For example:

- which independent variables should be used for model building,

- how they should be preprocessed; e.g., which of them should be standardised,

- if an algorithm has some tunable parameters, what is the best combination thereof; for instance, which $k$ should we use in the $k$-nearest neighbours search.

At initial stages of data analysis, we usually tune them up by trial and error. Later, but this is already beyond the scope of this introductory course, we are used to exploring all the possible combinations thereof (exhaustive grid search) or making use of some local search-based heuristics (e.g., greedy optimisers such as hill climbing).

These always involve verifying the performance of *many* different classifiers, for example, 1-, 3-, 9, and 15-nearest neighbours-based ones. For each of them, we need to compute separate quality metrics, e.g., F-measures. Then, the classifier which yields the highest score is picked as the best. Unfortunately, if we do it recklessly, this can lead to *overfitting*, this time with respect to the test set. The obtained metrics might be too optimistic and can poorly reflect the real performance of the solution on future data.

Assuming that our dataset carries a decent number of observations, to overcome this problem, we can perform a random *training/validation/test split*:

- *training sample* (e.g., 60% of randomly chosen rows) – for model construction,

- *validation sample* (e.g., 20%) – used to tune the hyperparameters of many classifiers and to choose the best one,

- *test (hold-out) sample* (e.g., the remaining 20%) – used to assess the goodness of fit of the best classifier.

This common-sense approach is not limited to classification. We can validate different regression models in the same way.

**Important** We would like to obtain a good estimate of a classifier's performance on previously unobserved data. For this reason, the test (hold-out) sample must neither be used in the training nor the validation phase.

---

**Exercise 12.17**  *Determine the best parameter setting for the k-nearest neighbour classification of the `color` variable based on standardised versions of some physicochemical features (chosen columns) of wines in the `wine_quality_all`[11] dataset. Create a 60/20/20% dataset split. For each $k = 1, 3, 5, 7, 9$, compute the corresponding F-measure on the validation test. Evaluate the quality of the best classifier on the test set.*

---

**Note**  (*) Instead of a training/validation/test split, we can use various *cross-validation* techniques, especially on smaller datasets. For instance, in a *5-fold cross-validation*, we split the original training set randomly into five disjoint parts: $A, B, C, D, E$ (more or less of the same size). We use each combination of four chunks as training sets and the remaining part as the validation set, for which we generate the predictions and then compute, say, the F-measure:

| training set | validation set | F-measure |
|---|---|---|
| $B \cup C \cup D \cup E$ | $A$ | $F_A$ |
| $A \cup C \cup D \cup E$ | $B$ | $F_B$ |
| $A \cup B \cup D \cup E$ | $C$ | $F_C$ |
| $A \cup B \cup C \cup E$ | $D$ | $F_D$ |
| $A \cup B \cup C \cup D$ | $E$ | $F_E$ |

In the end, we can determine the average F-measure, $(F_A + F_B + F_C + F_D + F_E)/5$, as a basis for assessing different classifiers' quality.

Once the best classifier is chosen, we can use the whole training sample to fit the final model and then consider the separate test sample to assess its quality.

Furthermore, for highly imbalanced labels, some form of stratified sampling might be necessary. Such problems are typically explored in more advanced courses in statistical learning.

---

**Exercise 12.18**  (**) *Redo the above exercise (assessing the wine colour classifiers), but this time maximise the F-measure obtained by a 5-fold cross-validation.*

---

## 12.4  Clustering Tasks

So far, we have been implicitly assuming that either each dataset comes from a single homogeneous distribution, or we have a categorical variable that naturally defines the groups that we can split the dataset into. Nevertheless, it might be the case that we are given a sample coming from a distribution mixture, where some subsets behave differently, but a grouping variable has not been provided at all (e.g., we have height and weight data but no information about the subjects' sexes).

---

[11] https://github.com/gagolews/teaching-data/raw/master/other/wine_quality_all.csv

*Clustering* (also known as segmentation or quantisation; see, e.g., [86]) methods can be used to partition a dataset into groups based only on the spatial structure of the points' relative densities. In the *k*-means method, which we discuss below, the cluster structure is determined based on the points' proximity to *k* carefully chosen group centroids; compare Section 8.4.2.

### 12.4.1 *K*-Means Method

Fix $k \geq 2$. In the *k*-means method[12], we seek $k$ pivot points, $c_1, c_2, \ldots, c_k \in \mathbb{R}^m$, such that the sum of squared distances between the input points in $\mathbf{X} \in \mathbb{R}^{n \times m}$ and their closest pivots is minimised:

$$\text{minimise} \sum_{i=1}^{n} \min \left\{ \|\mathbf{x}_{i,\cdot} - c_1\|^2, \|\mathbf{x}_{i,\cdot} - c_2\|^2, \ldots, \|\mathbf{x}_{i,\cdot} - c_k\|^2 \right\} \qquad \text{w.r.t. } c_1, c_2, \ldots, c_k.$$

Let us introduce the *label vector $l$* such that:

$$l_i = \arg\min_{j} \|\mathbf{x}_{i,\cdot} - c_j\|^2,$$

i.e., it is the index of the pivot closest to $\mathbf{x}_{i,\cdot}$.

We will consider all the points $\mathbf{x}_{i,\cdot}$ with $i$ such that $l_i = j$ as belonging to the same, $j$-th, *cluster* (point group). This way $l$ defines a *partition* of the original dataset into $k$ nonempty, mutually disjoint subsets.

Now, the above optimisation task can be equivalently rewritten as:

$$\text{minimise} \sum_{i=1}^{n} \|\mathbf{x}_{i,\cdot} - c_{l_i}\|^2 \qquad \text{w.r.t. } c_1, c_2, \ldots, c_k.$$

And this is why we refer to the above objective function as the (total) *within-cluster sum of squares* (WCSS). This problem looks easier, but let us not be tricked; $l_i$s depend on $c_j$s. They vary *together*. We have just made it less explicit.

It can be shown that given a fixed label vector $l$ representing a partition, $c_j$ must be the centroid (Section 8.4.2) of the points assigned thereto:

$$c_j = \frac{1}{n_j} \sum_{i:l_i=j} \mathbf{x}_{i,\cdot},$$

---

[12] We do not have to denote the number of clusters with $k$: we could be speaking about the 2-means, 3-means, $l$-means, or $\ddot{u}$-means method too. Nevertheless, some mainstream practitioners consider $k$-means as a kind of a brand name, let us thus refrain from adding to their confusion. Interestingly, another widely known algorithm is called fuzzy (weighted) $c$-means [4].

where $n_j = |\{i : l_i = j\}|$ gives the number of $i$s such that $l_i = j$, i.e., the size of the $j$-th cluster.

Here is an example dataset (see below for a scatterplot):

```
X = np.loadtxt("https://raw.githubusercontent.com/gagolews/" +
    "teaching-data/master/marek/blobs1.txt", delimiter=",")
```

We can call **scipy.cluster.vq.kmeans2** to find $k = 2$ clusters:

```
import scipy.cluster.vq
C, l = scipy.cluster.vq.kmeans2(X, 2)
```

The discovered cluster centres are stored in a matrix with $k$ rows and $m$ columns, i.e., the $j$-th row gives $\mathbf{c}_j$.

```
C
## array([[ 0.99622971,  1.052801  ],
##        [-0.90041365, -1.08411794]])
```

The label vector is:

```
l
## array([1, 1, 1, ..., 0, 0, 0], dtype=int32)
```

As usual in Python, indexing starts at 0. So for $k = 2$ we only obtain the labels 0 and 1.

Figure 12.10 depicts the two clusters together with the cluster centroids. We use `l` as a colour selector in `my_colours[l]` (this is a clever instance of the integer vector-based indexing). It seems that we correctly discovered the very natural partitioning of this dataset into two clusters.

```
plt.scatter(X[:, 0], X[:, 1], c=np.array(["black", "#DF536B"])[l])
plt.plot(C[:, 0], C[:, 1], "yX")
plt.axis("equal")
plt.show()
```

Here are the cluster sizes:

```
np.bincount(l)  # or, e.g., pd.Series(l).value_counts()
## array([1017, 1039])
```

The label vector `l` can be added as a new column in the dataset. Here is a preview:

```
Xl = pd.DataFrame(dict(X1=X[:, 0], X2=X[:, 1], l=l))
Xl.sample(5, random_state=42)  # some randomly chosen rows
##                X1        X2  l
```

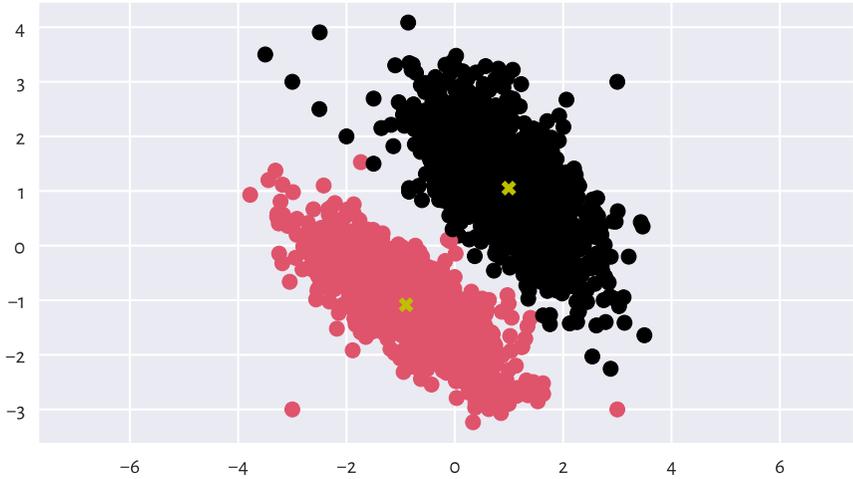

Figure 12.10: Two clusters discovered by the *k*-means method; cluster centroids are marked separately

```
## 184   -0.973736  -0.417269  1
## 1724   1.432034   1.392533  0
## 251   -2.407422  -0.302862  1
## 1121   2.158669  -0.000564  0
## 1486   2.060772   2.672565  0
```

We can now enjoy all the techniques for processing data in groups that we have discussed so far. In particular, computing the columnwise means gives nothing else than the above cluster centroids:

```
X1.groupby("l").mean()
##           X1         X2
## l
## 0   0.996230   1.052801
## 1  -0.900414  -1.084118
```

The label vector `l` can be recreated by computing the distances between all the points and the centroids and then picking the indexes of the closest pivots:

```
l_test = np.argmin(scipy.spatial.distance.cdist(X, C), axis=1)
np.all(l_test == l)   # verify they are identical
## True
```

---

**Important**  By construction[13], the *k*-means method can only detect clusters of convex shapes (such as Gaussian blobs).

---

**Exercise 12.19**  *Perform the clustering of the* `wut_isolation`[14] *dataset and notice how non-sensical, geometrically speaking, the returned clusters are.*

**Exercise 12.20**  *Determine a clustering of the* `wut_twosplashes`[15] *dataset and display the results on a scatterplot. Compare them with those obtained on the standardised version of the dataset. Recall what we said about the Euclidean distance and its perception being disturbed when a plot's aspect ratio is not 1:1.*

---

**Note**    (*) An even simpler classifier than the *k*-nearest neighbours one described above builds upon the concept of the nearest centroids. Namely, it first determines the centroids (componentwise arithmetic means) of the points in each class. Then, a new point (from the test set) is assigned to the class whose centroid is the closest thereto. The implementation of such a classifier is left as a rather straightforward exercise to the reader. As an application, we recommend using it to extrapolate the results generated by the *k*-means method (for different *k*s) to previously unobserved data, e.g., all points on a dense equidistant grid.

---

### 12.4.2    Solving *K*-means Is Hard

Unfortunately, the *k*-means method – the identification of label vectors/cluster centres that minimise the total within-cluster sum of squares – relies on solving a computationally hard combinatorial optimisation problem (e.g., [52]). In other words, the search for the *truly* (i.e., globally) optimal solution takes, for larger *n* and *k*, an impractically long time.

As a consequence, we must rely on some approximate algorithms which all have one drawback in common. Namely, whatever they return be *suboptimal*. Hence, they can constitute a possibly meaningless solution.

The documentation of `scipy.cluster.vq.kmeans2` is of course honest about it. It states that the method *attempts to minimise the Euclidean distance between observations and centroids*. Further, `sklearn.cluster.KMeans`, implementing a similar algorithm, mentions that the procedure *is very fast [...], but it falls in local minima. That is why it can be useful to restart it several times.*

To understand what it all means, it will be very educational to study this issue in more detail. This is because the discussed approach to clustering is not the only hard problem in data science (selecting an optimal set of independent variables with respect to AIC or BIC in linear regression is another example).

---

[13] (*) And its relation to Voronoi diagrams.
[14] https://github.com/gagolews/teaching-data/raw/master/clustering/wut_isolation.csv
[15] https://github.com/gagolews/teaching-data/raw/master/clustering/wut_twosplashes.csv

### 12.4.3    Lloyd's Algorithm

Technically, there is no such thing as *the k-*means *algorithm*. There are many procedures, based on numerous different heuristics, that attempt to solve the *k-*means *problem*. Unfortunately, neither of them is perfect. This is not possible.

Perhaps the most widely known and easiest to understand method is traditionally attributed to Lloyd [55]. It is based on the fixed-point iteration and. For a given $\mathbf{X} \in \mathbb{R}^{n \times m}$ and $k \geq 2$:

1. Pick initial cluster centres $c_1, \dots, c_k$, for example, randomly.

2. For each point in the dataset, $\mathbf{x}_{i,\cdot}$, determine the index of its closest centre $l_i$:

$$l_i = \arg\min_j \|\mathbf{x}_{i,\cdot} - c_j\|^2.$$

3. Compute the centroids of the clusters defined by the label vector $l$, i.e., for every $j = 1, 2, \dots, k$:

$$c_j = \frac{1}{n_j} \sum_{i:l_i=j} \mathbf{x}_{i,\cdot},$$

where $n_j = |\{i : l_i = j\}|$ gives the size of the *j*-th cluster.

4. If the objective function (total within-cluster sum of squares) has not changed significantly since the last iteration (say, the absolute value of the difference between the last and the current loss is less than $10^{-9}$), then stop and return the current $c_1, \dots, c_k$ as the result. Otherwise, go to Step 2.

**Exercise 12.21** (*) *Implement the Lloyd algorithm in the form of a function* `kmeans(X, C)`, *where* `X` *is the data matrix* (n-by-m) *and where the rows in* `C`, *being a* k-by-m *matrix, give the initial cluster centres.*

### 12.4.4    Local Minima

The way the above algorithm is constructed implies what follows.

---

**Important**   Lloyd's method guarantees that the centres $c_1, \dots, c_k$ it returns cannot be significantly improved any further by repeating Steps 2 and 3 of the algorithm. Still, it does not necessarily mean that they yield the *globally* optimal (the best possible) WCSS. We might as well get stuck in a *local* minimum, where there is no better positioning thereof in the *neighbourhoods* of the current cluster centres; compare Figure 12.11. Yet, had we looked beyond them, we could have found a superior solution.

---

A variant of the Lloyd method is implemented in `scipy.cluster.vq.kmeans2`, where the initial cluster centres are picked at random. Let us test its behaviour by analysing three chosen country-wise categories from the 2016 Sustainable Society Indices[16] dataset.

---

[16] https://ssi.wi.th-koeln.de/

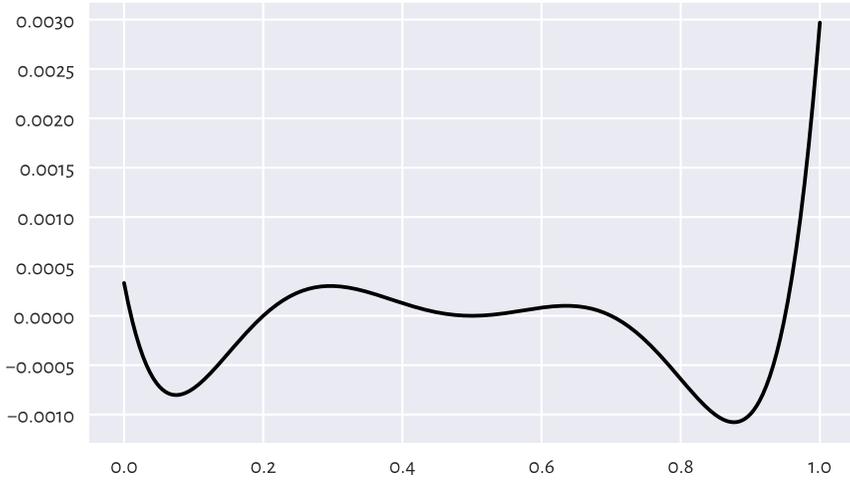

Figure 12.11: An example function (of only one variable; our problem is much higher-dimensional) with many local minima; how can we be sure there is no better minimum outside of the depicted interval?

```
ssi = pd.read_csv("https://raw.githubusercontent.com/gagolews/" +
    "teaching-data/master/marek/ssi_2016_categories.csv",
    comment="#")
X = ssi.set_index("Country").loc[:,
    ["PersonalDevelopmentAndHealth", "WellBalancedSociety", "Economy"]
].rename({
        "PersonalDevelopmentAndHealth": "Health",
        "WellBalancedSociety": "Balance",
        "Economy": "Economy"
    }, axis=1)  # rename columns
n = X.shape[0]
X.loc[["Australia", "Germany", "Poland", "United States"], :]  # preview
##                  Health   Balance   Economy
## Country
## Australia       8.590927  6.105539  7.593052
## Germany         8.629024  8.036620  5.575906
## Poland          8.265950  7.331700  5.989513
## United States   8.357395  5.069076  3.756943
```

It is a three-dimensional dataset, where each point (row) corresponds to a different country. Let us find a partition into $k = 3$ clusters.

```
k = 3
```

```python
np.random.seed(123)  # reproducibility matters
C1, l1 = scipy.cluster.vq.kmeans2(X, k)
C1
## array([[7.99945084, 6.50033648, 4.36537659],
##        [7.6370645 , 4.54396676, 6.89893746],
##        [6.24317074, 3.17968018, 3.60779268]])
```

The objective function (total within-cluster sum of squares) at the returned cluster centres is equal to:

```python
import scipy.spatial.distance
def get_wcss(X, C):
    D = scipy.spatial.distance.cdist(X, C)**2
    return np.sum(np.min(D, axis=1))

get_wcss(X, C1)
## 446.5221283436733
```

Is it good or not necessarily? We are unable to tell. What we can do, however, is to run the algorithm again, this time from a different starting point:

```python
np.random.seed(1234)  # different seed - different initial centres
C2, l2 = scipy.cluster.vq.kmeans2(X, k)
C2
## array([[7.80779013, 5.19409177, 6.97790733],
##        [6.31794579, 3.12048584, 3.84519706],
##        [7.92606993, 6.35691349, 3.91202972]])
get_wcss(X, C2)
## 437.51120966832775
```

It is a better solution (we are lucky; it might as well have been worse). But is it the best possible? Again, we cannot tell, alone in the dark.

Does a potential suboptimality affect the way the data points are grouped? It is indeed the case here. Let us look at the contingency table for the two label vectors:

```python
pd.DataFrame(dict(l1=l1, l2=l2)).value_counts().unstack(fill_value=0)
## l2    0    1    2
## l1
## 0     8    0   43
## 1    39    6    0
## 2     0   57    1
```

---

**Important** Clusters are essentially unordered. The label vector $(1, 1, 2, 2, 1, 3)$ represents the same clustering as the label vectors $(3, 3, 2, 2, 3, 1)$ and $(2, 2, 3, 3, 2, 1)$.

---

By looking at the contingency table, we see that clusters 0, 1, and 2 in l1 correspond, respectively, to clusters 2, 0, and 1 in l2 (via a kind of majority voting). We can relabel the elements in l1 to get a more readable result:

```
l1p = np.array([2, 0, 1])[l1]
pd.DataFrame(dict(l1p=l1p, l2=l2)).value_counts().unstack(fill_value=0)
## l2    0    1    2
## l1p
## 0    39    6    0
## 1     0   57    1
## 2     8    0   43
```

Much better. It turns out that 8+6+1 countries are categorised differently. We would definitely not want to initiate any diplomatic crisis because of our not knowing that the above algorithm might return suboptimal solutions.

**Exercise 12.22** (*) *Determine which countries are affected.*

### 12.4.5 Random Restarts

There will never be any guarantees, but we can increase the probability of generating a satisfactory solution by simply restarting the method multiple times from many randomly chosen points and picking the best[17] solution (the one with the smallest WCSS) identified as the result.

Let us make 1,000 such *restarts*:

```
wcss, Cs = [], []
for i in range(1000):
    C, l = scipy.cluster.vq.kmeans2(X, k, seed=i)
    Cs.append(C)
    wcss.append(get_wcss(X, C))
```

The best of the local minima (no guarantee that it is the global one, again) is:

```
np.min(wcss)
## 437.5112096832775
```

It corresponds to the cluster centres:

---

[17] If we have many different heuristics, each aiming to approximate a solution to the $k$-means problem, from the practical point of view it does not really matter which one returns the best solution – they are merely our tools to achieve a higher goal. Ideally, we should run all of them many times and get the result that corresponds to the smallest WCSS. It is crucial to *do our best* to find the optimal set of cluster centres – the more approaches we test, the better the chance of success.

```
Cs[np.argmin(wcss)]
## array([[7.80779013, 5.19409177, 6.97790733],
##         [7.92606993, 6.35691349, 3.91202972],
##         [6.31794579, 3.12048584, 3.84519706]])
```

They are the same as `C2` above (up to a permutation of labels). We were lucky[18], after all.

It is very educational to look at the distribution of the objective function at the identified local minima to see that, proportionally, in the case of this dataset it is not rare to end up in a quite bad solution; see Figure 12.12.

```
plt.hist(wcss, bins=100)
plt.show()
```

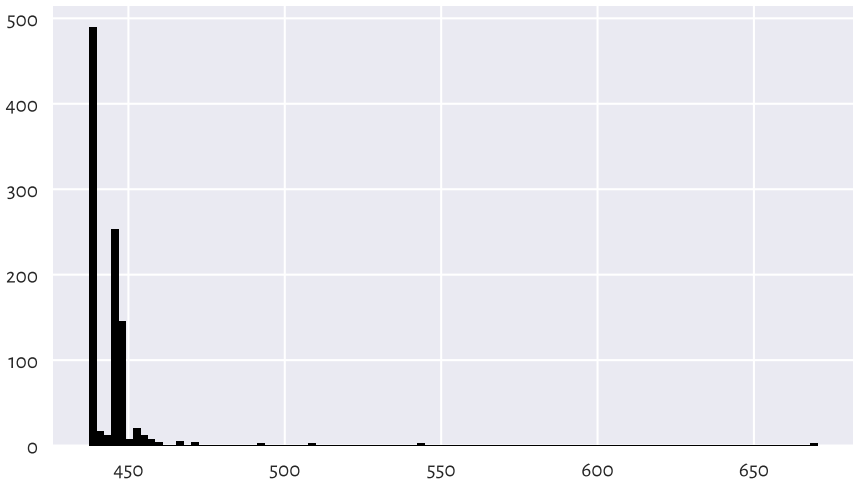

Figure 12.12: Within-cluster sum of squares at the results returned by different runs of the *k*-means algorithm; sometimes we might be very unlucky

Also, Figure 12.13 depicts all the cluster centres to which the algorithm converged. We see that we should not be trusting the results generated by a single run of a heuristic solver to the *k*-means problem.

**Example 12.23** (\*) *The* `scikit-learn` *package implements an algorithm that is similar to the Lloyd's one. The method is equipped with the* `n_init` *parameter (which defaults to 10) which automatically applies the aforementioned restarting.*

---

[18] Mind who is the benevolent dictator of the pseudorandom number generator's seed.

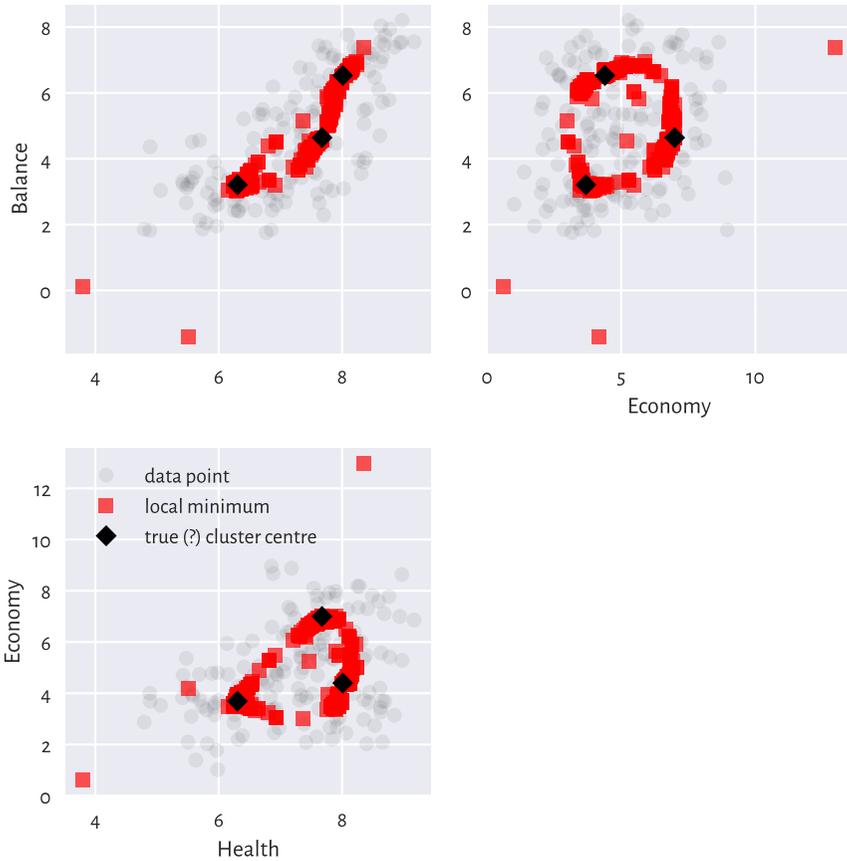

Figure 12.13: Traces of different cluster centres our k-means algorithm converged to; some are definitely not optimal, and therefore the method must be restarted a few times to increase the likelihood of pinpointing the true solution

```
import sklearn.cluster
np.random.seed(123)
km = sklearn.cluster.KMeans(k)  # KMeans(k, n_init=10)
km.fit(X)
## KMeans(n_clusters=3)
km.inertia_  # WCSS - not optimal!
## 437.5467188958928
```

Still, there are no guarantees: the solution is suboptimal too. As an exercise, pass *n_init=100*, *n_init=1000*, and *n_init=10000* and determine the returned WCSS.

**Note** It is theoretically possible that a developer from the **scikit-learn** team, when they see the above result, will make a tweak in the algorithm so that after an update to the package, the returned minimum will be better. This cannot be deemed a bug fix, though, as there are no bugs here. Improving the behaviour of the method in this example will lead to its degradation in others. There is no free lunch in optimisation.

**Note** Some datasets are more well-behaving than others. The $k$-means method is *overall* quite usable, but we must always be cautious.

We recommend always performing at least 100 random restarts. Also, if a report from data analysis does not say anything about the number of tries performed, we should assume that the results are gibberish[19]. People will complain about our being a pain, but we know better; compare Rule#9.

**Exercise 12.24** *Run the k-means method, $k = 8$, on the* sipu_unbalance[20] *dataset from many random sets of cluster centres. Note the value of the total within-cluster sum of squares. Also, plot the cluster centres discovered. Do they make sense? Compare these to the case where you start the method from the following cluster centres which are close to the global minimum.*

$$\mathbf{C} = \begin{bmatrix} -15 & 5 \\ -12 & 10 \\ -10 & 5 \\ 15 & 0 \\ 15 & 10 \\ 20 & 5 \\ 25 & 0 \\ 25 & 10 \end{bmatrix}.$$

## 12.5 Further Reading

An overall good introduction to classification is [42] and [6]. Nevertheless, as we said earlier, we recommend going through a solid course in matrix algebra and mathematical statistics first, e.g., [18, 34] and [19, 33, 35]. For advanced theoretical (probabilistic, information-theoretic) results, see, e.g., [7, 20].

Hierarchical clustering algorithms (see, e.g., [28, 58]) are also noteworthy, because they do not require asking for a fixed number of clusters. Furthermore, density-based algorithms (DBSCAN and its variants) [10, 22, 53] utilise the notion of fixed-radius search that we have introduced in Section 8.4.4.

---

[19] For instance, R's **stats::kmeans** automatically uses nstart=1. It is not rare, unfortunately, that data analysts only stick with the default arguments.

[20] https://github.com/gagolews/teaching-data/raw/master/clustering/sipu_unbalance.csv

There are quite a few ways that aim to assess the quality of clustering results, but their meaningfulness is somewhat limited; see [31] for discussion.

## 12.6   Exercises

**Exercise 12.25**   *Name the data type of the objects that the* `DataFrame.groupby` *method returns.*

**Exercise 12.26**   *What is the relationship between the* `GroupBy`, `DataFrameGroupBy`, *and* `SeriesGroupBy` *classes?*

**Exercise 12.27**   *What are relative z-scores and how can we compute them?*

**Exercise 12.28**   *Why and when the accuracy score might not be the best way to quantify a classifier's performance?*

**Exercise 12.29**   *What is the difference between recall and precision, both in terms of how they are defined and where they are the most useful?*

**Exercise 12.30**   *Explain how the k-nearest neighbour classification and regression algorithms work. Why do we say that they are model-free?*

**Exercise 12.31**   *In the context of k-nearest neighbour classification, why it might be important to resolve the potential ties at random when computing the mode of the neighbours' labels?*

**Exercise 12.32**   *What is the purpose of a training/test and a training/validation/test set split?*

**Exercise 12.33**   *Give the formula for the total within-cluster sum of squares.*

**Exercise 12.34**   *Are there any cluster shapes that cannot be detected by the k-means method?*

**Exercise 12.35**   *Why do we say that solving the k-means problem is hard?*

**Exercise 12.36**   *Why restarting Lloyd's algorithm many times is necessary? Why are reports from data analysis that do not mention the number of restarts not trustworthy?*

# 13

## *Accessing Databases*

**pandas** is convenient for working with data that fit into memory and which can be stored in individual CSV files. Still, larger information banks in a shared environment will often be made available to us via relational (structured) databases such as PostgreSQL or MariaDB, or a wide range of commercial products.

Most commonly, we use SQL (Structured Query Language) to define the data chunks[1] we wish to analyse. Then, we fetch them from the database driver in the form of a **pandas** data frame. This enables us to perform the operations we are already familiar with, e.g., various transformations or visualisations.

Below we make a quick introduction to the basics of SQL using SQLite[2], which is a lightweight, flat-file, and server-less open-source database management system. Overall, SQLite is a sensible choice for data of even hundreds or thousands of gigabytes in size that fit on a single computer's disk. This is more than enough for playing with our data science projects or prototyping more complex solutions.

---

**Important**   In this chapter, we will learn that the syntax of SQL is very readable: it is modelled after the natural (English) language. The purpose of this introduction is not to write own queries nor to design own databanks: this should be covered by a separate course on database systems; see, e.g., [13, 17].

---

## 13.1   Example Database

In this chapter, we will be working with a simplified data dump of the Q&A site Travel Stack Exchange[3], which we downloaded[4] on 2017-10-31. It consists of five separate data frames.

---

[1] Technically, there are ways to use **pandas** with data that do not fit into memory. Still, SQL is usually a more versatile choice. If we have too much data, we can always fetch random samples (this is what statistics is for) thereof or pre-aggregate the information on the server size. This should be sufficient for most intermediate-level users.

[2] https://sqlite.org

[3] https://travel.stackexchange.com

[4] https://archive.org/details/stackexchange

First, `Tags` gives, amongst others, topic categories (`TagName`) and how many questions mention them (`Count`):

```
Tags = pd.read_csv("https://raw.githubusercontent.com/gagolews/" +
    "teaching-data/master/travel_stackexchange_com_2017/Tags.csv.gz",
    comment="#")
Tags.head(3)
##    Count  ExcerptPostId  Id    TagName  WikiPostId
## 0    104         2138.0   1   cruising      2137.0
## 1     43          357.0   2  caribbean       356.0
## 2     43          319.0   4  vacations       318.0
```

Second, `Users` provides information on the registered users.

```
Users = pd.read_csv("https://raw.githubusercontent.com/gagolews/" +
    "teaching-data/master/travel_stackexchange_com_2017/Users.csv.gz",
    comment="#")
Users.head(3)
##    AccountId   Age              CreationDate  ... Reputation  UpVotes  Views
## 0       -1.0   NaN  2011-06-21T15:16:44.253  ...        1.0   2472.0    0.0
## 1        2.0  40.0  2011-06-21T20:10:03.720  ...      101.0      1.0   31.0
## 2     7598.0  32.0  2011-06-21T20:11:02.490  ...      101.0      1.0   14.0
##
## [3 rows x 11 columns]
```

Third, `Badges` recalls all rewards handed to the users (`UserId`) for their engaging in various praiseworthy activities:

```
Badges = pd.read_csv("https://raw.githubusercontent.com/gagolews/" +
    "teaching-data/master/travel_stackexchange_com_2017/Badges.csv.gz",
    comment="#")
Badges.head(3)
##    Class                     Date  Id           Name  TagBased  UserId
## 0      3  2011-06-21T20:16:48.910   1  Autobiographer     False       2
## 1      3  2011-06-21T20:16:48.910   2  Autobiographer     False       3
## 2      3  2011-06-21T20:16:48.910   3  Autobiographer     False       4
```

Fourth, `Posts` lists all the questions and answers (the latter do not have `ParentId` set to `NaN`).

```
Posts = pd.read_csv("https://raw.githubusercontent.com/gagolews/" +
    "teaching-data/master/travel_stackexchange_com_2017/Posts.csv.gz",
    comment="#")
Posts.head(3)
##    AcceptedAnswerId  ...  ViewCount
## 0             393.0  ...      419.0
```

*(continued from previous page)*

```
## 1                NaN  ...      1399.0
## 2                NaN  ...        NaN
##
## [3 rows x 17 columns]
```

Fifth, `Votes` list all the up-votes (`VoteTypeId` equal to 2) and down-votes (`VoteTypeId` of 3) to all the posts.

```
Votes = pd.read_csv("https://raw.githubusercontent.com/gagolews/" +
    "teaching-data/master/travel_stackexchange_com_2017/Votes.csv.gz",
    comment="#")
Votes.head(3)
##    BountyAmount             CreationDate  Id  PostId  UserId  VoteTypeId
## 0          NaN  2011-06-21T00:00:00.000   1       1     NaN           2
## 1          NaN  2011-06-21T00:00:00.000   2       1     NaN           2
## 2          NaN  2011-06-21T00:00:00.000   3       2     NaN           2
```

**Exercise 13.1** *See the README[5] file for a detailed description of each column. Note that are rows are uniquely defined by their respective* `Id`*s. They are relations between the data frames, e.g.,* `Users.Id` *vs* `Badges.UserId`*,* `Posts.Id` *vs* `Votes.PostId`*, etc. Moreover, for privacy reasons, some* `UserId`*s might be missing. In such a case, they are encoded with a not-a-number; compare Chapter 15.*

## 13.2 Exporting Data to a Database

Let us establish a connection with the to-be SQLite database. In our case, this will be an ordinary file stored on the computer's disk:

```
import tempfile, os.path
dbfile = os.path.join(tempfile.mkdtemp(), "travel.db")
print(dbfile)
## /tmp/tmpgo9f11ht/travel.db
```

The above defines the file path (compare Section 13.6.1) where the database is going to be stored. We use a randomly generated filename inside the local file system's (we are on Linux) temporary directory, `/tmp`. This is just a pleasant exercise, and we will not be using this database afterwards. The reader might prefer setting a filename relative to the current working directory (as given by **os.getcwd**), e.g., `dbfile = "travel.db"`.

We can now connect to the database:

```
import sqlite3
conn = sqlite3.connect(dbfile)
```

The database might now be queried: we can add new tables, insert new rows, and re-trieve records.

---

**Important**  In the end, we should remember to call `conn.close()`.

---

Our data are already in the form of **pandas** data frames. Therefore, exporting them to the database is straightforward. We only need to make a series of calls to the **pandas.DataFrame.to_sql** method.

```
Tags.to_sql("Tags", conn, index=False)
Users.to_sql("Users", conn, index=False)
Badges.to_sql("Badges", conn, index=False)
Posts.to_sql("Posts", conn, index=False)
Votes.to_sql("Votes", conn, index=False)
```

---

**Note**  (*) It is possible to export data that do not fit into memory by reading them in chunks of considerable, but not too large, sizes. In particular **pandas.read_csv** has the `nrows` argument that lets us read several rows from a file connection; see Section 13.6.4. Then, **pandas.DataFrame.to_sql(..., if_exists="append")** can be used to append new rows to an existing table.

Exporting data can of course be done without **pandas** as well, e.g., when they are to be fetched from XML or JSON files (compare Section 13.5) and processed manually, row by row. Intermediate-level SQL users can call `conn.execute("CREATE TABLE t...")`, followed by `conn.executemany("INSERT INTO t VALUES(?, ?, ?)", l)`, and then `conn.commit()`. This will create a new table (here: named t) populated by a list of records (e.g., in the form of tuples or **numpy** vectors). For more details, see the manual[6] of the **sqlite3** package.

---

## 13.3  Exercises on SQL vs pandas

We can use **pandas** to fetch the results of any SQL query in the form of a data frame. For example:

---

[6] https://docs.python.org/3/library/sqlite3.html

```
pd.read_sql_query("""
    SELECT * FROM Tags LIMIT 3
""", conn)
##     Count  ExcerptPostId  Id    TagName  WikiPostId
## 0    104         2138.0   1   cruising      2137.0
## 1     43          357.0   2  caribbean       356.0
## 2     43          319.0   4  vacations       318.0
```

The above query selected all columns (`SELECT *`) and the first three rows (`LIMIT 3`) from the `Tags` table.

**Exercise 13.2** *For the above and all the following SQL queries, write the equivalent Python code that generates the same result using **pandas** functions and methods. In each case, there might be more than one equally fine solution. In case of any doubt about the meaning of the queries, please refer to the SQLite documentation[7]. Example solutions are provided at the end of this section.*

**Example 13.3** *For a reference query:*

```
res1a = pd.read_sql_query("""
    SELECT * FROM Tags LIMIT 3
""", conn)
```

*The equivalent **pandas** implementation might look like:*

```
res1b = Tags.head(3)
```

*To verify that the results are equal, we can call:*

```
pd.testing.assert_frame_equal(res1a, res1b)  # no error == OK
```

*No error message means that the test is passed. The cordial thing about the **assert_frame_equal** function is that it ignores small round-off errors introduced by arithmetic operations.*

*Nonetheless, the results generated by **pandas** might be the same up to the reordering of rows. In such a case, before calling **pandas.testing.assert_frame_equal**, we can invoke **DataFrame.sort_values** on both data frames to sort them with respect to 1 or 2 chosen columns.*

### 13.3.1 Filtering

**Exercise 13.4** *From `Tags`, select two columns `TagName` and `Count` and rows for which `TagName` is equal to one of the three choices provided.*

```
res2a = pd.read_sql_query("""
    SELECT TagName, Count
    FROM Tags
```

*(continues on next page)*

---

[7] https://sqlite.org/lang.html

```
    WHERE TagName IN ('poland', 'australia', 'china')
""", conn)
res2a
##       TagName  Count
## 0      china    443
## 1  australia    411
## 2     poland    139
```

Hint: use **pandas.Series.isin**.

**Exercise 13.5** *Select a set of columns from `Posts` whose rows fulfil a given set of conditions.*

```
res3a = pd.read_sql_query("""
    SELECT Title, Score, ViewCount, FavoriteCount
    FROM Posts
    WHERE PostTypeId=1 AND
        ViewCount>=10000 AND
        FavoriteCount BETWEEN 35 AND 100
""", conn)
res3a
##                                              Title  ...  FavoriteCount
## 0  When traveling to a country with a different c...  ...          35.0
## 1         How can I do a "broad" search for flights?  ...          49.0
## 2  Tactics to avoid getting harassed by corrupt p...  ...          42.0
## 3  Flight tickets: buy two weeks before even duri...  ...          36.0
## 4  OK we're all adults here, so really, how on ea...  ...          79.0
## 5  How to intentionally get denied entry to the U...  ...          53.0
## 6  How do you know if Americans genuinely/literal...  ...          79.0
## 7  OK, we are all adults here, so what is a bidet...  ...          38.0
## 8          How to cope with too slow Wi-Fi at hotel?  ...          41.0
##
## [9 rows x 4 columns]
```

### 13.3.2 Ordering

**Exercise 13.6** *Select the `Title` and `Score` columns from `Posts` where `ParentId` is missing (i.e., the post is in fact a question) and `Title` is well-defined. Then, sort the results by the `Score` column, decreasingly (descending order). Finally, return only the first five rows (e.g., top five scoring questions).*

```
res4a = pd.read_sql_query("""
    SELECT Title, Score
    FROM Posts
    WHERE ParentId IS NULL AND Title IS NOT NULL
    ORDER BY Score DESC
```

```
    LIMIT 5
""", conn)
res4a
##                                          Title  Score
## 0  OK we're all adults here, so really, how on ea...    306
## 1  How do you know if Americans genuinely/literal...    254
## 2  How to intentionally get denied entry to the U...    219
## 3  Why are airline passengers asked to lift up wi...    210
## 4                        Why prohibit engine braking?    178
```

*Hint: use **pandas.DataFrame.sort_values** and **numpy.isnan** or **pandas.isnull**.*

### 13.3.3 Removing Duplicates

**Exercise 13.7** *Get all unique badge names for the user with Id=23.*

```
res5a = pd.read_sql_query("""
    SELECT DISTINCT Name
    FROM Badges
    WHERE UserId=23
""", conn)
res5a
##                 Name
## 0          Supporter
## 1            Student
## 2            Teacher
## 3            Scholar
## 4               Beta
## 5      Nice Question
## 6             Editor
## 7        Nice Answer
## 8           Yearling
## 9    Popular Question
## 10         Taxonomist
## 11   Notable Question
```

*Hint: use **pandas.DataFrame.drop_duplicates**.*

**Exercise 13.8** *For each badge handed to the user with Id=23, extract the award year store it in a new column named Year. Then, select only the unique pairs (Name, Year).*

```
res6a = pd.read_sql_query("""
    SELECT DISTINCT
        Name,
        CAST(strftime('%Y', Date) AS FLOAT) AS Year
```

```
    FROM Badges
    WHERE UserId=23
""", conn)
res6a
##                    Name    Year
## 0             Supporter  2011.0
## 1               Student  2011.0
## 2               Teacher  2011.0
## 3               Scholar  2011.0
## 4                  Beta  2011.0
## 5         Nice Question  2011.0
## 6                Editor  2012.0
## 7           Nice Answer  2012.0
## 8              Yearling  2012.0
## 9         Nice Question  2012.0
## 10        Nice Question  2013.0
## 11             Yearling  2013.0
## 12      Popular Question  2014.0
## 13             Yearling  2014.0
## 14            Taxonomist  2014.0
## 15     Notable Question  2015.0
## 16        Nice Question  2017.0
```

*Hint: use* `Badges.Date.astype("datetime64").dt.strftime("%Y").astype("float")`; *see Chapter 16.*

### 13.3.4  Grouping and Aggregating

**Exercise 13.9**  *Count how many badges of each type the user with* `Id=23` *won. Also, for each badge type, compute the minimal, average, and maximal receiving year. Return only the top four badges (with respect to the counts).*

```
res7a = pd.read_sql_query("""
    SELECT
        Name,
        COUNT(*) AS Count,
        MIN(CAST(strftime('%Y', Date) AS FLOAT)) AS MinYear,
        AVG(CAST(strftime('%Y', Date) AS FLOAT)) AS MeanYear,
        MAX(CAST(strftime('%Y', Date) AS FLOAT)) AS MaxYear
    FROM Badges
    WHERE UserId=23
    GROUP BY Name
    ORDER BY Count DESC
    LIMIT 4
```

```
""", conn)
res7a
##                   Name  Count   MinYear  MeanYear  MaxYear
## 0    Nice Question      4    2011.0   2013.25   2017.0
## 1          Yearling     3    2012.0   2013.00   2014.0
## 2  Popular Question     3    2014.0   2014.00   2014.0
## 3  Notable Question     2    2015.0   2015.00   2015.0
```

**Exercise 13.10** *Count how many unique combinations of pairs* `(Name, Year)` *for the badges won by the user with* `Id=23` *are there. Then, return only the rows having* `Count` *greater than 1 and order the results by* `Count` *decreasingly. In other words, list the badges received more than once in any given year.*

```
res8a = pd.read_sql_query("""
    SELECT
        Name,
        CAST(strftime('%Y', Date) AS FLOAT) AS Year,
        COUNT(*) AS Count
    FROM Badges
    WHERE UserId=23
    GROUP BY Name, Year
    HAVING Count > 1
    ORDER BY Count DESC
""", conn)
res8a
##                 Name    Year   Count
## 0  Popular Question  2014.0       3
## 1  Notable Question  2015.0       2
```

Note that `WHERE` is performed before `GROUP BY`, and `HAVING` is applied thereafter.

### 13.3.5 Joining

**Exercise 13.11** *Join (merge)* `Tags`, `Posts`, *and* `Users` *for all posts with* `OwnerUserId` *not equal to -1 (i.e., the tags which were created by "alive" users). Return the top six records with respect to* `Tags.Count`.

```
res9a = pd.read_sql_query("""
    SELECT Tags.TagName, Tags.Count, Posts.OwnerUserId,
        Users.Age, Users.Location, Users.DisplayName
    FROM Tags
    JOIN Posts ON Posts.Id=Tags.WikiPostId
    JOIN Users ON Users.AccountId=Posts.OwnerUserId
    WHERE OwnerUserId != -1
    ORDER BY Tags.Count DESC, Tags.TagName ASC
```

```
    LIMIT 6
""", conn)
res9a
##          TagName  Count  ...          Location       DisplayName
## 0          canada    802  ...      Mumbai, India             hitec
## 1          europe    681  ...  Philadelphia, PA       Adam Tuttle
## 2   visa-refusals    554  ...      New York, NY  Benjamin Pollack
## 3       australia    411  ...      Mumbai, India             hitec
## 4              eu    204  ...  Philadelphia, PA       Adam Tuttle
## 5   new-york-city    204  ...      Mumbai, India             hitec
##
## [6 rows x 6 columns]
```

**Exercise 13.12** *First, create an auxiliary (temporary) table named* `UpVotesTab`*, where we store the information about the number of up-votes (*`VoteTypeId=2`*) that each post has received. Then, join (merge) this table with* `Posts` *and fetch some details about the five questions (*`PostTypeId=1`*) with the most up-votes.*

```
res10a = pd.read_sql_query("""
    SELECT UpVotesTab.*, Posts.Title FROM
    (
        SELECT PostId, COUNT(*) AS UpVotes
        FROM Votes
        WHERE VoteTypeId=2
        GROUP BY PostId
    ) AS UpVotesTab
    JOIN Posts ON UpVotesTab.PostId=Posts.Id
    WHERE Posts.PostTypeId=1
    ORDER BY UpVotesTab.UpVotes DESC LIMIT 5
""", conn)
res10a
##     PostId  UpVotes                                        Title
## 0     3080      307  OK we're all adults here, so really, how on ea...
## 1    38177      254  How do you know if Americans genuinely/literal...
## 2    24540      221  How to intentionally get denied entry to the U...
## 3    20207      211  Why are airline passengers asked to lift up wi...
## 4    96447      178                      Why prohibit engine braking?
```

### 13.3.6 Solutions to Exercises

In this section we provide examples of solutions to the above exercises.

**Example 13.13** *To obtain a result equivalent to* `res2a`*, we need basic filtering only:*

```
res2b = (
    Tags.
    loc[
        Tags.TagName.isin(["poland", "australia", "china"]),
        ["TagName", "Count"]
    ].
    reset_index(drop=True)
)
```

Let us verify whether the two data frames are identical:

```
pd.testing.assert_frame_equal(res2a, res2b)  # no error == OK
```

**Example 13.14**  *To generate res3a with **pandas** only, we need some more complex filtering with* **loc[...]**:

```
res3b = (
    Posts.
    loc[
        (Posts.PostTypeId == 1) & (Posts.ViewCount >= 10000) &
        (Posts.FavoriteCount >= 35) & (Posts.FavoriteCount <= 100),
        ["Title", "Score", "ViewCount", "FavoriteCount"]
    ].
    reset_index(drop=True)
)
pd.testing.assert_frame_equal(res3a, res3b)  # no error == OK
```

**Example 13.15**  *For res4a, some filtering and sorting is all we need:*

```
res4b = (
    Posts.
    loc[
        Posts.ParentId.isna() & (~Posts.Title.isna()),
        ["Title", "Score"]
    ].
    sort_values("Score", ascending=False).
    head(5).
    reset_index(drop=True)
)
pd.testing.assert_frame_equal(res4a, res4b)  # no error == OK
```

**Example 13.16**  *The key to res5a is the **pandas.DataFrame.drop_duplicates** method:*

```
res5b = (
    Badges.
    loc[Badges.UserId == 23, ["Name"]].
```

```
    drop_duplicates().
    reset_index(drop=True)
)
pd.testing.assert_frame_equal(res5a, res5b)  # no error == OK
```

**Example 13.17**  *For* `res6a`*, we first need to add a new column to the copy of Badges:*

```
Badges2 = Badges.copy()  # otherwise we would destroy the original object
Badges2.loc[:, "Year"] = (
    Badges2.Date.astype("datetime64").dt.strftime("%Y").astype("float")
)
```

*Then, we apply some basic filtering and the removal of duplicated rows:*

```
res6b = (
    Badges2.
    loc[Badges2.UserId == 23, ["Name", "Year"]].
    drop_duplicates().
    reset_index(drop=True)
)
pd.testing.assert_frame_equal(res6a, res6b)  # no error == OK
```

**Example 13.18**  *For* `res7a`*, we can use* ***pandas.DataFrameGroupBy.aggregate****:*

```
Badges2 = Badges.copy()
Badges2.loc[:, "Year"] = (
    Badges2.Date.astype("datetime64").dt.strftime("%Y").astype("float")
)
res7b = (
    Badges2.
    loc[Badges2.UserId == 23, ["Name", "Year"]].
    groupby("Name")["Year"].
    aggregate([len, np.min, np.mean, np.max]).
    sort_values("len", ascending=False).
    head(4).
    reset_index()
)
res7b.columns = ["Name", "Count", "MinYear", "MeanYear", "MaxYear"]
```

*Had we not converted* `Year` *to* `float`*, we would obtain a meaningless average year, without any warning.*

*Unfortunately, the rows in* `res7a` *and* `res7b` *are ordered differently. For testing, we need to reorder them in the same way:*

```
pd.testing.assert_frame_equal(
    res7a.sort_values(["Name", "Count"]).reset_index(drop=True),
    res7b.sort_values(["Name", "Count"]).reset_index(drop=True)
)  # no error == OK
```

**Example 13.19**  *For res8a, we first count the number of values in each group:*

```
Badges2 = Badges.copy()
Badges2.loc[:, "Year"] = (
    Badges2.Date.astype("datetime64").dt.strftime("%Y").astype("float")
)
res8b = (
    Badges2.
    loc[ Badges2.UserId == 23, ["Name", "Year"] ].
    groupby(["Name", "Year"]).
    size().
    rename("Count").
    reset_index()
)
```

*The HAVING part is performed after WHERE and GROUP BY.*

```
res8b = (
    res8b.
    loc[ res8b.Count > 1, : ].
    sort_values("Count", ascending=False).
    reset_index(drop=True)
)
pd.testing.assert_frame_equal(res8a, res8b)  # no error == OK
```

**Example 13.20**  *To obtain a result equivalent to res9a, we need to merge Posts with Tags, and then merge the result with Users:*

```
res9b = pd.merge(Posts, Tags, left_on="Id", right_on="WikiPostId")
res9b = pd.merge(Users, res9b, left_on="AccountId", right_on="OwnerUserId")
```

*Then, some filtering and sorting will do the trick:*

```
res9b = (
    res9b.
    loc[
        (res9b.OwnerUserId != -1) & (~res9b.OwnerUserId.isna()),
        ["TagName", "Count", "OwnerUserId", "Age", "Location", "DisplayName"]
    ].
    sort_values(["Count", "TagName"], ascending=[False, True]).
    head(6).
```

```
    reset_index(drop=True)
)
```

*In SQL, "not equals to -1" implies* IS NOT NULL.

```
pd.testing.assert_frame_equal(res9a, res9b)  # no error == OK
```

**Example 13.21** *To obtain a result equivalent to* res10a, *we first need to create an auxiliary data frame that corresponds to the subquery.*

```
UpVotesTab = (
    Votes.
    loc[Votes.VoteTypeId==2, :].
    groupby("PostId").
    size().
    rename("UpVotes").
    reset_index()
)
```

*And now:*

```
res10b = pd.merge(UpVotesTab, Posts, left_on="PostId", right_on="Id")
res10b = (
    res10b.
    loc[res10b.PostTypeId==1, ["PostId", "UpVotes", "Title"]].
    sort_values("UpVotes", ascending=False).
    head(5).
    reset_index(drop=True)
)
pd.testing.assert_frame_equal(res10a, res10b)  # no error == OK
```

## 13.4 Closing the Database Connection

We said we should not forget about:

```
conn.close()
```

This gives some sense of closure. Such a relief.

## 13.5   Common Data Serialisation Formats for the Web

CSV files are an all-round way to exchange *tabular* data between different programming and data analysis environments.

For unstructured or non-tabularly-structured data, XML and JSON (and its superset, YAML) are the common formats of choice, especially for communicating with different Web APIs.

It is recommended that we solve some of the following exercises to make sure we can fetch data in these formats. Sadly, often this will require some quite tedious labour, neither art nor science; see also [81] and [16].

**Exercise 13.22** *Consider the Web API for accessing*[8] *the on-street parking bay sensor data in Melbourne, VIC, Australia. Using, for example, the* `json` *package, convert the data*[9] *in the JSON format to a data frame.*

**Exercise 13.23** *Australian Radiation Protection and Nuclear Safety Agency publishes*[10] *UV data for different Aussie cities. Using, for example, the* `xml` *package, convert this XML dataset*[11] *to a data frame.*

**Exercise 13.24** *(\*) Check out the English Wikipedia article featuring a list of 20th-century classical composers*[12]*. Using* `pandas.read_html`*, convert the* Climate Data *table included therein to a data frame.*

**Exercise 13.25** *(\*) Using, for example, the* `lxml` *package, write a function that converts each bullet list featured in a given Wikipedia article (e.g., this one*[13]*), to a list of strings.*

**Exercise 13.26** *(\*\*) Import an archived version of a Stack Exchange*[14] *site that you find interesting and store it in an SQLite database. You can find the relevant data dumps here*[15]*.*

**Exercise 13.27** *(\*\*) Download*[16] *and then import an archived version of one of the wikis hosted by the Wikimedia Foundation*[17] *(e.g., the whole English Wikipedia) so that it can be stored in an SQLite database.*

---

[8] https://data.melbourne.vic.gov.au/Transport/On-street-Parking-Bay-Sensors/vh2v-4nfs
[9] https://data.melbourne.vic.gov.au/resource/vh2v-4nfs.json
[10] https://www.arpansa.gov.au/our-services/monitoring/ultraviolet-radiation-monitoring/ultraviolet-radation-data-information
[11] https://uvdata.arpansa.gov.au/xml/uvvalues.xml
[12] https://en.wikipedia.org/wiki/List_of_20th-century_classical_composers
[13] https://en.wikipedia.org/wiki/Category:Fr%C3%A9d%C3%A9ric_Chopin
[14] https://stackexchange.com/
[15] https://archive.org/details/stackexchange
[16] https://meta.wikimedia.org/wiki/Data_dumps
[17] https://wikimediafoundation.org/

## 13.6 Working with Many Files

For the mass-processing of many files, it is worth knowing *of* the most basic functions for dealing with file paths, searching for files, etc. Usually, we will be looking up ways to complete specific tasks at hand, e.g., how to read data from a ZIP-like archive, on the internet. This is because – contrary to the basic operations of vectors, matrices, and data frames – these are not amongst the actions that we perform that often.

Good development practices related to data storage are described in [41].

### 13.6.1 File Paths

UNIX-like operating systems, including GNU/Linux and macOS, use slashes, `` `/` ``, as path separators, e.g., `"/home/marek/file.csv"`. Windows, however, uses backslashes, `` `\` ``, which have a special meaning in character strings (escape sequences; see Section 2.1.3). Therefore, they should be input as, e.g., `"c:\\users\\marek\\file.csv"`. Alternatively, we can use *raw* strings, where the backslash is treated literally, e.g., `r"c:\users\marek\file.csv"`.

When constructing file paths programmatically, it is thus best to rely on **os.path.join**, which takes care of the system-specific nuances.

```python
import os.path
os.path.join("~", "Desktop", "file.csv")  # we are on GNU/Linux
## '~/Desktop/file.csv'
```

The tilde, `` `~` ``, denotes the current user's *home* directory.

For storing auxiliary data, we can use the system's temporary directory. See the **tempfile** module for functions that generate appropriate file paths therein. For instance, a subdirectory inside the temporary directory can be created via a call to **tempfile.mkdtemp**.

---

**Important** We will frequently be referring to file paths relative to the working directory of the currently executed Python session (e.g., from which IPython/Jupyter notebook server was started); see **os.getcwd**.

---

All non-absolute file names (ones that do not start with `` `~` ``, `` `/` ``, `` `c:\\` ``, and the like), for example, `"filename.csv"` or **os.path.join**(`"subdir"`, `"filename.csv"`) are always *relative* to the current working directory.

For instance, if the working directory is `"/home/marek/projects/python"`, then `"filename.csv"` refers to `"/home/marek/projects/python/filename.csv"`.

Also, `` `..` `` denotes the current working directory's parent directory. Thus, `"../filename2.csv"` resolves to `"/home/marek/projects/filename2.csv"`.

**Exercise 13.28**  *Print the current working directory by calling* **os.getcwd***. Next, download the file* `air_quality_2018_param`[18] *and save it in the current Python session's working directory (e.g., in your web browser, right-click on the web page's canvas and select* Save Page As...*). Load with* **pandas.read_csv** *by passing* `"air_quality_2018_param.csv"` *as the input path.*

**Exercise 13.29**  *(\*) Download the aforementioned file programmatically (if you have not done so yet) using the* **requests** *module.*

## 13.6.2   File Search

**glob.glob** and **os.listdir** can generate a list of files in a given directory (and possibly all its subdirectories).

**os.path.isdir** and **os.path.isfile** can be used to determine the type of a given object in the file system.

**Exercise 13.30**  *Write a function that computes the total size of all the files in a given directory and all its subdirectories.*

## 13.6.3   Exception Handling

Accessing resources on the disk or the internet can lead to errors, for example when the file is not found. The **try..except** statement can be used if we want to be able to react to any of the envisaged errors

```python
try:
    # statements to execute
    x = pd.read_csv("file_not_found.csv")
    print(x.head())  # this will not be executed if the above raises an error
except OSError:
    # if an exception occurs, we can handle it here
    print("File has not been found")
## File has not been found
```

For more details, refer to the documentation[19].

## 13.6.4   File Connections (\*)

Basic ways of opening and reading from/writing to file connections are described in the documentation[20]. In Section 14.3.5, we show an example where we create a Markdown file manually.

They may be useful if we wish to process large files on a chunk-by-chunk basis. In particular, **pandas.read_csv** accepts a file handler (see **open**). Then, passing the `nrows` argument we can indicate the number of rows to fetch.

---

[18] https://github.com/gagolews/teaching-data/raw/master/marek/air_quality_2018_param.csv
[19] https://docs.python.org/3/tutorial/errors.html
[20] https://docs.python.org/3/tutorial/inputoutput.html

## 13.7 Exercises

**Exercise 13.31** *Find an example of an XML and JSON file. Which one is more human-readable? Do they differ in terms of capabilities?*

**Exercise 13.32** *What is wrong with constructing file paths like* `"~" + "\\" + "filename.csv"`*?*

**Exercise 13.33** *What are the benefits of using a SQL database management system in data science activities?*

**Exercise 13.34** *(\*) How can we populate a database with gigabytes of data read from many CSV files?*

# Part V

# Other Data Types

# 14

## *Text Data*

In [29] it is noted that effective processing of character strings is needed at various stages of data analysis pipelines: from data cleansing and preparation, through information extraction, to report generation; compare, e.g., [81] and [16]. *Pattern searching, string collation and sorting, normalisation, transliteration, and formatting are ubiquitous in text mining, natural language processing, and bioinformatics.* Means for the handling of string data *should be included in each statistician's or data scientist's repertoire to complement their numerical computing and data wrangling skills.*

In this chapter, we discuss the handiest string operations in base Python, together with their vectorised versions in `numpy` and `pandas`. We also mention some more advanced features of the Unicode `ICU` library.

## 14.1 Basic String Operations

Recall from Section 2.1.3 that the `str` class represents individual character strings:

```
x = "spam"
type(x)
## <class 'str'>
```

There are a few binary operators overloaded for strings, e.g., `` `+` `` stands for string concatenation:

```
x + " and eggs"
## 'spam and eggs'
```

`` `*` `` duplicates a given string:

```
x * 3
## 'spamspamspam'
```

In Chapter 3, we noted that `str` is a sequential type. As a consequence, we can extract individual code points and create substrings using the index operator:

```
x[-1]    # last letter
## 'ɱ'
```

Strings are immutable, but parts thereof can always be reused in conjunction with the concatenation operator:

```
x[:2] + "ecial"
## 'special'
```

### 14.1.1   Unicode as the Universal Encoding

It is worth knowing that all strings in Python (from version 3.0) use Unicode[1], which is a universal encoding capable of representing ca. 150,000 characters covering letters and numbers in contemporary and historic alphabets/scripts, mathematical, political, phonetic, and other symbols, emojis, etc.

---

**Note** Despite the wide support for Unicode, sometimes our own or other readers' display (e.g., web browsers when viewing an HTML version of the output report) might not be able to *render* all code points properly, e.g., due to missing fonts. Still, we should rest assured that they are processed correctly if string functions are applied thereon.

---

### 14.1.2   Normalising Strings

Dirty text data are a pain, especially if similar (semantically) tokens are encoded in many different ways. For the sake of string matching, we might want, e.g., the German "groß", "GROSS", and " gross " to compare all equal.

**str.strip** removes whitespaces (spaces, tabs, newline characters) at both ends of strings (see also **str.lstrip** and **str.rstrip** for their nonsymmetric versions).

**str.lower** and **str.upper** change letter case. For caseless comparison/matching, **str.casefold** might be a slightly better option as it unfolds many more code point sequences:

```
"Groß".lower(), "Groß".upper(), "Groß".casefold()
## ('groß', 'GROSS', 'gross')
```

---

**Note** (*) More advanced string transliteration[2] can be performed by means of the **ICU**[3]

---

[1] (*) More precisely, Python strings are UTF-8-encoded. Most web pages and APIs are nowadays served in UTF-8. But we can still occasionally encounter files encoded in ISO-8859-1 (Western Europe), Windows-1250 (Eastern Europe), Windows-1251 (Cyrillic), GB18030 and Big5 (Chinese), EUC-KR (Korean), Shift-JIS and EUC-JP (Japanese), amongst others. They can be converted using the **str.decode** method.

[2] https://unicode-org.github.io/icu/userguide/transforms/general/

[3] https://icu.unicode.org/

(International Components for Unicode) library, which the **PyICU** package provides wrappers for.

For instance, converting all code points to ASCII (English) might be necessary when identifiers are expected to miss some diacritics that would normally be included (as in "Gągolewski" vs "Gagolewski"):

```python
import icu  # PyICU package
(icu.Transliterator
    .createInstance("Lower; Any-Latin; Latin-ASCII")
    .transliterate(
        "Χαίρετε! Groß gżegżółka – © La Niña – köszönöm – Gągolewski"
    )
)
## 'chairete! gross gzegzolka - (C) la nina - koszonom - gagolewski'
```

Converting between different Unicode Normalisation Forms[4] (also available in the **unicodedata** package and via **pandas.Series.str.normalize**) might be used for the removal of some formatting nuances:

```python
icu.Transliterator.createInstance("NFKD; NFC").transliterate("¼qr²")
## '¼qr2'
```

### 14.1.3   Substring Searching and Replacing

Determining if a string features a particular fixed substring can be done in several ways.

For instance, the **in** operator verifies whether a particular substring occurs at least once:

```python
food = "bacon, spam, spam, srapatapam, eggs, and spam"
"spam" in food
## True
```

The **str.count** method determines the number of occurrences of a substring:

```python
food.count("spam")
## 3
```

To locate the first pattern appearance, we call **str.index**:

```python
food.index("spam")
## 7
```

---

[4] https://www.unicode.org/faq/normalization.html

**str.replace** substitutes matching substrings with new content:

```
food.replace("spam", "veggies")
## 'bacon, veggies, veggies, srapatapam, eggs, and veggies'
```

**Exercise 14.1** *Read the manual of the following methods: **str.startswith**, **str.endswith**, **str.find**, **str.rfind**, **str.rindex**, **str.removeprefix**, and **str.removesuffix**.*

The splitting of long strings at specific fixed delimiter strings can be done via:

```
food.split(", ")
## ['bacon', 'spam', 'spam', 'srapatapam', 'eggs', 'and spam']
```

See also **str.partition**. The **str.join** method implements the inverse operation:

```
", ".join(["spam", "bacon", "eggs", "spam"])
## 'spam, bacon, eggs, spam'
```

Moreover, in Section 14.4, we will discuss pattern matching with regular expressions, which can be useful in, amongst others, extracting more abstract data chunks (numbers, URLs, email addresses, IDs) from within strings.

### 14.1.4 Locale-Aware Services in ICU (*)

Recall that relational operators such as `<` and `>=` perform the lexicographic comparing of strings (like in a dictionary or an encyclopedia):

```
"spam" > "egg"
## True
```

We have: "a" < "aa" < "aaaaaaaaaaaaa" < "ab" < "aba" < "abb" < "b" < "ba" < "baaaaaaa" < "bb" < "Spanish Inquisition".

The lexicographic ordering (character-by-character, from left to right) is not necessarily appropriate for strings featuring numerals:

```
"a9" < "a123"   # 1 is greater than 9
## False
```

Additionally, it only takes into account the numeric codes (see Section 14.4.3) corresponding to each Unicode character. Consequently, it does not work well with non-English alphabets:

```
"MIELONECZKĄ" < "MIELONECZKI"
## False
```

In Polish, *A with ogonek* (Ą) should sort after *A* and before *B*, let alone *I*. However, their corresponding numeric codes in the Unicode table are: 260 (Ą), 65 (A), 66 (B), and 73 (I). The resulting ordering is thus incorrect, as far as natural language processing is concerned.

It is best to perform string collation using the services provided by **ICU**. Here is an example of German phone book-like collation where "ö" is treated the same as "oe":

```
c = icu.Collator.createInstance(icu.Locale("de_DE@collation=phonebook"))
c.setStrength(0)  # ignore case and some diacritics
c.compare("Löwe", "loewe")
## 0
```

A result of 0 means that the strings are deemed equal.

In some languages, contractions occur, e.g., in Slovak and Czech, two code points "ch" are treated as a single entity and are sorted after "h":

```
icu.Collator.createInstance(icu.Locale("sk_SK")).compare("chladný", "hladný")
## 1
```

This means that we have "chladný" > "hladný" (the 1st argument is greater than the 2nd one). Compare the above to something similar in Polish:

```
icu.Collator.createInstance(icu.Locale("pl_PL")).compare("chłodny", "hardy")
## -1
```

That is, "chłodny" < "hardy" (the first argument is less than the 2nd one).

Also, with **ICU**, numeric collation is possible:

```
c = icu.Collator.createInstance()
c.setAttribute(
    icu.UCollAttribute.NUMERIC_COLLATION,
    icu.UCollAttributeValue.ON
)
c.compare("a9", "a123")
## -1
```

Which is the correct result: "a9" is less than "a123" (compare the above to the example where we used the ordinary `<`).

## 14.1.5 String Operations in pandas

String sequences in `pandas.Series` are by default[5] using the broadest possible `object` data type:

---

[5] https://pandas.pydata.org/pandas-docs/stable/user_guide/text.html

```
pd.Series(["spam", "bacon", "spam"])
## 0     spam
## 1    bacon
## 2     spam
## dtype: object
```

This allows for the encoding of missing values by means of the None object (which is of type None, not str); compare Section 15.1.

Vectorised versions of base string operations are available via the **pandas.Series.str** accessor. We thus have **pandas.Series.str.strip**, **pandas.Series.str.split**, **pandas.Series.str.find**, and so forth. For instance:

```
x = pd.Series(["spam", "bacon", None, "buckwheat", "spam"])
x.str.upper()
## 0         SPAM
## 1        BACON
## 2         None
## 3    BUCKWHEAT
## 4         SPAM
## dtype: object
```

But there is more. For example, a function to compute the length of each string:

```
x.str.len()
## 0    4.0
## 1    5.0
## 2    NaN
## 3    9.0
## 4    4.0
## dtype: float64
```

Vectorised concatenation of strings can be performed using the overloaded `+` operator:

```
x + " and spam"
## 0         spam and spam
## 1        bacon and spam
## 2                   NaN
## 3    buckwheat and spam
## 4         spam and spam
## dtype: object
```

To concatenate all items into a single string, we call:

```
x.str.cat(sep="; ")
## 'spam; bacon; buckwheat; spam'
```

Conversion to numeric:

```python
pd.Series(["1.3", "-7", None, "3523"]).astype(float)
## 0        1.3
## 1       -7.0
## 2        NaN
## 3     3523.0
## dtype: float64
```

Select substrings:

```python
x.str.slice(2, -1)  # like x.iloc[i][2:-1] for all i
## 0         a
## 1        co
## 2      None
## 3    ckwhea
## 4         a
## dtype: object
```

Replace substrings:

```python
x.str.slice_replace(0, 2, "tofu")  # like x.iloc[i][2:-1] = "tofu"
## 0        tofuam
## 1       tofucon
## 2          None
## 3    tofuckwheat
## 4        tofuam
## dtype: object
```

**Exercise 14.2** *Consider the* `nasaweather_glaciers`[6] *data frame. All glaciers are assigned 11/12-character unique identifiers as defined by the WGMS convention that forms the glacier ID number by combining the following five elements.*

1. *2-character political unit (first two letters of the ID),*

2. *1-digit continent code (the third letter),*

3. *4-character drainage code (next four),*

4. *2-digit free position code (next two),*

5. *2- or 3-digit local glacier code (the remaining ones).*

*Extract the five chunks and store them as independent columns in the data frame.*

_________________________________

[6] https://github.com/gagolews/teaching-data/raw/master/other/nasaweather_glaciers.csv

## 14.1.6  String Operations in `numpy` (*)

There is a huge overlap between the **numpy** and **pandas** capabilities for string handling, with the latter being more powerful. After all, **numpy** is a workhorse for *numerical* computing. Still, some readers might find the following useful.

As mentioned in our introduction to **numpy** vectors, objects of type ndarray can store not only numeric and logical data, but also character strings. For example:

```
x = np.array(["spam", "bacon", "egg"])
x
## array(['spam', 'bacon', 'egg'], dtype='<U5')
```

Here, the data type "<U5" means that we deal with Unicode strings of length no greater than 5. Unfortunately, replacing elements with too long a content will result in truncated strings:

```
x[2] = "buckwheat"
x
## array(['spam', 'bacon', 'buckw'], dtype='<U5')
```

To remedy this, we first need to recast the vector manually:

```
x = x.astype("<U10")
x[2] = "buckwheat"
x
## array(['spam', 'bacon', 'buckwheat'], dtype='<U10')
```

Conversion from/to numeric is also possible:

```
np.array(["1.3", "-7", "3523"]).astype(float)
## array([ 1.300e+00, -7.000e+00,  3.523e+03])
np.array([1, 3.14, -5153]).astype(str)
## array(['1.0', '3.14', '-5153.0'], dtype='<U32')
```

The **numpy.char**[7] module includes several vectorised versions of string routines, most of which we discussed above. For example:

```
x = np.array([
    "spam", "spam, bacon, and spam",
    "spam, eggs, bacon, spam, spam, and spam"
])
np.char.split(x, ", ")
## array([list(['spam']), list(['spam', 'bacon', 'and spam']),
##        list(['spam', 'eggs', 'bacon', 'spam', 'spam', 'and spam'])],
##       dtype=object)
```

*(continues on next page)*

---

[7] https://numpy.org/doc/stable/reference/routines.char.html

```
np.char.count(x, "spam")
## array([1, 2, 4])
```

Vectorised operations that we would normally perform through the binary operators (i.e., `` `+` ``, `` `*` ``, `` `<` ``, etc.) are available through standalone functions:

```
np.char.add(["spam", "bacon"], " and spam")
## array(['spam and spam', 'bacon and spam'], dtype='<U14')
np.char.equal(["spam", "bacon", "spam"], "spam")
## array([ True, False,  True])
```

The function that returns the length of each string is also noteworthy:

```
np.char.str_len(x)
## array([ 4, 21, 39])
```

## 14.2 Working with String Lists

**pandas** nicely supports lists of strings of varying lengths. For instance:

```
x = pd.Series([
    "spam",
    "spam, bacon, spam",
    "potatoes",
    None,
    "spam, eggs, bacon, spam, spam"
])
xs = x.str.split(", ", regex=False)
xs
## 0                              [spam]
## 1                 [spam, bacon, spam]
## 2                          [potatoes]
## 3                                None
## 4     [spam, eggs, bacon, spam, spam]
## dtype: object
```

And now, e.g., looking at the last element:

```
xs.iloc[-1]
## ['spam', 'eggs', 'bacon', 'spam', 'spam']
```

reveals that it is indeed a list of strings.

There are a few vectorised operations that enable us to work with such variable length lists, such as concatenating all strings:

```
xs.str.join("; ")
## 0                             spam
## 1                spam; bacon; spam
## 2                         potatoes
## 3                             None
## 4    spam; eggs; bacon; spam; spam
## dtype: object
```

selecting, say, the first string in each list:

```
xs.str.get(0)
## 0        spam
## 1        spam
## 2    potatoes
## 3        None
## 4        spam
## dtype: object
```

or slicing:

```
xs.str.slice(0, -1)  # like xs.iloc[i][0:-1] for all i
## 0                          []
## 1               [spam, bacon]
## 2                          []
## 3                        None
## 4    [spam, eggs, bacon, spam]
## dtype: object
```

**Exercise 14.3**  (*) *Using* ***pandas.merge***, *join the* `countries`[8], `world_factbook_2020`[9], *and* `ssi_2016_dimensions`[10] *datasets based on the country names. Note that some manual data cleansing will be necessary beforehand.*

**Exercise 14.4**  (**) *Given a* `Series` *object* `xs` *featuring lists of strings, convert it to a 0/1 representation.*

1. *Determine the list of all unique strings; let us call it* `xu`.

2. *Create a data frame* `x` *with* `xs.shape[0]` *rows and* `len(xu)` *columns such that* `x.iloc[i, j]` *is equal to 1 if* `xu[j]` *is amongst* `xs.loc[i]` *and equal to 0 otherwise. Set the column names to* `xs`.

3. *Given* `x` *(and only* `x`*: neither* `xs` *nor* `xu`*), perform the inverse operation.*

---

*For example, for the above xs object, x should look like:*

```
##      bacon   eggs  potatoes  spam
## 0       0      0         0     1
## 1       1      0         0     1
## 2       0      0         1     0
## 3       0      0         0     0
## 4       1      1         0     1
```

## 14.3 Formatted Outputs for Reproducible Report Generation

Some good development practices related to reproducible report generation are discussed in [73, 87, 88]. Note that the paradigm of literate programming was introduced by D. Knuth in [49].

Reports from data analysis can be prepared, e.g., in Jupyter Notebooks or by writing directly to Markdown files which we can later compile to PDF or HTML. Below we briefly discuss how to output nicely formatted objects programmatically.

### 14.3.1 Formatting Strings

Inclusion of textual representation of data stored in existing objects can easily be done using f-strings (formatted string literals; see Section 2.1.3) of type f"...{expression}...". For instance:

```
pi = 3.14159265358979323846
f"π = {pi:.2f}"
## 'π = 3.14'
```

creates a string showing the value of the variable `pi` formatted as a `float` rounded to two places after the decimal separator.

**Note** (**) Similar functionality can be achieved using the **str.format** method:

```
"π = {:.2f}".format(pi)
## 'π = 3.14'
```

as well as the `%` operator overloaded for strings, which uses **sprintf**-like value placeholders known to some readers from other programming languages (such as C):

```
"π = %.2f" % pi
## 'π = 3.14'
```

### 14.3.2 `str` and `repr`

The **str** and **repr** functions can create string representations of many objects:

```python
x = np.array([1, 2, 3])
str(x)
## '[1 2 3]'
repr(x)
## 'array([1, 2, 3])'
```

The former is more human-readable, and the latter is slightly more technical. Note that **repr** often returns an output that can be interpreted as executable Python code with no or few adjustments. Nonetheless, **pandas** objects are amongst the many exceptions to this rule.

### 14.3.3 Aligning Strings

**str.center**, **str.ljust**, **str.rjust** can be used to centre-, left-, or right-align a string so that it is of at least given width. This might make the display thereof more aesthetic. Very long strings, possibly containing whole text paragraphs can be dealt with using the **wrap** and **shorten** functions from the **textwrap** package.

### 14.3.4 Direct Markdown Output in Jupyter

Further, with IPython/Jupyter, we can output strings that will be directly interpreted as Markdown-formatted:

```python
import IPython.display
x = 2+2
out = f"*Result*: $2^2=2\\cdot 2={x}$."  # LaTeX math
IPython.display.Markdown(out)
```

*Result*: $2^2 = 2 \cdot 2 = 4$.

Recall from Section 1.2.5 that Markdown is a very flexible markup[11] language that allows us to define itemised and numbered lists, mathematical formulae, tables, images, etc.

On a side note, data frames can be nicely prepared for display in a report using **pandas. DataFrame.to_markdown**.

### 14.3.5 Manual Markdown File Output (*)

We can also generate Markdown code programmatically in the form of standalone `.md` files:

---

[11] (*) Markdown is amongst many markup languages. Other learn-worthy ones include HTML (for the Web) and LaTeX (especially for beautiful typesetting of maths, print-ready articles, and books, e.g., PDF; see [61] for a good introduction).

```python
import tempfile, os.path
filename = os.path.join(tempfile.mkdtemp(), "test-report.md")
f = open(filename, "w")  # open for writing (overwrite if exists)
f.write("**Yummy Foods** include, but are not limited to:\n\n")
x = ["spam", "bacon", "eggs", "spam"]
for e in x:
    f.write(f"* {e}\n")
f.write("\nAnd now for something *completely* different:\n\n")
f.write("Rank | Food\n")
f.write("-----|-----\n")
for i in range(len(x)):
    f.write(f"{i+1:4} | {x[i][::-1]:10}\n")
f.close()
```

Here is the resulting raw Markdown source file:

```python
with open(filename, "r") as f:  # will call f.close() automatically
    out = f.read()
print(out)
## **Yummy Foods** include, but are not limited to:
##
## * spam
## * bacon
## * eggs
## * spam
##
## And now for something *completely* different:
##
## Rank | Food
## -----|-----
##    1 | maps
##    2 | nocab
##    3 | sgge
##    4 | maps
```

We can convert it to other formats, including HTML, PDF, EPUB, ODT, and even presentations by running[12] the **pandoc**[13] tool. We may also embed it directly inside an IPython/Jupyter notebook:

```python
IPython.display.Markdown(out)
```

**Yummy Foods** include, but are not limited to:

- spam

---

[12] External programs can be executed using `subprocess.run`.
[13] https://pandoc.org/

- bacon

- eggs

- spam

And now for something *completely* different:

| Rank | Food |
|------|------|
| 1    | maps |
| 2    | nocab |
| 3    | sgge |
| 4    | maps |

**Note**  Figures created in `matplotlib` can be exported to PNG, SVG, or PDF files using the `matplotlib.pyplot.savefig` function. We can include them manually in a Markdown document using the `` syntax.

**Note**  (*) IPython/Jupyter Notebooks can be converted to different formats using the `jupyter-nbconvert`[14] command line tool. `jupytext`[15] can create notebooks from ordinary text files. Literate programming with mixed R and Python is possible with the R packages `knitr`[16] and `reticulate`[17]. See [65] for an overview of many more options.

## 14.4  Regular Expressions (*)

This section contains large excerpts from yours truly's other work [29].

*Regular expressions (regexes) provide concise grammar for defining systematic patterns which can be sought in character strings. Examples of such patterns include: specific fixed substrings, emojis of any kind, stand-alone sequences of lower-case Latin letters ("words"), substrings that can be interpreted as real numbers (with or without fractional parts, also in scientific notation), telephone numbers, email addresses, or URLs.*

*Theoretically, the concept of regular pattern matching dates to the so-called regular languages and finite state automata [48]; see also [68] and [44]. Regexes in the form as we know it today were already present in one of the pre-Unix implementations of the command-line text editor **qed** [69] (the predecessor of the well-known **sed**).*

---

[14] https://pypi.org/project/nbconvert/
[15] https://jupytext.readthedocs.io/en/latest/
[16] https://yihui.org/knitr/
[17] https://rstudio.github.io/reticulate/

### 14.4.1 Regex Matching with `re`

In Python, the **`re`** module implements a regular expression matching engine. It accepts patterns that follow similar syntax to the ones available in the Perl language.

As a matter of fact, most programming languages and text editors (including **Kate**[18], **Eclipse**[19], and **VSCodium**[20]) support finding and replacing patterns with regexes. This is why they should be amongst the instruments at every data scientist's disposal.

Before we proceed with a detailed discussion on how to read and write regexes, let us first review some of the methods for identifying the matching substrings. Below we use the `r"\bni+\b"` regex as an example, which catches `"n"` followed by at least one `"i"` that begins and ends at a *word boundary*. In other words, we seek `"ni"`, `"nii"`, `"niii"`, etc. which may be considered standalone words.

In particular, **`re.findall`** extracts all non-overlapping matches to a given regex:

```python
import re
x = "We're the knights who say ni! niiiii! ni! niiiiiiiii!"
re.findall(r"\bni+\b", x)
## ['ni', 'niiiii', 'ni', 'niiiiiiiii']
```

The order of arguments is *(look for what, where)*, not the other way around.

---

**Important** We used the `r"..."` prefix to input a string so that "\b" is not treated as an escape sequence which denotes the backspace character. Otherwise, the above would have to be written as "\\bni+\\b".

---

If we had not insisted on matching at the word boundaries (i.e., if we used the simple `"ni+"` regex instead), we would also match the `"ni"` in `"knights"`.

The **`re.search`** function returns an object of class `re.Match` that enables us to get some more information about the first match:

```python
r = re.search(r"\bni+\b", x)
r.start(), r.end(), r.group()
## (26, 28, 'ni')
```

The above includes the start and end position (index) and the match itself. If the regex contains *capture groups* (see below for more details), we can also pinpoint the matches thereto.

Moreover, **`re.finditer`** returns an iterable object that includes the same details, but now about all the matches:

---

[18] https://kate-editor.org/
[19] https://www.eclipse.org/ide/
[20] https://vscodium.com/

```
rs = re.finditer(r"\bni+\b", x)
for r in rs:
    print((r.start(), r.end(), r.group()))
## (26, 28, 'ni')
## (30, 36, 'niiiii')
## (38, 40, 'ni')
## (42, 52, 'niiiiiiiii')
```

**re.split** divides a string into chunks separated by matches to a given regex:

```
re.split(r"!\s+", x)
## ["We're the knights who say ni", 'niiiii', 'ni', 'niiiiiiiii!']
```

The "!\s*" regex matches the exclamation mark followed by one or more whitespace characters.

Using **re.sub**, each match can be replaced with a given string:

```
re.sub(r"\bni+\b", "nu", x)
## "We're the knights who say nu! nu! nu! nu!"
```

---

**Note**   (\*\*) More flexible replacement strings can be generated by passing a custom function as the second argument:

```
re.sub(r"\bni+\b", lambda m: "n" + "u"*(m.end()-m.start()-1), x)
## "We're the knights who say nu! nuuuuu! nu! nuuuuuuuuu!"
```

---

### 14.4.2   Regex Matching with **pandas**

The **pandas.Series.str** accessor also defines a number of vectorised functions that utilise the **re** package's matcher.

Example Series object:

```
x = pd.Series(["ni!", "niiii, ni, nii!", None, "spam, bacon", "nii, ni!"])
x
## 0                 ni!
## 1     niiii, ni, nii!
## 2                None
## 3         spam, bacon
## 4            nii, ni!
## dtype: object
```

Here are the most notable functions:

```python
x.str.contains(r"\bni+\b")
## 0     True
## 1     True
## 2     None
## 3    False
## 4     True
## dtype: object
x.str.count(r"\bni+\b")
## 0    1.0
## 1    3.0
## 2    NaN
## 3    0.0
## 4    2.0
## dtype: float64
x.str.replace(r"\bni+\b", "nu", regex=True)
## 0             nu!
## 1    nu, nu, nu!
## 2           None
## 3    spam, bacon
## 4        nu, nu!
## dtype: object
x.str.findall(r"\bni+\b")
## 0               [ni]
## 1    [niiii, ni, nii]
## 2               None
## 3                 []
## 4          [nii, ni]
## dtype: object
x.str.split(r",\s+")  # a comma, one or more whitespaces
## 0              [ni!]
## 1    [niiii, ni, nii!]
## 2               None
## 3       [spam, bacon]
## 4          [nii, ni!]
## dtype: object
```

In the two last cases, we get lists of strings as results.

Also, later we will mention **pandas.Series.str.extract** and **pandas.Series.str.extractall** which work with regexes that include capture groups.

---

**Note** (*) If we intend to seek matches to the same pattern in many different strings without the use of **pandas**, it might be faster to pre-compile a regex first, and then use the **re.Pattern.findall** method instead or **re.findall**:

```
p = re.compile(r"\bni+\b")   # returns an object of class `re.Pattern`
p.findall("We're the Spanish Inquisition ni! ni! niiiii! nininiiiiiiiii!")
## ['ni', 'ni', 'niiiii']
```

### 14.4.3  Matching Individual Characters

In the following subsections, we review the most essential elements of the regex syntax as we did in [29]. One general introduction to regexes is [25]. The **re** module flavour is summarised in the official manual[21], see also [51].

We begin by discussing different ways to define character sets. In this part, determining the length of all matching substrings will be quite straightforward.

**Important**   The following characters have special meaning to the regex engine: ".", "\", "|", "(", ")", "[", "]", "{", "}", "^", "$", "*", "+", and "?".

Any regular expression that contains none of the above behaves like a fixed pattern:

```
re.findall("spam", "spam, eggs, spam, bacon, sausage, and spam")
## ['spam', 'spam', 'spam']
```

There are three occurrences of a pattern that is comprised of four code points, "s" followed by "p", then by "a", and ending with "m".

If we wish to include a special character as part of a regular expression so that it is treated literally, we will need to escape it with a backslash, "\".

```
re.findall(r"\.", "spam...")
## ['.', '.', '.']
```

#### Matching Any Character

The (unescaped) dot, ".", matches any code point except the newline.

```
x = "Spam, ham,\njam, SPAM, eggs, and spam"
re.findall("..am", x, re.IGNORECASE)
## ['Spam', ' ham', 'SPAM', 'spam']
```

The above extracts non-overlapping length-4 substrings that end with "am", case-insensitively.

The dot's insensitivity to the newline character is motivated by the need to maintain compatibility with tools such as **grep** (when searching within text files in a line-by-line manner). This behaviour can be altered by setting the DOTALL flag.

---

[21] https://docs.python.org/3/library/re.html

```
re.findall("..am", x, re.DOTALL|re.IGNORECASE)  # `|` is the bitwise OR
## ['Spam', ' ham', '\njam', 'SPAM', 'spam']
```

### Defining Character Sets

Sets of characters can be introduced by enumerating their members within a pair of square brackets. For instance, "[abc]" denotes the set *{a, b, c}* – such a regular expression matches one (and only one) symbol from this set. Moreover, in:

```
re.findall("[hj]am", x)
## ['ham', 'jam']
```

the "[hj]am" regex matches: "h" or "j", followed by "a", followed by "m". In other words, "ham" and "jam" are the only two strings that are matched by this pattern (unless matching is done case-insensitively).

---

**Important**  The following characters, if used within square brackets, may be treated non-literally: "\", "[", "]", "^", "-", "&", "~", and "|".

---

To include them as-is in a character set, the backslash-escape must be used. For example, "[\[\]\\]" matches a backslash or a square bracket.

### Complementing Sets

Including "^" (the caret) after the opening square bracket denotes a set's complement. Hence, "[^abc]" matches any code point except "a", "b", and "c". Here is an example where we seek any substring that consists of four non-spaces:

```
x = "Nobody expects the Spanish Inquisition!"
re.findall("[^ ][^ ][^ ][^ ]", x)
## ['Nobo', 'expe', 'Span', 'Inqu', 'isit', 'ion!']
```

### Defining Code Point Ranges

Each Unicode character can be referenced by its unique numeric code[22]. For instance, "a" is assigned code U+0061 and "z" is mapped to U+007A. In the pre-Unicode era (mostly with regard to the ASCII codes, ≤ U+007F, representing English letters, decimal digits, as well as some punctuation and control characters), we were used to relying on specific code ranges. For example, "[a-z]" denotes the set comprised of all characters with codes between U+0061 and U+007A, i.e., lowercase letters of the English (Latin) alphabet.

---

[22] https://www.unicode.org/charts/

```
re.findall("[0-9A-Za-z]", "Gągolewski")
## ['G', 'g', 'o', 'l', 'e', 'w', 's', 'k', 'i']
```

The above pattern denotes the union of three code ranges: ASCII upper- and lower-case letters and digits. Nowadays, in the processing of text in natural languages, this notation should be avoided. Note the missing "ą" (Polish "a" with ogonek) in the result.

**Using Predefined Character Sets**

Consider the following string:

```
x = "aąbßÆAĄB⬚12⬚⬚,.;'! \t-+=\n[]©⬚⬚""„"
```

Some glyphs are not available in the PDF version of this book (because we did not install the required fonts, e.g., the Arabic digit 4 or left and right arrows; this is an educational example), but they are well-defined at the program level.

Noteworthy Unicode-aware code point classes include the "word" characters:

```
re.findall(r"\w", x)
## ['a', 'ą', 'b', 'ß', 'Æ', 'A', 'Ą', 'B', '⬚', '1', '2', '⬚', '⬚']
```

decimal digits:

```
re.findall(r"\d", x)
## ['1', '2', '⬚', '⬚']
```

and whitespaces:

```
re.findall(r"\s", x)
## [' ', '\t', '\n']
```

Moreover, e.g., "\W" is equivalent to "[^\w]" , i.e., denotes the set's complement.

### 14.4.4   Alternating and Grouping Subexpressions

**Alternation Operator**

The alternation operator, "|" (the pipe or bar), matches either its left or its right branch, for instance:

```
x = "spam, egg, ham, jam, algae, and an amalgam of spam, all al dente"
re.findall("spam|ham", x)
## ['spam', 'ham', 'spam']
```

**Grouping Subexpressions**

The "|" operator has very low precedence (otherwise, we would match "spamam" or "spaham" above instead). If we wish to introduce an alternative of *sub*expressions, we

need to group them using the "(?:...)" syntax. For instance, "(?:sp|h)am" matches either "spam" or "ham".

Notice that the bare use of the round brackets, "(...)" (i.e., without the "?:" part), has the side-effect of creating new capturing groups; see below for more details.

Also, matching is always done left-to-right, on a first-come, first-served (greedy) basis. Consequently, if the left branch is a subset of the right one, the latter will never be matched. In particular, "(?:al|alga|algae)" can only match "al". To fix this, we can write "(?:algae|alga|al)".

**Non-grouping Parentheses**

Some parenthesised subexpressions – those in which the opening bracket is followed by the question mark – have a distinct meaning. In particular, "(?#...)" denotes a free-format comment that is ignored by the regex parser:

```
re.findall(
  "(?# match 'sp' or 'h')(?:sp|h)(?# and 'am')am|(?# or match 'egg')egg",
  x
)
## ['spam', 'egg', 'ham', 'spam']
```

This is just horrible. Luckily, constructing more sophisticated regexes by concatenating subfragments thereof is more readable:

```
re.findall(
      "(?:sp|h)" +    # match either 'sp' or 'h'
      "am" +          # followed by 'am'
    "|" +           # ... or ...
      "egg",          # just match 'egg'
    x
)
## ['spam', 'egg', 'ham', 'spam']
```

What is more, e.g., "(?i)" enables the case-insensitive mode.

```
re.findall("(?i)spam", "Spam spam SPAMITY spAm")
## ['Spam', 'spam', 'SPAM', 'spAm']
```

## 14.4.5 Quantifiers

More often than not, a variable number of instances of the same subexpression needs to be captured or its presence should be made optional. This can be achieved by means of the following quantifiers:

- "?" matches 0 or 1 time;
- "*" matches 0 or more times;

- "+" matches 1 or more times;

- "{n,m}" matches between n and m times;

- "{n,}" matches at least n times;

- "{n}" matches exactly n times.

These operators are applied onto the directly preceding atoms. For example, "ni+" captures "ni", "nii", "niii", etc., but neither "n" alone nor "ninini" altogether.

By default, the quantifiers are greedy – they match the repeated subexpression as many times as possible. The "?" suffix (forming quantifiers such as "??", "*?", "+?", and so forth) tries with as few occurrences as possible (to obtain a match still).

Greedy:

```
x = "sp(AM)(maps)(SP)am"
re.findall(r"\(.+\)", x)
## ['(AM)(maps)(SP)']
```

Lazy:

```
re.findall(r"\(.+?\)", x)
## ['(AM)', '(maps)', '(SP)']
```

Greedy (but clever):

```
re.findall(r"\([^)]+\)", x)
## ['(AM)', '(maps)', '(SP)']
```

The first regex is greedy: it matches an opening bracket, then as many characters as possible (including ")") that are followed by a closing bracket. The two other patterns terminate as soon as the first closing bracket is found.

More examples:

```
x = "spamamamnomnomnomammmmmmmm"
re.findall("sp(?:am|nom)+", x)
## ['spamamamnomnomnomam']
re.findall("sp(?:am|nom)+?", x)
## ['spam']
```

And:

```
re.findall("sp(?:am|nom)+?m*", x)
## ['spam']
re.findall("sp(?:am|nom)+?m+", x)
## ['spamamamnomnomnomammmmmmmm']
```

Let us stress that the quantifier is applied to the subexpression that stands directly before it. Grouping parentheses can be used in case they are needed.

```
x = "12, 34.5, 678.901234, 37...629, ..."
re.findall(r"\d+\.\d+", x)
## ['34.5', '678.901234']
```

matches digits, a dot, and another series of digits.

```
re.findall(r"\d+(?:\.\d+)?", x)
## ['12', '34.5', '678.901234', '37', '629']
```

finds digits which are possibly (but not necessarily) followed by a dot and a digit sequence.

**Exercise 14.5** *Write a regex that extracts all #hashtags from a string #omg #SoEasy.*

### 14.4.6 Capture Groups and References Thereto (**)

Round-bracketed subexpressions (without the "?:" prefix) form the so-called *capture groups* that can be extracted separately or be referred to in other parts of the same regex.

**Extracting Capture Group Matches**

The above is evident when we use `re.findall`:

```
x = "name='Sir Launcelot', quest='Seek Grail', favcolour='blue'"
re.findall(r"(\w+)='(.+?)'", x)
## [('name', 'Sir Launcelot'), ('quest', 'Seek Grail'), ('favcolour', 'blue')]
```

This returned the matches to the individual capture groups, not the whole matching substrings.

`re.find` and `re.finditer` can pinpoint each component:

```
r = re.search(r"(\w+)='(.+?)'", x)
print("whole (0):", (r.start(), r.end(), r.group()))
print("      1 :", (r.start(1), r.end(1), r.group(1)))
print("      2 :", (r.start(2), r.end(2), r.group(2)))
## whole (0): (0, 20, "name='Sir Launcelot'")
##       1 : (0, 4, 'name')
##       2 : (6, 19, 'Sir Launcelot')
```

Here is a vectorised version of the above from **pandas**, returning the first match:

```
y = pd.Series([
    "name='Sir Launcelot'",
```

```
    "quest='Seek Grail'",
    "favcolour='blue', favcolour='yel.. Aaargh!'"
])
y.str.extract(r"(\w+)='(.+?)'")
##            0             1
## 0      name  Sir Launcelot
## 1     quest     Seek Grail
## 2  favcolour          blue
```

We see that the findings are conveniently presented in the data frame form. The first column gives the matches to the first capture group. All matches can be extracted too:

```
y.str.extractall(r"(\w+)='(.+?)'")
##              0               1
##   match
## 0 0        name  Sir Launcelot
## 1 0       quest     Seek Grail
## 2 0   favcolour          blue
##   1   favcolour  yel.. Aaargh!
```

Recall that if we just need the grouping part of "(...)", i.e., without the capturing feature, "(?:...)" can be applied.

Also, named capture groups defined like "(?P<name>...)" are supported.

```
y.str.extract("(?:\\w+)='(?P<value>.+?)'")
##           value
## 0  Sir Launcelot
## 1     Seek Grail
## 2          blue
```

### Replacing with Capture Group Matches

When using **re.sub** and **pandas.Series.str.replace**, matches to particular capture groups can be recalled in replacement strings. The match in its entirety is denoted with "\g<0>", then "\g<1>" stores whatever was caught by the first capture group, and "\g<2>" is the match to the second capture group, etc.

```
re.sub(r"(\w+)='(.+?)'", r"\g<2> is a \g<1>", x)
## 'Sir Launcelot is a name, Seek Grail is a quest, blue is a favcolour'
```

Named capture groups can be referred to too:

```
re.sub(r"(?P<key>\w+)='(?P<value>.+?)'",
  r"\g<value> is a \g<key>", x)
## 'Sir Launcelot is a name, Seek Grail is a quest, blue is a favcolour'
```

**Back-Referencing**

Matches to capture groups can also be part of the regexes themselves. In such a context, e.g., "\1" denotes whatever has been consumed by the first capture group.

In general, parsing HTML code with regexes is not recommended, unless it is well-structured (which might be the case if it is generated programmatically; but we can always use the `lxml` package). Despite this, let us consider the following examples:

```
x = "<p><em>spam</em></p><code>eggs</code>"
re.findall(r"<[a-z]+>.*?</[a-z]+>", x)
## ['<p><em>spam</em>', '<code>eggs</code>']
```

This did not match the correct closing HTML tag. But we can make this happen by writing:

```
re.findall(r"(<([a-z]+)>.*?</\2>)", x)
## [('<p><em>spam</em></p>', 'p'), ('<code>eggs</code>', 'code')]
```

This regex guarantees that the match will include all characters between the opening "<tag>" and the corresponding (not: any) closing "</tag>".

Named capture groups can be referenced using the "(?P=name)" syntax:

```
re.findall(r"(<(?P<tagname>[a-z]+)>.*?</(?P=tagname)>)", x)
## [('<p><em>spam</em></p>', 'p'), ('<code>eggs</code>', 'code')]
```

The angle brackets are part of the token.

### 14.4.7 Anchoring

Lastly, let us mention the ways to match a pattern at a given abstract position within a string.

**Matching at the Beginning or End of a String**

"^" and "$" match, respectively, start and end of the string (or each line within a string, if the `re.MULTILINE` flag is set).

```
x = pd.Series(["spam egg", "bacon spam", "spam", "egg spam bacon", "milk"])
rs = ["spam", "^spam", "spam$", "spam$|^spam", "^spam$"]  # regexes to test
```

The five regular expressions match "spam", respectively, anywhere within the string,

at the beginning, at the end, at the beginning or end, and in strings that are equal to the pattern itself. We can check this by calling:

```
pd.concat([x.str.contains(r) for r in rs], axis=1, keys=rs)
##      spam  ^spam  spam$  spam$|^spam  ^spam$
## 0   True   True  False         True   False
## 1   True  False   True         True   False
## 2   True   True   True         True    True
## 3   True  False  False        False   False
## 4  False  False  False        False   False
```

**Exercise 14.6** *Write a regex that does the same job as* ***str.strip***.

**Matching at Word Boundaries**

What is more, "\b" matches at a "word boundary", e.g., near spaces, punctuation marks, or at the start/end of a string (i.e., wherever there is a transition between a word, "\w", and a non-word character, "\W", or vice versa).

In the following example, we match all stand-alone numbers (this regular expression is imperfect, though):

```
re.findall(r"[-+]?\b\d+(?:\.\d+)?\b", "+12, 34.5, -5.3243")
## ['+12', '34.5', '-5.3243']
```

**Looking Behind and Ahead (\*\*)**

There is a way to guarantee that a pattern occurrence begins or ends with a match to a subexpression: "(?<=...)..." is the so-called *look-behind*, whereas "...(?=...)" denotes a *look-ahead*.

```
x = "I like spam, spam, eggs, and spam."
re.findall(r"\b\w+\b(?=[,.])", x)
## ['spam', 'spam', 'eggs', 'spam']
```

This regex captured words that end with a comma or a dot

Moreover, "(?<!...)..." and "...(?!...)" are their *negated* versions (negative look-behind/ahead).

```
re.findall(r"\b\w+\b(?![,.])", x)
## ['I', 'like', 'and']
```

This time, we matched the words that end with neither a comma nor a dot.

## 14.5    Exercises

**Exercise 14.7**  *List some ways to normalise character strings.*

**Exercise 14.8**  *(\*\*) What are the challenges of processing non-English text?*

**Exercise 14.9**  *What are the problems with the* `"[A-Za-z]"` *and* `"[A-z]"` *character sets?*

**Exercise 14.10**  *Name the two ways to turn on case-insensitive regex matching.*

**Exercise 14.11**  *What is a word boundary?*

**Exercise 14.12**  *What is the difference between the* `"^"` *and* `"$"` *anchors?*

**Exercise 14.13**  *When would we prefer using* `"[0-9]"` *instead of* `"\d"`?

**Exercise 14.14**  *What is the difference between the* `"?"`, `"??"`, `"*"`, `"*?"`, `"+"`, *and* `"+?"` *quantifiers?*

**Exercise 14.15**  *Does* `"."` *match all the characters?*

**Exercise 14.16**  *What are named capture groups and how can we refer to the matches thereto in* `re.sub`?

**Exercise 14.17**  *Write a regex that extracts all standalone numbers accepted by Python, including* `12.123`, `-53`, `+1e-9`, `-1.2423e10`, `4.` *and* `.2`.

**Exercise 14.18**  *Write a regex that matches all email addresses.*

**Exercise 14.19**  *Write a regex that matches all URLs starting with* `http://` *or* `https://`.

**Exercise 14.20**  *Cleanse the* `warsaw_weather`[23] *dataset so that it contains analysable numeric data.*

---

[23] https://github.com/gagolews/teaching-data/raw/master/marek/warsaw_weather.csv

# 15

## *Missing, Censored, and Questionable Data*

Up to now, we have been mostly assuming that observations are of decent quality, i.e., trustworthy. It would be nice if that was always the case, but it is not.

In this chapter, we briefly address the most basic methods for dealing with *suspicious* observations: outliers, missing, censored, imprecise, and incorrect data.

### 15.1 Missing Data

Let us consider an excerpt from National Health and Nutrition Examination Survey that we played with in Chapter 12:

```
nhanes = pd.read_csv("https://raw.githubusercontent.com/gagolews/" +
    "teaching-data/master/marek/nhanes_p_demo_bmx_2020.csv",
    comment="#")
nhanes.loc[:, ["BMXWT", "BMXHT", "RIDAGEYR", "BMIHEAD", "BMXHEAD"]].head()
##     BMXWT  BMXHT  RIDAGEYR  BMIHEAD  BMXHEAD
## 0    NaN    NaN         2      NaN      NaN
## 1   42.2  154.7        13      NaN      NaN
## 2   12.0   89.3         2      NaN      NaN
## 3   97.1  160.2        29      NaN      NaN
## 4   13.6    NaN         2      NaN      NaN
```

Some of the columns feature `NaN` (not-a-number) values. They are used here to encode *missing* (not available) data. Previously, we decided not to be bothered by them: a shy call to **dropna** resulted in their removal. But we are curious now.

The reasons behind why some items are missing might be numerous, for instance:

- a participant did not know the answer to a given question;

- someone refused to answer a given question;

- a person did not take part in the study anymore (attrition, death, etc.);

- an item was not applicable (e.g., number of minutes spent cycling weekly when someone answered they did not learn to ride a bike yet);

- a piece of information was not collected, e.g., due to the lack of funding or a failure of a piece of equipment.

### 15.1.1    Representing and Detecting Missing Values

Sometimes missing values are specially encoded, especially in CSV files, e.g., with -1, 0, 9999, numpy.inf, -numpy.inf, or None, strings such as "NA", "N/A", "Not Applicable", "---" – we should always inspect our datasets carefully. To assure consistent representation, we can convert them to NaN (as in: numpy.nan) in numeric (floating-point) columns or to Python's None otherwise.

Vectorised functions such as **numpy.isnan** (or, more generally, **numpy.isfinite**) and **pandas.isnull** as well as **isna** methods for the DataFrame and Series classes can be used to verify whether an item is missing or not.

For instance, here are the counts and proportions of missing values in selected columns of nhanes:

```
nhanes.isna().apply([np.sum, np.mean]).T.nlargest(5, "sum")  # top 5 only
##              sum      mean
## BMIHEAD   14300.0  1.000000
## BMIRECUM  14257.0  0.996993
## BMIHT     14129.0  0.988042
## BMXHEAD   13990.0  0.978322
## BMIHIP    13924.0  0.973706
```

Looking at the column descriptions on the data provider's website[1], for example, BMIHEAD stands for "Head Circumference Comment", whereas BMXHEAD is "Head Circumference (cm)", but these were only collected for infants.

**Exercise 15.1**    *Read the column descriptions (refer to the comments in the CSV file for the relevant URLs) to identify the possible reasons for some of the records in nhanes being missing.*

**Exercise 15.2**    *Learn about the difference between the **pandas.DataFrameGroupBy.size** and **pandas.DataFrameGroupBy.count** methods.*

### 15.1.2    Computing with Missing Values

Our using NaN to denote a missing piece of information is merely an ugly (but functional) hack[2]. The original use case for not-a-number is to represent the results of incorrect operations, e.g., logarithms of negative numbers or subtracting two infinite entities. We thus need extra care when handling them.

Generally, arithmetic operations on missing values yield a result that is undefined as well:

---

[1] https://wwwn.cdc.gov/Nchs/Nhanes/2017-2018/P_BMX.htm
[2] (*) The R environment, on the other hand, has built-in seamless support for missing values. It quite consistently honours the rule that operations on missing values yield an undefined result.

```
np.nan + 2   # "don't know" + 2 == "don't know"
## nan
np.mean([1, np.nan, 2, 3])
## nan
```

There are versions of certain aggregation functions that ignore missing values whatsoever: **numpy.nanmean**, **numpy.nanmin**, **numpy.nanmax**, **numpy.nanpercentile**, **numpy.nanstd**, etc.

```
np.nanmean([1, np.nan, 2, 3])
## 2.0
```

Regrettably, running these aggregation functions directly on `Series` objects ignores missing entities by default. Compare an application of **numpy.mean** on a `Series` instance vs on a vector:

```
x = nhanes.head().loc[:, "BMXHT"]   # some example Series, whatever
np.mean(x), np.mean(np.array(x))
## (134.73333333333332, nan)
```

This is quite unfortunate behaviour as this way we might miss (sic!) the presence of missing values. Therefore, it is crucial to have the dataset carefully pre-inspected.

Also, `NaN` is of the floating-point type. As a consequence, it cannot be present in, amongst others, logical vectors.

```
x   # preview
## 0       NaN
## 1     154.7
## 2      89.3
## 3     160.2
## 4       NaN
## Name: BMXHT, dtype: float64
y = (x > 100)
y
## 0    False
## 1     True
## 2    False
## 3     True
## 4    False
## Name: BMXHT, dtype: bool
```

Unfortunately, comparisons against missing values yield `False`, instead of the more semantically valid missing value. Hence, if we want to retain the missingness information (we do not know if a missing value is greater than 100), we need to do it manually:

```
y = y.astype("object")  # required for numpy vectors, not for pandas Series
y[np.isnan(x)] = None
y
## 0     None
## 1     True
## 2    False
## 3     True
## 4     None
## Name: BMXHT, dtype: object
```

**Exercise 15.3**  *Read the **pandas** documentation[3] about missing value handling.*

### 15.1.3   Missing at Random or Not?

At a general level (from the mathematical modelling perspective), we may distinguish between a few missingness patterns; see [72]:

- *missing completely at random*: reasons are unrelated to data and probabilities of cases' being missing are all the same;

- *missing at random*: there are different probabilities of being missing within distinct groups (e.g., ethical data scientists might tend to refuse to answer specific questions);

- *missing not at random*: due to reasons unknown to us (e.g., data was collected at different times, there might be significant differences within the groups that we cannot easily identify, e.g., amongst participants with a background in mathematics where we did not ask about education or occupation).

It is important to try to determine the reason for missingness, because this will usually imply the kinds of techniques that are suitable in specific cases.

### 15.1.4   Discarding Missing Values

We may try removing (discarding) the rows or columns that feature at least one, some, or too many missing values. Nonetheless, such a scheme will obviously not work for small datasets, where each observation is precious[4].

Also, we should not exercise data removal in situations where missingness is conditional (e.g., data only available for infants) or otherwise group-dependent (not completely at random), as, for example, it might result in an imbalanced dataset.

**Exercise 15.4**  *With the `nhanes_p_demo_bmx_2020`[5] dataset, perform what follows.*

---

[3] https://pandas.pydata.org/pandas-docs/stable/user_guide/missing_data.html

[4] On the other hand, if we want to infer from small datasets, we should ask ourselves whether this is a good idea at all… It might be better to refrain from any data analysis than to come up with conclusions that are likely to be unjustified.

[5] https://github.com/gagolews/teaching-data/raw/master/marek/nhanes_p_demo_bmx_2020.csv

1. *Remove all columns that are comprised of missing values only.*

2. *Remove all columns that are made of more than 20% missing values.*

3. *Remove all rows that only consist of missing values.*

4. *Remove all rows that bear at least one missing value.*

5. *Remove all columns that feature at least one missing value.*

Hint: `pandas.DataFrame.dropna` might be useful in the simplest cases, and `numpy.isnan` or `pandas.DataFrame.isna` with `loc[...]` or `iloc[...]` can be applied otherwise.

### 15.1.5 Mean Imputation

When we cannot afford or it is inappropriate/inconvenient to proceed with the removal of missing observations or columns, we may try applying some missing value *imputation* techniques. Let us be clear, though: this is merely a replacement thereof by some *hopefully* adequate guesstimates.

---

**Important**   In all kinds of reports from data analysis, we need to be explicit about the way we handle the missing values. This is because sometimes they might strongly affect the results.

---

Let us consider an example vector with missing values, comprised of heights of the adult participants of the NHANES study.

```
x = nhanes.loc[nhanes.loc[:, "RIDAGEYR"] >= 18, "BMXHT"]
```

The simplest approach is to replace each missing value with the corresponding column's mean. This does not change the overall average but decreases the variance.

```
xi = x.copy()
xi[np.isnan(xi)] = np.nanmean(xi)
```

Similarly, we could consider replacing missing values with the median, or – in the case of categorical data – the mode.

Furthermore, we expect heights to differ, on average, between sexes. Consequently, another basic imputation option is to replace the missing values with the corresponding within-group averages:

```
xg = x.copy()
g = nhanes.loc[nhanes.loc[:, "RIDAGEYR"] >= 18, "RIAGENDR"]
xg[np.isnan(xg) & (g == 1)] = np.nanmean(xg[g == 1])  # male
xg[np.isnan(xg) & (g == 2)] = np.nanmean(xg[g == 2])  # female
```

Unfortunately, whichever imputation method we choose, will artificially distort the data distribution and introduce some kind of bias; see Figure 15.1 for the histograms of x, xi, and xg. These effects can be obscured if we increase the histogram bins' widths, but they will still be present in the data. No surprise here: we added to the sample many identical values.

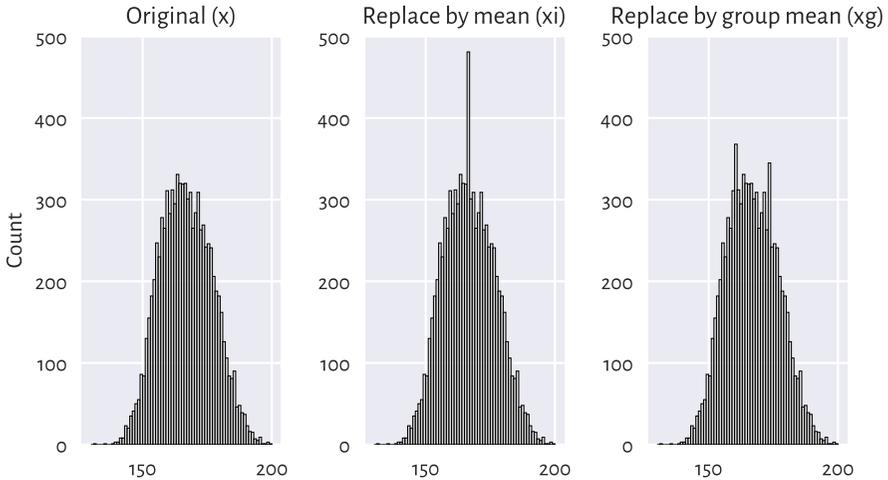

Figure 15.1: Mean imputation distorts the data distribution

**Exercise 15.5** *With the* nhanes_p_demo_bmx_2020[6] *dataset, perform what follows.*

1. *For each numerical column, replace all missing values with the column averages.*

2. *For each categorical column, replace all missing values with the column modes.*

3. *For each numerical column, replace all missing values with the averages corresponding to a patient's sex (as given by the* RIAGENDR *column).*

### 15.1.6  Imputation by Classification and Regression (*)

We can easily implement a missing value imputer based on averaging data from an observation's non-missing nearest neighbours; compare Section 9.2.1 and Section 12.3.1. This is an extension of the simple idea of finding the most *similar* observation (with respect to chosen criteria) to a given one and then borrowing non-missing measurements from it.

More generally, different regression or classification models can be built on non-missing data (training sample) and then the missing observations can be replaced by the values predicted by those models.

---

[6] https://github.com/gagolews/teaching-data/raw/master/marek/nhanes_p_demo_bmx_2020.csv

**Note** (\*\*) Rubin (e.g., in [54]) suggests the use of a procedure called *multiple imputation* (see also [80]), where copies of the original datasets are created, missing values are imputed by sampling from some estimated distributions, the inference is made, and then the results are aggregated. An example implementation of such an algorithm is available in `sklearn.impute.IterativeImputer`.

## 15.2 Censored and Interval Data (\*)

Censored data frequently appear in the context of reliability, risk analysis, and biostatistics, where the observed objects might *fail* (e.g., break down, die, withdraw; compare, e.g., [57]). Our introductory course cannot obviously cover everything, but a beginner analyst should at least be aware of such data being a thing. In particular:

- *right-censored* data: we only know that the actual value is above the recorded one (e.g., we stopped the experiment on the reliability of light bulbs after 1,000 hours, so those which still work will not have their time-of-failure precisely known);

- *left-censored* data: the true observation is below the recorded one, e.g., we observe a component's failure, but we do not know for how long it has been in operation before the study has started.

In such cases, the recorded datum of, say, 1,000, can essentially mean $[1000, \infty)$, $[0, 1000]$, or $(-\infty, 1000]$.

There might also be instances where we know that a value is in some interval $[a, b]$. There are numerical libraries that deal with *interval computations*, and some data analysis methods exist for dealing with such a scenario.

## 15.3 Incorrect Data

*Missing data* can already be marked in a given sample. But we also might be willing to mark some existing values as missing, e.g., when they are incorrect. For example:

- for text data, misspelled words;

- for spatial data, GPS coordinates of places out of this world, non-existing zip codes, or invalid addresses;

- for date-time data, misformatted date-time strings, incorrect dates such as "29 February 2011", an event's start date being after the end date;

- for physical measurements, observations that do not meet specific constraints, e.g., negative ages, or heights of people over 300 centimetres;

- IDs of entities that simply do not exist (e.g., unregistered or deleted clients' accounts);

and so forth.

To be able to identify and handle incorrect data, we need specific knowledge of a particular domain. Optimally, basic data validation techniques should already be employed on the data collection stage, for instance when a user submits an online form.

There can be many tools that can assist us with identifying erroneous observations, e.g., spell checkers such as `hunspell`[7].

For smaller datasets, observations can also be inspected manually. In other cases, we might have to develop our own algorithms for detecting such bugs in data.

**Exercise 15.6**  *Given some data frame with numeric columns only, perform what follows.*

1. *Check if all numeric values in each column are between 0 and 1,000.*

2. *Check if all values in each column are unique.*

3. *Verify that all the rowwise sums add up to 1.0 (up to a small numeric error).*

4. *Check if the data frame consists of 0s and 1s only. Provided that this is the case, verify that for each row, if there is a 1 in some column, then all the columns to the right are filled with 1s too.*

Many data validation methods can be reduced to operations on strings; see Chapter 14. They may be as simple as writing a single regular expression or checking if a label is in a dictionary of possible values but also as difficult as writing your own parser for a custom context-sensitive grammar.

**Exercise 15.7**  *Once we import the data fetched from dirty sources, relevant information will have to be extracted from raw text, e.g., strings like "1" should be converted to floating-point numbers. Below we suggest several tasks that can aid in developing data validation skills involving some operations on text.*

*Given an example data frame with text columns (manually invented, please be creative), perform what follows.*

1. *Remove trailing and leading whitespaces from each string.*

2. *Check if all strings can be interpreted as numbers, e.g., "23.43".*

3. *Verify if a date string in the YYYY-MM-DD format is correct.*

4. *Determine if a date-time string in the YYYY-MM-DD hh:mm:ss format is correct.*

5. *Check if all strings are of the form (+NN) NNN-NNN-NNN or (+NN) NNNN-NNN-NNN, where N denotes any digit (valid telephone numbers).*

---

[7] https://hunspell.github.io/

6. *Inspect whether all strings are valid country names.*

7. *(\*) Given a person's date of birth, sex, and Polish ID number PESEL[8], check if that ID is correct.*

8. *(\*) Determine if a string represents a correct International Bank Account Number (IBAN[9]) (note that IBANs feature two check digits).*

9. *(\*) Transliterate text to ASCII, e.g.,* `"żółty ©"` *to* `"zolty (C)"`.

10. *(\*\*) Using an external spell checker, determine if every string is a valid English word.*

11. *(\*\*) Using an external spell checker, ascertain that every string is a valid English noun in the singular form.*

12. *(\*\*) Resolve all abbreviations by means of a custom dictionary, e.g.,* `"Kat."` → `"Katherine"`, `"Gr."` → `"Grzegorz"`.

---

## 15.4  Outliers

Another group of inspectionworthy observations consists of *outliers*. We can define them as the samples that reside in the areas of substantially lower density than their neighbours.

Outliers might be present due to an error, or their being otherwise anomalous, but they may also simply be interesting, original, or novel. After all, statistics does not give any meaning to data items; humans do.

What we do with outliers is a separate decision. We can get rid of them, correct them, replace them with a missing value (and then possibly impute), or analyse them separately. In particular, there is a separate subfield in statistics called extreme value theory that is interested in predicting the distribution of very large observations (e.g., for modelling floods, extreme rainfall, or temperatures); see, e.g., [3]. But this is a topic for a more advanced course. By then, let us stick with simpler settings.

### 15.4.1  The 3/2 IQR Rule for Normally-Distributed Data

For unidimensional data (or individual columns in matrices and data frames), the first few smallest and largest observations should usually be inspected manually. It might be, for instance, the case that someone accidentally entered a patient's height in metres instead of centimetres – such cases are easily detectable. A data scientist is like a detective.

Let us recall the rule of thumb discussed in the section on box-and-whisker plots (Section 5.1.4). For data that are expected to come from a normal distribution, everything that does not fall into the interval $[Q_1 - 1.5\text{IQR}, Q_3 + 1.5\text{IQR}]$ can be considered suspicious. This definition is based on quartiles only, so it should not be affected by potential outliers (they are robust aggregates). Plus, the magic constant 1.5 is nicely round and thus easy to memorise (good for some practitioners). It is not too small and not too large; for the normal distribution $N(\mu, \sigma)$, the above interval corresponds to roughly $[\mu - 2.698\sigma, \mu + 2.698\sigma]$, and the probability of obtaining a value outside of it is ca. 0.7%. In other words, for a sample of size 1,000 that is *truly* normally distributed (not contaminated by anything), only seven observations will be flagged. It is not a problem to inspect them by hand.

---

[8] https://en.wikipedia.org/wiki/PESEL
[9] https://en.wikipedia.org/wiki/International_Bank_Account_Number

---

**Note**   (*) We can of course choose a different threshold. For instance, for the normal distribution $N(10, 1)$, even though the probability of observing a value greater than 15 is *theoretically* non-zero, it is smaller 0.000029%, so it is sensible to treat this observation as suspicious. On the other hand, we do not want to mark too many observations as outliers because inspecting them manually will be too labour-intense.

---

**Exercise 15.8**   *For each column in* `nhanes_p_demo_bmx_2020`[10], *inspect a few smallest and largest observations and see if they make sense.*

**Exercise 15.9**   *Perform the above separately for data in each group as defined by the* `RIAGENDR` *column.*

### 15.4.2   Unidimensional Density Estimation (*)

For skewed distributions such as the ones representing incomes, there might be nothing wrong, at least statistically speaking, with very large isolated observations.

For well-separated multimodal distributions on the real line, outliers may sometimes also fall in between the areas of high density.

**Example 15.10**   *That neither box plots themselves, nor the* 1.5IQR *rule might not be ideal tools for multimodal data is exemplified in Figure 15.2, where we have a mixture of N(10, 1) and N(25, 1) samples and 4 potential outliers at 0, 15, 45, and 50.*

```
x = np.loadtxt("https://raw.githubusercontent.com/gagolews/" +
    "teaching-data/master/marek/blobs2.txt")
plt.subplot(1, 2, 1)
sns.boxplot(data=x, orient="h", color="lightgray")
plt.yticks([])
plt.subplot(1, 2, 2)
sns.histplot(x, binwidth=1, color="lightgray")
plt.show()
```

Fixed-radius search techniques, which we discussed in Section 8.4, can be used for estimating the underlying probability density function. Given a data sample $x =$

---

[10] https://github.com/gagolews/teaching-data/raw/master/marek/nhanes_p_demo_bmx_2020.csv

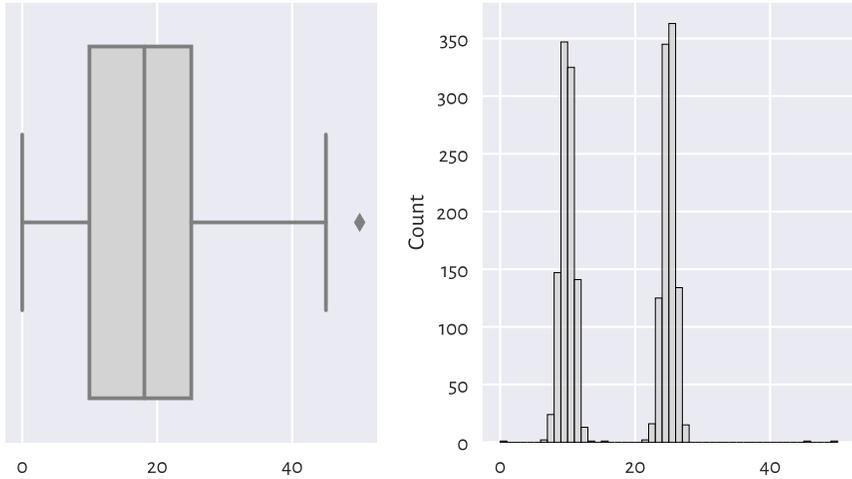

Figure 15.2: With box plots, we may fail to detect some outliers

$(x_1, \ldots, x_n)$, let us consider[11]:

$$\hat{f}_r(z) = \frac{1}{2rn} \sum_{i=1}^{n} |B_r(z)|,$$

where $|B_r(z)|$ denotes the number of observations from $\boldsymbol{x}$ whose distance to $z$ is not greater than $r$, i.e., fall into the interval $[z - r, z + r]$.

```
n = len(x)
r = 1  # radius – feel free to play with different values
import scipy.spatial
t = scipy.spatial.KDTree(x.reshape(-1, 1))
dx = pd.Series(t.query_ball_point(x.reshape(-1, 1), r)).str.len() / (2*r*n)
dx[:6]  # preview
## 0    0.000250
## 1    0.116267
## 2    0.116766
## 3    0.166667
## 4    0.076098
## 5    0.156188
## dtype: float64
```

Then, points in the sample lying in low-density regions (i.e., all $x_i$ such that $\hat{f}_r(x_i)$ is small) can be flagged for further inspection:

---

[11] This is an instance of a kernel density estimator, with the simplest kernel – a rectangular one.

```
x[dx < 0.001]
## array([ 0.       , 13.57157922, 15.       , 45.       , 50.       ])
```

See Figure 15.3 for an illustration of $\hat{f}_r$. Of course, $r$ should be chosen with care – just like the number of bins in a histogram.

```
sns.histplot(x, binwidth=1, stat="density", color="lightgray")
z = np.linspace(np.min(x)-5, np.max(x)+5, 1001)
dz = pd.Series(t.query_ball_point(z.reshape(-1, 1), r)).str.len() / (2*r*n)
plt.plot(z, dz, label=f"density estimator ($r={r}$)")
plt.show()
```

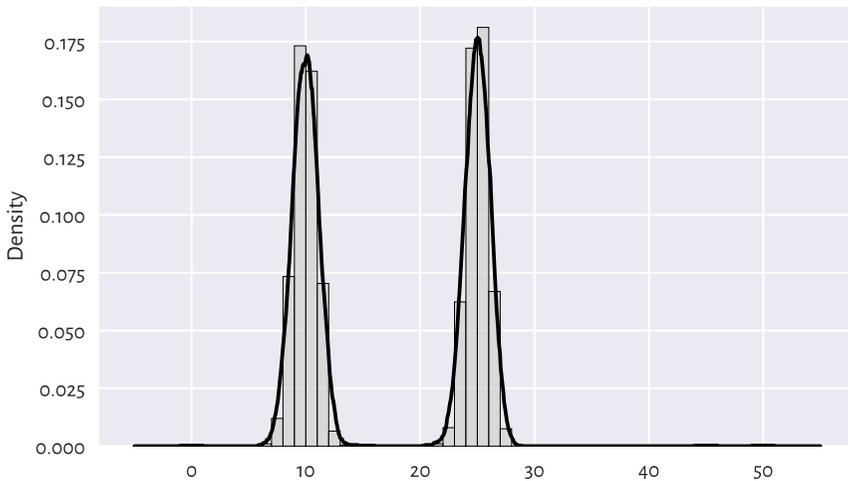

Figure 15.3: Density estimation based on fixed-radius search

### 15.4.3 Multidimensional Density Estimation (*)

By far we should have become used to the fact that unidimensional data projections might lead to our losing too much information. Some values can seem perfectly fine when they are considered in isolation, but already plotting them in 2D reveals that the reality is more complex than that.

Consider the following example dataset and the depiction of the distributions of its two natural projections in Figure 15.4.

```
X = np.loadtxt("https://raw.githubusercontent.com/gagolews/" +
    "teaching-data/master/marek/blobs1.txt", delimiter=",")
plt.figure(figsize=(plt.rcParams["figure.figsize"][0], )*2)  # width=height
plt.subplot(2, 2, 1)
```

```
sns.boxplot(data=X[:, 0], orient="h", color="lightgray")
plt.yticks([0], ["X[:, 0]"])
plt.subplot(2, 2, 2)
sns.histplot(X[:, 0], bins=20, color="lightgray")
plt.title("X[:, 0]")
plt.subplot(2, 2, 3)
sns.boxplot(data=X[:, 1], orient="h", color="lightgray")
plt.yticks([0], ["X[:, 1]"])
plt.subplot(2, 2, 4)
sns.histplot(X[:, 1], bins=20, color="lightgray")
plt.title("X[:, 1]")
plt.show()
```

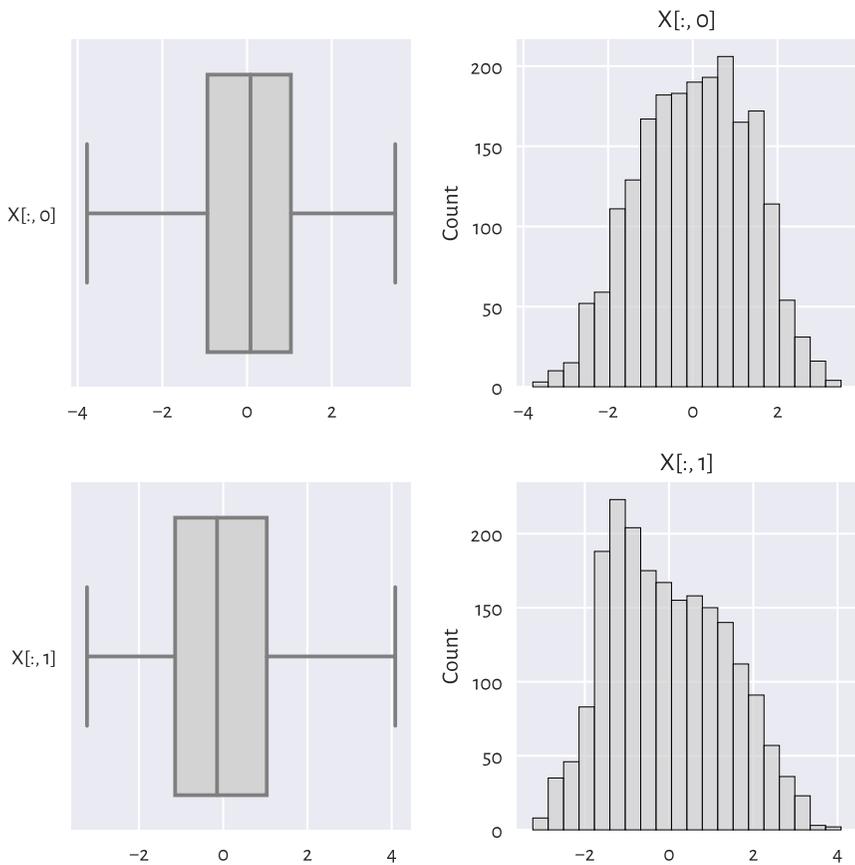

Figure 15.4: One-dimensional projections of the `blobs1` dataset

There is nothing suspicious here. Or is there?

The scatterplot in Figure 15.5 reveals that the data consist of two quite well-separable blobs:

```
plt.plot(X[:, 0], X[:, 1], "o")
plt.axis("equal")
plt.show()
```

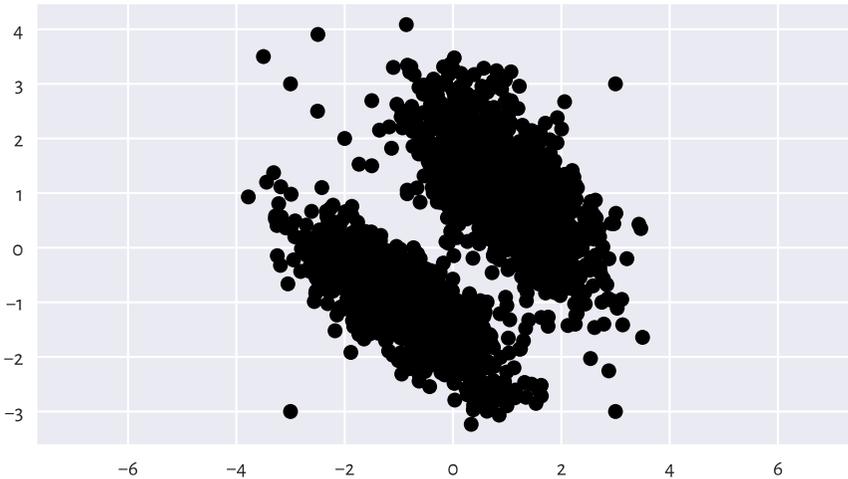

Figure 15.5: Scatterplot of the `blobs1` dataset

There are a few observations that we might mark as outliers. The truth is that yours truly injected eight junk points at the very end of the dataset. Ha.

```
X[-8:, :]
## array([[-3. ,  3. ],
##        [ 3. ,  3. ],
##        [ 3. , -3. ],
##        [-3. , -3. ],
##        [-3.5,  3.5],
##        [-2.5,  2.5],
##        [-2. ,  2. ],
##        [-1.5,  1.5]])
```

Handling multidimensional data requires slightly more sophisticated methods. A quite straightforward approach is to check if there are any points within an observation's radius of some assumed size $r > 0$. If that is not the case, we may consider it an outlier. This is a variation on the aforementioned unidimensional density estimation approach[12].

**Example 15.11** *Consider the following code chunk:*

```
t = scipy.spatial.KDTree(X)
n = t.query_ball_point(X, 0.2)  # r=0.2 (radius) – play with it yourself
c = np.array(pd.Series(n).str.len())
c[[0, 1, -2, -1]]  # preview
## array([42, 30,  1,  1])
```

*c[i] gives the number of points within X[i, :]'s r-radius (with respect to the Euclidean distance), including the point itself. Consequently, c[i]==1 denotes a potential outlier; see Figure 15.6 for an illustration.*

```
plt.plot(X[c > 1, 0], X[c > 1, 1], "o", label="normal point")
plt.plot(X[c == 1, 0], X[c == 1, 1], "v", label="outlier")
plt.axis("equal")
plt.legend()
plt.show()
```

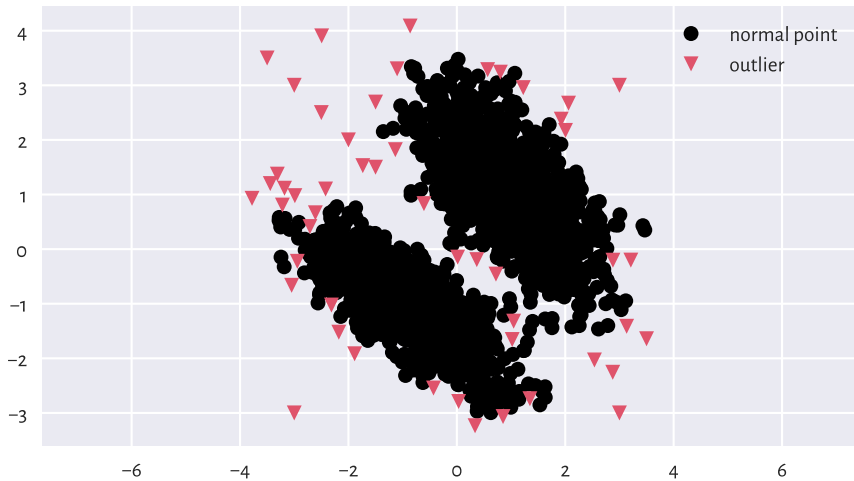

Figure 15.6: Outlier detection based on a fixed-radius search for the `blobs1` dataset

---

[12] (**) We can easily normalise the outputs to get a true 2D kernel density estimator, but multivariate statistics is beyond the scope of this course. In particular, that data might have fixed marginal distributions (projections onto 1D) but their multidimensional images might be very different is beautifully described by the copula theory; see [59].

## 15.5 Exercises

**Exercise 15.12** *How can missing values be represented in **numpy** and **pandas**?*

**Exercise 15.13** *Explain some basic strategies for dealing with missing values in numeric vectors.*

**Exercise 15.14** *Why we should be very explicit about the way we handle missing and other suspicious data? Is it a good idea to mark as missing (or remove completely) the observations that we dislike or otherwise deem* inappropriate, controversial, dangerous, incompatible with our political views, *etc.?*

**Exercise 15.15** *Is replacing missing values with the sample arithmetic mean for income data (as in, e.g., the* `uk_income_simulated_2020`[13] *dataset) a sensible strategy?*

**Exercise 15.16** *What are the differences between data missing completely at random, missing at random, and missing not at random?*

**Exercise 15.17** *List some basic strategies for dealing with data that might contain outliers.*

---

[13] https://github.com/gagolews/teaching-data/raw/master/marek/uk_income_simulated_2020.txt

# 16

## *Time Series*

So far, we have been using `numpy` and `pandas` mostly for storing:

- *independent* measurements, where each row gives, e.g., weight, height, … records of a different subject; we often consider these a sample of a representative subset of one or more populations, each recorded at a particular point in time;

- data summaries to be reported in the form of tables or figures, e.g., frequency distributions giving counts for the corresponding categories or labels.

In this chapter, we will explore the most basic concepts related to the wrangling of *time series*, i.e., signals indexed by discrete time. Usually, a time series is a sequence of measurements sampled at equally spaced moments, e.g., a patient's heart rate probed every second, daily average currency exchange rates, or highest yearly temperatures recorded in some location.

## 16.1 Temporal Ordering and Line Charts

Consider the midrange[1] daily temperatures in degrees Celsius at the Spokane International Airport (Spokane, WA, US) between 1889-08-01 (first observation) and 2021-12-31 (last observation).

```
temps = np.loadtxt("https://raw.githubusercontent.com/gagolews/" +
    "teaching-data/master/marek/spokane_temperature.txt")
```

Let us preview the December 2021 data:

```
temps[-31:]  # last 31 days
## array([ 11.9,    5.8,    0.6,    0.8,   -1.9,   -4.4,   -1.9,    1.4,   -1.9,
##         -1.4,    3.9,    1.9,    1.9,   -0.8,   -2.5,   -3.6,  -10. ,   -1.1,
##         -1.7,   -5.3,   -5.3,   -0.3,    1.9,   -0.6,   -1.4,   -5. ,   -9.4,
##        -12.8,  -12.2,  -11.4,  -11.4])
```

Here are some data aggregates for the whole sample. First, the popular quantiles:

---

[1] Note that midrange, being the mean of the lowest and the highest observed temperature on a given day, is not a particularly good estimate of the average daily reading. This dataset is considered for illustrational purposes only.

```
np.quantile(temps, [0, 0.25, 0.5, 0.75, 1])
## array([-26.9,   2.2,    8.6,   16.4,   33.9])
```

Then, the arithmetic mean and standard deviation:

```
np.mean(temps), np.std(temps)
## (8.990273958441023, 9.16204388619955)
```

A graphical summary of the data distribution is depicted in Figure 16.1.

```
sns.violinplot(data=temps, orient="h", color="lightgray")
plt.yticks([])
plt.show()
```

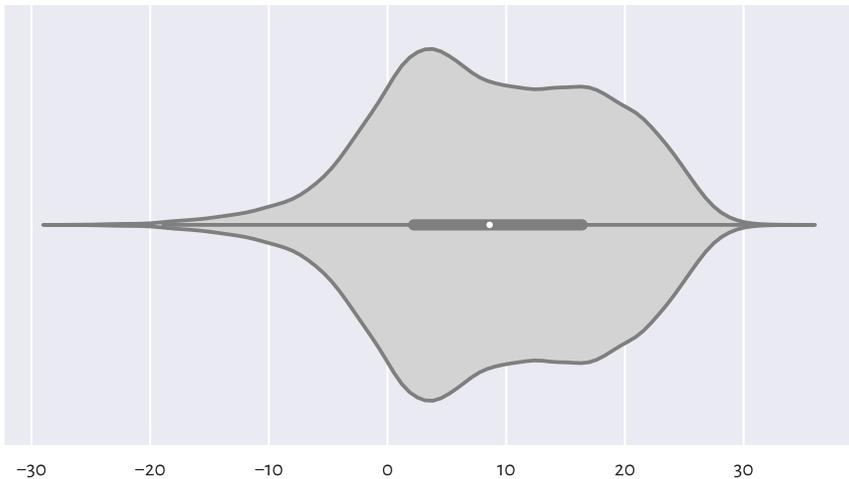

Figure 16.1: Distribution of the midrange daily temperatures in Spokane in the period 1889–2021; observations are treated as a bag of unrelated items (temperature on a "randomly chosen day" in a version of planet Earth where there is no climate change)

When computing data aggregates or plotting histograms, the order of elements does not matter. Contrary to the case of the *independent* measurements, vectors representing time series do not have to be treated simply as mixed bags of unrelated items.

**Important**   In time series, for any given item $x_i$, its neighbouring elements $x_{i-1}$ and $x_{i+1}$ denote the recordings occurring directly before and after it. We can use this *temporal ordering* to model how *consecutive* measurements *depend* on each other, de-scribe how they change over time, forecast future values, detect seasonal and long-time trends, and so forth.

In Figure 16.2, we depict the data for 2021, plotted as a function of time. What we see is often referred to as a *line chart*: data points are connected by straight line segments. There are some visible seasonal variations, such as, well, obviously, that winter is colder than summer. There is also some natural variability on top of seasonal patterns typical for the Northern Hemisphere.

```
plt.plot(temps[-365:])
plt.xticks([0, 181, 364], ["2021-01-01", "2021-07-01", "2021-12-31"])
plt.show()
```

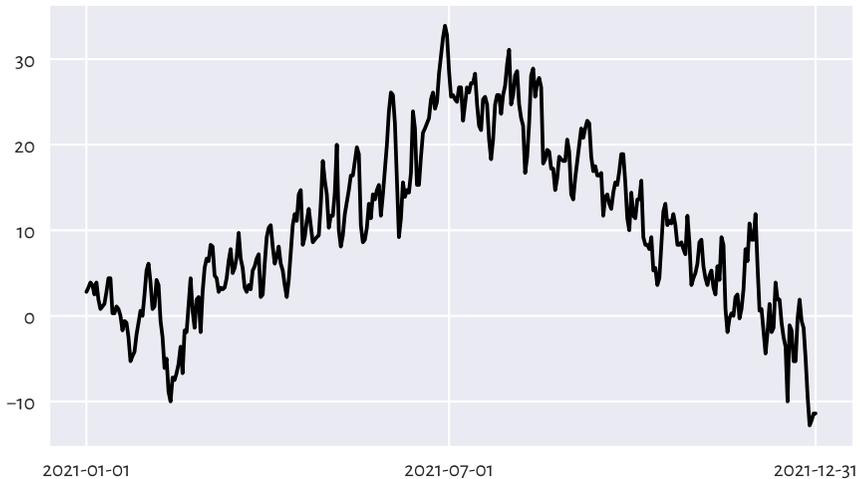

Figure 16.2: Line chart of midrange daily temperatures in Spokane for 2021

## 16.2  Working with Datetimes and Timedeltas

### 16.2.1  Representation: The UNIX Epoch

`numpy.datetime64`[2] is a type to represent datetimes. Usually, we will be creating dates from strings, for instance:

---

[2] https://numpy.org/doc/stable/reference/arrays.datetime.html

```
d = np.array([
    "1889-08-01", "1970-01-01", "1970-01-02", "2021-12-31", "today"
], dtype="datetime64")
d
## array(['1889-08-01', '1970-01-01', '1970-01-02', '2021-12-31',
##         '2022-08-18'], dtype='datetime64[D]')
```

Similarly with datetimes:

```
dt = np.array(["1970-01-01T02:01:05", "now"], dtype="datetime64")
dt
## array(['1970-01-01T02:01:05', '2022-08-18T07:55:43'],
##         dtype='datetime64[s]')
```

---

**Important**  Internally, the above are represented as the number of days or seconds since the so-called *Unix Epoch*, 1970-01-01T00:00:00 in the UTC time zone.

---

Let us verify the above statement:

```
d.astype(float)
## array([-2.9372e+04,  0.0000e+00,  1.0000e+00,  1.8992e+04,  1.9222e+04])
dt.astype(float)
## array([7.26500000e+03, 1.66080934e+09])
```

When we think about it for a while, this is exactly what we expected.

**Exercise 16.1**  *(\*) Write a regular expression that extracts all dates in the YYYY-MM-DD format from a (possibly long) string and converts them to datetime64.*

### 16.2.2   Time Differences

Computing datetime differences is possible thanks to the `numpy.timedelta64` objects:

```
d - np.timedelta64(1, "D")   # minus 1 Day
## array(['1889-07-31', '1969-12-31', '1970-01-01', '2021-12-30',
##         '2022-08-17'], dtype='datetime64[D]')
dt + np.timedelta64(12, "h")   # plus 12 hours
## array(['1970-01-01T14:01:05', '2022-08-18T19:55:43'],
##         dtype='datetime64[s]')
```

Also, **numpy.arange** (and **pandas.date_range**) can be used to generate sequences of equidistant datetimes:

```
dates = np.arange("1889-08-01", "2022-01-01", dtype="datetime64[D]")
dates[:3]   # preview
```

```
## array(['1889-08-01', '1889-08-02', '1889-08-03'], dtype='datetime64[D]')
dates[-3:]  # preview
## array(['2021-12-29', '2021-12-30', '2021-12-31'], dtype='datetime64[D]')
```

### 16.2.3 Datetimes in Data Frames

Dates and datetimes can of course be emplaced in **pandas** data frames:

```
spokane = pd.DataFrame(dict(
    date=np.arange("1889-08-01", "2022-01-01", dtype="datetime64[D]"),
    temp=temps
))
spokane.head()
##          date  temp
## 0 1889-08-01  21.1
## 1 1889-08-02  20.8
## 2 1889-08-03  22.2
## 3 1889-08-04  21.7
## 4 1889-08-05  18.3
```

Interestingly, if we ask the `date` column to become the data frame's index (i.e., row labels), we will be able select date ranges quite easily with **loc**[...] and string slices (refer to the manual of `pandas.DateTimeIndex` for more details).

```
spokane.set_index("date").loc["2021-12-25":, :].reset_index()
##          date  temp
## 0 2021-12-25  -1.4
## 1 2021-12-26  -5.0
## 2 2021-12-27  -9.4
## 3 2021-12-28 -12.8
## 4 2021-12-29 -12.2
## 5 2021-12-30 -11.4
## 6 2021-12-31 -11.4
```

**Example 16.2** *Based on the above, we can plot the data for the last five years quite easily; see Figure 16.3. Note that the x-axis labels are generated automatically.*

```
x = spokane.set_index("date").loc["2017-01-01":, "temp"]
plt.plot(x)
plt.show()
```

The **pandas.to_datetime** function can also convert arbitrarily formatted date strings, e.g., "MM/DD/YYYY" or "DD.MM.YYYY" to Series of datetime64s.

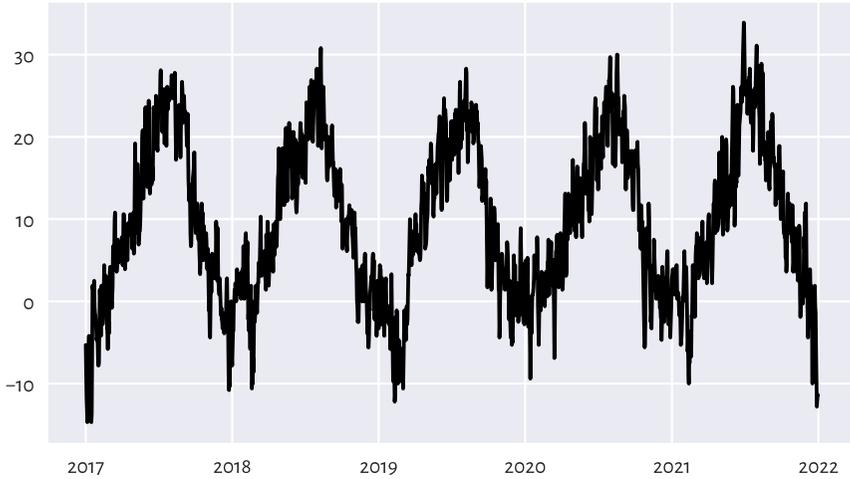

Figure 16.3: Line chart of midrange daily temperatures in Spokane for 2017–2021

```
dates = ["05.04.1991", "14.07.2022", "21.12.2042"]
dates = pd.Series(pd.to_datetime(dates, format="%d.%m.%Y"))
dates
## 0    1991-04-05
## 1    2022-07-14
## 2    2042-12-21
## dtype: datetime64[ns]
```

**Exercise 16.3** *From the* `birth_dates`[3] *dataset, select all people less than 18 years old (as of the current day).*

Several datetime functions and related properties can be referred to via the **pandas. Series.dt** accessor (similarly to **pandas.Series.str** discussed in Chapter 14). In particular, they deliver a convenient means for extracting different date or time fields, such as:

```
dates_ymd = pd.DataFrame(dict(
    year  = dates.dt.year,
    month = dates.dt.month,
    day   = dates.dt.day
))
dates_ymd
##    year  month  day
## 0  1991      4    5
```

*(continues on next page)*

```
## 1   2022    7   14
## 2   2042   12   21
```

Interestingly, **pandas.to_datetime** can also convert data frames with columns named year, month, day, etc., back to datetime objects directly:

```
pd.to_datetime(dates_ymd)
## 0    1991-04-05
## 1    2022-07-14
## 2    2042-12-21
## dtype: datetime64[ns]
```

**Example 16.4**  *Let us extract the month and year parts of dates to compute the average monthly temperatures it the last 50-ish years:*

```
x = spokane.set_index("date").loc["1970":, ].reset_index()
mean_monthly_temps = x.groupby([
    x.date.dt.year.rename("year"),
    x.date.dt.month.rename("month")
]).temp.mean().unstack()
mean_monthly_temps.head().round(1)  # preview
## month    1    2    3    4     5     6     7     8     9    10   11   12
## year
## 1970   -3.4  2.3  2.8  5.3  12.7  19.0  22.5  21.2  12.3  7.2  2.2 -2.4
## 1971   -0.1  0.8  1.7  7.4  13.5  14.6  21.0  23.4  12.9  6.8  1.9 -3.5
## 1972   -5.2 -0.7  5.2  5.6  13.8  16.6  20.0  21.7  13.0  8.4  3.5 -3.7
## 1973   -2.8  1.6  5.0  7.8  13.6  16.7  21.8  20.6  15.4  8.4  0.9  0.7
## 1974   -4.4  1.8  3.6  8.0  10.1  18.9  19.9  20.1  15.8  8.9  2.4 -0.8
```

*Figure 16.4 depicts these data on a heatmap. We rediscover the ultimate truth that winters are cold, whereas in the summertime the living is easy, what a wonderful world.*

```
sns.heatmap(mean_monthly_temps)
plt.show()
```

## 16.3  Basic Operations

### 16.3.1  Iterative Differences and Cumulative Sums Revisited

Recall from Section 5.5.1 the **numpy.diff** function and its almost-inverse, **numpy. cumsum**. The former can turn a time series into a vector of *relative changes* (*deltas*), $\Delta_i = x_{i+1} - x_i$.

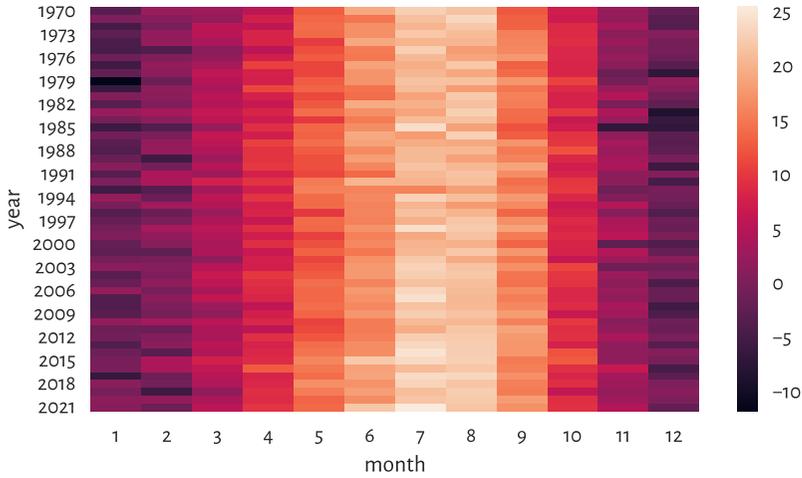

Figure 16.4: Average monthly temperatures

```
x = temps[-7:]  # last 7 days
x
## array([ -1.4,   -5. ,   -9.4, -12.8, -12.2, -11.4, -11.4])
```

The iterative differences (deltas) are:

```
d = np.diff(x)
d
## array([-3.6, -4.4, -3.4,  0.6,  0.8,  0. ])
```

For instance, between the second and the first day of the last week, the midrange temperature dropped by -3.6°C.

The other way around, here the cumulative sums of the deltas:

```
np.cumsum(d)
## array([ -3.6,   -8. , -11.4, -10.8, -10. , -10. ])
```

This turned deltas back to a shifted version of the original series. But we will need the first (root) observation therefrom to restore the dataset in full:

```
x[0] + np.append(0, np.cumsum(d))
## array([ -1.4,   -5. ,   -9.4, -12.8, -12.2, -11.4, -11.4])
```

**Exercise 16.5** *Consider the `euraud-20200101-20200630-no-na`[4] dataset which lists daily EUR/AUD exchange rates in the first half of 2020 (remember COVID-19?), with missing observations removed. Using `numpy.diff`, compute the minimum, median, average, and maximum daily price changes. Also, draw a box and whisker plot for these deltas.*

**Example 16.6** *(\*) The exponential distribution family is sometimes used for the modelling of times between different events (deltas). It might be a sensible choice under the assumption that a system generates a constant number of events on average and that they occur independently of each other, e.g., for the times between requests to a cloud service during peak hours, wait times for the next pedestrian to appear at a crossing near the Southern Cross Station in Melbourne, or the amount of time it takes a bank teller to interact with a customer (there is a whole branch of applied mathematics called queuing theory that deals with this type of modelling).*

*An exponential family is identified by the scale parameter $s > 0$, being at the same time its expected value. The probability density function of Exp(s) is given for $x \geq 0$ by:*

$$f(x) = \frac{1}{s} e^{-x/s},$$

*and $f(x) = 0$ otherwise. We should be careful: some textbooks choose the parametrisation by $\lambda = 1/s$ instead of $s$. The **scipy** package also uses this convention.*

*Here is a pseudorandom sample where there are five events per minute on average:*

```
np.random.seed(123)
λ = 60/5  # 5 events per 60 seconds on average
d = scipy.stats.expon.rvs(size=1200, scale=λ)
np.round(d[:8], 3)  # preview
## array([14.307,  4.045,  3.087,  9.617, 15.253,  6.601, 47.412, 13.856])
```

*This gave us the wait times between the events, in seconds.*

*A natural sample estimator of the scale parameter is:*

```
np.mean(d)
## 11.839894504211724
```

*The result is close to what we expected, i.e., $s = 12$ seconds between the events.*

*We can convert the above to datetime (starting at a fixed calendar date) as follows. Note that we will measure the deltas in milliseconds so that we do not loose precision; `datetime64` is based on integers, not floating-point numbers.*

```
t0 = np.array("2022-01-01T00:00:00", dtype="datetime64[ms]")
d_ms = np.round(d*1000).astype(int)  # in milliseconds
```

*(continues on next page)*

---

[4] https://github.com/gagolews/teaching-data/raw/master/marek/euraud-20200101-20200630-no-na.txt

```
t = t0 + np.array(np.cumsum(d_ms), dtype="timedelta64[ms]")
t[:8]  # preview
## array(['2022-01-01T00:00:14.307', '2022-01-01T00:00:18.352',
##        '2022-01-01T00:00:21.439', '2022-01-01T00:00:31.056',
##        '2022-01-01T00:00:46.309', '2022-01-01T00:00:52.910',
##        '2022-01-01T00:01:40.322', '2022-01-01T00:01:54.178'],
##       dtype='datetime64[ms]')
t[-2:]  # preview
## array(['2022-01-01T03:56:45.312', '2022-01-01T03:56:47.890'],
##       dtype='datetime64[ms]')
```

As an exercise, let us apply binning and count how many events occur in each hour:

```
b = np.arange(  # four 1-hour interval (five time points)
    "2022-01-01T00:00:00", "2022-01-01T05:00:00",
    1000*60*60,  # number of milliseconds in 1 hour
    dtype="datetime64[ms]"
)
np.histogram(t, bins=b)[0]
## array([305, 300, 274, 321])
```

We expect 5 events per second, i.e., 300 of them per hour. On a side note, from a course in statistics we know that for exponential inter-event times, the number of events per unit of time follows a Poisson distribution.

**Exercise 16.7** *(\*) Consider the* `wait_times`[5] *dataset that gives the times between some consecutive events, in seconds. Estimate the event rate per hour. Draw a histogram representing the number of events per hour.*

**Exercise 16.8** *(\*) Consider the* `btcusd_ohlcv_2021_dates`[6] *dataset which gives the daily BTC/USD exchange rates in 2021:*

```
btc = pd.read_csv("https://raw.githubusercontent.com/gagolews/" +
    "teaching-data/master/marek/btcusd_ohlcv_2021_dates.csv",
    comment="#").loc[:, ["Date", "Close"]]
btc["Date"] = btc["Date"].astype("datetime64[D]")
btc.head(12)
##          Date      Close
## 0  2021-01-01  29374.152
## 1  2021-01-02  32127.268
## 2  2021-01-03  32782.023
## 3  2021-01-04  31971.914
## 4  2021-01-05  33992.430
```

---

[5] https://github.com/gagolews/teaching-data/raw/master/marek/wait_times.txt
[6] https://github.com/gagolews/teaching-data/raw/master/marek/btcusd_ohlcv_2021_dates.csv

```
## 5  2021-01-06  36824.363
## 6  2021-01-07  39371.043
## 7  2021-01-08  40797.609
## 8  2021-01-09  40254.547
## 9  2021-01-10  38356.441
## 10 2021-01-11  35566.656
## 11 2021-01-12  33922.961
```

Write a function that converts it to a lagged representation, *being a convenient form for some machine learning algorithms.*

1. *Add the* `Change` *column that gives by how much the price changed since the previous day.*

2. *Add the* `Dir` *column indicating if the change was positive or negative.*

3. *Add the* `Lag1, ..., Lag5` *columns which give the* `Changes` *in the five preceding days.*

*The first few rows of the resulting data frame should look like this (assuming we do not want any missing values):*

```
##         Date  Close   Change Dir     Lag1     Lag2     Lag3    Lag4    Lag5
## 2021-01-07 39371  2546.68 inc  2831.93  2020.52 -810.11  654.76 2753.12
## 2021-01-08 40798  1426.57 inc  2546.68  2831.93 2020.52 -810.11  654.76
## 2021-01-09 40255  -543.06 dec  1426.57  2546.68 2831.93 2020.52 -810.11
## 2021-01-10 38356 -1898.11 dec  -543.06  1426.57 2546.68 2831.93 2020.52
## 2021-01-11 35567 -2789.78 dec -1898.11  -543.06 1426.57 2546.68 2831.93
## 2021-01-12 33923 -1643.69 dec -2789.78 -1898.11 -543.06 1426.57 2546.68
```

*In the 6th row (representing 2021-01-12),* `Lag1` *corresponds to* `Change` *on 2021-01-11,* `Lag2` *gives the* `Change` *on 2021-01-10, and so forth.*

*To spice things up, make sure your code can generate any number (as defined by another parameter to the function) of lagged variables.*

### 16.3.2 Smoothing with Moving Averages

With time series it makes sense to consider processing whole batches of consecutive points, because there is a time dependence between them. In particular, we can consider computing different aggregates inside *rolling windows* of a particular size. For instance, the *k-moving average* of a given sequence $(x_1, x_2, \dots, x_n)$ is a vector $(y_1, y_2, \dots, y_{n-k+1})$ such that:

$$y_i = \frac{1}{k}\left(x_i + x_{i+1} + \dots + x_{i+k-1}\right) = \frac{1}{k}\sum_{j=1}^{k} x_{i+j-1},$$

i.e., the arithmetic mean of $k \le n$ consecutive observations starting at $x_i$.

For example, here are the temperatures in the last 7 days of December 2011:

```
x = spokane.set_index("date").iloc[-7:, :]
x
##              temp
## date
## 2021-12-25  -1.4
## 2021-12-26  -5.0
## 2021-12-27  -9.4
## 2021-12-28 -12.8
## 2021-12-29 -12.2
## 2021-12-30 -11.4
## 2021-12-31 -11.4
```

The 3-moving (rolling) average:

```
x.rolling(3, center=True).mean().round(2)
##              temp
## date
## 2021-12-25    NaN
## 2021-12-26  -5.27
## 2021-12-27  -9.07
## 2021-12-28 -11.47
## 2021-12-29 -12.13
## 2021-12-30 -11.67
## 2021-12-31    NaN
```

We get, in this order: the mean of the first three observations; the mean of the 2nd, 3rd, and 4th items; then the mean of the 3rd, 4th, and 5th; and so forth. Notice that the observations were centred in such a way that we have the same number of missing values at the start and end of the series. This way, we treat the first 3-day moving average (the average of the temperatures on the first three days) as representative of the 2nd day.

And now for something completely different; the 5-moving average:

```
x.rolling(5, center=True).mean().round(2)
##              temp
## date
## 2021-12-25    NaN
## 2021-12-26    NaN
## 2021-12-27  -8.16
## 2021-12-28 -10.16
## 2021-12-29 -11.44
## 2021-12-30    NaN
## 2021-12-31    NaN
```

Applying the moving average has the nice effect of *smoothing* out all kinds of broadly conceived noise. To illustrate this, compare the temperature data for the last five years in Figure 16.3 to their averaged versions in Figure 16.5.

```python
x = spokane.set_index("date").loc["2017-01-01":, "temp"]
x30 = x.rolling(30, center=True).mean()
x100 = x.rolling(100, center=True).mean()
plt.plot(x30, label="30-day moving average")
plt.plot(x100, "r--", label="100-day moving average")
plt.legend()
plt.show()
```

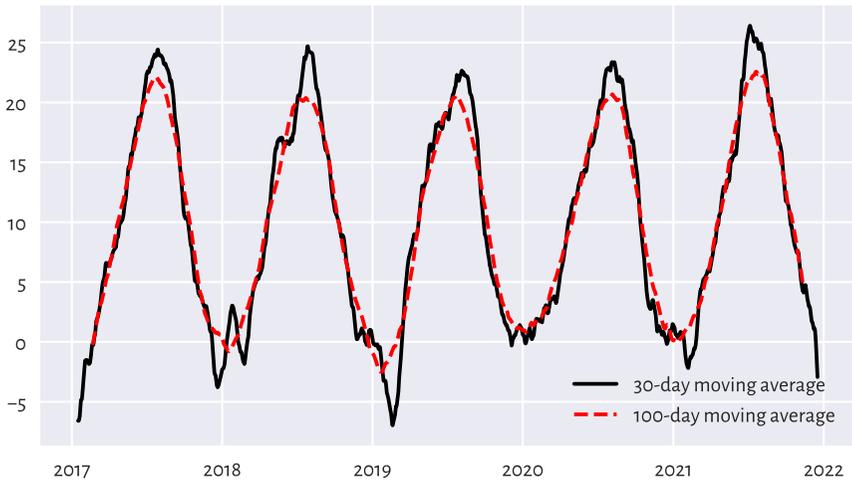

Figure 16.5: Line chart of 30- and 100-moving averages of the midrange daily temperatures in Spokane for 2017-2021

**Exercise 16.9**   (*) *Other aggregation functions can be applied in rolling windows as well. Draw, in the same figure, the plots of the 1-year moving minimums, medians, and maximums.*

### 16.3.3    Detecting Trends and Seasonal Patterns

Thanks to windowed aggregation, we can also detect general trends (when using longish windows). For instance, below we compute the 10-year moving averages for the last 50-odd years' worth of data:

```python
x = spokane.set_index("date").loc["1970-01-01":, "temp"]
x10y = x.rolling(3653, center=True).mean()
```

Based on this, we can compute the detrended series:

```
xd = x - x10y
```

Seasonal patterns can be revealed by smoothening out the detrended version of the data, e.g., using a 1-year moving average:

```
xd1y = xd.rolling(365, center=True).mean()
```

Figure 16.6 illustrates this.

```
plt.plot(x10y, label="trend")
plt.plot(xd1y, "r--", label="seasonal pattern")
plt.legend()
plt.show()
```

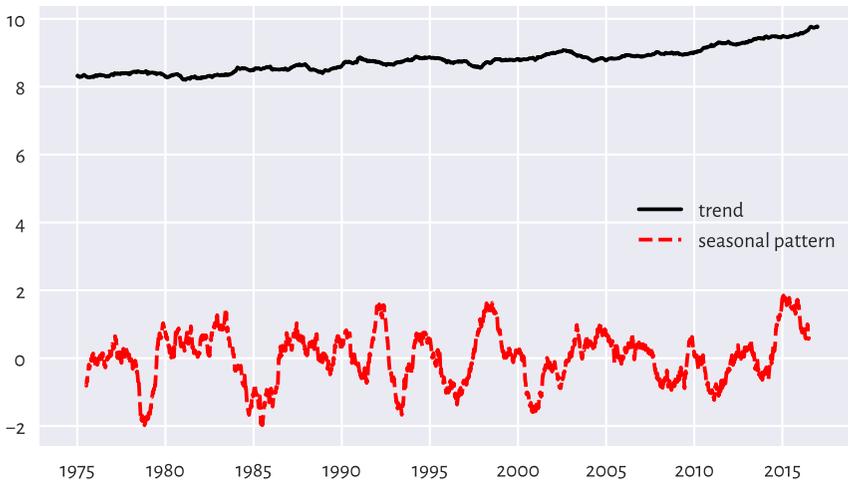

Figure 16.6: Trend and seasonal pattern for the Spokane temperatures in recent years

Also, if we know the length of the seasonal pattern (in our case, 365-ish days), we can draw a seasonal plot, where we have a separate curve for each season (here: year) and where all the series share the same x-axis (here: the day of the year); see Figure 16.7.

```
from matplotlib import cm
cmap=cm.get_cmap("coolwarm")
x = spokane.set_index("date").loc["1970-01-01":, :].reset_index()
for year in range(1970, 2022, 5):                     # selected years only
    y = x.loc[x.date.dt.year == year, :]
    plt.plot(y.date.dt.day_of_year, y.temp,
        c=cmap((year-1970)/(2021-1970)), alpha=0.3,
        label=year if year % 10 == 0 else None)
```

*(continues on next page)*

```
avex = x.temp.groupby(x.date.dt.day_of_year).mean()
plt.plot(avex.index, avex, "g-", label="Average")    # all years
plt.legend()
plt.xlabel("Day of year")
plt.ylabel("Temperature")
plt.show()
```

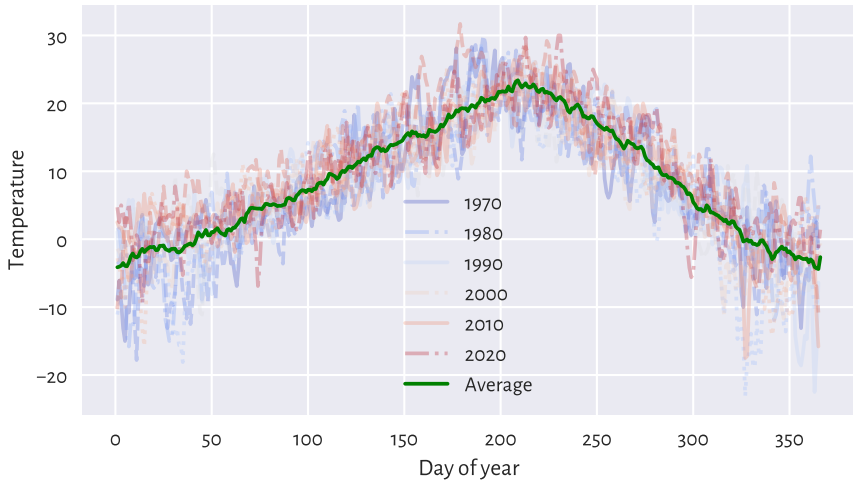

Figure 16.7: Seasonal plot: Temperatures in Spokane vs the day of the year; years between 1970 and 2021

**Exercise 16.10** *Draw a similar plot for the whole data range, i.e., 1889–2021.*

### 16.3.4  Imputing Missing Values

Missing values in time series can be imputed based on the information from the neighbouring non-missing observations. After all, it is usually the case that, e.g., today's weather is "similar" to yesterday's and tomorrow's.

The most straightforward ways for dealing with missing values in time series are:

- *forward-fill* – propagate the last non-missing observation,

- *backward-fill* – get the next non-missing value,

- *linearly interpolate between two adjacent non-missing values* – in particular, a single missing value will be replaced by the average of its neighbours.

**Example 16.11** *The classic* `air_quality_1973`[7] *dataset gives some daily air quality measure-*

---

[7] https://github.com/gagolews/teaching-data/raw/master/r/air_quality_1973.csv

ments in New York, between May and September 1973. Let us impute the first few observations in the solar radiation column:

```
air = pd.read_csv("https://raw.githubusercontent.com/gagolews/" +
    "teaching-data/master/r/air_quality_1973.csv",
    comment="#")
x = air.loc[:, "Solar.R"].iloc[:12]
pd.DataFrame(dict(
    original=x,
    ffilled=x.fillna(method="ffill"),
    bfilled=x.fillna(method="bfill"),
    interpolated=x.interpolate(method="linear")
))
##     original  ffilled  bfilled  interpolated
## 0     190.0    190.0    190.0     190.000000
## 1     118.0    118.0    118.0     118.000000
## 2     149.0    149.0    149.0     149.000000
## 3     313.0    313.0    313.0     313.000000
## 4       NaN    313.0    299.0     308.333333
## 5       NaN    313.0    299.0     303.666667
## 6     299.0    299.0    299.0     299.000000
## 7      99.0     99.0     99.0      99.000000
## 8      19.0     19.0     19.0      19.000000
## 9     194.0    194.0    194.0     194.000000
## 10      NaN    194.0    256.0     225.000000
## 11    256.0    256.0    256.0     256.000000
```

**Exercise 16.12** (*) With the `air_quality_2018`[8] dataset:

1. Based on the hourly observations, compute the daily mean PM2.5 measurements for Melbourne CBD and Morwell South.

   For Melbourne CBD, if some hourly measurement is missing, linearly interpolate between the preceding and following non-missing data, e.g., a PM2.5 sequence of [..., 10, NaN, NaN, 40, ...] (you need to manually add the NaN rows to the dataset) should be transformed to [..., 10, 20, 30, 40, ...].

   For Morwell South, impute the readings with the averages of the records in the nearest air quality stations, which are located in Morwell East, Moe, Churchill, and Traralgon.

2. Present the daily mean PM2.5 measurements for Melbourne CBD and Morwell South on a single plot. The x-axis labels should be human-readable and intuitive.

3. For the Melbourne data, determine the number of days where the average PM2.5 was greater than in the preceding day.

4. Find five most air-polluted days for Melbourne.

---

[8] https://github.com/gagolews/teaching-data/raw/master/marek/air_quality_2018.csv.gz

### 16.3.5    Plotting Multidimensional Time Series

Multidimensional time series stored in the form of an *n*-by-*m* matrix are best viewed as *m* time series – possibly but not necessarily related to each other – all sampled at the same *n* points in time (e.g., *m* different stocks on *n* consecutive days).

Consider the currency exchange rates for the first half of 2020:

```python
eurxxx = np.loadtxt("https://raw.githubusercontent.com/gagolews/" +
    "teaching-data/master/marek/eurxxx-20200101-20200630-no-na.csv",
    delimiter=",")
eurxxx[:6, :]  # preview
## array([[1.6006 , 7.7946 , 0.84828, 4.2544 ],
##        [1.6031 , 7.7712 , 0.85115, 4.2493 ],
##        [1.6119 , 7.8049 , 0.85215, 4.2415 ],
##        [1.6251 , 7.7562 , 0.85183, 4.2457 ],
##        [1.6195 , 7.7184 , 0.84868, 4.2429 ],
##        [1.6193 , 7.7011 , 0.85285, 4.2422 ]])
```

This gives EUR/AUD (how many Australian Dollars we pay for 1 Euro), EUR/CNY (Chinese Yuans), EUR/GBP (British Pounds), and EUR/PLN (Polish Złotys), in this order. Let us draw the four time series; see Figure 16.8.

```python
dates = np.loadtxt("https://raw.githubusercontent.com/gagolews/" +
    "teaching-data/master/marek/euraud-20200101-20200630-dates.txt",
    dtype="datetime64")
labels = ["AUD", "CNY", "GBP", "PLN"]
styles = ["solid", "dotted", "dashed", "dashdot"]
for i in range(eurxxx.shape[1]):
    plt.plot(dates, eurxxx[:, i], ls=styles[i], label=labels[i])
plt.legend(loc="upper right", bbox_to_anchor=(1, 0.9))  # a bit lower
plt.show()
```

Unfortunately, they are all on different scales. This is why the plot is not necessarily readable. It would be better to draw these time series on four separate plots (compare the trellis plots in Section 12.2.5).

Another idea is to depict the currency exchange rates *relative* to the prices on some day, say, the first one; see Figure 16.9.

```python
for i in range(eurxxx.shape[1]):
    plt.plot(dates, eurxxx[:, i]/eurxxx[0, i],
        ls=styles[i], label=labels[i])
plt.legend()
plt.show()
```

This way, e.g., a relative EUR/AUD rate of ca. 1.15 in mid-March means that if an Aussie bought some Euros on the first day, and then sold them three-ish months later, they would have 15% more wealth (the Euro become 15% stronger relative to AUD).

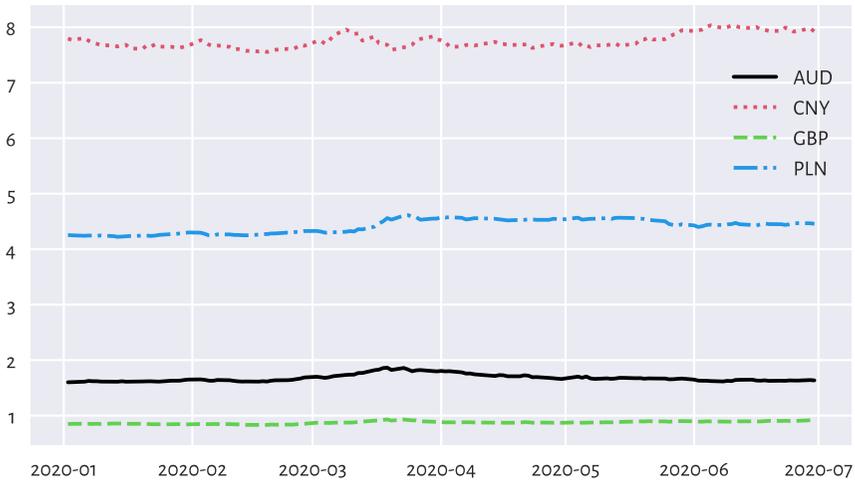

Figure 16.8: EUR/AUD, EUR/CNY, EUR/GBP, and EUR/PLN exchange rates in the first half of 2020

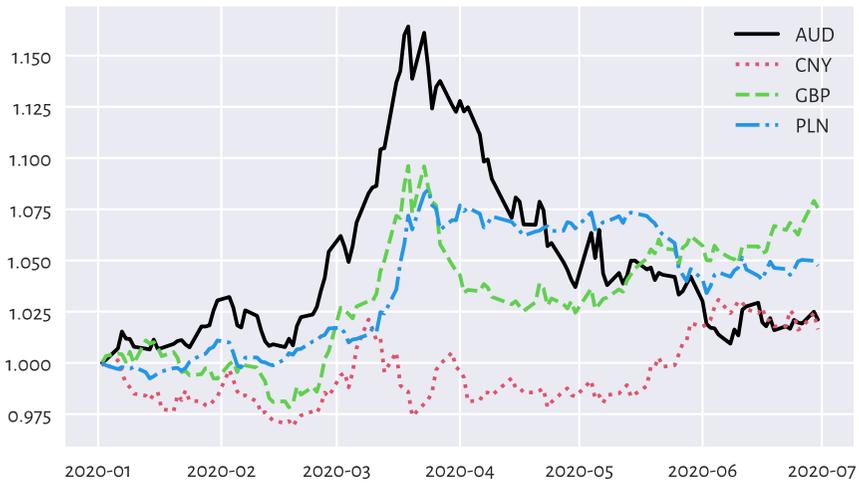

Figure 16.9: EUR/AUD, EUR/CNY, EUR/GBP, and EUR/PLN exchange rates relative to the prices on the first day

**Exercise 16.13**  *Based on the EUR/AUD and EUR/PLN records, compute and plot the AUD/PLN as well as PLN/AUD rates.*

**Exercise 16.14**  *(\*) Draw the EUR/AUD and EUR/GBP rates on a single plot, but where each series has its own[9] y-axis.*

**Exercise 16.15**  *(\*) Draw the EUR/xxx rates for your favourite currencies over a larger period. Use data[10] downloaded from the European Central Bank. Add a few moving averages. For each year, identify the lowest and the highest rate.*

### 16.3.6   Candlestick Plots (*)

Consider the BTC/USD data for 2021:

```
btcusd = np.loadtxt("https://raw.githubusercontent.com/gagolews/" +
    "teaching-data/master/marek/btcusd_ohlcv_2021.csv",
    delimiter=",")
btcusd[:6, :4]  # preview (we skip the Volume column for readability)
## array([[28994.01 , 29600.627, 28803.586, 29374.152],
##        [29376.455, 33155.117, 29091.182, 32127.268],
##        [32129.408, 34608.559, 32052.316, 32782.023],
##        [32810.949, 33440.219, 28722.756, 31971.914],
##        [31977.041, 34437.59 , 30221.188, 33992.43 ],
##        [34013.613, 36879.699, 33514.035, 36824.363]])
```

This gives the open, high, low, and close (OHLC) prices on the 365 consecutive days, which is a common way to summarise daily rates.

The `mplfinance`[11] package (`matplotlib-finance`) features a few functions related to the plotting of financial data. Here, let us briefly describe the well-known candlestick plot.

```
import mplfinance as mpf
dates = np.arange("2021-01-01", "2022-01-01", dtype="datetime64[D]")
mpf.plot(pd.DataFrame(
    btcusd,
    columns=["Open", "High", "Low", "Close", "Volume"]
).set_index(dates).iloc[:31, :], type="candle")
# plt.show() # not needed...
```

Figure 16.10 depicts the January 2021 data. Let us stress that this is not a box and whisker plot. The candlestick body denotes the difference in the market opening and the closing price. The wicks (shadows) give the range (high to low). White candlesticks represent bullish days – where the closing rate is greater than the opening one (uptrend). Black candles are bearish (decline).

---

[9] https://matplotlib.org/stable/gallery/subplots_axes_and_figures/secondary_axis.html
[10] https://www.ecb.europa.eu/stats/policy_and_exchange_rates/euro_reference_exchange_rates/html/index.en.html
[11] https://github.com/matplotlib/mplfinance

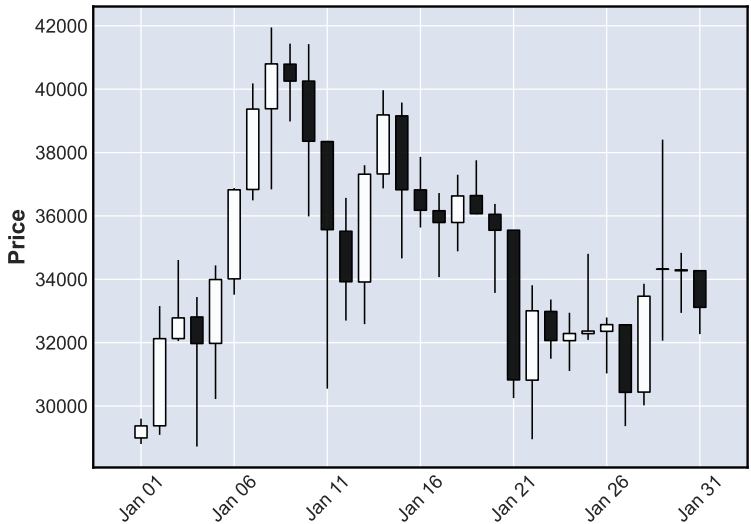

Figure 16.10: Candlestick plot for the BTC/USD exchange rates in January 2021

**Exercise 16.16** *Draw the BTC/USD rates for the entire year and add the 10-day moving averages.*

**Exercise 16.17** *(\*) Draw a candlestick plot manually, without using the `mplfinance` package. Hint: `matplotlib.pyplot.fill` might be helpful.*

**Exercise 16.18** *(\*) Using `matplotlib.pyplot.fill_between` add a semi-transparent polygon that fills the area bounded between the Low and High prices on all the days.*

## 16.4   Further Reading

Data science classically deals with information that is or can be represented in tabular form and where particular observations (which can be multidimensional) are usually independent from but still to some extent similar to each other. We often treat them as samples from different larger populations which we would like to describe or compare at some level of generality (think: health data on patients being subject to two treatment plans that we wish to evaluate).

From this perspective, time series are already quite distinct, because there is some dependence observed in the time domain: a price of a stock that we observe today is influenced by what was happening yesterday. There might also be some seasonal pat-terns or trends under the hood. For a good introduction to forecasting; see [46, 63]. Also, for data of this kind, employing statistical modelling techniques (*stochastic processes*) can make a lot of sense; see, e.g., [76].

*Signals* such as audio, images, and video are different, because *structured randomness* does not play a dominant role there (unless it is a noise that we would like to filter out). Instead, what is happening in the frequency (think: perceiving pitches when listening to music) or spatial (seeing green grass and sky in a photo) domain will play a key role there.

Signal processing thus requires a distinct set of tools, e.g., Fourier analysis and finite impulse response (discrete convolution) filters. This course obviously cannot be about everything (also because it requires some more advanced calculus skills that we did not assume the reader to have at this time); but see, e.g., [74, 75].

Nevertheless, we should keep in mind that these are not completely independent domains. For example, we can extract various features of audio signals (e.g., overall loudness, timbre, and danceability of each recording in a large song database) and then treat them as tabular data to be analysed using the techniques described in this course. Moreover, machine learning (e.g., convolutional neural networks) algorithms may also be used for tasks such as object detection on images or optical character recognition; see, e.g., [37].

## 16.5    Exercises

**Exercise 16.19**  *Assume we have a time series with* n *observations. What is a 1- and an n-moving average? Which one is smoother, a* (0.01n)- *or a* (0.1n)- *one?*

**Exercise 16.20**  *What is the Unix Epoch?*

**Exercise 16.21**  *How can we recreate the original series when we are given its* `numpy.diff`-*transformed version?*

**Exercise 16.22**  *(\*) In your own words, describe the key elements of a candlestick plot.*

# Part VI

# Appendix

# A

## *Changelog*

**Important**  Any bug/typos reports/fixes[1] are appreciated.

Below is the list of the most noteworthy changes.

- **under development (v1.0.2.9xxx):**
  - Numeric reference style; updated bibliography.
  - (...) to do (...)
- **2022-08-24 (v1.0.2):**
  - First printed (paperback) version can be ordered from Amazon[2].
  - Fix page margins and headers.
  - Minor typesetting and other fixes.
- **2022-08-12 (v1.0.1):**
  - Cover.
  - ISBN 978-0-6455719-1-2 assigned.
- **2022-07-16 (v1.0.0):**
  - Preface complete.
  - Handling tied observations.
  - Plots look better when printed in black and white.
  - Exception handling.
  - File connections.
  - Other minor extensions and material reordering: more aggregation functions, **pandas.unique**, **pandas.factorize**, probability vectors representing binary categorical variables, etc.
  - Final proof-reading.

---

[1] https://github.com/gagolews/datawranglingpy/issues
[2] https://www.amazon.com/dp/0645571911

- **2022-06-13 (v0.5.1):**
    - The Kolmogorov–Smirnov Test (one and two sample).
    - The Pearson Chi-Squared Test (one and two sample and for independence).
    - Dealing with round-off and measurement errors.
    - Adding white noise (jitter).
    - Lambda expressions.
    - Matrices are iterable.
- **2022-05-31 (v0.4.1):**
    - The Rules.
    - Matrix multiplication, dot products.
    - Euclidean distance, few-nearest-neighbour and fixed-radius search.
    - Aggregation of multidimensional data.
    - Regression with $k$-nearest neighbours.
    - Least squares fitting of linear regression models.
    - Geometric transforms; orthonormal matrices.
    - SVD and dimensionality reduction/PCA.
    - Classification with $k$-nearest neighbours.
    - Clustering with $k$-means.
    - Text Processing and Regular Expression chapters merged.
    - Unidimensional Data Aggregation and Transformation chapters merged.
    - `pandas.GroupBy` objects are iterable.
    - Semitransparent histograms.
    - Contour plots.
    - Argument unpacking and variadic arguments (`*args`, `**kwargs`).
- **2022-05-23 (v0.3.1):**
    - More lightweight mathematical notation.
    - Some equalities related to the mathematical functions we rely on (the natural logarithm, cosine, etc.).
    - A way to compute the most correlated pair of variables.
    - A note on modifying elements in an array and on adding new rows and columns.
    - An example seasonal plot in the time series chapter.

- Solutions to the SQL exercises added; to ignore small round-off errors, use `pandas.testing.assert_frame_equal` instead of `pandas.DataFrame.equals`.

- More details on file paths.

- **2022-04-12 (v0.2.1)**:

  - Many chapters merged or relocated.

  - Added captions to all figures.

  - Improved formatting of elements (information boxes such as *note*, *important*, *exercise*, *example*).

- **2022-03-27 (v0.1.1)**:

  - First public release – most chapters are drafted, more or less.

  - Using `Sphinx` for building.

- **2022-01-05 (v0.0.0)**:

  - Project started.